\RequirePackage{fix-cm}
\documentclass[twocolumn]{svjour3}          
\smartqed  
\usepackage{graphicx}
\usepackage{enumerate}
\usepackage{color, colortbl}
\definecolor{LightCyan}{rgb}{0.88,1,1}
\usepackage{bm}
\usepackage{amsbsy}

\usepackage{mathptmx}      
\usepackage{amsmath,amssymb} 
\usepackage{caption}

\def\bbbz{{\mathbb Z}}

\newcommand{\sign}{\operatorname{sign}}

 \journalname{To appear in Journal of Mathematical Imaging and Vision}
\begin{document}


\title{Dynamic texture recognition using time-causal and time-recursive spatio-temporal receptive fields \thanks{The support from the Swedish Research Council 
             (Contract 2014-4083) and Stiftelsen Olle Engkvist
             Byggm{\"a}stare (Contract 2015/465) is gratefully acknowledged.}
}


\author{Ylva Jansson and Tony Lindeberg
}


\institute{Ylva Jansson \and
	       Tony Lindeberg  \\	
	        \email{yjansson@kth.se \and tony@kth.se}  \\
		 \at
	      Computational Brain Science Lab \\
	      Department of Computational Science and Technology \\ 
	      KTH Royal Institute of Technology, Stockholm, Sweden 
}


\maketitle

\begin{abstract}
This work presents a first evaluation of using spatio-temporal receptive fields from a recently proposed time-causal spatio-temporal scale-space framework as primitives for video analysis. We propose a new family of video descriptors based on regional statistics of spatio-temporal receptive field responses and evaluate this approach on the problem of dynamic texture recognition. Our approach generalises a previously used method, based on joint histograms of receptive field responses, from the spatial to the spatio-temporal domain and from object recognition to dynamic texture recognition. 
The time-recursive formulation enables computationally efficient time-causal recognition.

The experimental evaluation demonstrates competitive performance compared to state-of-the-art. Especially, it is shown that binary versions of our dynamic texture descriptors achieve improved performance compared to a large range of similar methods using different primitives either handcrafted or learned from data. Further, our qualitative and quantitative investigation into parameter choices and the use of different sets of receptive fields highlights the robustness and flexibility of our approach. Together, these results support the descriptive power of this family of time-causal spatio-temporal receptive fields, validate our approach for dynamic texture recognition 
and point towards the possibility of designing a range of video analysis methods based on these new time-causal spatio-temporal primitives.


\end{abstract}

\section{Introduction}
\label{sec:intro}

The ability to derive properties of the surrounding world from
time-varying visual input is a key function of a general purpose computer
vision system and necessary for any artificial or biological agent
that is to use vision for interpreting a dynamic environment. 
Motion provides additional cues for understanding a scene and some tasks
by necessity require motion information, e.g. distinguishing between events or actions with similar spatial appearance or estimating the speed of moving objects. 
Motion cues are also helpful when other visual cues are weak or contradictory.
Thus, understanding how to represent and incorporate motion information 
for different video analysis tasks -- including the question of what constitute meaningful spatio-temporal features -- 
is a key factor for successful applications in areas such as action recognition, automatic surveillance, 
dynamic texture and scene understanding, video-indexing and retrieval, autonomous driving, etc. 

Challenges when dealing with spatio-temporal image data are similar to those present in the static case, that is high-dimensional data with large intraclass
variability caused both by the diverse appearance of conceptually similar objects and the presence of 
a large number of identity preserving visual transformations. 
From the spatial domain, we inherit the basic image transformations: translations, scalings, rotations, nonlinear perspective transformations and illumination transformations. In addition to these, spatio-temporal data will contain additional sources of variability: differences in motion patterns for conceptually similar motions, events occurring faster or slower, velocity transformations caused by camera motion and time-varying illumination.
Moreover, the larger data dimensionality 
compared to static images presents additional computational challenges. 

For biological vision, local image measurements in terms of receptive fields
constitute the first processing layers (Hubel and Wiesel \cite{HubWie05-book}; DeAngelis et al. \cite{DeAngOhzFre95-TINS}).
In the area of scale-space theory, it has been shown that Gaussian derivatives and related operators constitute a canonical model for 
visual receptive fields (Iijima \cite{Iij62-TR}; 
Witkin \cite{Wit83}; 
Koenderink \cite{Koe84-BC,Koe88-BC}; 
Koenderink and van Doorn \cite{KoeDoo87-BC,KoeDoo92-PAMI}; 
Lindeberg \cite{Lin93-Dis,Lin10-JMIV};  
Florack \cite{florack:97}; 
Sporring et al.\ \cite{SpoNieFloJoh96-SCSPTH}; 
Weickert et al.\ \cite{WeiIshImi99-JMIV}; 
ter Haar Romeny \mbox{et al.\ \cite{Haa04-book,RomFloNie01-SCSP}).}
In computer vision, spatial receptive fields based on the Gaussian
scale-space concept have been demonstrated to be a powerful front-end for solving a
large range of visual tasks. Theoretical properties of scale-space
filters enable the design of methods invariant or robust to natural image transformations (Lindeberg \cite{Lin10-JMIV,Lin13-BICY,Lin-PONE2013,Lin16-JMIV}). Also, early processing layers based on such ``ideal"
receptive fields can be shared among multiple tasks and thus free resources both for learning
higher level features from data and during on-line processing. 

The most straightforward extension of the Gaussian scale-space concept 
to the spatio-temporal domain is to use Gaussian smoothing also over the temporal domain. 
However, for real-time processing or when modelling biological vision this would violate the fundamental \emph{temporal causality} constraint present for real-time tasks: It is simply not possible to access the future. 
The ad hoc solution of using a time-delayed truncated Gaussian temporal kernel would instead imply unnecessarily long temporal delays, which would make the framework less suitable for time-critical applications.  

The preferred option is to use truly time-casual visual operators. 
Recently, a new time-causal spatio-temporal 
scale-space framework, leading to a new family of \emph{time-causal spatio-temporal receptive fields}, has been introduced by Lindeberg \cite{Lin16-JMIV}. In addition to temporal causality, the time-recursive
formulation of the temporal smoothing operation offers computational advantages
compared to Gaussian filtering over time in terms of fewer computations and a 
compact recursively updated memory of the past.

The idealised spatio-temporal receptive fields derived\\ within that framework also have a strong connection to
biology in the sense of very well modelling both spatial and spatio-temporal receptive field shapes
of neurons in the LGN and V1 (Lindeberg \cite{Lin13-BICY,Lin16-JMIV}). 
This similarity further motivates designing algorithms based on these primitives. 
They provably work well in a biological system, which points towards 
their applicability also for artificial vision. An additional motivation, although not actively pursued here,
is that designing methods based on primitives similar to biological receptive fields 
can enable a better understanding of information processing in biological systems. 

The purpose of this study is a first evaluation of using this new family of time-causal
spatio-temporal receptive fields as visual primitives for video
analysis. As a first application, we have chosen the problem of dynamic texture recognition. 
A dynamic texture or spatio-temporal texture is an extension of texture
to the spatio-temporal domain and can be naively defined as 
``texture + motion'' or more formally as a spatio-temporal pattern
that exhibits certain stationarity or self-similar properties
over both space and time (Chetverikov and P{\'e}teri \cite{ChePet-book2005}). Examples of dynamic textures are
windblown vegetation, fire, waves, a flock of flying birds or
a flag flapping in the wind. Recognising different types of dynamic
textures is important for visual tasks such as
automatic surveillance (e.g. detecting forest fires), video indexing and
retrieval (e.g. return all images set on the sea) and to enable
artificial agents to understand and interact with the world
by interpreting different environments. 

In this work, we start by presenting a new family of video
descriptors in the form of joint histograms of spatio-temporal receptive field
responses. Our approach generalises a previous method by Linde and
Lindeberg \cite{LinLin12-CVIU} from the spatial to the spatio-temporal
domain and from object recognition to dynamic texture recognition. We subsequently perform an experimental evaluation on two commonly used
dynamic texture datasets and present results on:
\begin{enumerate}[(i)]
\item Qualitative and quantitative effects from varying model parameters
including the spatial and the temporal scales and the number of principal components 
and the number of bins used in the histogram descriptor.
\item A comparison between the performance of descriptors 
constructed from different sets of receptive fields.
\item An extensive comparison with state-of-the-art dynamic texture recognition methods.
\end{enumerate}
Our benchmark results demonstrate competitive performance compared to state-of-the-art dynamic texture recognition methods. Notably, we achieve very good results using a conceptually simple method that only makes use of local information and with the additional constraint of using time-causal operators. 
Especially, it is shown that binary versions of our dynamic texture descriptors achieve improved performance compared to a large range of similar methods using different primitives either handcrafted or learned from data. Further, our qualitative and quantitative investigation into parameter choices and the use of different sets of receptive fields highlights the robustness and flexibility of our approach. 

Together, these results support the descriptive power of this family of time-causal spatio-temporal receptive fields, validate our approach for dynamic texture recognition and point towards the possibility of designing a range of video analysis methods based on these new time-causal spatio-temporal primitives.

This paper is an extended version of a conference paper \cite{JanLin-SSVM2017},
in which only a single descriptor version was evaluated. We here present a more extensive experimental evaluation,
including results on varying the descriptor parameters and of using different sets of receptive fields. We specifically present new video descriptors  with improved performance compared to the previous video descriptor used in \cite{JanLin-SSVM2017}.

\begin{figure*}[hbpt]

  \begin{center}
    \begin{tabular}{@{} c @{} c @{} c @{} c @{} c@{} c @{} c @{}}          
           \includegraphics[width=0.135\textwidth]{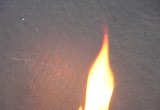} 
       & \includegraphics[width=0.135\textwidth]{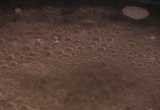} 
           & \includegraphics[width=0.135\textwidth]{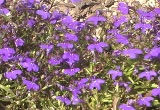} 
        & \includegraphics[width=0.135\textwidth]{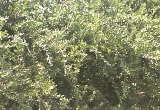}
          & \includegraphics[width=0.135\textwidth]{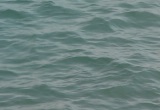} 
           & \includegraphics[width=0.135\textwidth]{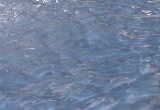}
        & \includegraphics[width=0.135\textwidth]{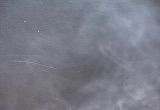}\\
        	 \includegraphics[width=0.135\textwidth]{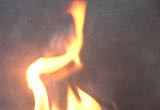} 
        & \includegraphics[width=0.135\textwidth]{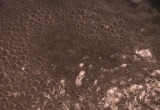} 
        &  \includegraphics[width=0.135\textwidth]{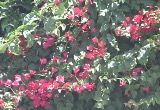} 
        & \includegraphics[width=0.135\textwidth]{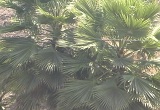} 
        & \includegraphics[width=0.135\textwidth]{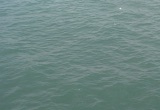}
        & \includegraphics[width=0.135\textwidth]{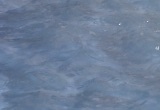}
 	& \includegraphics[width=0.135\textwidth]{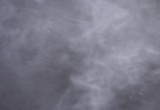}
  \end{tabular}
  \end{center}
  \caption{{\em Sample frames from the UCLA dataset}. The top and bottom rows show different dynamic texture instances from the same conceptual class as used in the UCLA8 and UCLA9 benchmarks. For the UCLA50 benchmark, these instances should instead be separated as different classes. The examples are from the conceptual classes (from left to right): ``fire", ``boiling", ``flowers", ``plants", ``sea", ``water" and ``smoke".}
  \label{fig:dataset-ucla}
  
  \begin{center}
    \begin{tabular}{@{} c @{} c @{} c @{} c @{} c@{} c @{} c @{}}
        \includegraphics[width=0.135\textwidth]{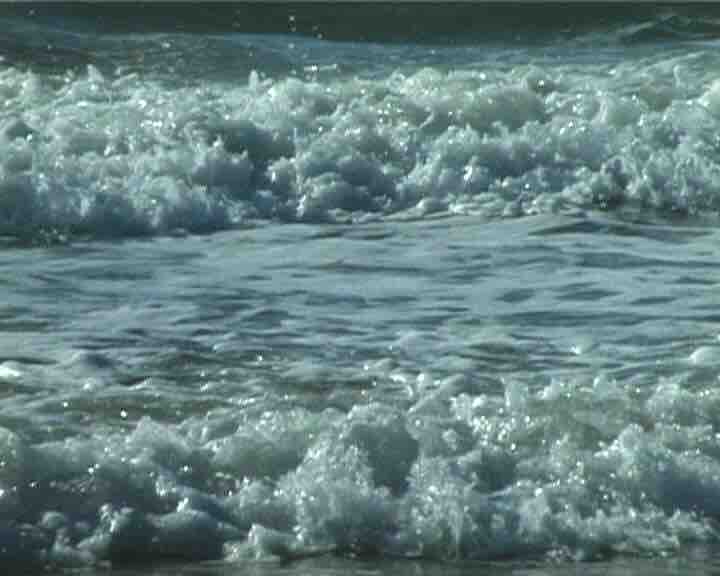} 
         & \includegraphics[width=0.135\textwidth]{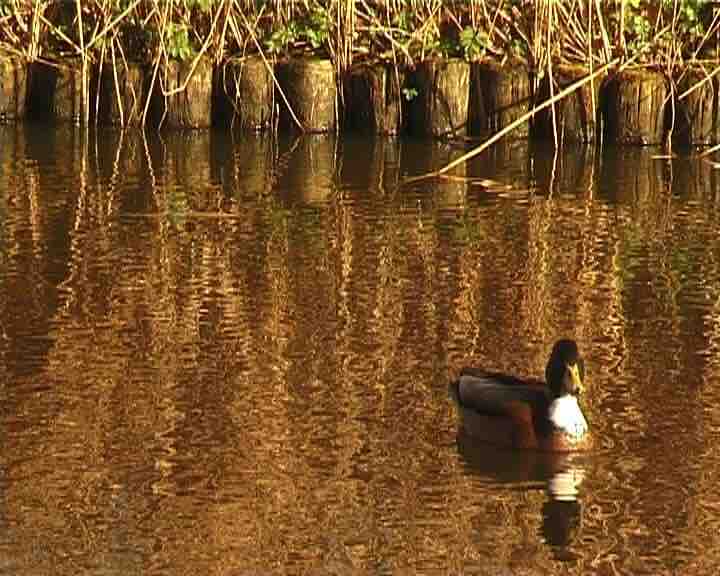}
         & \includegraphics[width=0.135\textwidth]{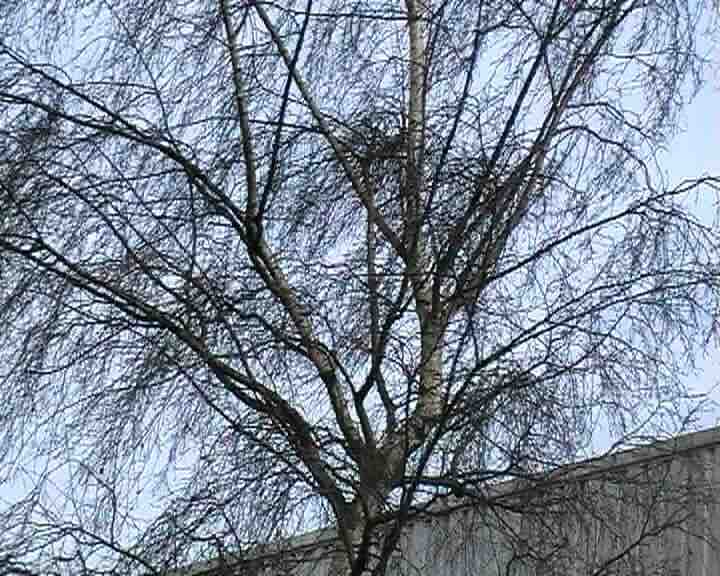}
         & \includegraphics[width=0.135\textwidth]{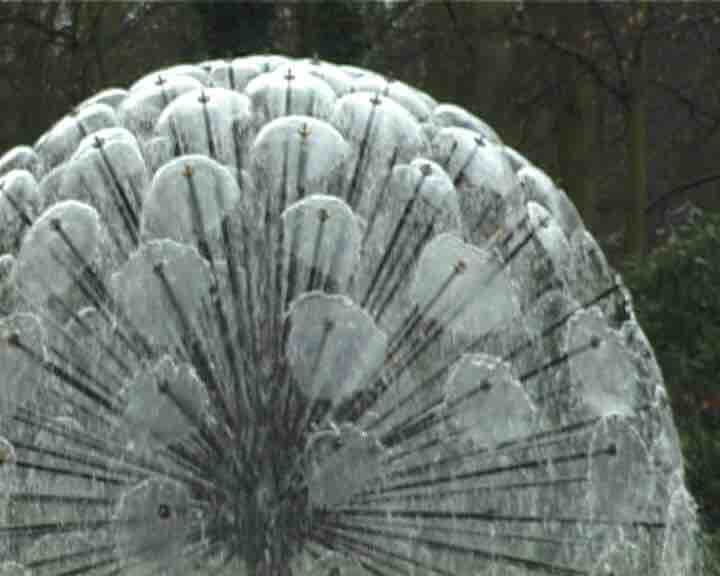} 
        &  \includegraphics[width=0.135\textwidth]{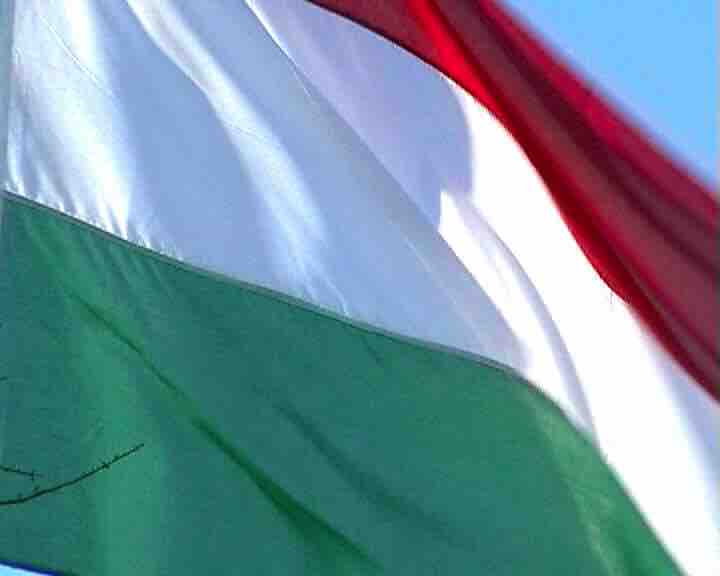}
        & \includegraphics[width=0.135\textwidth]{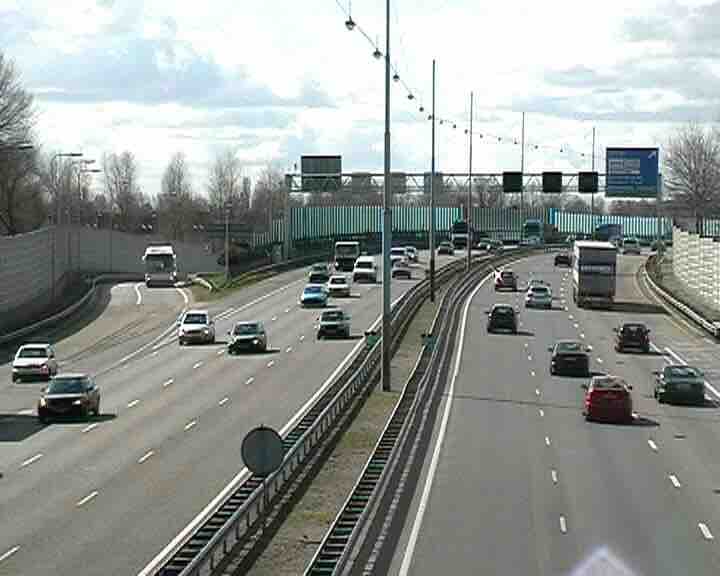} 
        & \includegraphics[width=0.135\textwidth]{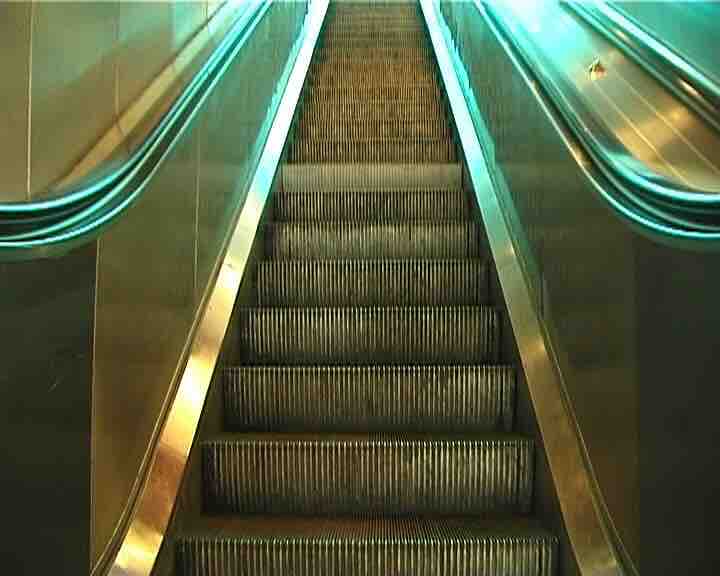} 
         \\

        \includegraphics[width=0.135\textwidth]{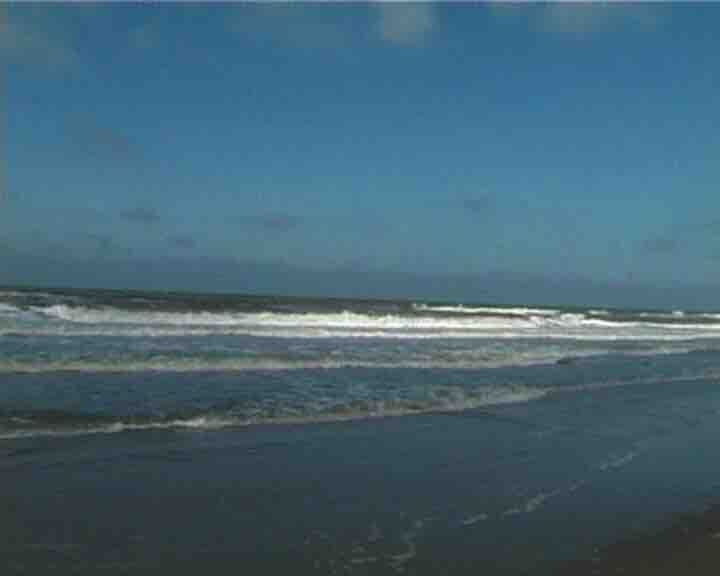}   
        & \includegraphics[width=0.135\textwidth]{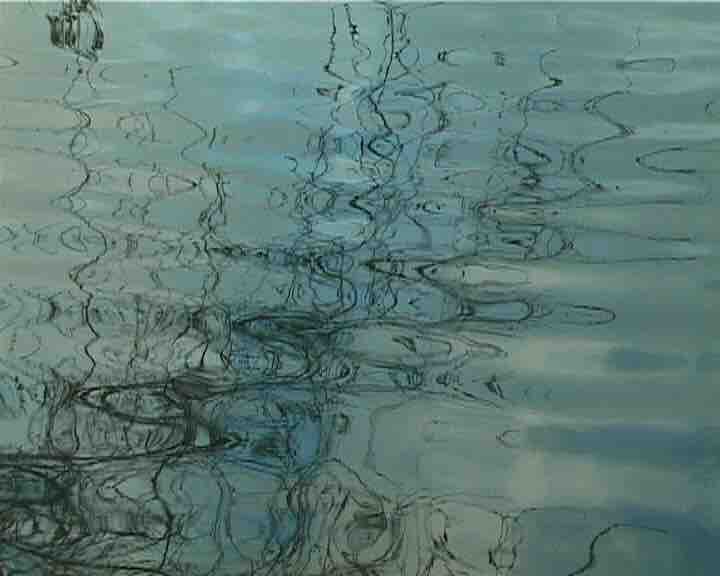} 
         & \includegraphics[width=0.135\textwidth]{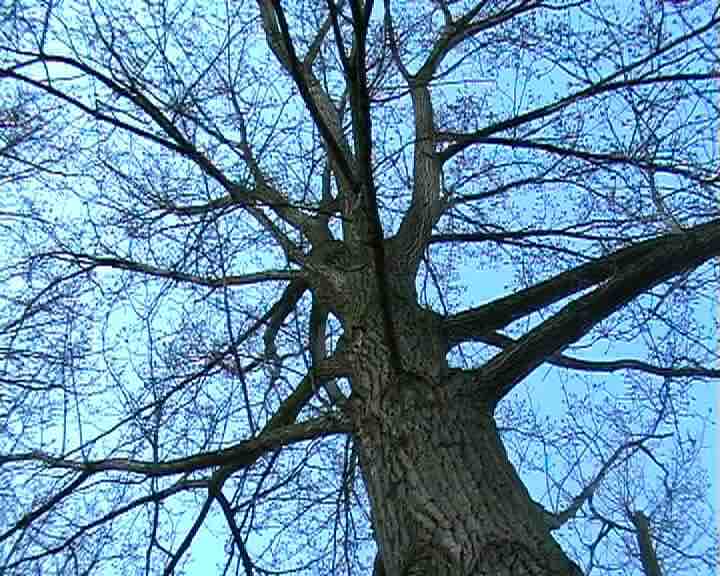}
         & \includegraphics[width=0.135\textwidth]{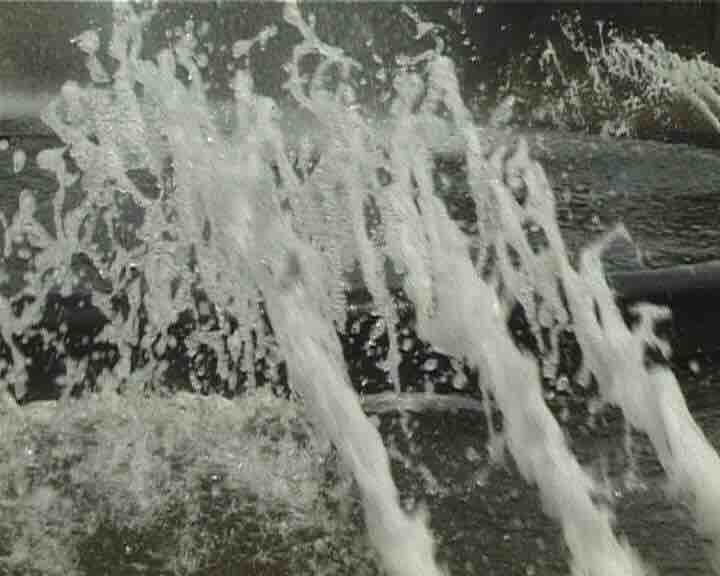}
        & \includegraphics[width=0.135\textwidth]{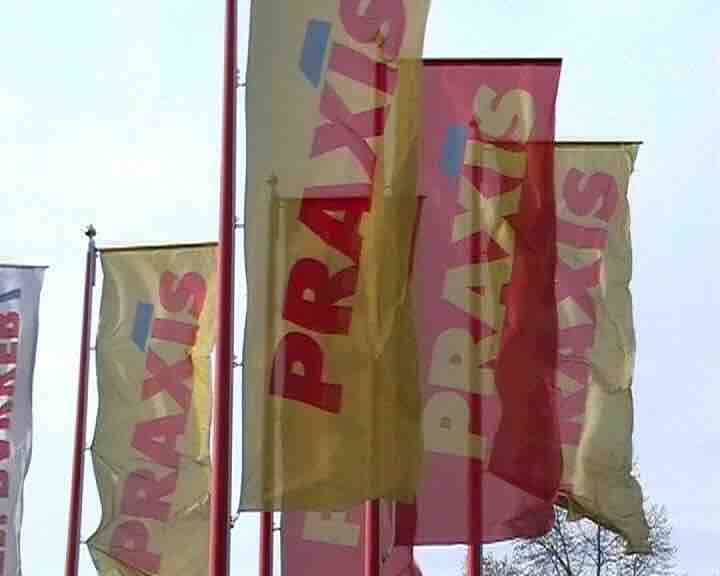}
        & \includegraphics[width=0.135\textwidth]{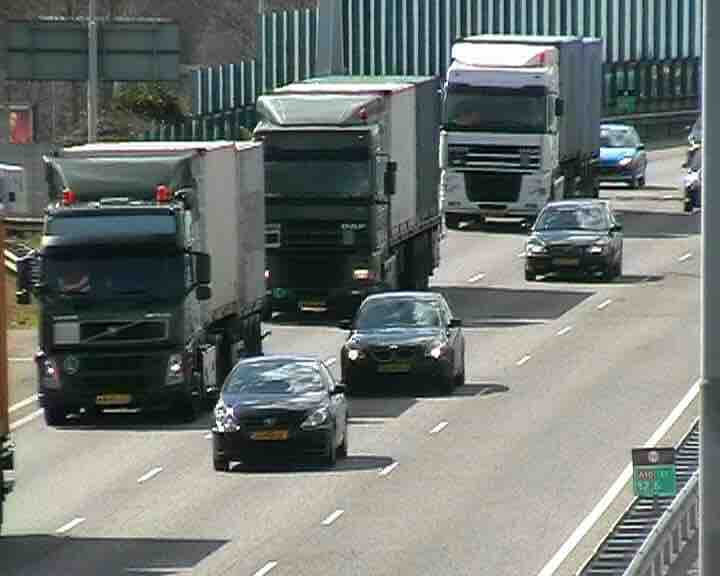} 
        & \includegraphics[width=0.135\textwidth]{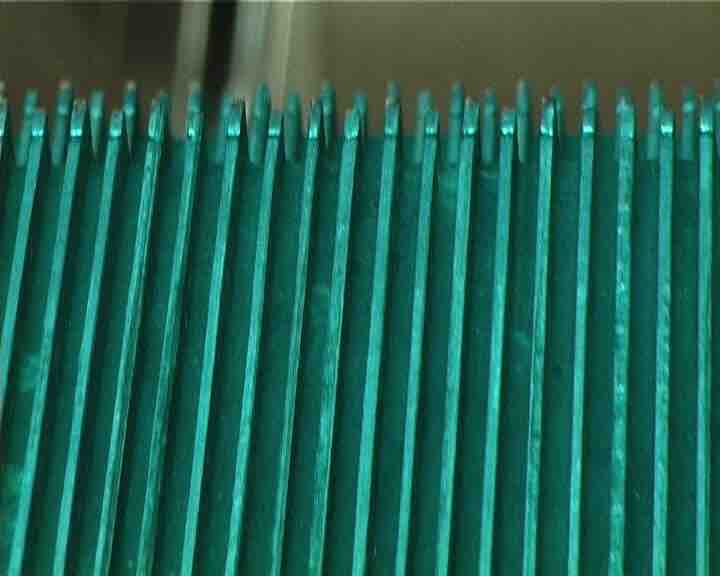} 
        
  \end{tabular}
  \end{center}
  \caption{{\em Sample frames from the DynTex
    dataset}. The top and bottom rows show different dynamic texture instances from the same conceptual class. The examples are from the conceptual classes (from left to right):  ``sea",``calm water", ``naked trees", ``fountains", ``flags", ``traffic" and ``escalator".}
  \label{fig:dataset-dyntex}
\end{figure*}

\section{Related work}
\label{sec:related-work}
For dynamic texture recognition, additional challenges compared to the static case 
include variabilities in motion patterns and speed and much higher data dimensionality. 
Although spatial texture descriptors can give reasonable performance also in the dynamic case (a human observer can typically distinguish dynamic textures based on a single frame), making use of dynamic information as well provides a consistent advantage over purely spatial descriptors (see e.g. Zhao et~al.\ \cite{ZhaGuoPie-TPAMI-2007}; Arashloo and Kittler \cite{AraKit-TOM2014}; Hong et~al. \cite{HonRyuetal-MSSP2016}; Qi et~al.\ \cite{QiLietal-NC2016} and  Andrearczyk and Whelan \cite{AndWhe-arXiV2017}). Thus, a large range of \emph{dynamic} texture recognition methods have been developed, exploring different options for representing motion information and combining this with spatial information.

Some of the first methods for dynamic texture recognition were based on \emph{optic flow}, see e.g. the pioneering work by Nelson and Polana \cite{NelPol-CVGIP1992} and later work by P{\'e}teri and Chetverikov \cite{PetChe-PRIA2005}, Lu et al.\ \cite{LuXiePeiHua-WMVC2005}, Fazekas and Chetverikov \cite{FazChe-SPIC2007}, Fazekas et al.\ \cite{FazAmiCheKir-IJCV2009} and Crivelli et al.\ \cite{CriCerBouYao-JIS2013}. 

Compared to the motion patterns of larger rigid objects, however, dynamic textures often feature chaotic non-smooth motions, multiple directions of motion at the same image point and intensity changes not mediated by rigid motion -- consider e.g. 
fluttering leaves, turbulent water and fire. Thus, the brightness constancy assumption, usually underlying optical flow estimation, is violated for many dynamic texture types. This implies difficulties estimating optic flow for dynamic textures, although alternative types of assumptions such as brightness conservation and color constancy, have later been explored e.g. in the context of dynamic texture detection and segmentation (Fazekas et~al. \cite{FazAmiCheKir-IJCV2009,CheFazHai-MVA2011}).
 
Another early approach, first proposed by Soatto et~al. \cite{SoaDorWu-ICCV2001}, is to model the time-evolving appearance of a dynamic texture as a \emph{linear dynamical system} (LDS). Although originally designed for synthesis, recognition can be done using  model parameters as features. A drawback of such global LDS models, however, is poor invariance to viewpoint and illumination conditions  (Woolfe and Fitzgibbon \cite{WolFit-ECCV2006}).
To circumvent the heavy dependence of the first LDS models on global spatial appearance, the \emph{bags of dynamical systems} (BoS) approach was developed by Ravichandran et~al. \cite{RavChaVid-CVPR2009}. Here, the global model is replaced by a set of codebook LDS models, each describing a small space-time cuboid, used as local descriptors in a bag-of-words framework. For additional LDS-based approaches, see also Chan and Vasconcelos \cite{ChaVas-CVPR2009}, Mumtaz et al.\ \cite{MumCovLanCha-TPAMI2015}, Qiao and Weng \cite{QiaWen-SPL2015}, Wang et al.\ \cite{WanLiuSun-NC2016} and Sagel and Kleinsteuber \cite{SagKle-arXiv2017}.

The use of \emph{histograms of local image measurements} as visual descriptors was pioneered by Swain and Ballard \cite{SwaBal-IJCV1991}, who proposed using 3D colour histograms for object recognition. Different types of histogram descriptors have subsequently proven highly useful for a large range of visual tasks (Schiele and Crowley \cite{SchCro00-IJCV}; Zelnik-Manor and Irani \cite{ZelIra01-CVPR}; Lowe \cite{Low04-IJCV}; Laptev and Lindeberg \cite{LapLin04-ECCVWS}; Linde and Lindeberg \cite{LinLin04-ICPR,LinLin12-CVIU}; Dalal et al.\ \cite{DalTri05-CVPR,DalTriSch06-ECCV}; Lazebnik et al.\ \cite{LazSchPon06-CVPR}; Kl{\"a}ser et al.\ \cite{KlaMarSch08-BMVC}; van de Sande et al.\ \cite{SanGevSno10-PAMI}; Hernandez et al.\ \cite{HerCroLuxPie-book2014}).

A large group of dynamic texture recognition methods based on statistics of local image primitives are \emph{local binary pattern} (LBP) based approaches. The original spatial LBP descriptor captures the joint binarized distribution of pixel values in local neighbourhoods. The spatio-temporal generalisations of LBP used for dynamic texture recognition do the same either for 3D space-time volumes (VLBP) (Zhao et~al. \cite{ZhaPie-WDV2006,ZhaGuoPie-TPAMI-2007}) or by applying
a two-dimensional LBP descriptor but on \emph{three orthogonal planes} (LBP-TOP) (Zhao et~al. \cite{ZhaGuoPie-TPAMI-2007}). 
Several extensions/versions of LBP-TOP have subsequently been presented, e.g. utilising averaging and principal histogram analysis to get more reliable statistics (Enhanced LBP) (Ren et~al. \cite{RenJiaYua-ICASSP2013}) or multi-clustering of salient features to identify and remove outlier frames (AFS-TOP) (Hong et~al. \cite{HonRyuetal-MSSP2016}). See also the completed local binary pattern approach (CVLBP) (Tiwari and Tyagi) \cite{TiwTya-MSSP2016} and multiresolution edge-weight\-ed local structure patterns (MEWLSP) (Tiwari and Tyagi \cite{TiwTya-CEE2016}).

Contrasting LBP-TOP with VLBP highlights a conceptual difference concerning the generalisation of spatial descriptors to the spatio-temporal domain: Whereas the VLBP descriptor is based on full space-time 2D+T features, the LBP-TOP descriptor applies 2D features originally designed for purely spatial tasks on several cross sections of a spatio-temporal volume. While the former is in some sense conceptually more appealing 
and implies more discriminative modelling of the space-time structure, 
this comes with the drawback of higher descriptor dimensionality and higher computational load. 

For LBP-based methods, using 2D descriptors on three orthogonal planes has so far been proven more successful and this approach is frequently used also by non LBP-based methods (Andrearczyk and Whelan \cite{AndWhe-arXiV2017}; Arashloo and Kittler \cite{AraKit-TOM2014}; Arashloo et al.\ \cite{AraAmiNor-JVCIR2017}; Xu et al.\ \cite{XuQuaetal-PR2015}). A variant of this idea proposed by Norouznezhad et al.\ \cite{NorHaretal-ECCV2012} is to replace the three orthogonal planes with \emph{nine spatio-temporal symmetry planes} and apply a histogram of oriented gradients on these planes. Similarly, the directional number transitional graph (DNG) by Rivera et al.\ \cite{RivCha-TPAMI2015} is evaluated on nine spatio-temporal cross sections of a video.

A different group of approaches similarly based on gathering statistics of local space-time structure, but using different primitives, is \emph{spatio-temporal filtering} based approaches. Examples are the oriented energy representations by Wildes and Bergen \cite{WilBer-ECCV2000} and Derpanis and Wildes \cite{DerWil-TPAMI2012}, where the latter represents \emph{pure dynamics} of spatio-temporal textures by capturing space-time orientation using 3D Gaussian derivative filters. Marginal histograms of spatio-temporal Gabor filters were proposed by Gon{\c{c}}alves et~al. \cite{GonNunetal-ArXiv2012}. 
Our proposed approach, which is based on \emph{joint histograms} of time-causal spatio-temporal receptive field responses, also fits into this category. 
However, in contrast to spatio-temporal filtering based
approaches using marginal histograms, the use of joint histograms, which can capture the covariation of different features,  enables distinguishing a larger number of local space-time patterns. Joint statistics of ideal scale-space filters have previously been used for spatial object recognition, see e.g. Schiele and Crowley \cite{SchCro00-IJCV} and the approach by Linde and Lindeberg \cite{LinLin04-ICPR,LinLin12-CVIU}, which we here generalise to the spatio-temporal domain. 
The use of \emph{time-causal} filters in our approach also contrasts with other spatio-temporal filtering based approaches and it should also be noted that the time-causal limit kernel used in this paper 
is \emph{time-recursive}, whereas no time-recursive formulation is known for the scale-time kernel previously proposed by Koenderink \cite{Koe88-BC}.

There also exist a number of related methods similarly based on joint statistics of local space-time structure but using \emph{filters learned from data} as primitives. \emph{Unsupervised learning} based approaches are e.g. multi-scale binarized statistical image features (MBSIF-TOP) by Arashloo and Kittler \cite{AraKit-TOM2014}, which learn filters by means of independent component analysis and PCANet-TOP by 
Arashloo et~al. \cite{AraAmiNor-JVCIR2017}, where two layers of hierarchical features are learnt by means of principal component analysis (PCA). Approaches instead based on \emph{sparse coding} for learning filters are orthogonal tensor dictionary learning (OTD) (Quan et~al. \cite{QuaHuaJi-ICCV2015}), equiangular kernel dictionary learning (SKDL) (Quan et al.\ 
 \cite{QuaCheHui-CVPR2016}) and manifold regularised slow feature analysis (MR-SFA) (Miao et al.\ \cite{MiaXuXinTao-arXiv2017}).

Recently, several \emph{deep learning} based approaches for dynamic texture recognition 
have been proposed. These are typically trained by \emph{supervised learning} and learn complex and abstract high-level features from data.
Andrearczyk and Whelan \cite{AndWhe-arXiV2017} train a dynamic texture convolutional neural network from scratch to extract features on three orthogonal planes (DT-CNN). Qi et~al.\ \cite{QiLietal-NC2016} instead use a network pretrained on an object recognition task for feature extraction (TCoF), whereas Hong et~al. \cite{HonRyuImYan-arXiv2017} use features from a pretrained network as the basis for a deep dual descriptor (D3). Additional approaches include the high-level feature approach by Wang and Hu \cite{WanHu-NC2015}, which is a hybrid method using deep learning in combination with chaotic features, and the early deep learning approach of Culibrk and Sebe \cite{CulSeb-ICM2014} based on temporal dropout of changes. It should be noted that none of these approaches are based on full \emph{spatio-temporal deep features}. They instead use features extracted from individual frames, differences between pairs of frames 
or on orthogonal space-time planes. 
The best deep learning approaches are among the best performing dynamic texture recognition methods, but this comes at the price of a lack of understanding and interpretability. The properties of the learned non-linear mappings and why these prove successful are not fully understood, neither in general nor for a network trained on a specific task. These highly complex "black box" methods may also suffer more from surprising and unintuitive failure modes. 
Compared to methods incorporating more prior information about the problem structure, deep learning also requires larger amounts of training data from the specific domain, or one similar enough for successful transfer learning. 

\emph{Spatio-temporal transforms} have also been applied for dynamic texture recognition. Ji et~al. \cite{JiYanetal-TIP2013} propose a method based on wavelet domain fractal analysis (WMFS), whereas Dubois et al.\ \cite{DubSloetal-SIVP2015} utilise the 2D+T curvelet transform and Smith et al.\ \cite{SmiLinNap-ICIP2002} use spatio-temporal wavelets for video texture indexing. \emph{Fractal dimension} based methods make use of the self-similarity properties of dynamic textures. Xu et~al.\ \cite{XuQuaetal-PR2015} create a descriptor from the fractal dimension of motion features (DFS), whereas Xu et al. \cite{XuHuaJuFer-CVIU2012} instead extract the power-law behaviour of local multi-scale gradient orientation histograms (3D-OTF) (see also Ghanem and Ahuja \cite{GhaAhu-ICCV2007} and Smith et al.\ \cite{SmiLinNap-ICIP2002}). 

Additional approaches include using average degrees of complex networks (Gon{\c{c}}alves et al.\ \cite{GonWesMacetal-NC2015}) and a total variation based approach by El Moubtahij et al.\ \cite{MouAugFeretal-ECQCA2015}. 
Wang and Hu\ \cite{WanHu-SC2016} instead create a descriptor from chaotic features. There are also approaches combining several different descriptor types or features, such as DL-PEGASOS by Ghanem and Ahuja \cite{GhaAhu-ECCV2010}, which combines LBP, PHOG and LDS descriptors with maximum margin distance learning, or Yang et~al. \cite{YanXiaetal-NC2016} who use ensemble SVMs to combine LBP, shape-invariant co-occurrence patterns (SCOPs) and chromatic information with dynamic information represented by LDS models.

\begin{figure}[hbpt]
  \begin{center}
    \begin{tabular}{ccc}
     \hspace{-2mm} {\footnotesize $T(x, t;\; s, \tau)$} \hspace{-2mm} 
     & \hspace{-2mm} {\footnotesize $T_t(x, t;\; s, \tau)$} \hspace{-2mm} 
     & \hspace{-2mm} {\footnotesize $T_{tt}(x, t;\; s, \tau)$} \hspace{-2mm} \\
      \hspace{-2mm} \includegraphics[width=0.15\textwidth]{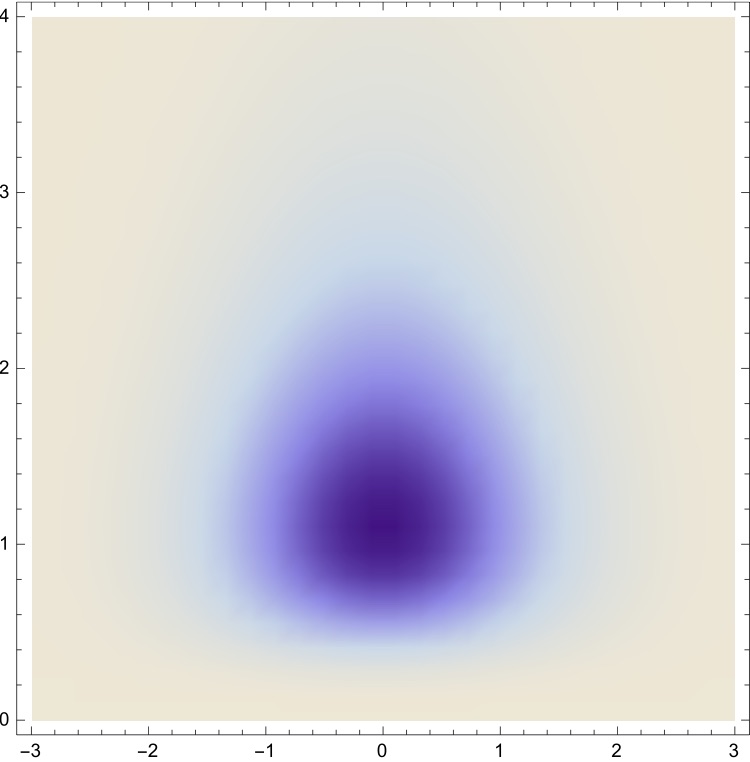}
      \hspace{-2mm} 
      & \hspace{-2mm} \includegraphics[width=0.15\textwidth]{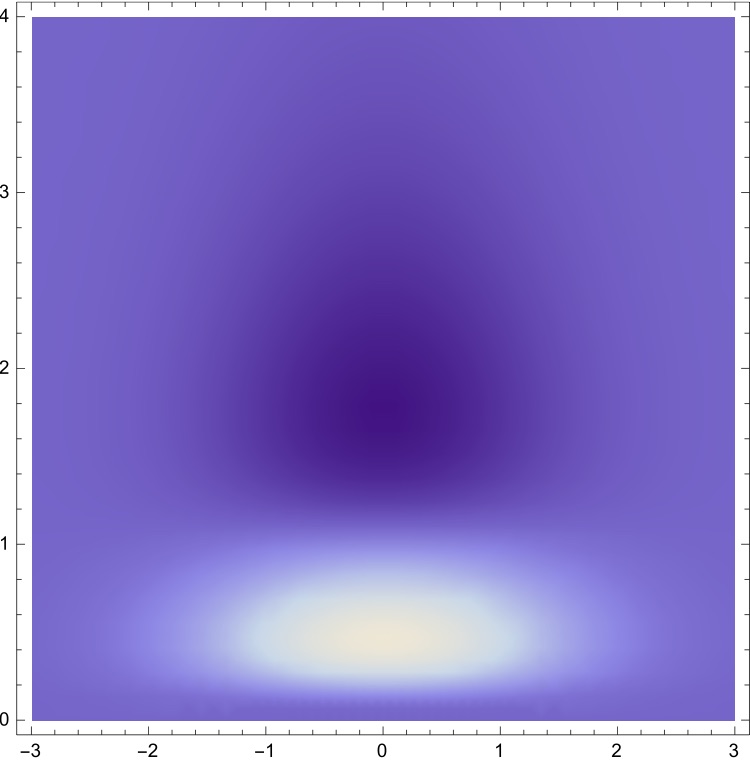} \hspace{-2mm} 
     & \hspace{-2mm} \includegraphics[width=0.15\textwidth]{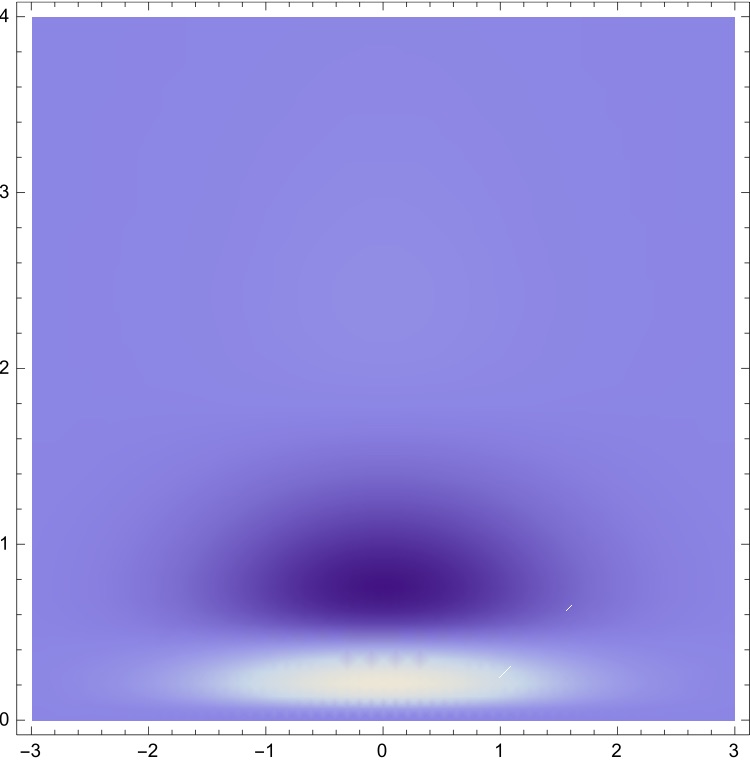} \hspace{-2mm} \\
    \end{tabular} 
  \end{center}
  \vspace{-6mm}
  \begin{center}
    \begin{tabular}{ccc}
      \hspace{-2mm} {\footnotesize $T_x(x, t;\; s, \tau)$} \hspace{-2mm} 
      & \hspace{-2mm} {\footnotesize $T_{xt}(x, t;\; s, \tau)$} \hspace{-2mm} 
      & \hspace{-2mm} {\footnotesize $T_{xtt}(x, t;\; s, \tau)$} \hspace{-2mm} \\
      \hspace{-2mm} \includegraphics[width=0.15\textwidth]{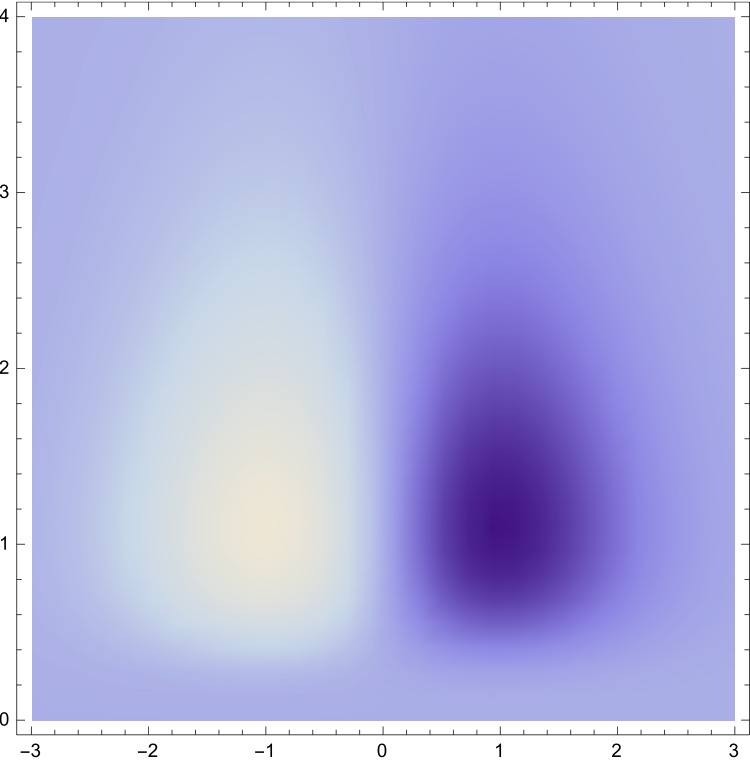} \hspace{-2mm} &
     \hspace{-2mm} \includegraphics[width=0.15\textwidth]{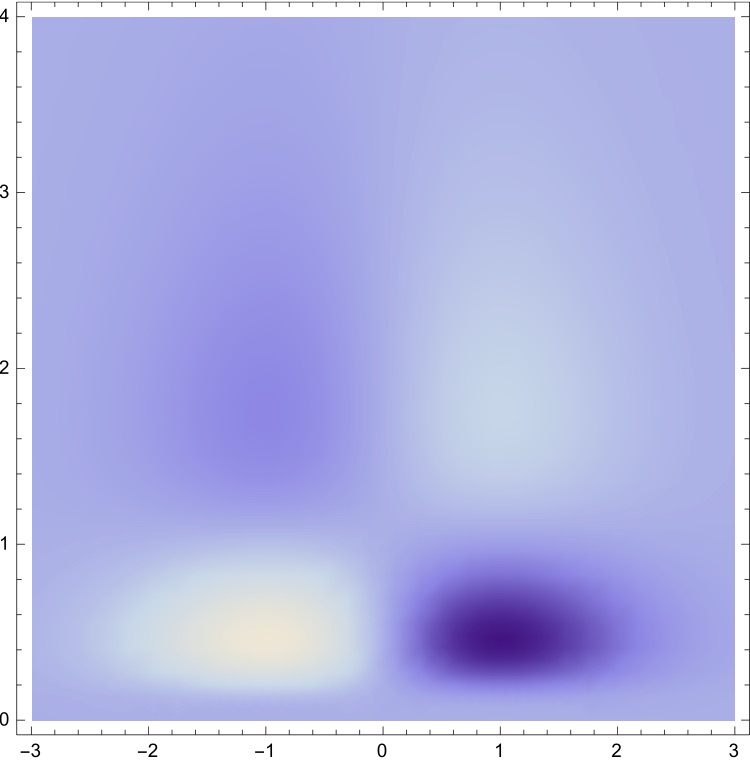} \hspace{-2mm} &
      \hspace{-2mm} \includegraphics[width=0.15\textwidth]{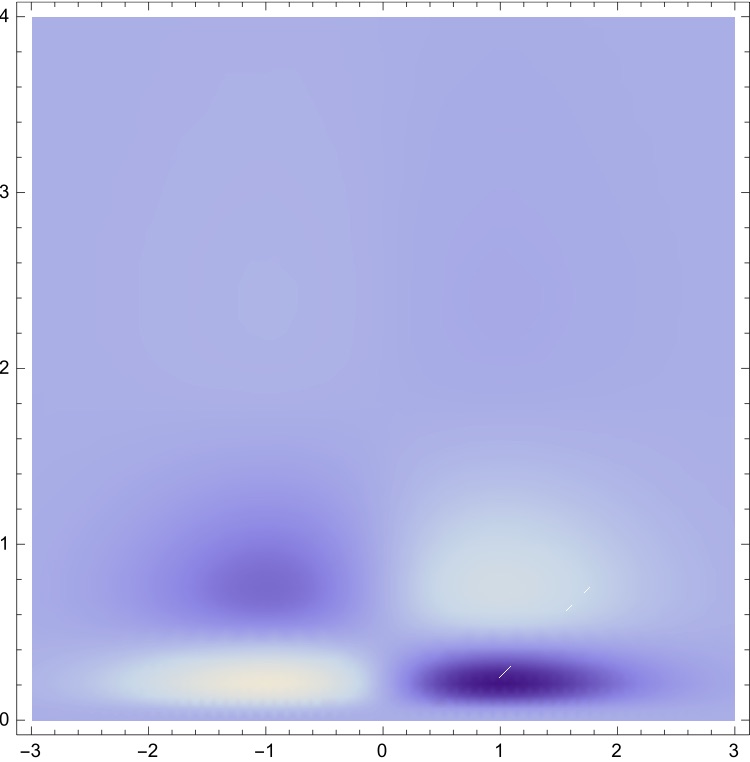} \hspace{-2mm} \\
    \end{tabular} 
  \end{center}
  \vspace{-6mm}
  \begin{center}
    \begin{tabular}{ccc}
      \hspace{-2mm} {\footnotesize $T_{xx}(x, t;\; s, \tau)$} \hspace{-2mm} 
      & \hspace{-2mm} {\footnotesize $T_{xxt}(x, t;\; s, \tau)$} \hspace{-2mm} 
      & \hspace{-2mm} {\footnotesize $T_{xxtt}(x, t;\; s, \tau)$} \hspace{-2mm} \\
      \hspace{-2mm} \includegraphics[width=0.15\textwidth]{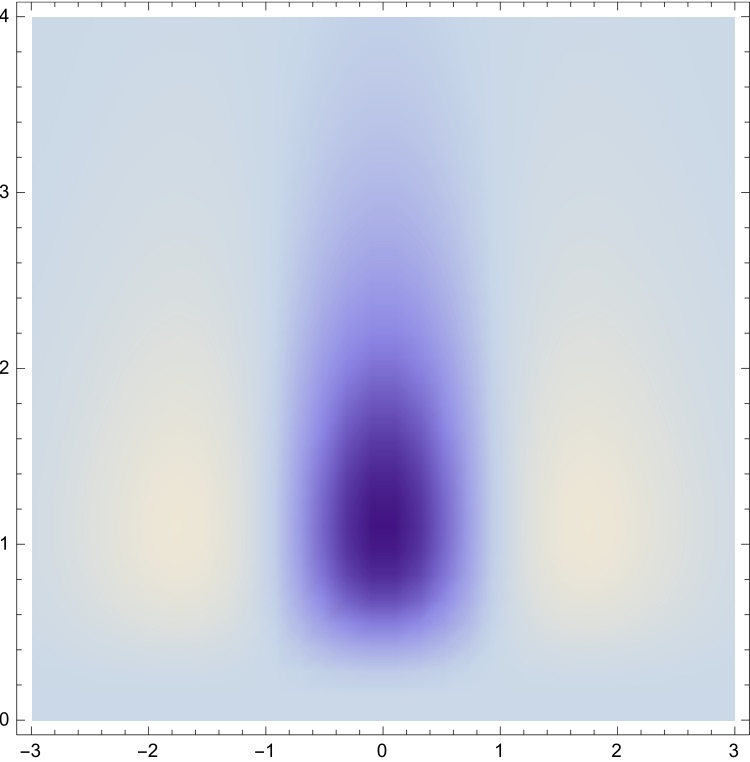} \hspace{-2mm} &
      \hspace{-2mm} \includegraphics[width=0.15\textwidth]{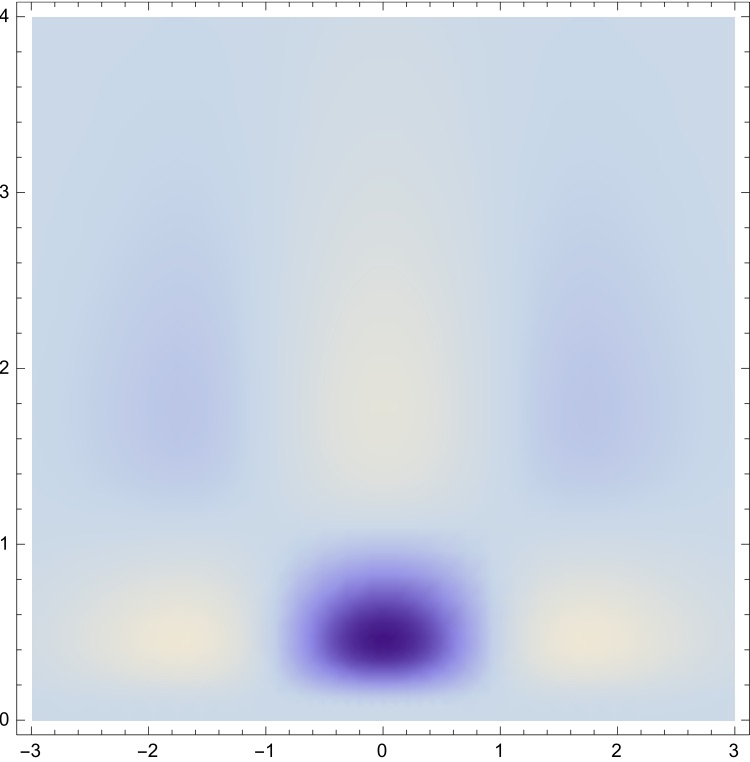} \hspace{-2mm} &
      \hspace{-2mm} \includegraphics[width=0.15\textwidth]{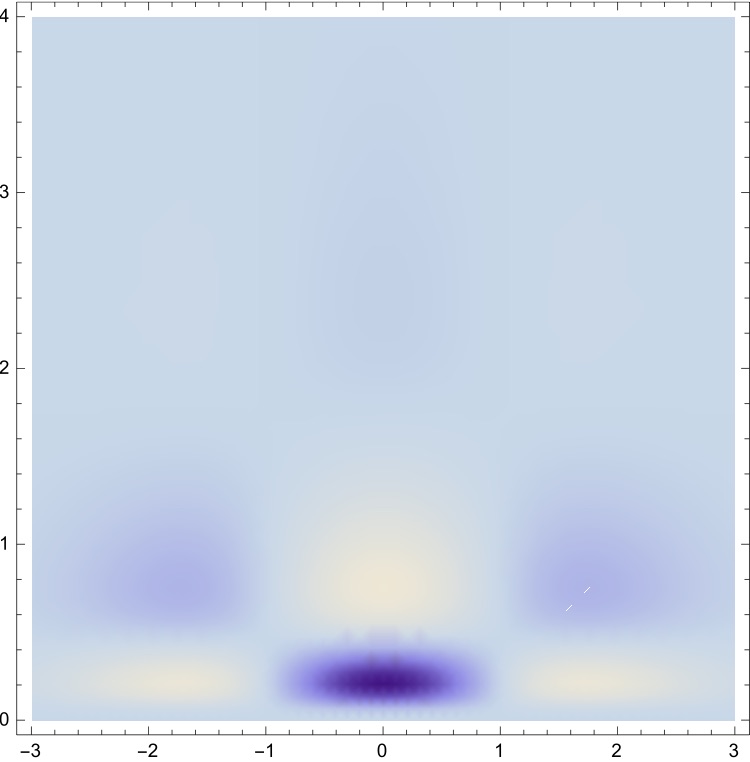} \hspace{-2mm} \\
    \end{tabular} 
  \end{center}
  \caption{{\em Space-time separable kernels\/}
           $T_{x^{m}t^{n}}(x, t;\; s, \tau) = \partial_{x^m t^n} (g(x;\; s) \, h(t;\; \tau))$ 
           for all combinations of spatial and temporal
           derivatives up to order two as obtained from the composition of Gaussian
           kernels over the spatial domain $x$ and the time-causal
           limit kernel over the temporal domain ($s = 1$, $\tau = 1$,
           $c = 2$).
           (Horizontal axis: space $x \in [-3, 3]$. Vertical axis:
           time $t \in [0, 4]$.)}
  \label{fig:spat-temp-rec-fields}
\end{figure}

\section{Spatio-temporal receptive field model}
\label{sec:spat-temp-RF}
We here give a brief description of the time-causal spatio-temporal receptive field framework of Lindeberg \cite{Lin16-JMIV}. This framework provides the time causal primitives for describing local space-time structure -- the \emph{spatio-temporal receptive fields} (or equivalently 2D+T scale-space filters) -- which our proposed family of video descriptors is based on. 

In this framework, the axiomatically derived scale-space kernel at spatial scale $s$ and temporal scale $\tau$ is of the form
\begin{equation}
       T(x, y, t;\; s, \tau, u, v, \Sigma)  
       = g(x - u t, y - v t;\; s, \Sigma) \, h(t;\; \tau)
      \label{eq:spat-temp-scale-space}
\end{equation}
where 
\begin{itemize}
\item $(x, y)$ denotes the image coordinates,
\item  $t$ denotes time,
\item $h(t;\; \tau)$ denotes a temporal smoothing kernel,
\item $ g(x - u t, y - vt;\; s, \Sigma)$ denotes a spatial affine Gaussian
kernel with spatial covariance matrix $\Sigma$ that moves with image velocity $(u, v)$.
\end{itemize}
Here, we restrict ourselves to rotationally symmetric Gaussian kernels over the spatial domain corresponding to $\Sigma = I$ and to smoothing kernels with image velocity $(u,v) = (0,0)$ leading to space-time separable receptive fields. 
This set of receptive fields is closed under spatial and temporal scaling transformations, but not under the group of Galilean transformations. 

A more general approach, and also more similar to biological vision, would be to include a family of velocity-adapted receptive fields over a range of image velocities $v$. Such a set of receptive fields would constitute a Galilean covariant representation enabling fully Galilean invariant recognition according to the general theory presented in \cite[Section 4.1.3-4.1.4]{Lin10-JMIV} and applied to motion recognition in \cite{LapLin04-IVC}. 
Exploring these possibilities is, however, left for future work. In this study, we choose to evaluate how far we can get using space-time separable receptive fields only.

Notably, also the family of space-time separable receptive fields can represent image velocities, since a set of spatial and temporal derivatives of different orderns with respect to space and time can implicitly encode optic flow. 
An interesting biological parallel is that the superior colliculus is able to perform basic visual tasks, although receiving its primary inputs from the lateral geniculate nucleus (LGN), where a majority of the receptive fields are space-time separable.

The temporal smoothing kernel $h(t;\; \tau)$ used here, is the time-causal kernel composed of truncated exponential functions coupled in
cascade: 
\begin{equation}	
   \label{eq:temp-kernel}
   h_{exp} (t; \mu_k)  = \begin{cases} 
   					\frac{1}{\mu_k} e^{-t/ \mu_k} \text{~~~} t \geq 0 \\ 
					0 \text{~~~~~~~~~~~~~} t < 0 \end
				{cases}
\end{equation}
\begin{equation}
    \label{eq:comp-temp-kernel}
   h(t; \tau) = \ast_{k=1}^{K} h_{exp}(-t; \mu_k)
\end{equation}
where the symbol $\ast_{k=1}^{K}$ represents convolution of a set of K kernels, in our case $K$ truncated exponentials with individual time constants $\mu_k$. 
The temporal variance of the composed time-causal kernel (\ref{eq:comp-temp-kernel}) will depend on the time constants $\mu_k$ as $\tau_K = \sum_{k=1}^{K} \mu_k^2$. Specifically, we choose the time constants $\mu_k$ is such a way that the temporal variance $\tau_k$ of the intermediate kernels obeys a logarithmic distribution, 
$\tau_k = c^{k-K}\tau_{K}$ for some $c > 1$. Here we choose $c = 2$ which gives a reasonable trade-off between the sampling density over temporal scales and 
the temporal delays of the time-causal smoothing kernels \cite{Lin16-JMIV}. 

In the limit of using an infinite number of temporal scale levels, the composed time-causal kernel (\ref{eq:comp-temp-kernel}) together with a logarithmic distribution of the intermediate scale levels leads to a \emph{scale-invariant limit kernel}, possessing true scale covariance and having a Fourier transform of the form (\cite[Eq. (38)]{Lin16-JMIV})
\begin{equation}
   \label{eq:FT-temp-kernel}
   \hat{\Psi}(\omega;\; \tau, c) 
   = \prod_{k=1}^{\infty} \frac{1}{1 + i \, c^{-k} \sqrt{c^2-1} \sqrt{\tau} \, \omega}
\end{equation}
For practical implementation, a finite number of intermediate scale levels need to be used. However, the composed kernel (\ref{eq:comp-temp-kernel}) converges rapidly to the limit kernel (\ref{eq:FT-temp-kernel}) with increasing $K$ (\cite[Table 5]{Lin16-JMIV}). The temporal smoothing kernels used in this work with temporal scale $\sigma_\tau = \sqrt{\tau}$ are such composed convolution kernels, where we use $K \geq 7$, which leads to very small deviations from the limit kernel.  

The \emph{spatio-temporal receptive fields} are in turn defined as partial 
derivatives of the spatio-temporal scale-space kernel $T$ for different orders $(m_1, m_2, n)$ of spatial and temporal differentiation computed at multiple spatio-temporal scales. The result of convolving a video $f(x,y,t)$ with one of these kernels
\begin{multline}
   \label{eq:spattempder}
   L_{x^{m_1}y^{m_2}t^n } (\cdot, \cdot, \cdot; s, \tau)   
   = \partial_{x^{m_1}y^{m_2}t^n} (T(\cdot, \cdot, \cdot;\; s, \tau)  \ast f(\cdot,\cdot, \cdot))
\end{multline}
is referred to as the \emph{receptive field response}. A set of receptive field responses will comprise a \emph{spatio-temporal $N$-jet} representation of the local space-time structure, in essence corresponding to a truncated Taylor expansion possibly at multiple spatial and temporal scales. Using this representation thus enables capturing more diverse information about the local space-time structure than what can be done using a single filter type.

One set of receptive fields considered in this work consists of combinations of the following sets of partial derivatives: (i) the first- and second-order spatial derivatives, (ii) the first- and second-order temporal derivatives of these and (iii) the first- and second-order temporal derivatives of the original smoothed video $L$:
\begin{multline}
   \label{eq:spattemp-njet}
  \{ \{L_x , L_y , L_{xx} , L_{xy}, L_{yy}\},   
    \{L_{xt}, L_{yt}, L_{xxt}, L_{xyt}, L_{yyt}\}, \\  \{L_{xtt}, L_{ytt} , L_{xxtt}, L_{xytt}, L_{yytt}\},   \{L_t, L_{tt}\} \}.
\end{multline}
A second set consists of spatio-temporal invariants defined from combinations of these partial derivatives. A subset of the receptive fields/scale-space derivative kernels are shown in Figure~\ref{fig:spat-temp-rec-fields}. (The kernels are three-dimensional over $(x,y,t)$, but are here illustrated using a single spatial dimension $x$). 

All video descriptors are expressed in terms of scale-normalised derivatives \cite{Lin16-JMIV}: 
\begin{equation}
  \partial_{\xi} = s^{\gamma_s/2} \partial_x, \quad
  \partial_{\eta} = s^{\gamma_s/2} \partial_y, \quad
   \partial_{\zeta} = \tau^{\gamma_{\tau}/2} \partial_t
\end{equation}
with the corresponding scale-normalised receptive fields computed as:
\begin{equation}
  L_{\xi^{m_1} \eta^{m_2} \zeta^n} 
  = s^{(m_1 + m_2) \gamma_s/2} \, \tau^{n \gamma_{\tau}/2} \, L_{x^{m_1} y^{m_2} t^n} 
  \end{equation}
where $\gamma_s > 0$ and $\gamma_{\tau} > 0$ are the scale normalization
powers of the spatial and the temporal domains, respectively. We here use $\gamma_s  = 1$ and $ \gamma_{\tau} = 1$ corresponding to the maximally scale-invariant case (as described in Appendix \ref{app:cov-properties}). 

It should be noted that the scale-space representation and receptive field responses
are computed ``on the fly" using only a compact temporal buffer. The time-recursive formulation for the temporal smoothing \cite[Eq (56)]{Lin16-JMIV} means that computing the scale-space representation for a new frame at temporal scale $\tau_{k}$ only requires information from the present moment $t$ and the scale-space representation for the preceding frame $t-1$ at temporal scales $\tau_{k}$ and  $\tau_{k -1}$ 
\begin{multline}
	L(\cdot,\cdot,t; \cdot, {\tau_k}) = L(\cdot,\cdot,t-1; \cdot, {\tau_k}) + \\
	\frac{1} {\mu_k} ( L(\cdot,\cdot,t; \cdot, \tau_{k -1}) - L(\cdot,\cdot,t-1; \cdot, \tau_{k -1})   )
\end{multline}
where $t$ represents time and $\tau_{k} > \tau_{k -1}$ are two adjacent temporal scale levels, where $k = 0$ corresponds to the original signal (the spatial coordinates and the spatial scale are here suppressed for brevity). 
This is a clear advantage compared to using a Gaussian kernel (possibly delayed and truncated) over the temporal domain, since it implies less computations and smaller temporal buffers.  

These spatio-temporal receptive field responses can be computed efficiently by space-time separable filtering. The spatial smoothing is done in by separable discrete Gaussian filtering over the spatial dimensions, with the spatial extent proportional to the spatial scale parameter in units of the standard deviation $\sigma_s = \sqrt{s}$. The temporal smoothing is performed by recursive filtering requiring only two additions and one multiplication per pixel and spatio-temporal scale level. Then, scale-space derivatives are computed by applying small support discrete derivative approximations to the spatio-temporal scale-space representation. 

For further details concerning the spatio-temporal scale-space representation and its discrete implementation, we refer to Lindeberg \cite{Lin16-JMIV}.

\begin{figure}

\centerline{ \includegraphics[width=0.39\textwidth]{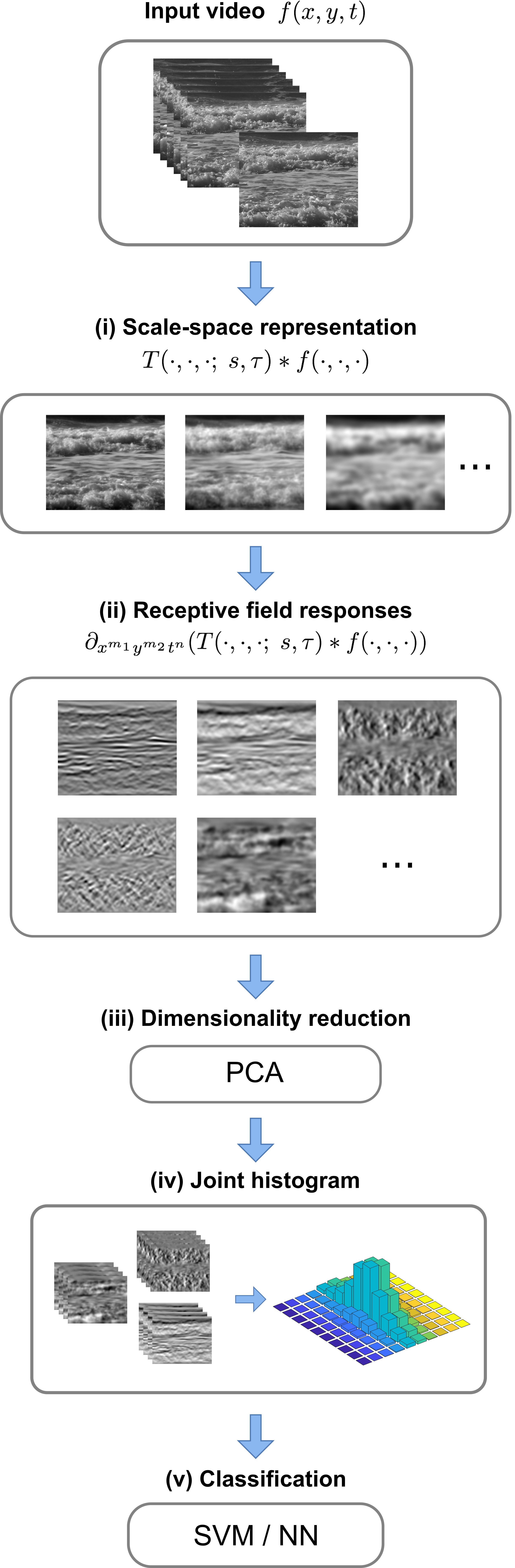} }
 \caption{Overview of the main steps in our dynamic texture recognition workflow: (i) Spatio-temporal scale-space representation of the current frame. (ii) Computation of local spatio-temporal receptive field responses. (iii) Dimensionality reduction of local receptive field responses using principal component analysis (PCA). (iv) Aggregation of joint statistics of receptive field responses over a space-time region into a joint multidimensional histogram (here illustrated in two dimensions). (v) Classification using an SVM or NN classifier.}
 \label{fig:workflow}
\end{figure}

\section{Video descriptors }
\label{sec:videodesc}

We here describe our proposed family of video descriptors. The histogram descriptor is based on regional statistics of time-causal spatio-temporal receptive field responses and the process of computing the video descriptor can be divided into four main steps:
\begin{enumerate}[(i)]
\item Computation of the spatio-temporal scale-space representation at the specified spatial and temporal scales (spatial and temporal smoothing). 
\item Computation of local spatio-temporal receptive field responses using discrete derivative approximation filters over space and time.
\item Dimensionality reduction of the local spatio-temporal receptive field responses/feature vectors using principal component analysis (PCA). 
\item Aggregation of joint statistics of receptive field responses over a space-time region into a multidimensional histogram.
\end{enumerate}
The resulting histogram video descriptor will describe a space-time region by the relative frequency of different local space-time patterns. 
Note that all computations are performed ``on the fly", one frame at a time. In the following, each of the above steps is described in more detail. A schematic illustration of our dynamic texture recognition workflow is given in Figure~\ref{fig:workflow}. 

\subsection{Spatio-temporal scale-space representation}
\label{sec:videodesc-scalespace}
The first processing step is spatial and temporal smoothing to compute the time-causal spatio-temporal
scale-space representation of the current frame at the chosen spatial and temporal scales. The spatial smoothing is done in a cascade over spatial scales and the temporal smoothing is performed recursively over time and in a cascade over temporal scales according to \cite{Lin16-JMIV}. The fully separable convolution kernels 
and the time-recursive implementation of the temporal smoothing imply that the smoothing can be done
in a computationally efficient way.

\subsection{Spatio-temporal receptive field responses}
\label{sec:videodesc-rfresponse}
After the smoothing step, the spatio-temporal receptive field responses $F = [F_1,F_2,... F_N]$ are computed \emph{densely} over the current frame for a set of $N$ chosen scale-space derivative filters. These filters will include partial derivatives over a set of different spatial and temporal differentiation orders at single or multiple
spatial and temporal scales. Alternatively, for a video descriptor based on differential invariants, the features will correspond to such differential invariants computed from the partial derivatives. Using filters of multiple scales enables capturing image structures of different spatial extent and temporal duration. All derivatives are scale-normalised. 

Previous methods utilising ``ideal" (as opposed to learned) spatio-temporal filters have used families of filters created from a single filter type e.g. by applying third-order filters in different directions in space-time (Derpanis and Wildes \cite{DerWil-TPAMI2012}) or using a Gabor filter extended by both spatial rotations and velocity transformations (Gon{\c{c}}alves et~al. \cite{GonNunetal-ArXiv2012}). 

We here take a different approach, using a richer family of receptive fields encompassing different orders of spatial and temporal differentiation, specifically including mixed spatio-temporal derivatives. Using such a set, enables representing local intensity variations of \emph{different orders}. This is not possible if restricting filters to a single base filter, even if the filters would be extended by spatio-temporal transformations. The spatio-temporal receptive field responses from such a family can thus be used to separate patterns based on a combination of e.g. first-order and second-order intensity variations. It will also enable  computing measures that place equal weight on the first- and second-order changes.

Biological vision does also comprise such richer sets of spatio-temporal receptive fields, which typically occur in pairs of odd-shaped and even-shaped receptive fields \cite{ValCotMahElfWil00-VR} and which can be well modelled by spatio-temporal scale-space derivatives for different orders of spatial and temporal differentiation \cite{Lin16-JMIV}. As previously discussed in Section~\ref{sec:spat-temp-RF}, a more general model would be to combine these spatio-temporal receptive fields with velocity adaptation over a set of Galilean transformations. This extension is, however, left for future work. 

A more detailed discussion concerning the choice of the specific receptive field sets that our video descriptors are based on is given in Section~\ref{sec:rfsets}.  

\subsection{Dimensionality reduction with PCA}
\label{sec:videodesc-pca}
When combining a large number of local image
measurements, most of the variability in a
dataset will be contained in a lower-dimensional subspace of the
original feature space. This opens up for dimensionality reduction of 
the local feature vectors to enable
summarising a larger set of receptive field responses
without resulting in a prohibitively large joint histogram descriptor.

For this purpose, a dimensionality reduction step in the form of principal component analysis is added before constructing
the joint histogram. The result will be a local feature vector $\tilde{F}(x,y,t) = [\tilde{F}_1, \tilde{F}_2,
... \tilde{F}_M] \in \mathbb{R}^M$ $M \leq N$, where the new components correspond to 
linear combinations of the filter responses from the original feature vector $F$.

The transformation properties of the spatio-temporal scale-space derivatives imply 
scale covariance for such linear combinations of receptive fields, as well as for the individual receptive field responses \emph{if using the proper scale-normalisation} with $\gamma_s  = 1$ and $ \gamma_{\tau} = 1$.
A more detailed treatment of the scale-covariance properties is given in Appendix \ref{app:cov-properties}.
The main reason for choosing PCA is simplicity and that it has empirically been shown to give good results for spatial images (Linde and Lindeberg \cite{LinLin12-CVIU}). The number of principal components $M$ can be adapted to requirements for descriptor size and need for
detail in modelling the local image structure. The dimensionality reduction step can also be skipped if working with a smaller number of receptive fields. 

\subsection{Joint receptive field histograms}
\label{sec:videodesc-hist}

The approach that we will follow to represent videos of dynamic textures is to use histograms of spatio-temporal receptive field responses. Notably, such a histogram representation will discard information about the spatial positions and the temporal moments that the feature responses originate from. Because the histograms are computed from spatio-temporal receptive field responses in terms of spatio-temporal derivatives, these feature responses will, however, implicitly code for partial spatial and temporal information, like pieces in a spatio-temporal jigsaw puzzle. By computing these receptive field responses over multiple spatial and temporal scales, we additionally capture primitives of different spatial size and temporal duration. For spatial recognition related histogram approaches have turned out to be highly successful, leading to approaches such as SIFT \cite{Low04-IJCV}, HOG \cite{DalTri05-CVPR} and HOF \cite{DalTriSch06-ECCV} and bag-of-words models. For the task of texture recognition, the loss of information caused by discarding spatial positions is also less critical, since many textures can be expected to possess certain stationarity properties. In this work, we use an extension of this paradigm of histogram-based image descriptors for building a conceptually simple system for video analysis. 

The multi-dimensional histogram of receptive field responses will represent a discretised version of the joint distribution of local receptive field responses over a space-time region. To construct the histogram, each feature
dimension is partitioned into $n_{bins}$ number of equidistant bins in the range
\begin{equation*}
[\operatorname{mean}(\tilde{F}_i) - d\, \operatorname{std}(\tilde{F}_i), ~\operatorname{mean}(\tilde{F}_i) + d \operatorname{std}(\tilde{F}_i)]
\end{equation*}
where the mean and the standard deviation are computed over the training set and $d$ is a parameter controlling the number of standard deviations that are spanned for each dimension. We here use $d=5$. 
If including $M$ principal components, the result is a multidimensional histogram with \\ $n_{cells} = (n_{bins})^M$ distinct histogram cells. Note the use of the parameter $n_{bins}$ (histogram bins) to refer to the number of partitions along each feature dimension and the parameter $n_{cells}$ (histogram cells) to refer to the number of cells in the joint histogram. 

The receptive field responses up to a certain order will represent information corresponding to the coefficients of a Taylor expansion around each point in space-time. Each histogram cell will correspond to a 
certain ``template'' local space-time structure, encoding joint conditions on the magnitudes of the spatio-temporal receptive field responses (here, image derivatives, differential invariants or PCA components).
This is somewhat similar to e.g. the templates used in VLBP \cite{ZhaGuoPie-TPAMI-2007} but notably represented and computed using different primitives. 
The normalised histogram video descriptor then captures \emph{the relative frequency} of
such space-time structures in a space-time region. The number of
different local ``templates'' will be decided by the number of receptive
fields/principal components and the number of bins. 

A \emph{joint} histogram video descriptor explicitly models the \emph{co-variation} of different types of image measurements, in contrast to the more common choice of descriptors based on marginal distributions or relative feature strength (see e.g. \cite{DerWil-TPAMI2012,GonNunetal-ArXiv2012}).
A simple example of this is that a joint histogram over $L_x$ and $L_y$ will reflect the orientations of gradients over image space, which would not be sufficiently captured by the corresponding marginal histograms. Using joint histograms similarly implies the ability to represent other types of patterns that correspond to certain \emph{relationships} between groups of features, such as how receptive field responses co-vary over different spatial and temporal scales.

In the general case, these histograms should be computed regionally, over different regions over space and/or time, e.g., to separate regions that contain different types of dynamic textures. 
For almost all experiments in this study, however, the \emph{space-time region} for the histogram descriptor will be chosen as the entire video, leading to \emph{a single global histogram descriptor} per dynamic texture video. The single exception is the experiment presented in Section~\ref{sec:expm-frames}, where we compute histograms over a smaller number of video frames. The reason for primarily using global histograms is that the videos in the DynTex and UCLA benchmarks are pre-segmented to contain a single dynamic texture class per video. Thereby, we can make this conceptual simplification for the experimental evaluation of our different types of video descriptors.

It should be noted that, if represented naively, a subset of the histograms descriptors evaluated here would be prohibitively large. However, for such high-dimensional descriptors the number of \emph{non-zero} histogram cells can be considerably lower than the maximum number of cells. This implies they can be efficiently represented using a computationally efficient sparse representation as outlined by Linde and Lindeberg \cite{LinLin12-CVIU}. 

\subsection{Covariance and invariance properties of the video descriptor}
\label{sec:videodesc-cov}

The scale-covariance properties of the spatio-temporal receptive fields and the PCA components, according to the theory presented in Appendix \ref{app:cov-properties}, imply that a histogram descriptor constructed from these primitives will be \emph{scale covariant} over all non-zero spatial scaling factors 
and for temporal scaling factors that are integer powers of the distribution parameter $c$
of the time-causal limit kernel. This means that our proposed video descriptors can be used as the building blocks of a \emph{scale-invariant} recognition framework. 

A straightforward option for this, is to use the video descriptors in a multi-scale recognition framework, where each video is represented by \emph{a set of descriptors} computed at multiple spatial and temporal scales, both during training and testing. A scale-covariant descriptor then implies that if training is performed for a video sequence at scale $(s_1,\tau_1)$ and if a corresponding video sequence is 
rescaled by spatial and temporal scaling factors $S_s$ and $S_\tau$, corresponding recognition can be performed at scale $(s_2, \tau_2)= (S_s^2 s_1, S_\tau^2 \tau_1)$. 
However, for the initial experiments performed in this work this option for
scale invariant recognition has not been explored. Instead, training and classification are performed at the same scale or the same set of scales and the outlined scale-invariant extension is left for future work.

\subsection{Choice of receptive fields and descriptor parameters }
\label{sec:videodesc-params}

The basic video descriptor described above will give rise to a family of video descriptors 
when varying the set of receptive fields and the descriptor parameters. 
Here, we describe the different options investigated in this work considering: (i) the set of receptive fields, (ii) the number of
bins and the number of principal components used for constructing the histogram and (iii) the spatial and temporal scales of the receptive fields.


\begin{figure*}[hpbt]   
 \begingroup
    \setlength\tabcolsep{0.5mm}

    \begin{center}
             \scriptsize

    \begin{tabular}{c c c c c}

       $f$ & $L$ & $L_t$ & $L_{tt}$ &  \\
       \includegraphics[width=0.15\textwidth]{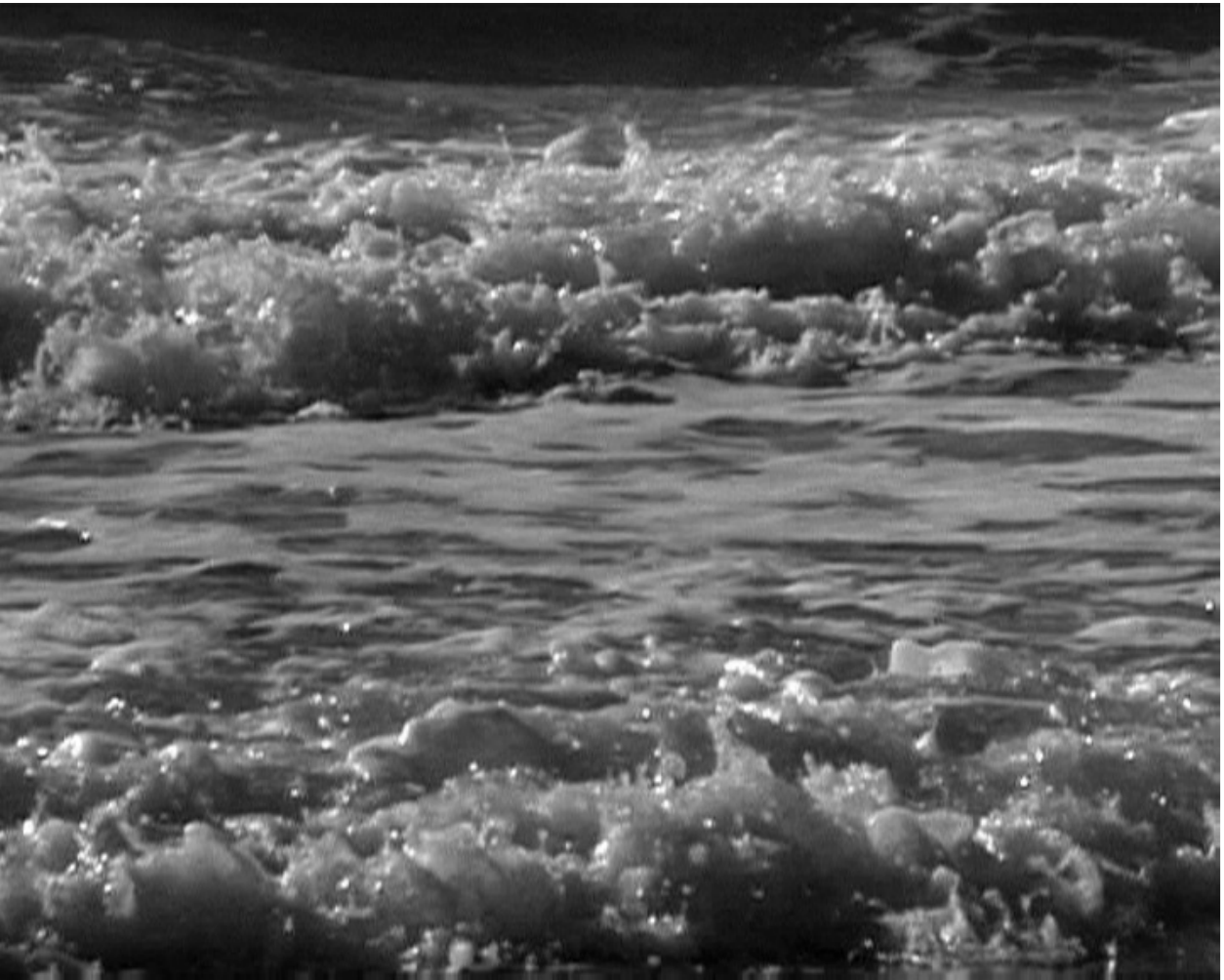}     &
       \includegraphics[width=0.15\textwidth]{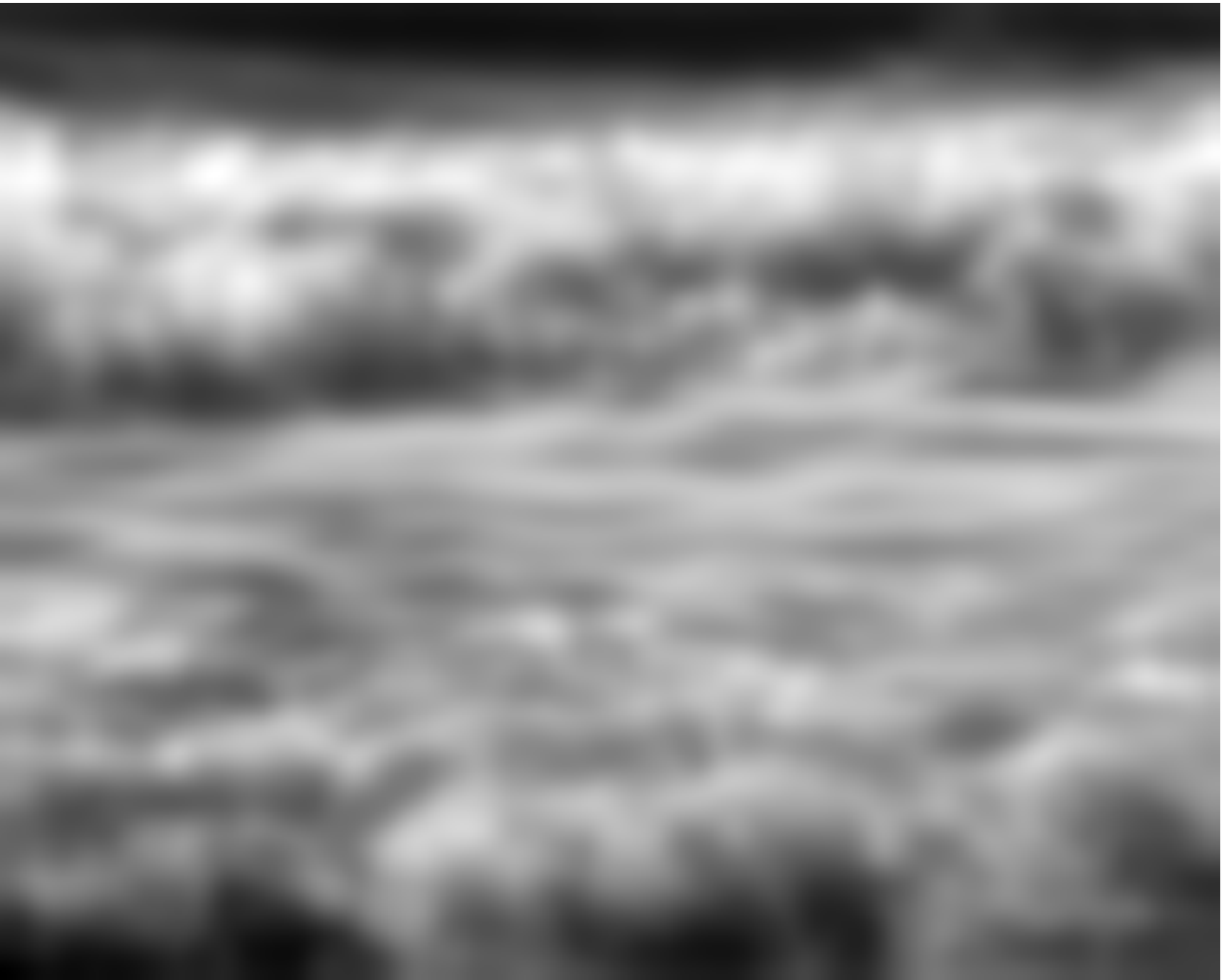}     &
       \includegraphics[width=0.15\textwidth]{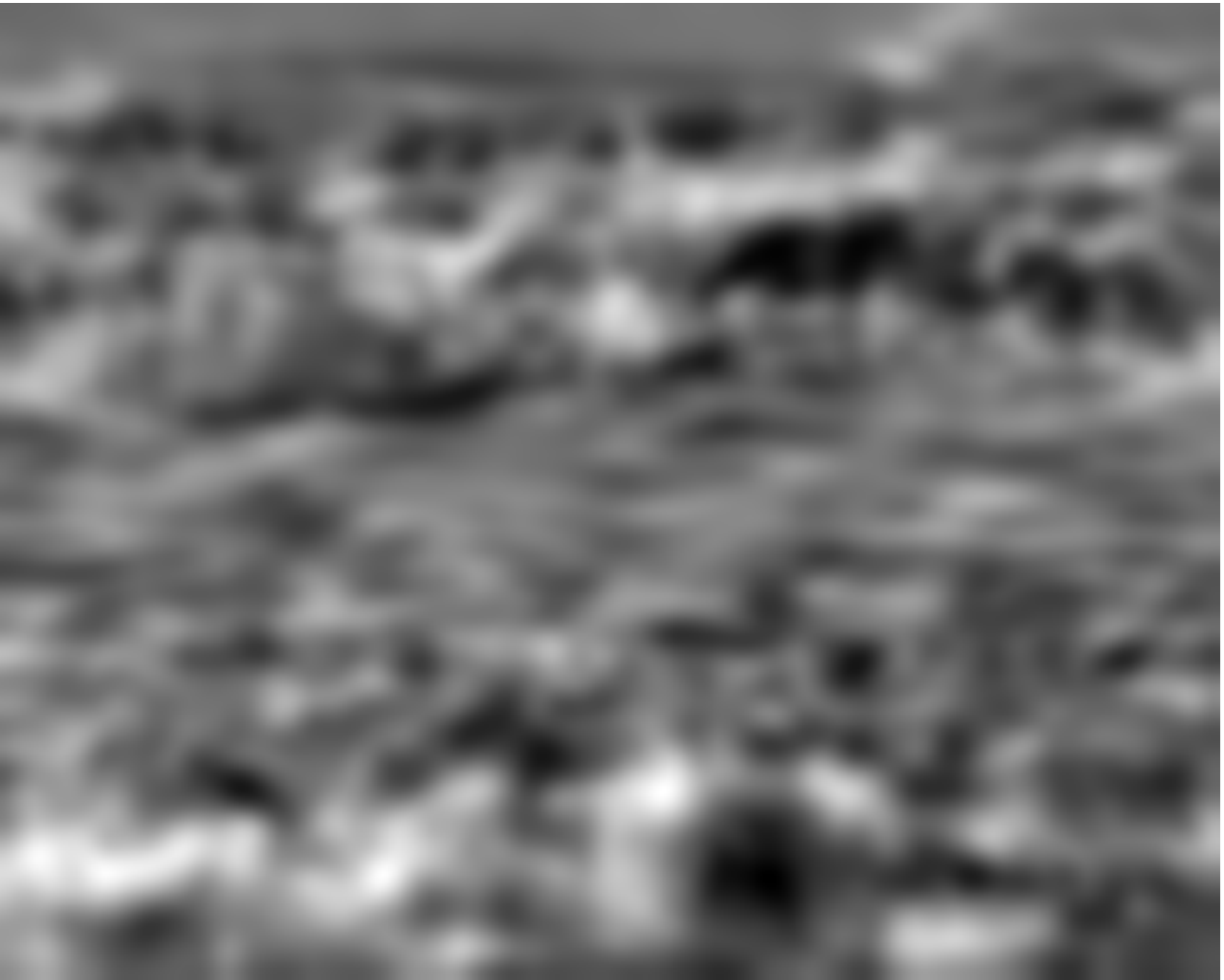}     &
       \includegraphics[width=0.15\textwidth]{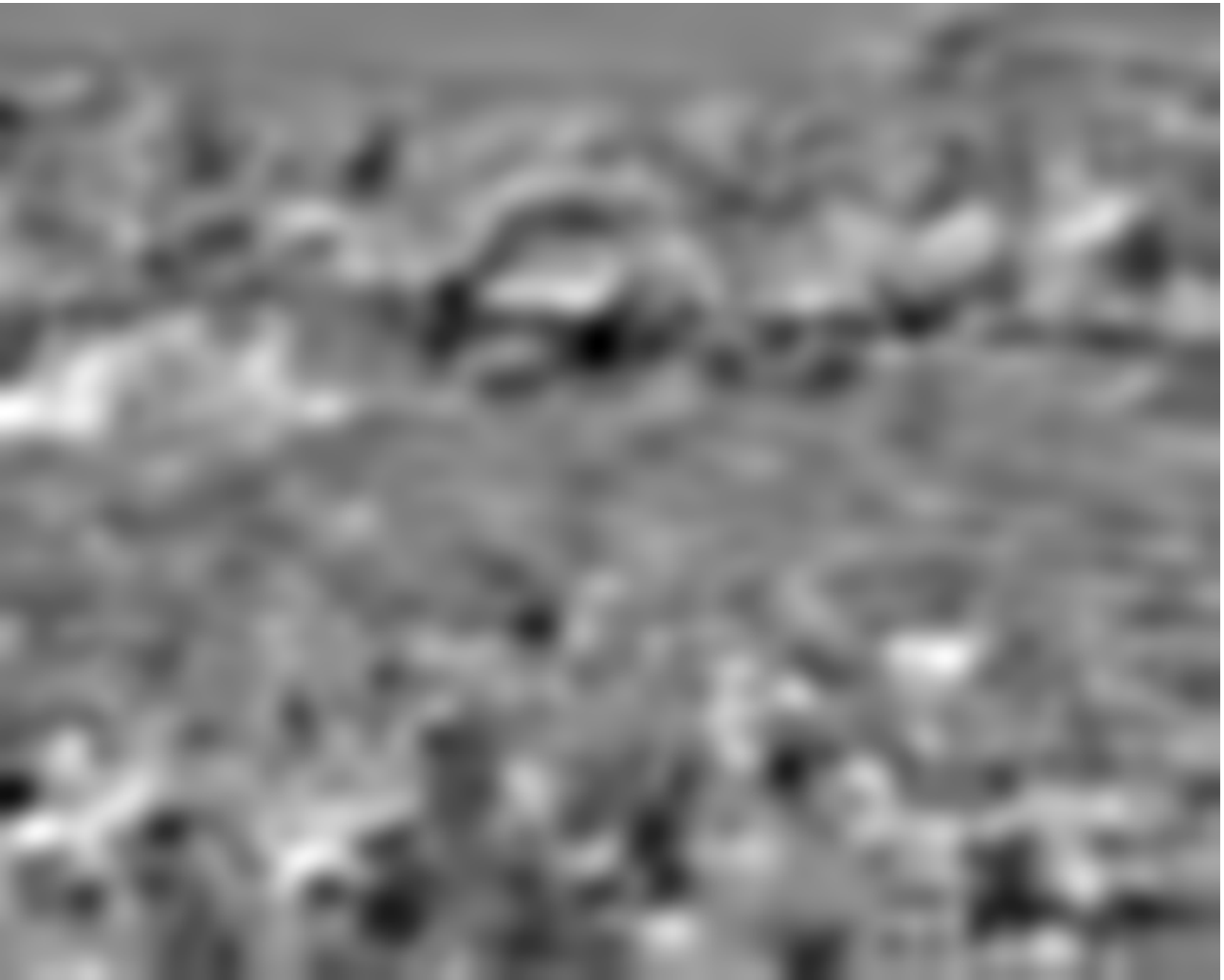}     & \\       
       $L_x$ & $L_y$ & $L_{xx}$ & $L_{xy}$ & $L_{yy}$\\
       \includegraphics[width=0.15\textwidth]{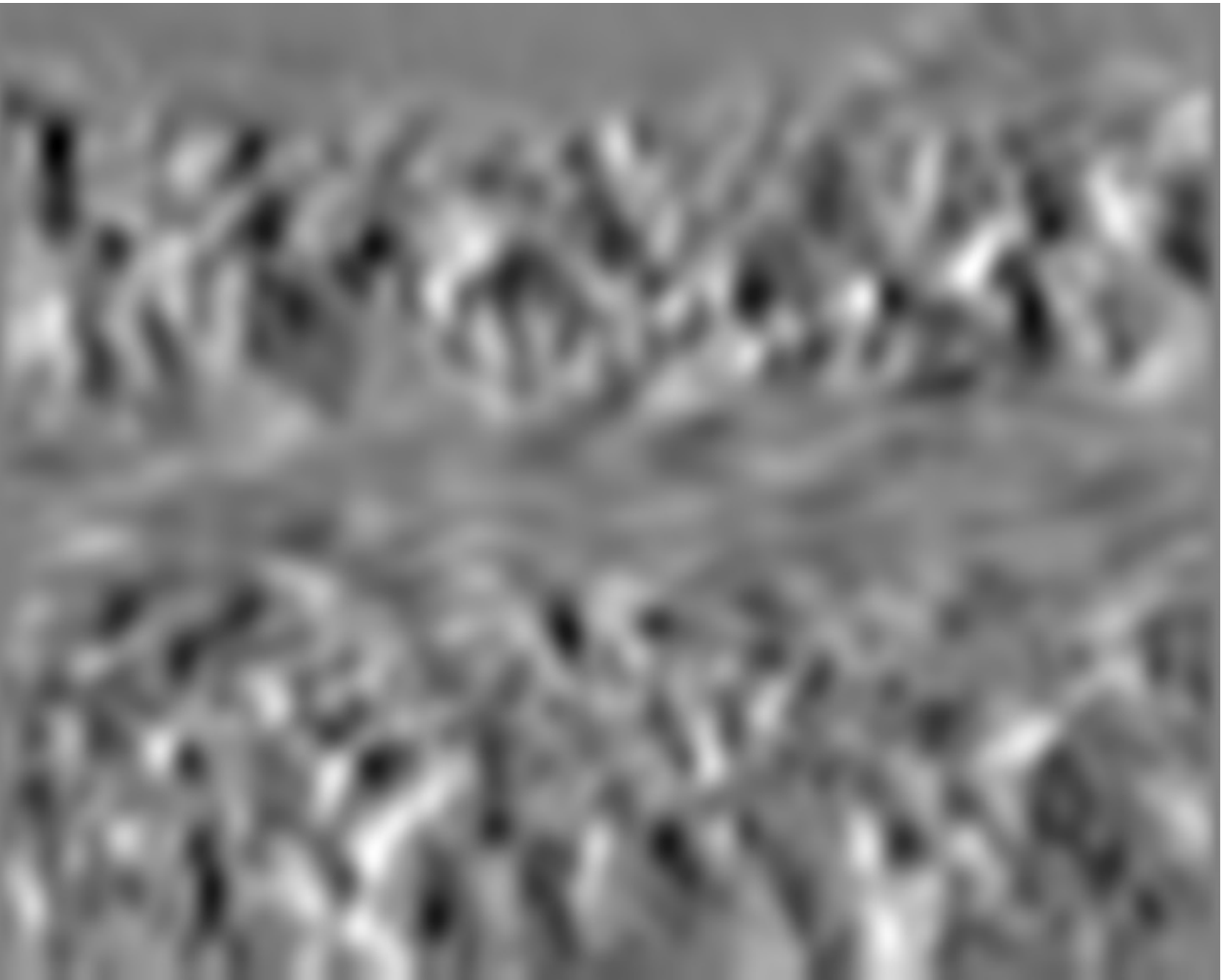}     &
       \includegraphics[width=0.15\textwidth]{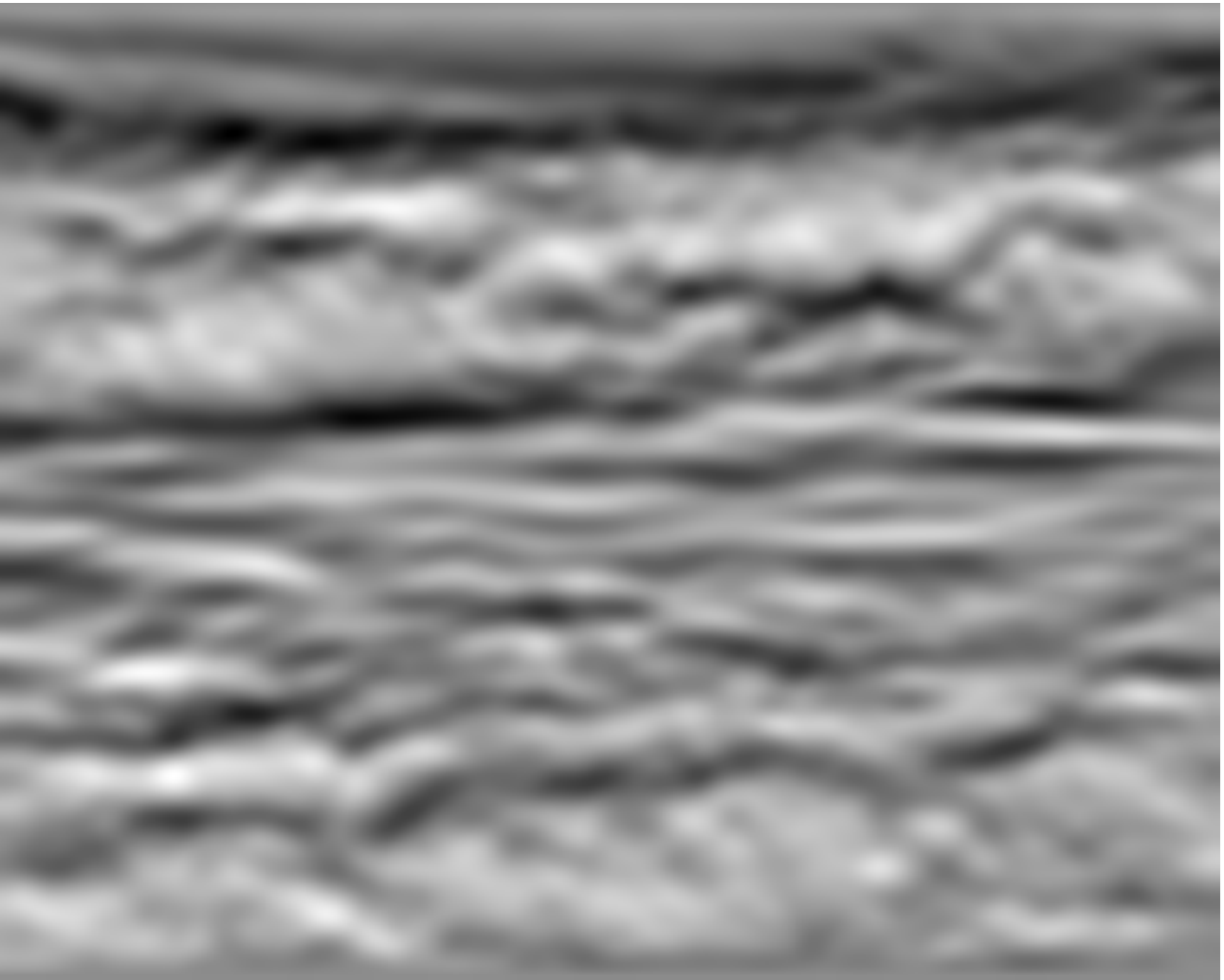}     &
       \includegraphics[width=0.15\textwidth]{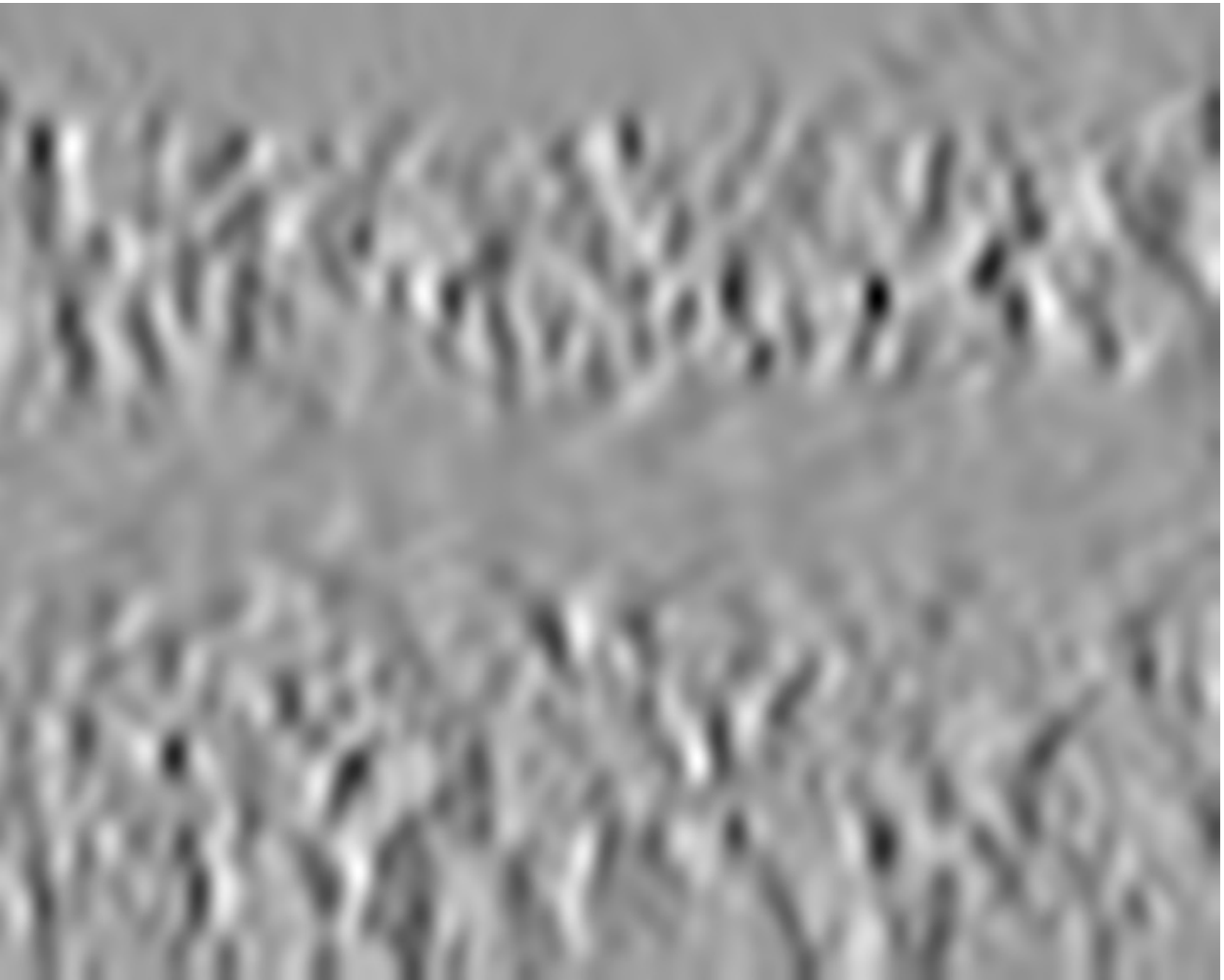}     &
       \includegraphics[width=0.15\textwidth]{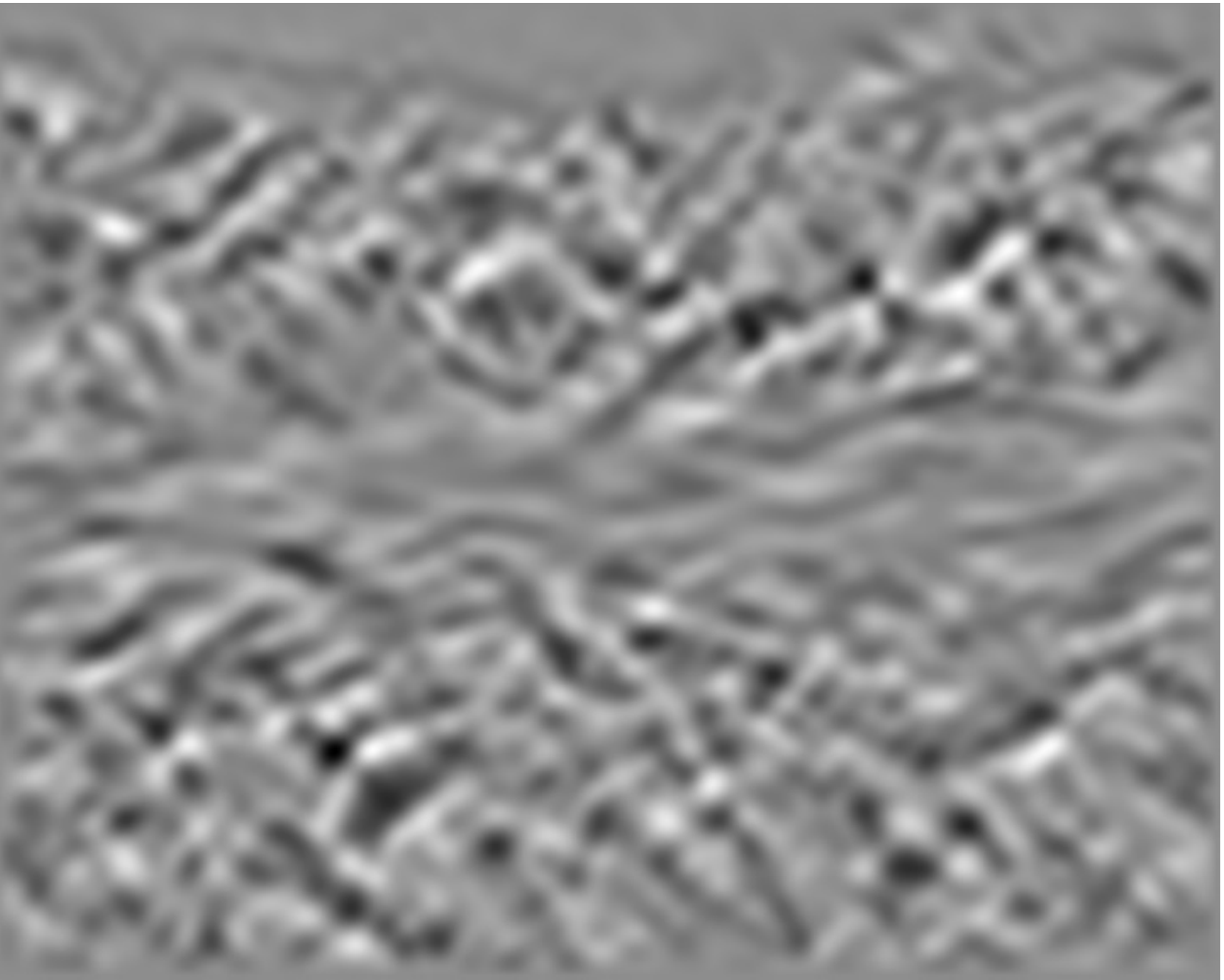}     &
              \includegraphics[width=0.15\textwidth]{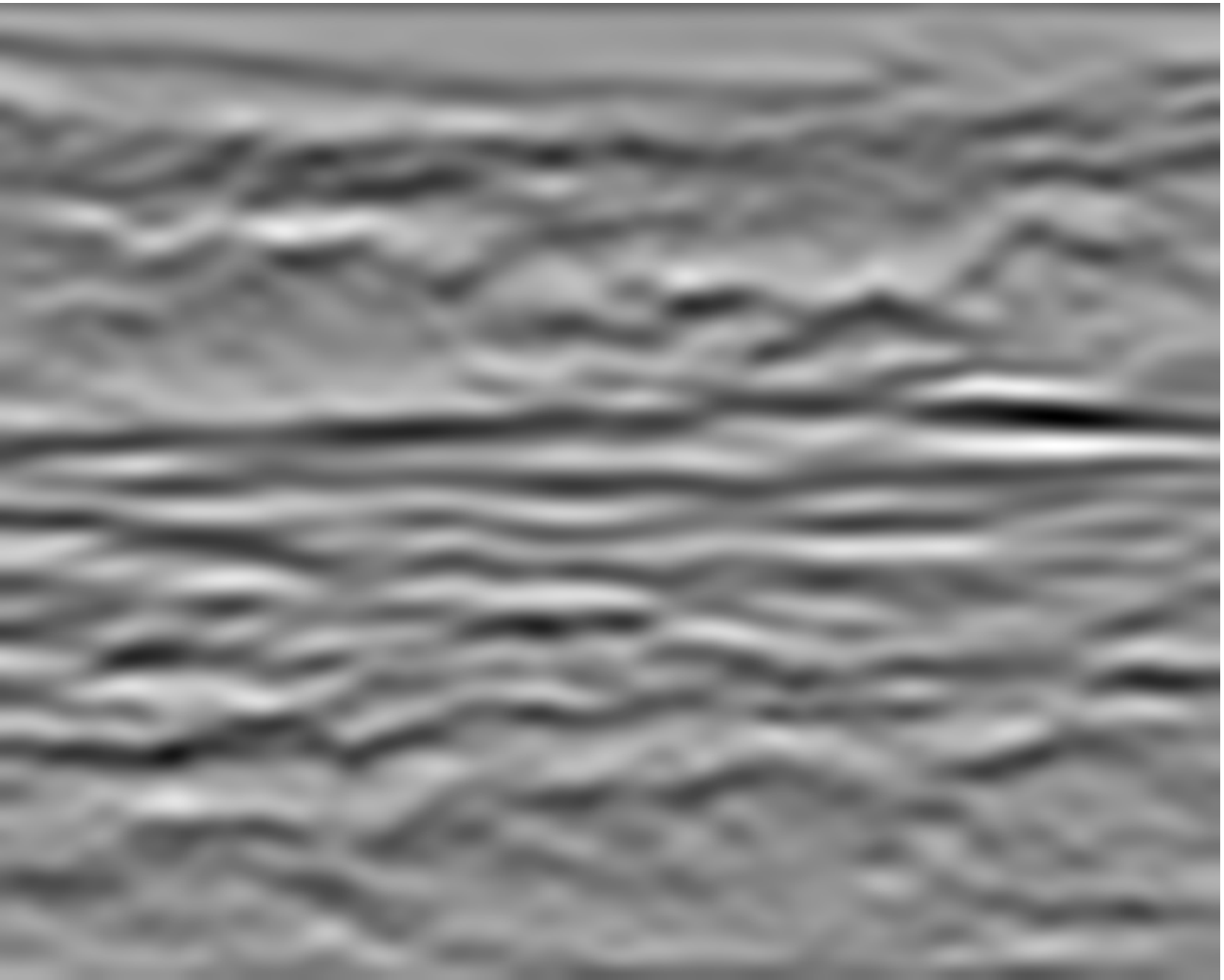}     \\

       $L_{xt}$ & $L_{yt}$ & $L_{xxt}$ & $L_{xyt}$ & $L_{yyt}$\\
       \includegraphics[width=0.15\textwidth]{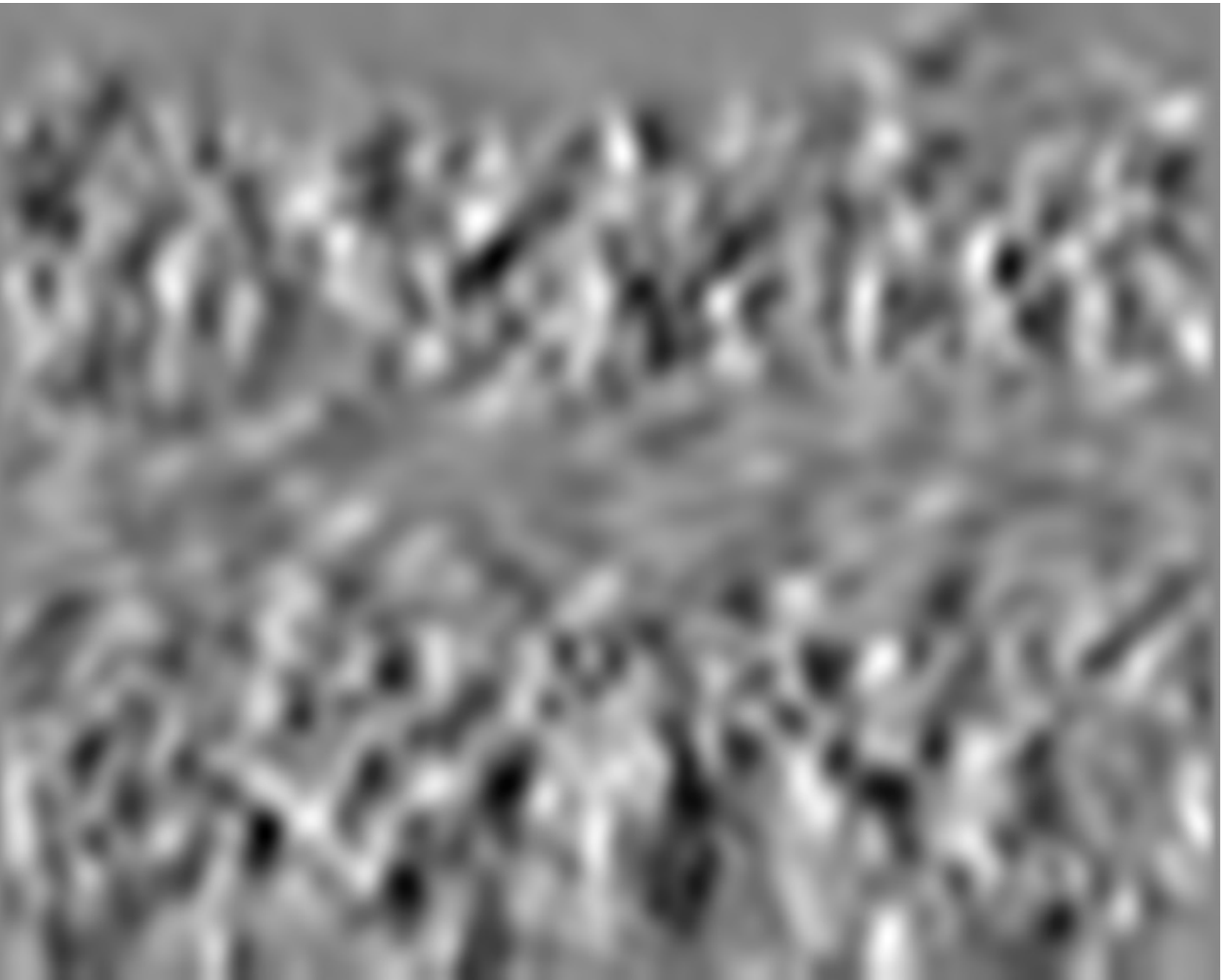}     &
       \includegraphics[width=0.15\textwidth]{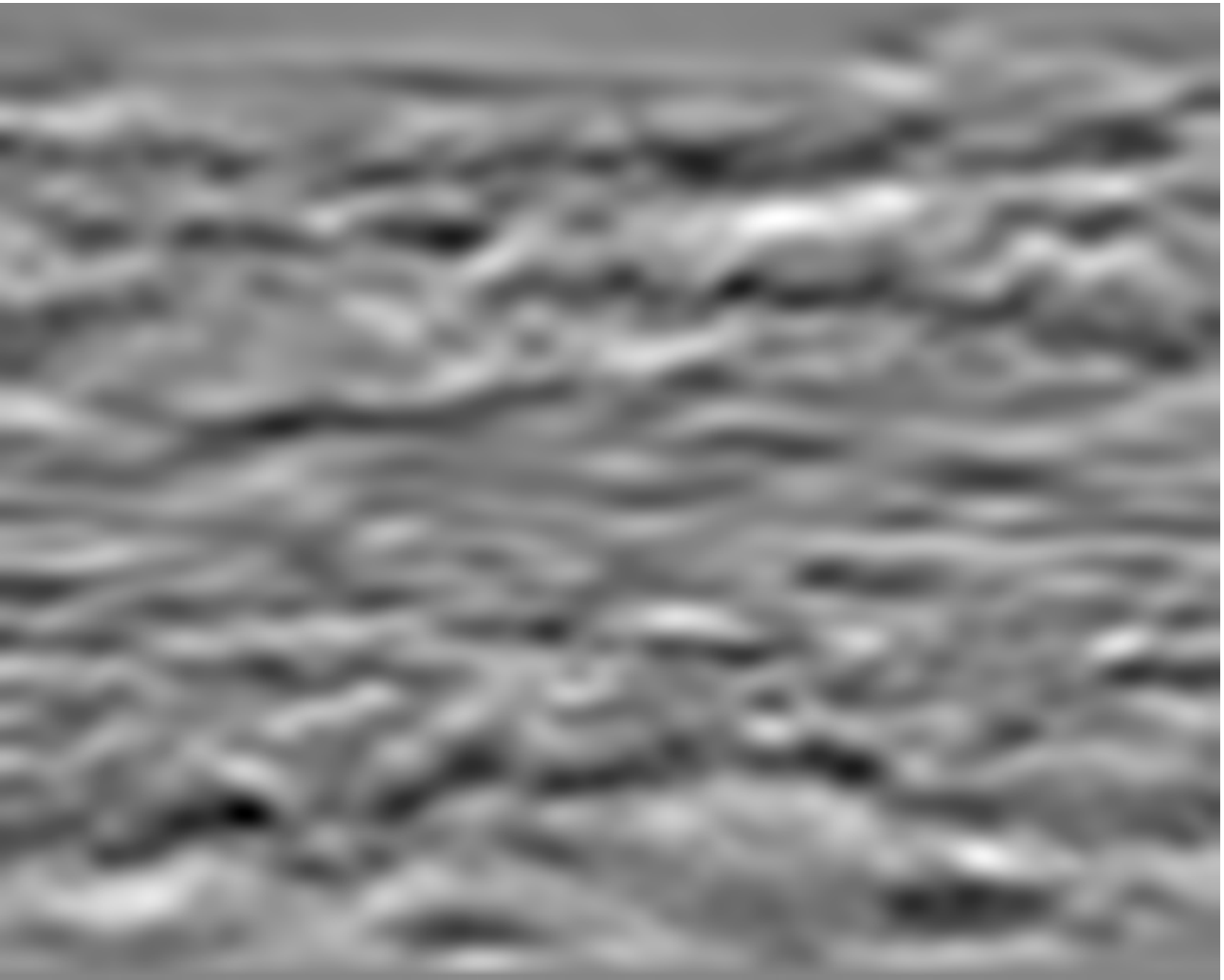}     &
       \includegraphics[width=0.15\textwidth]{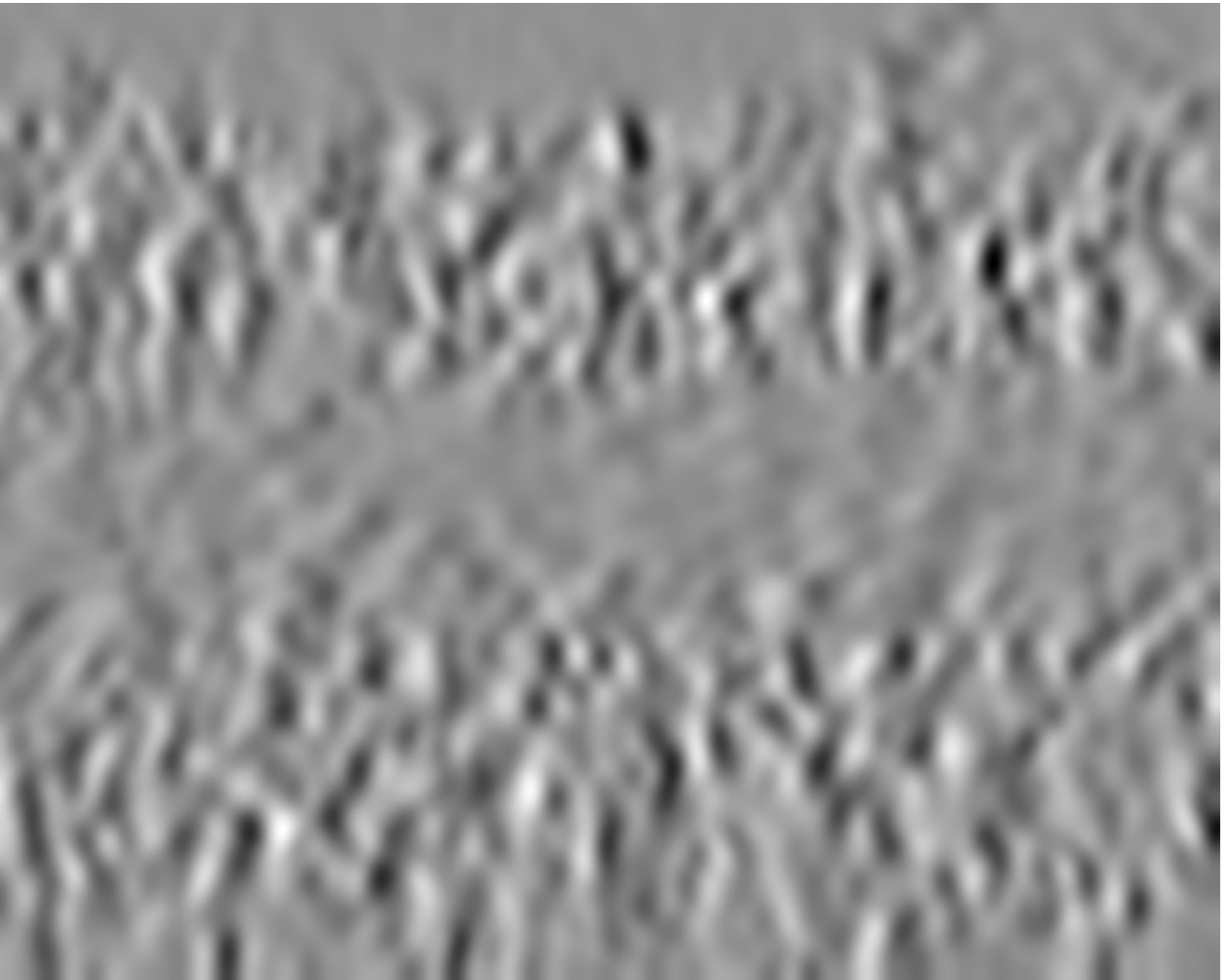}     &
       \includegraphics[width=0.15\textwidth]{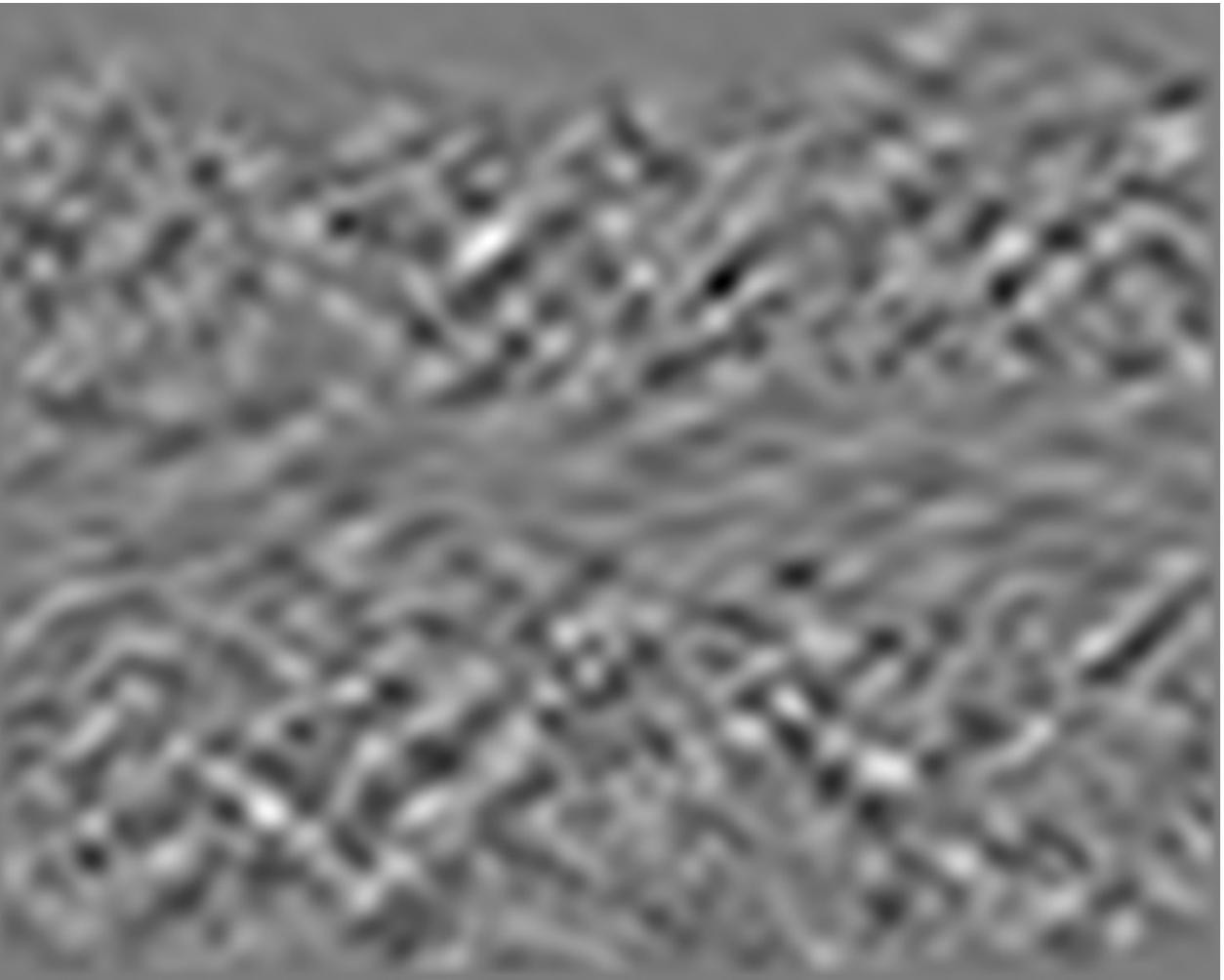}    &
              \includegraphics[width=0.15\textwidth]{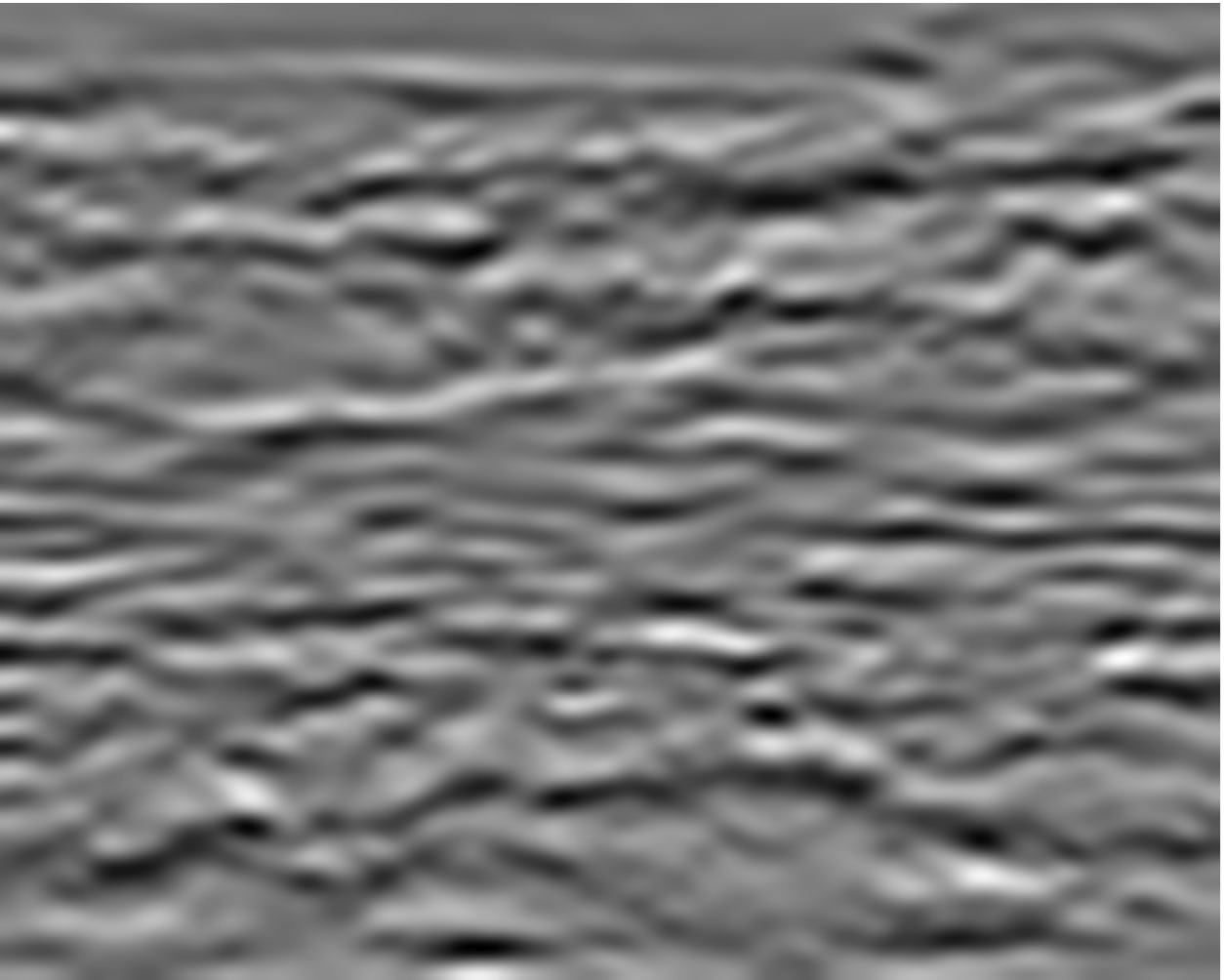}     \\

       $L_{xtt}$ & $L_{ytt}$ & $L_{xxtt}$ & $L_{xytt}$ & $L_{yytt}$\\
       \includegraphics[width=0.15\textwidth]{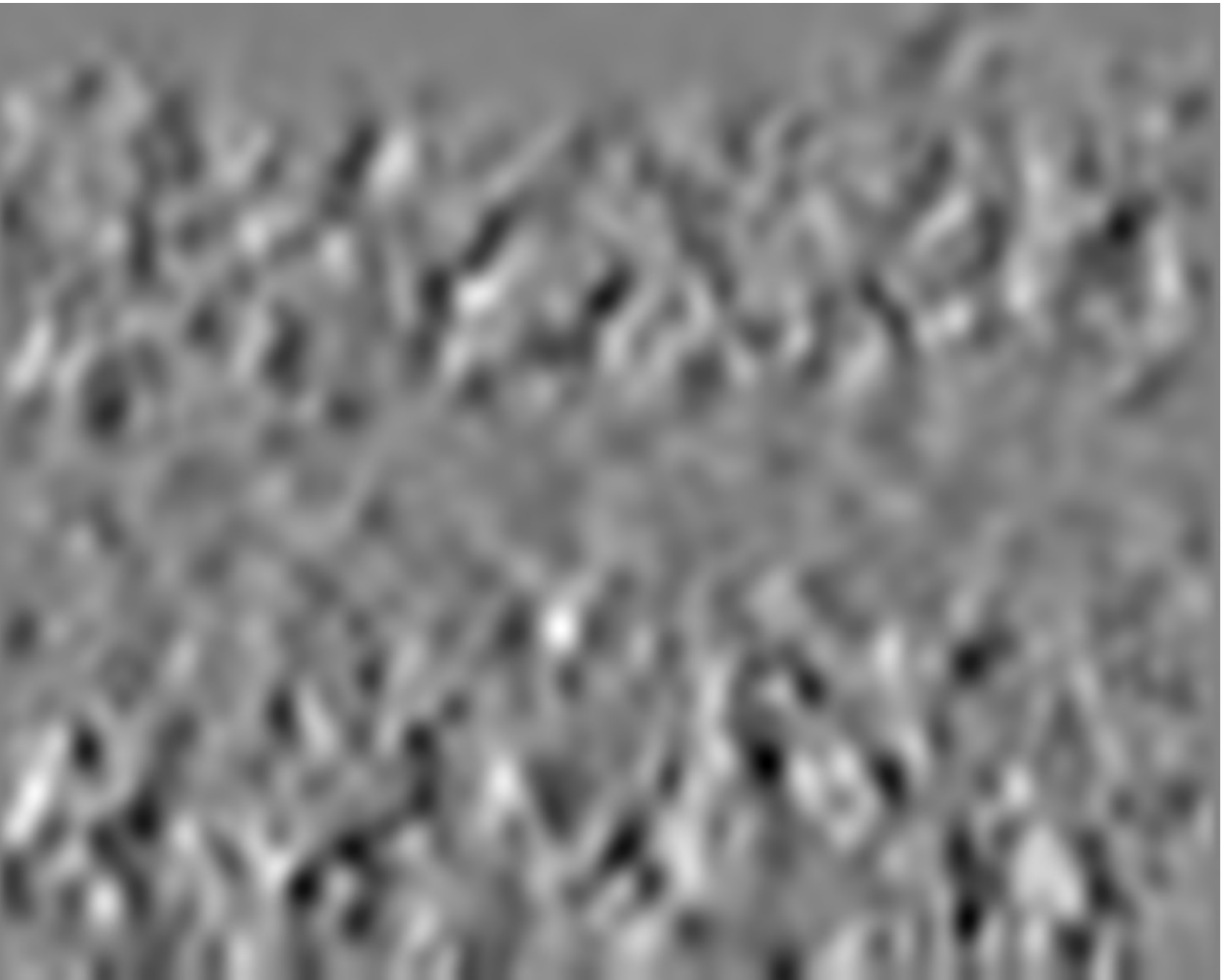}     &
       \includegraphics[width=0.15\textwidth]{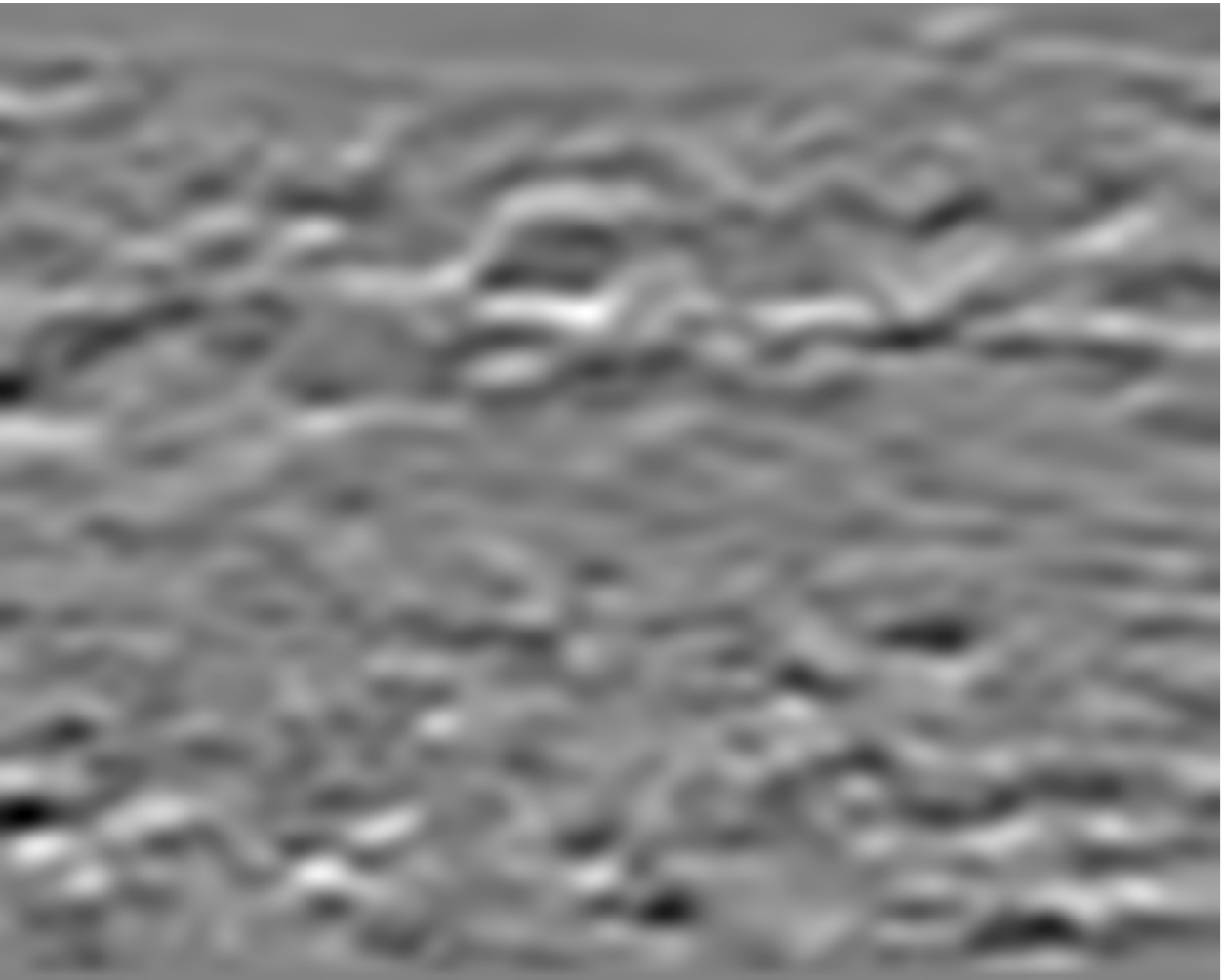}     &
       \includegraphics[width=0.15\textwidth]{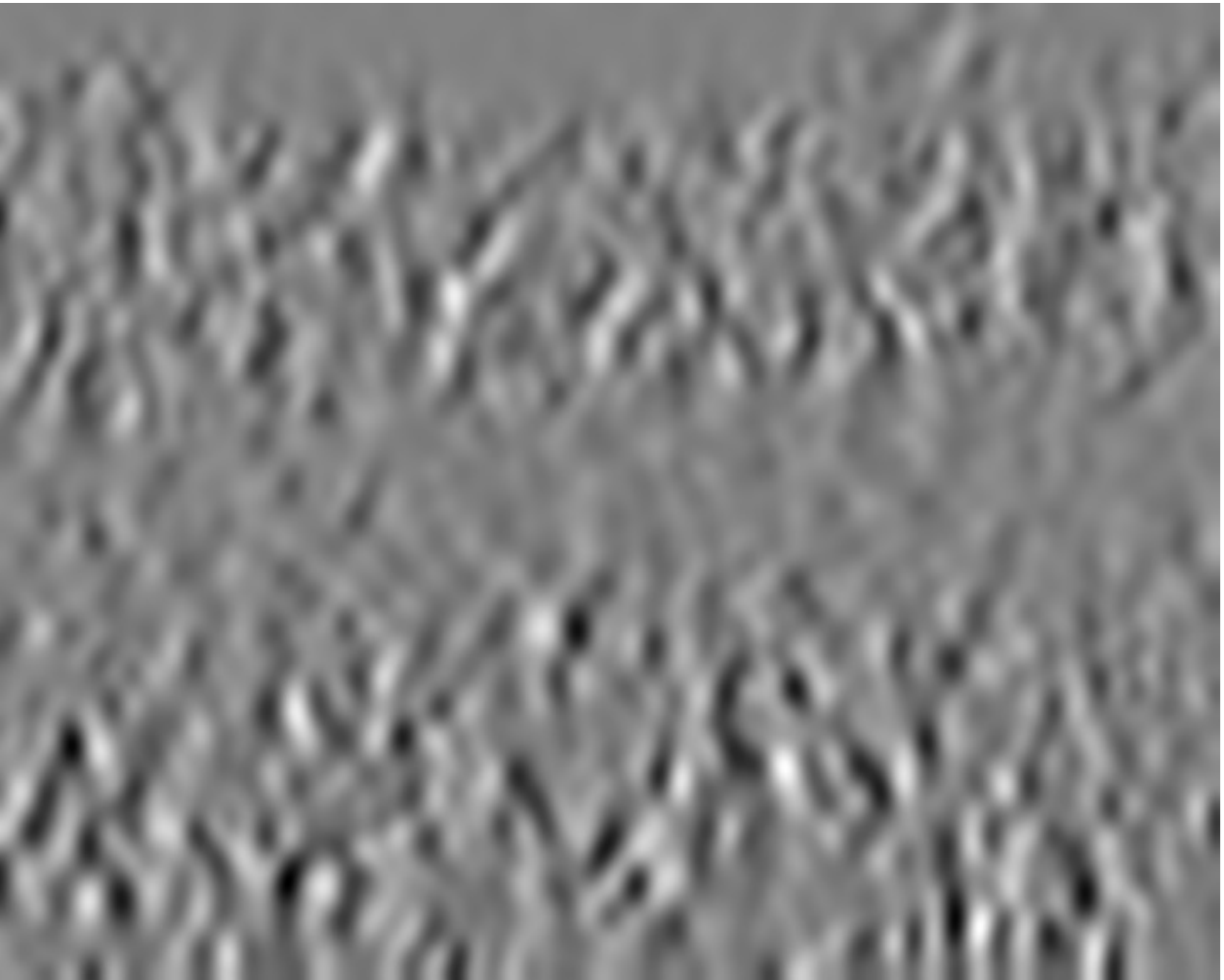}     &
       \includegraphics[width=0.15\textwidth]{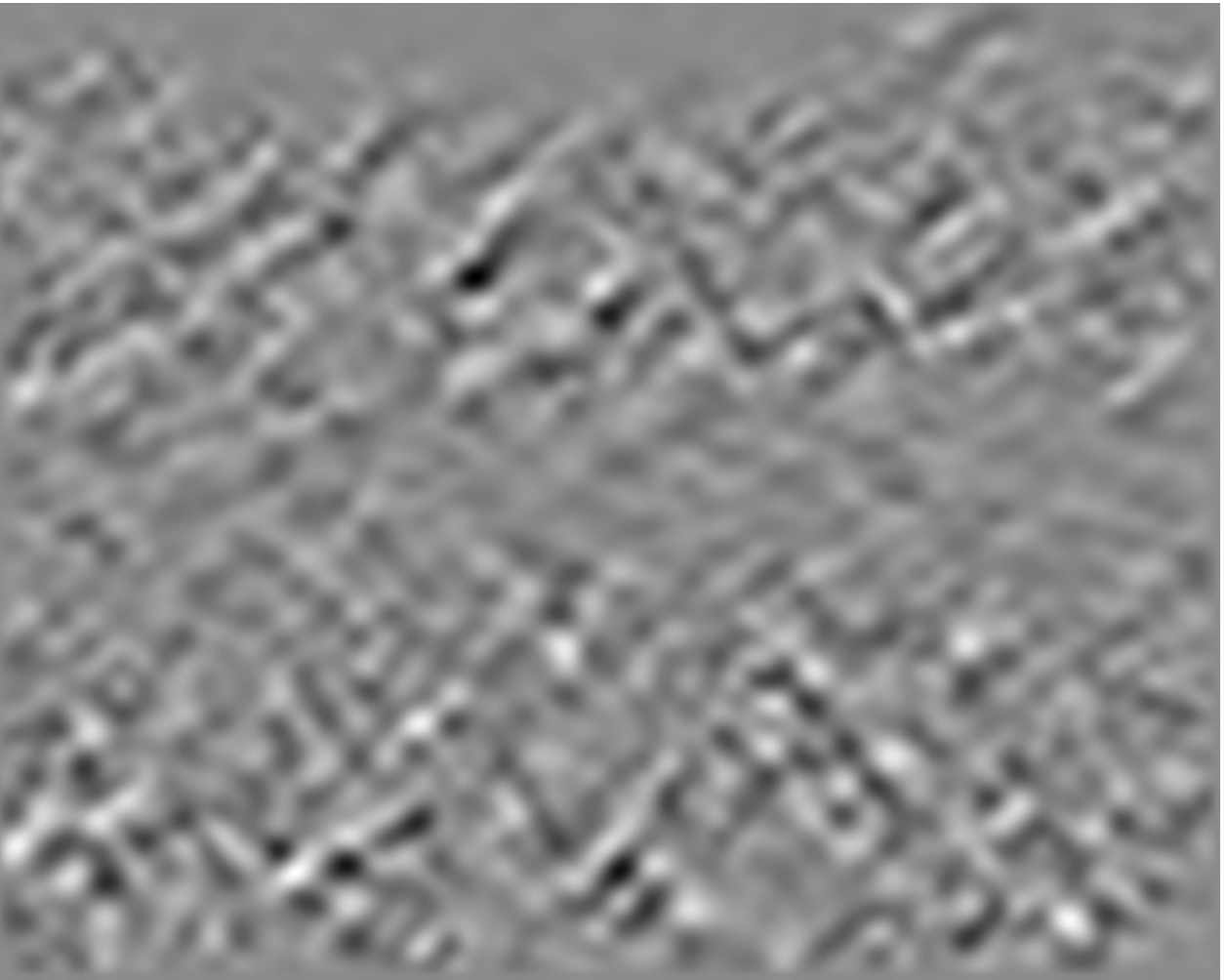} &
              \includegraphics[width=0.15\textwidth]{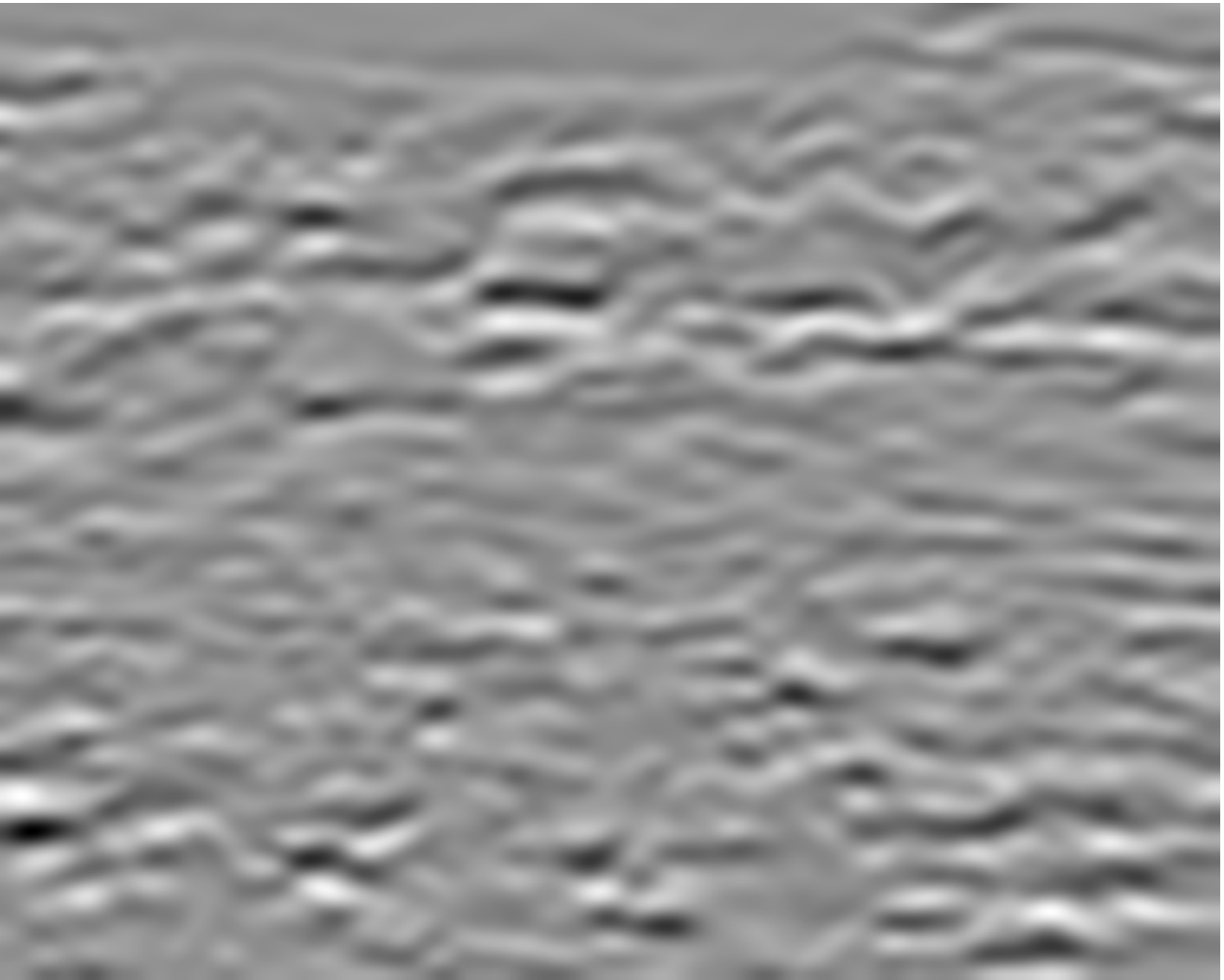}  \\

	\end{tabular}
	\end{center}
	  \vspace{2mm}

        \begin{center}
         \scriptsize
	   \begin{tabular}{c c c c c}
      $f$ & $L$ & $L_t$ & $L_{tt}$ &  \\
       \includegraphics[width=0.15\textwidth]{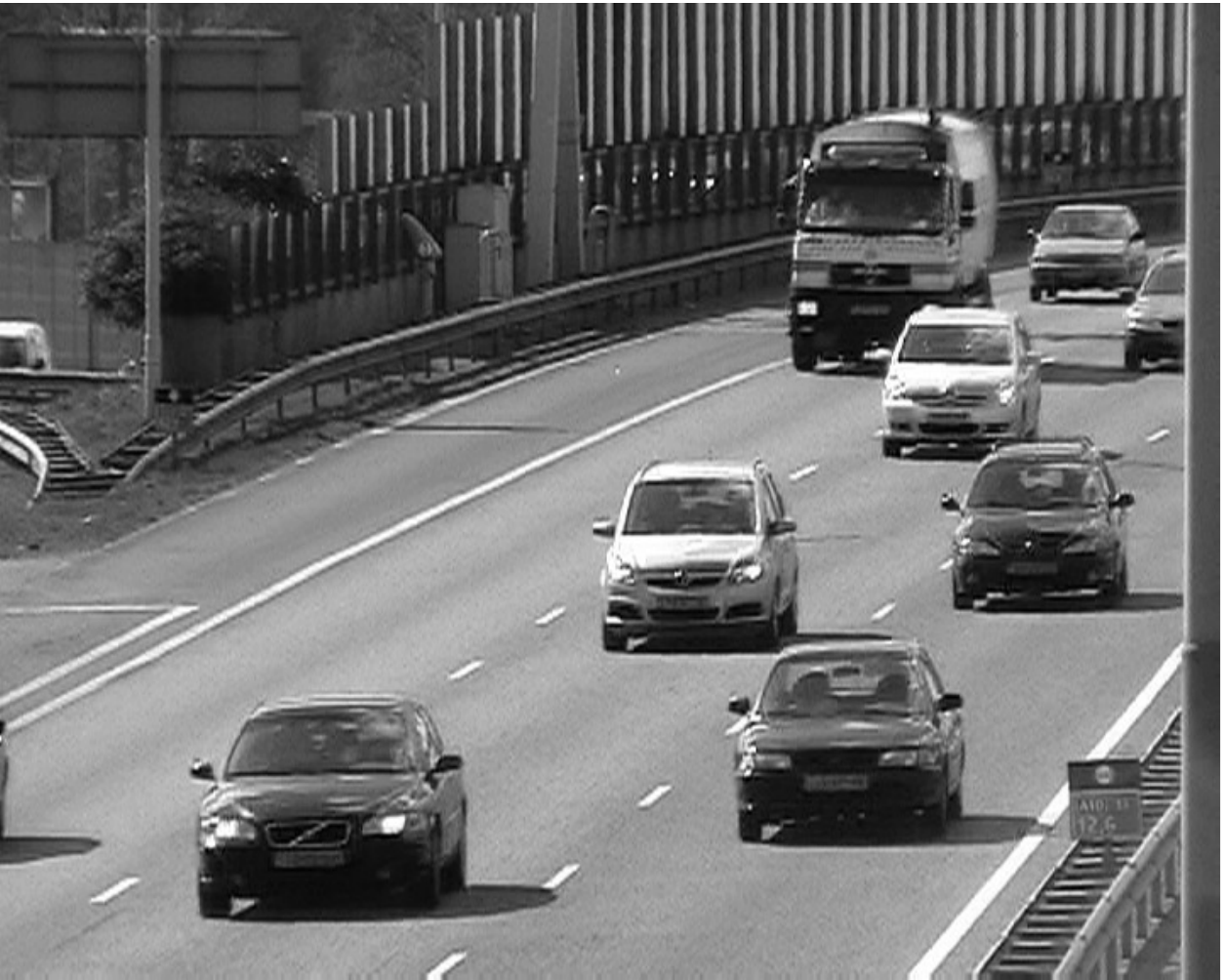}     &
       \includegraphics[width=0.15\textwidth]{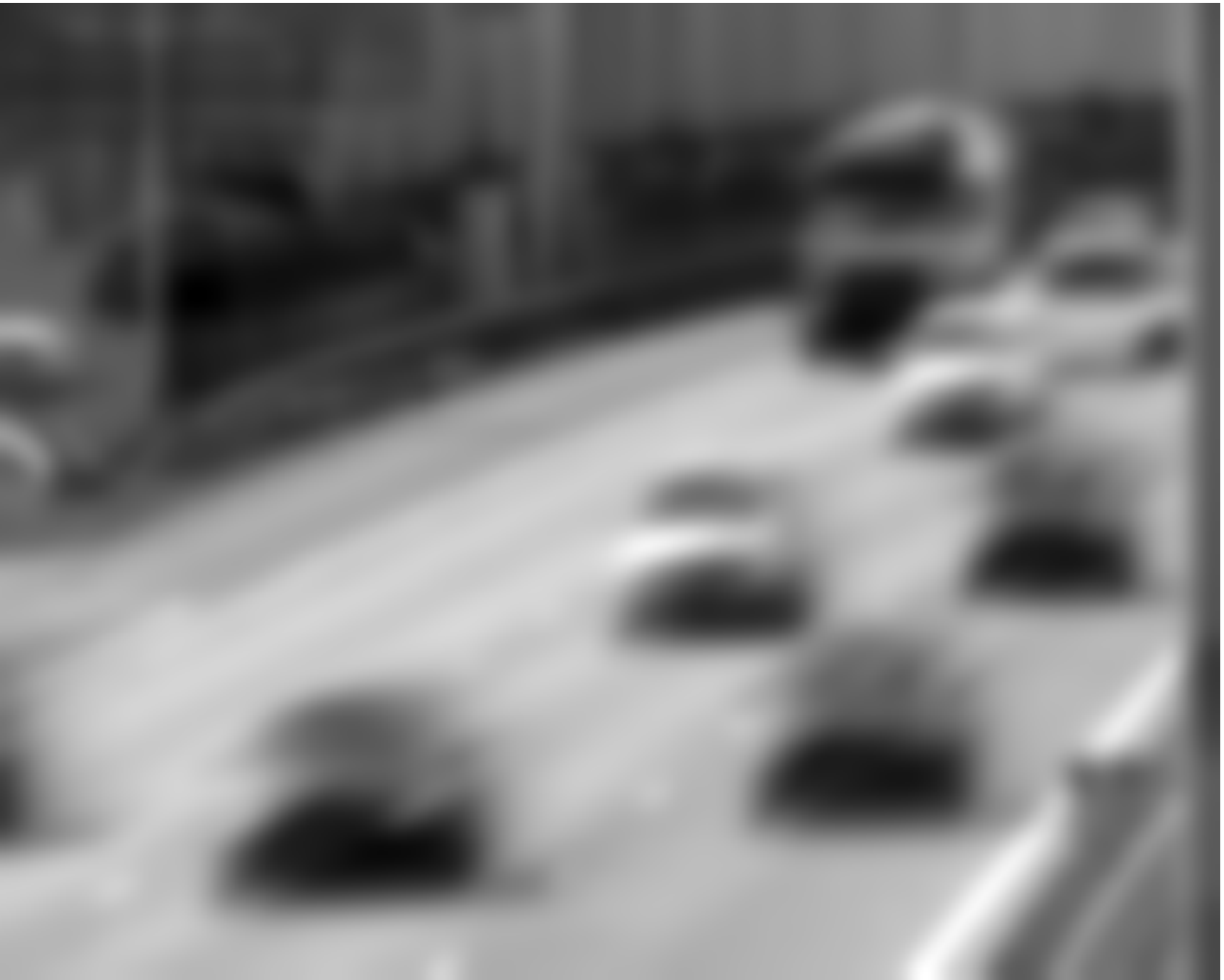}     &
       \includegraphics[width=0.15\textwidth]{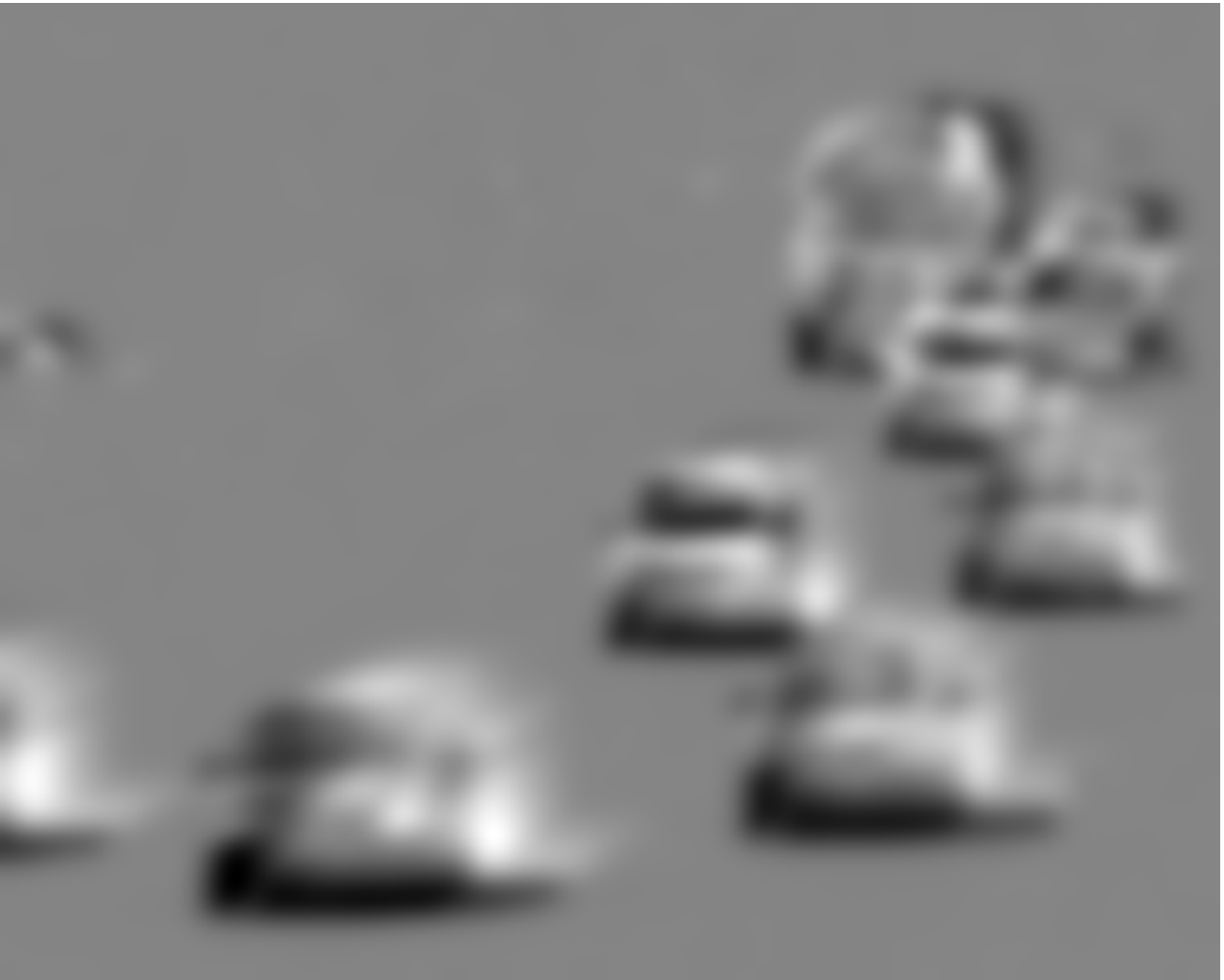}     &
       \includegraphics[width=0.15\textwidth]{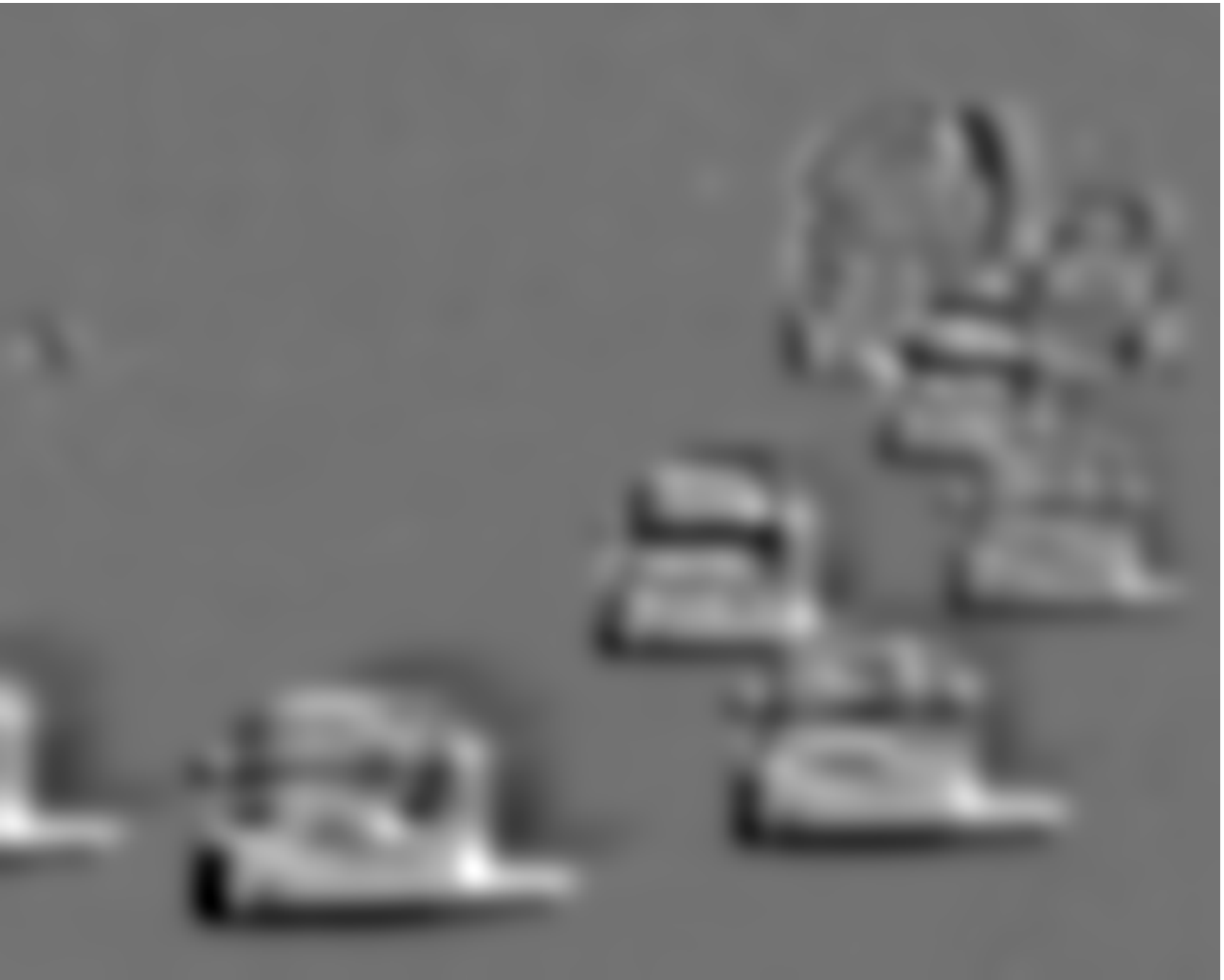}     & \\       
       $L_x$ & $L_y$ & $L_{xx}$ & $L_{xy}$ & $L_{yy}$\\
       
       \includegraphics[width=0.15\textwidth]{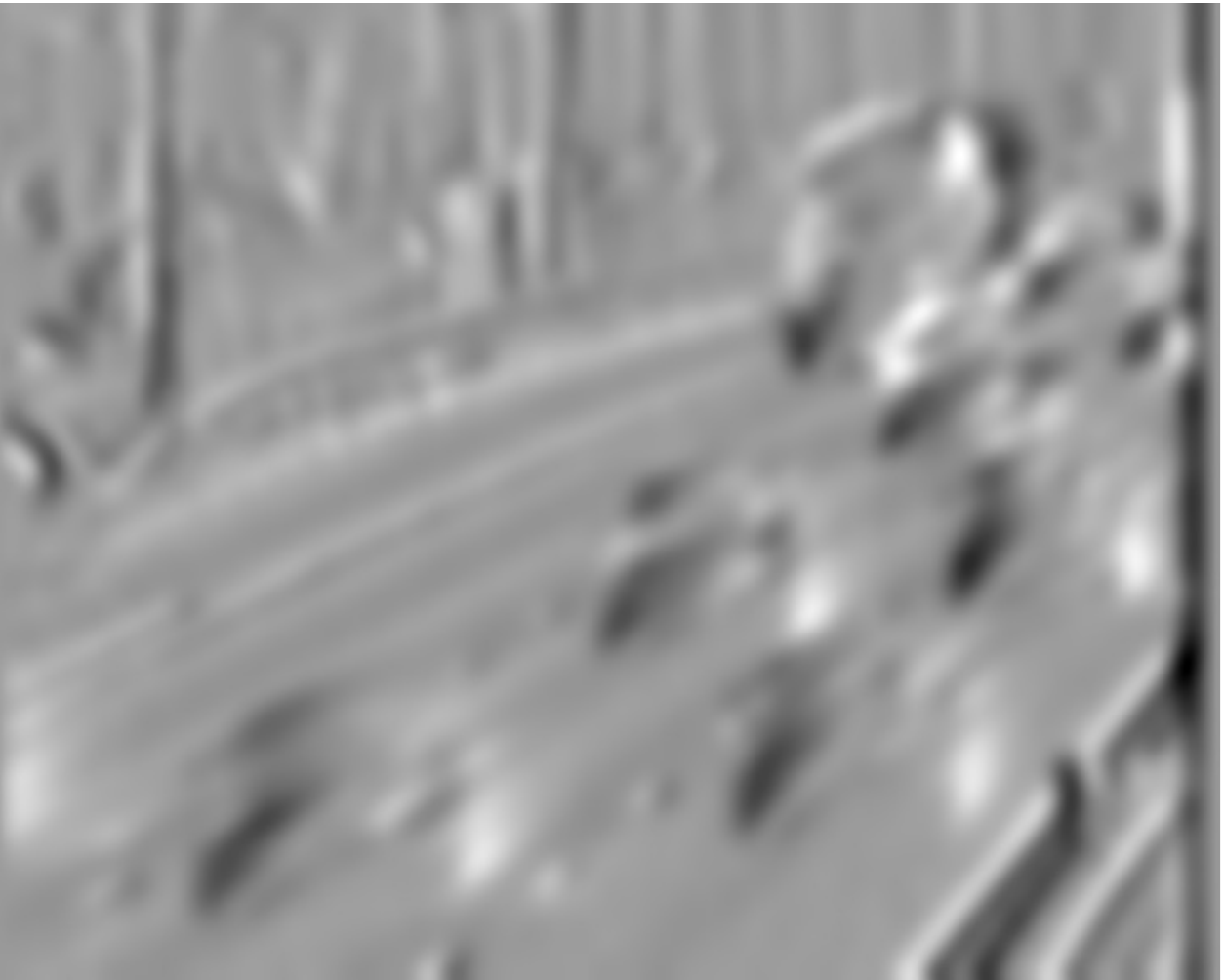}     &
       \includegraphics[width=0.15\textwidth]{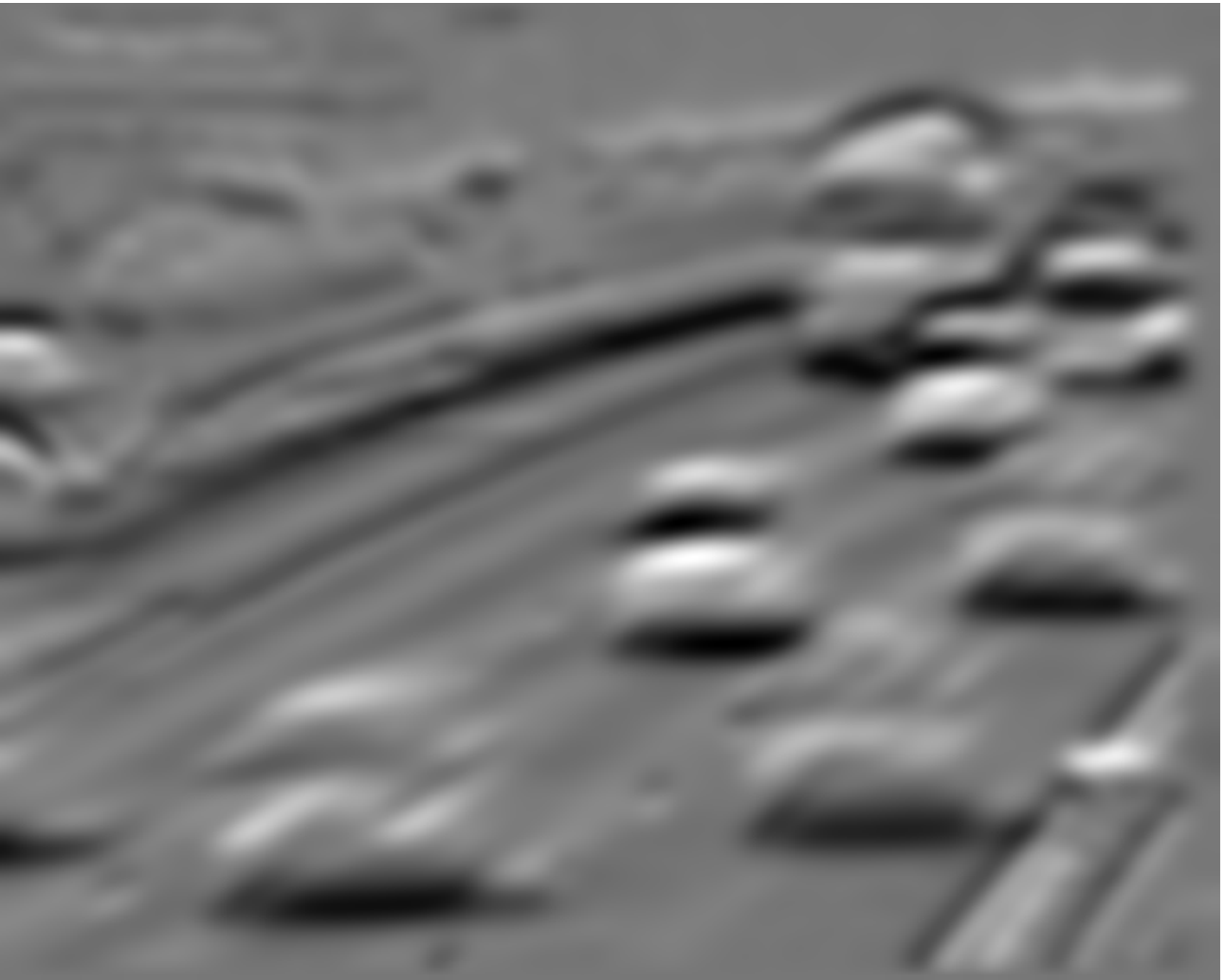}     &
       \includegraphics[width=0.15\textwidth]{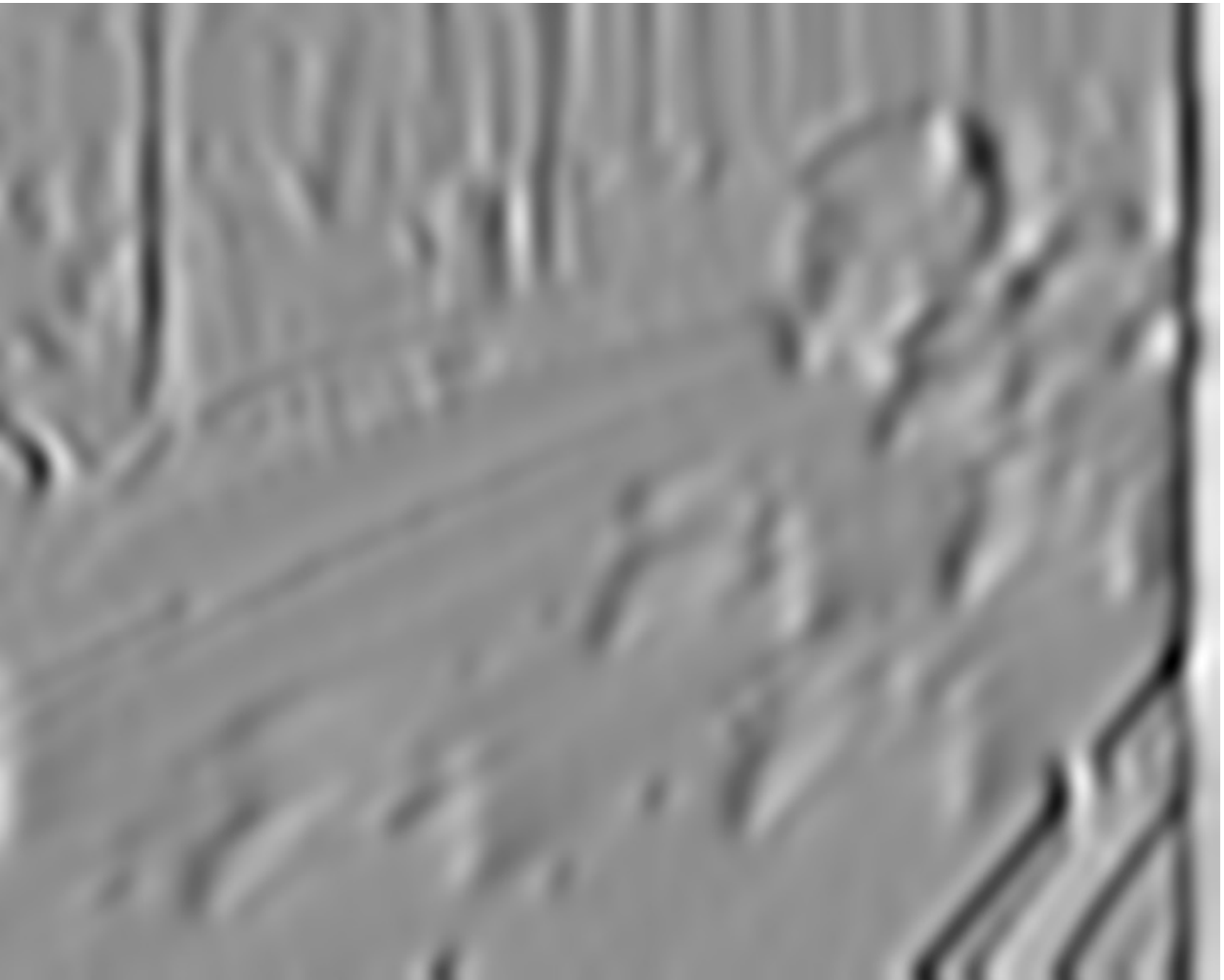}     &
       \includegraphics[width=0.15\textwidth]{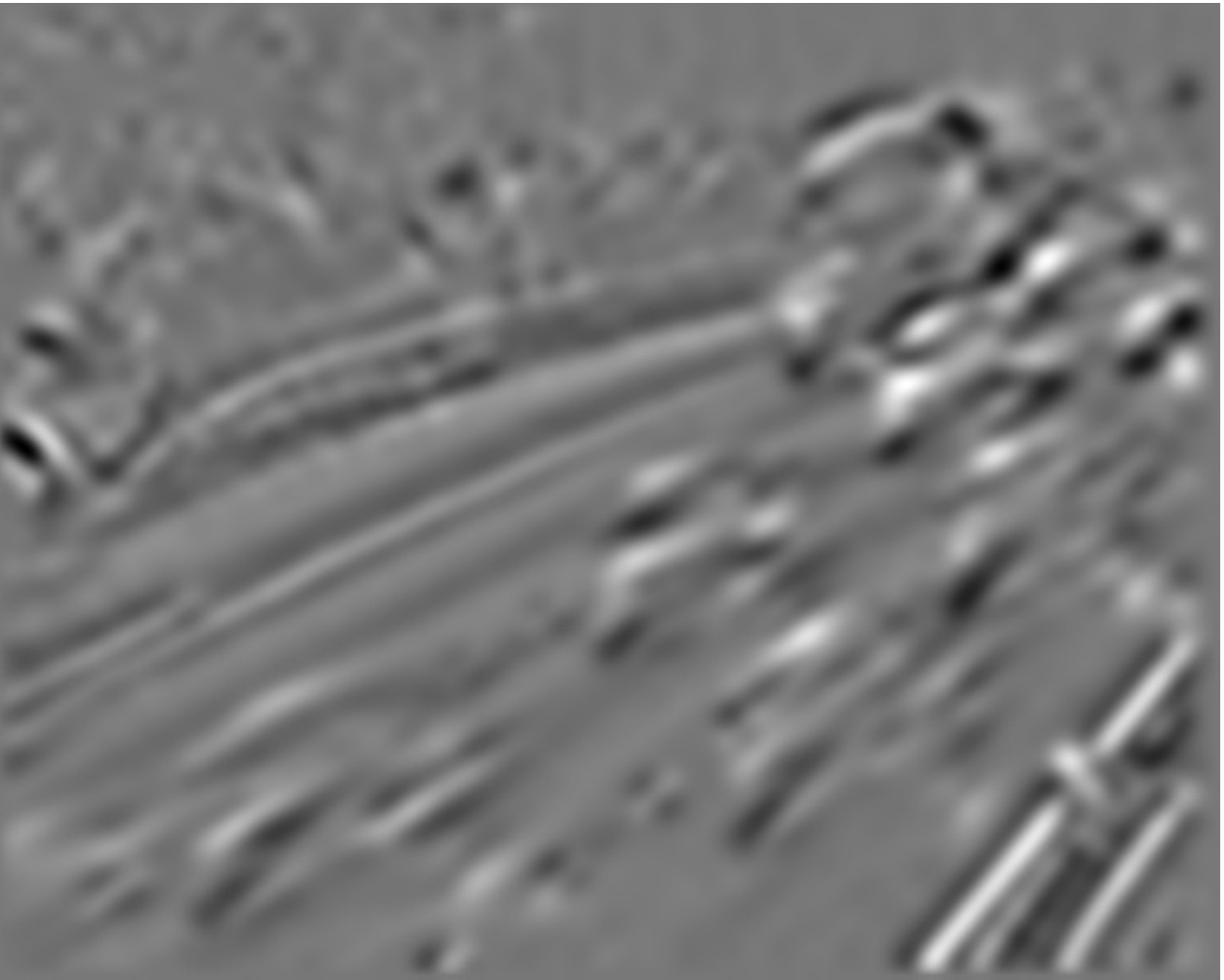}     &
        \includegraphics[width=0.15\textwidth]{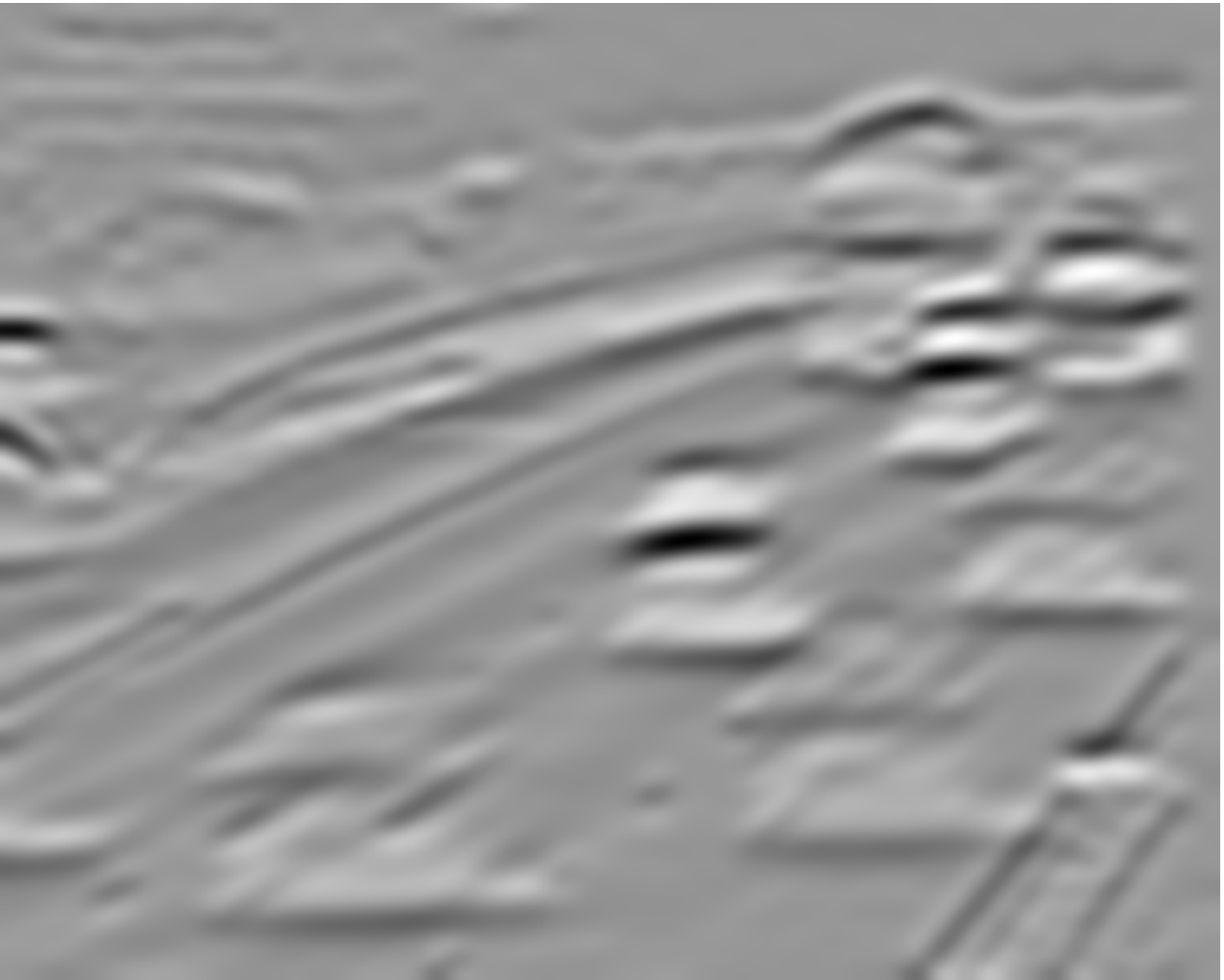}     \\

       $L_{xt}$ & $L_{yt}$ & $L_{xxt}$ & $L_{xyt}$ & $L_{yyt}$\\
       \includegraphics[width=0.15\textwidth]{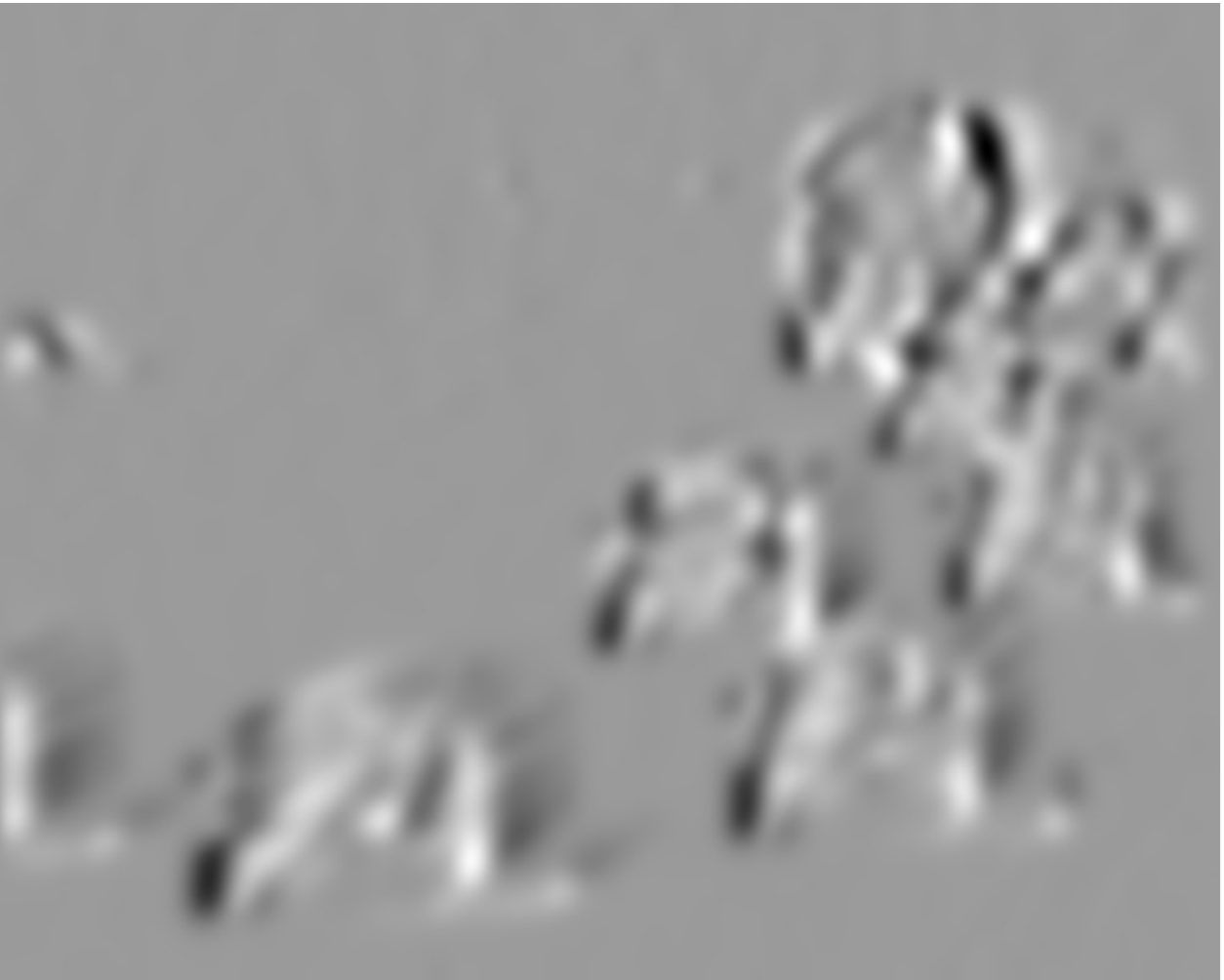}     &
       \includegraphics[width=0.15\textwidth]{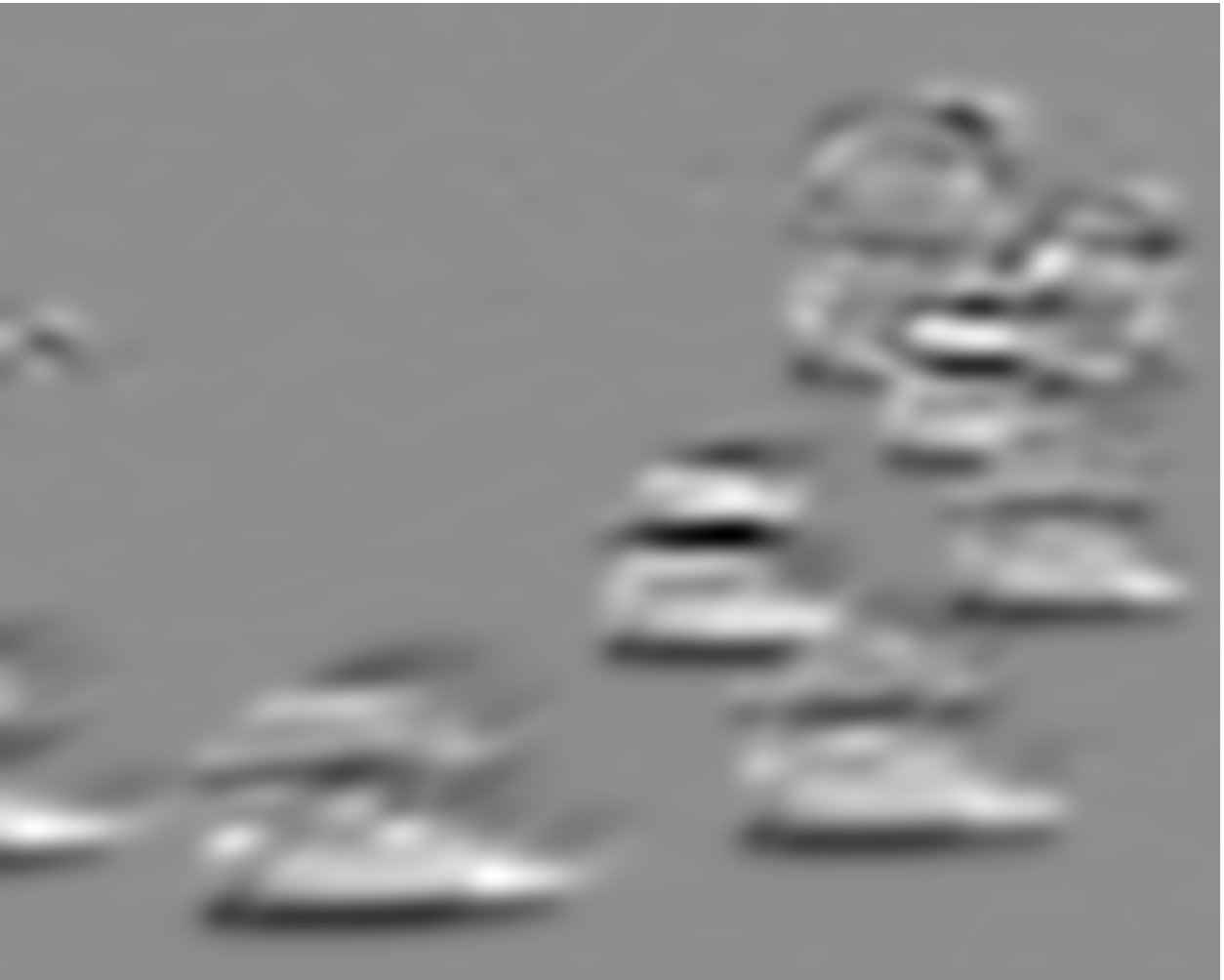}     &
       \includegraphics[width=0.15\textwidth]{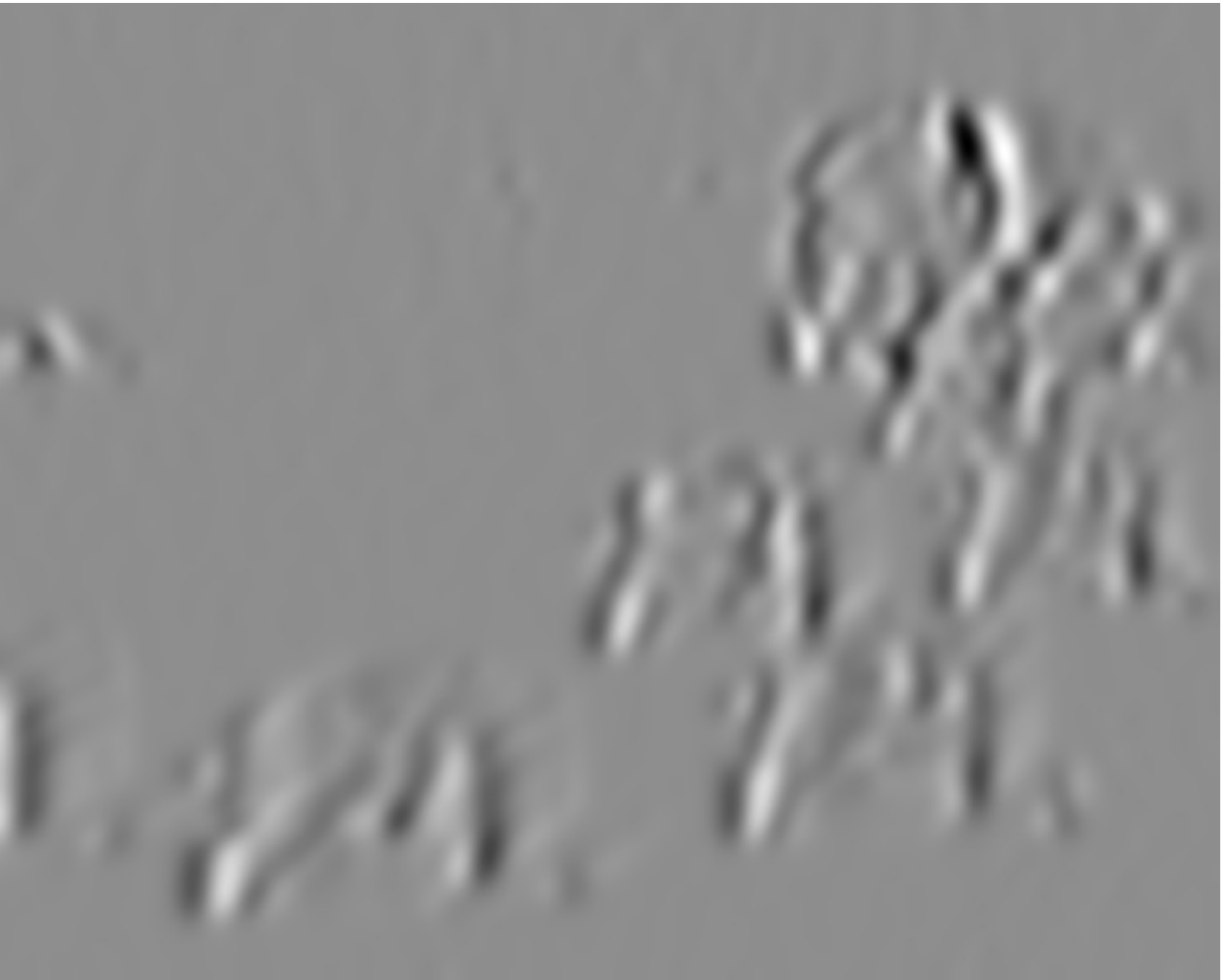}     &
       \includegraphics[width=0.15\textwidth]{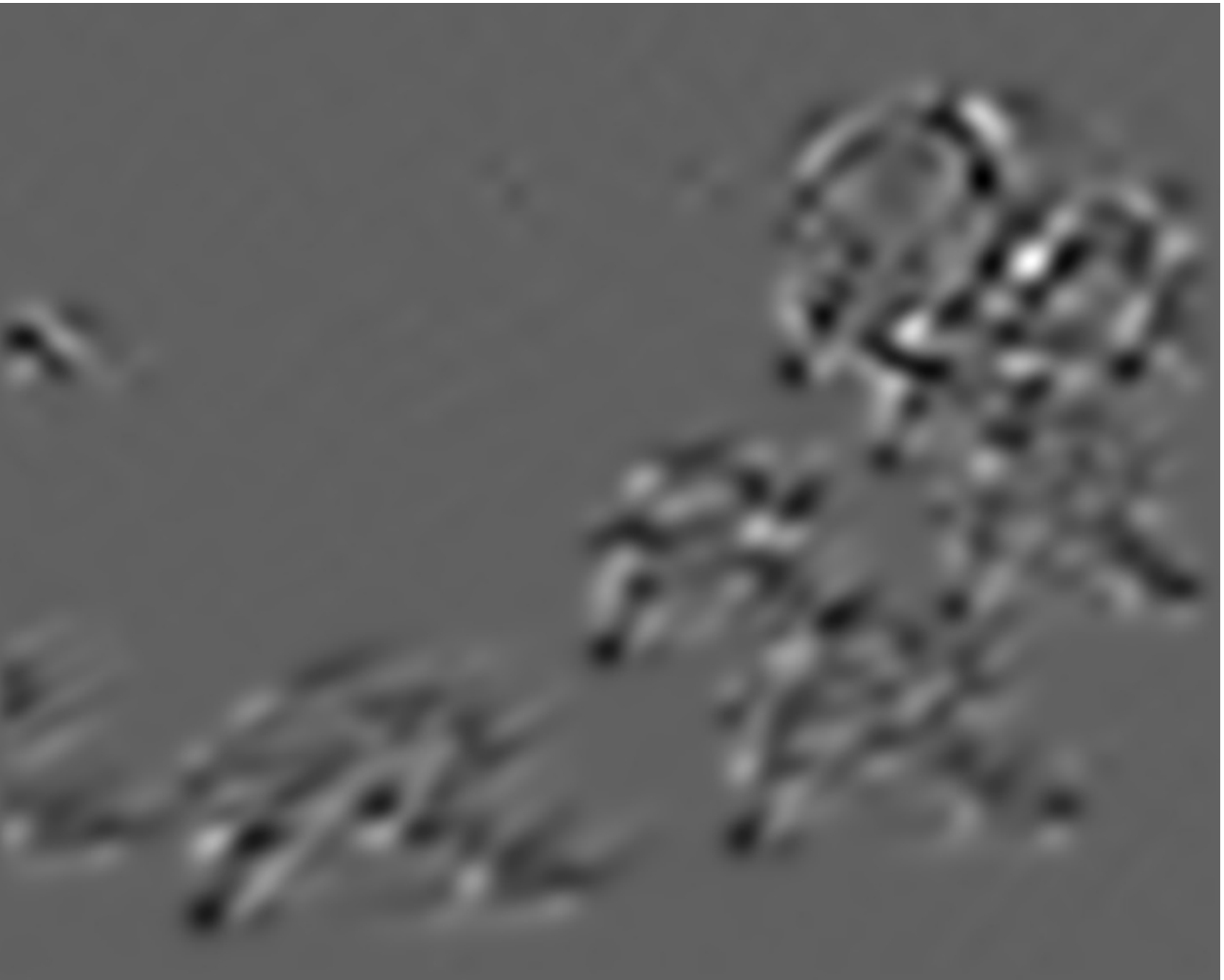}    &
       \includegraphics[width=0.15\textwidth]{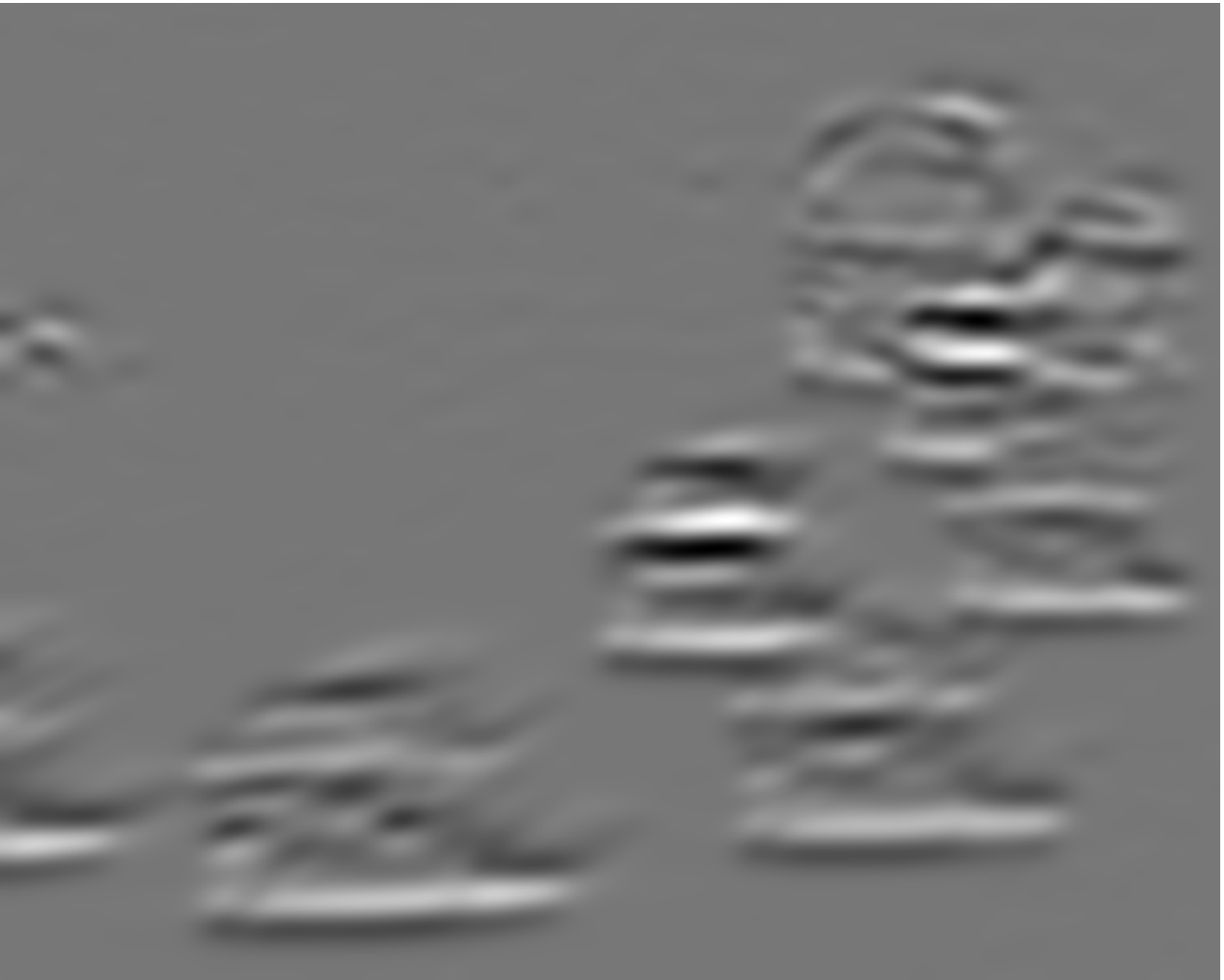}     \\

       $L_{xtt}$ & $L_{ytt}$ & $L_{xxtt}$ & $L_{xytt}$ & $L_{yytt}$\\
       \includegraphics[width=0.15\textwidth]{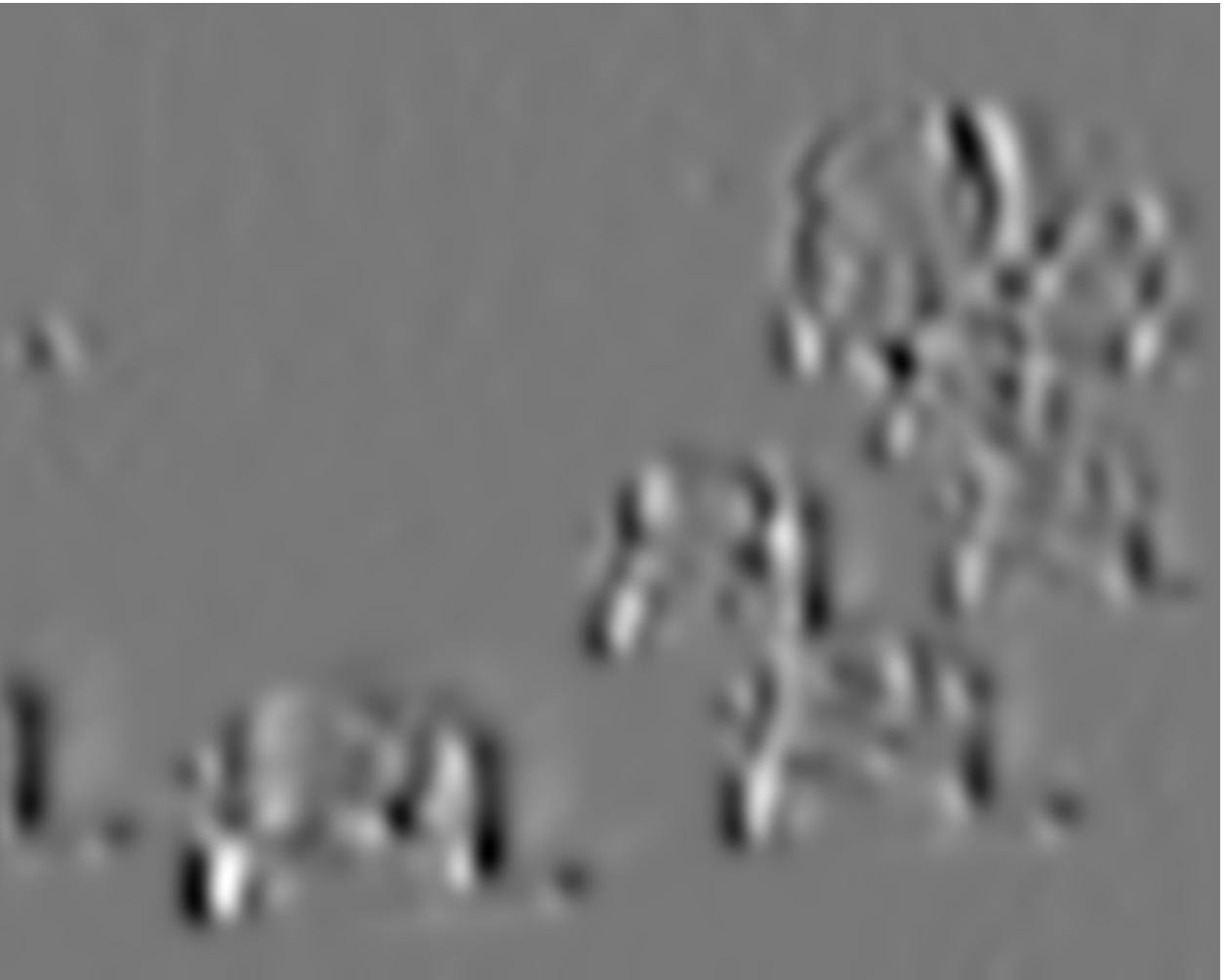}     &
       \includegraphics[width=0.15\textwidth]{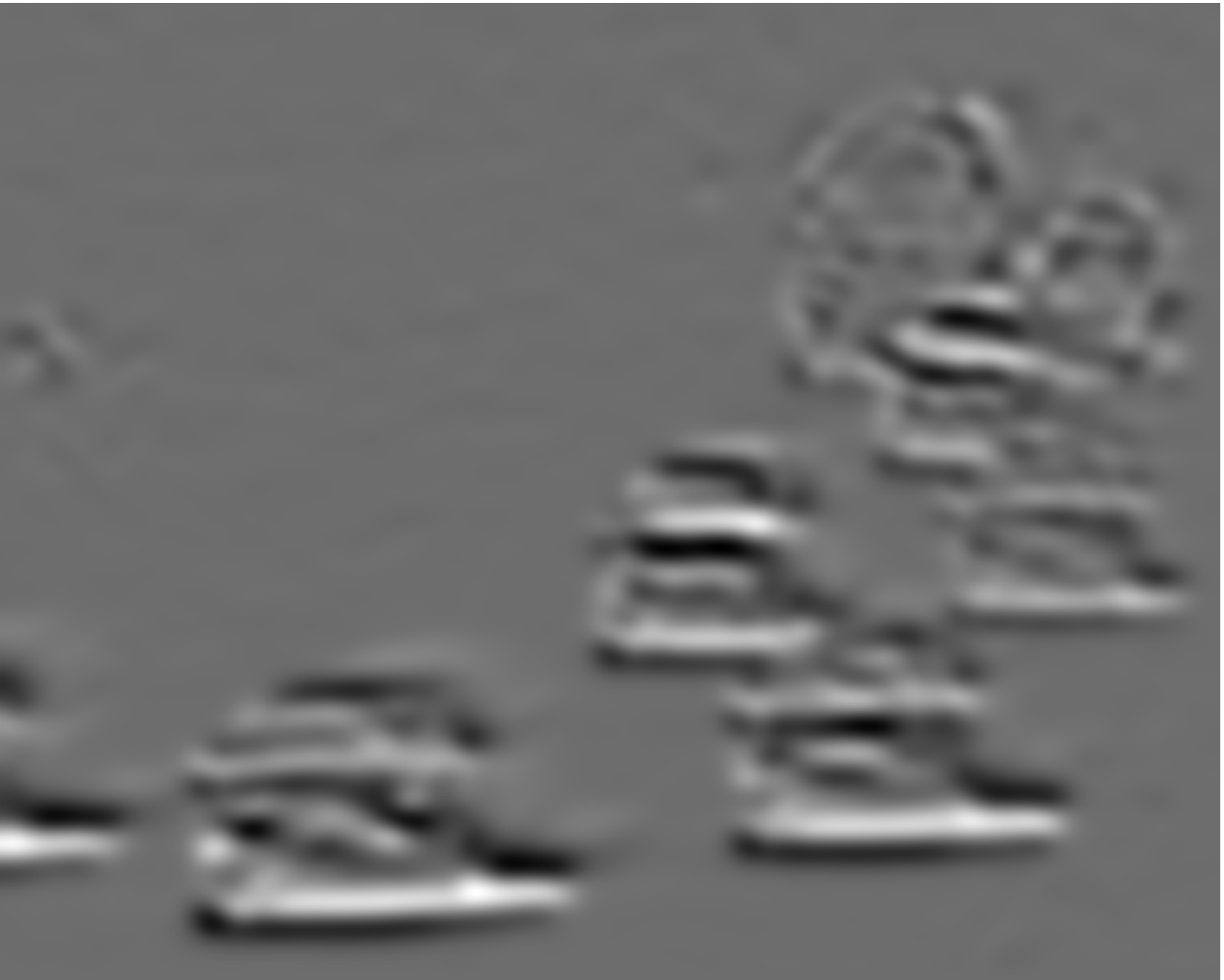}     &
       \includegraphics[width=0.15\textwidth]{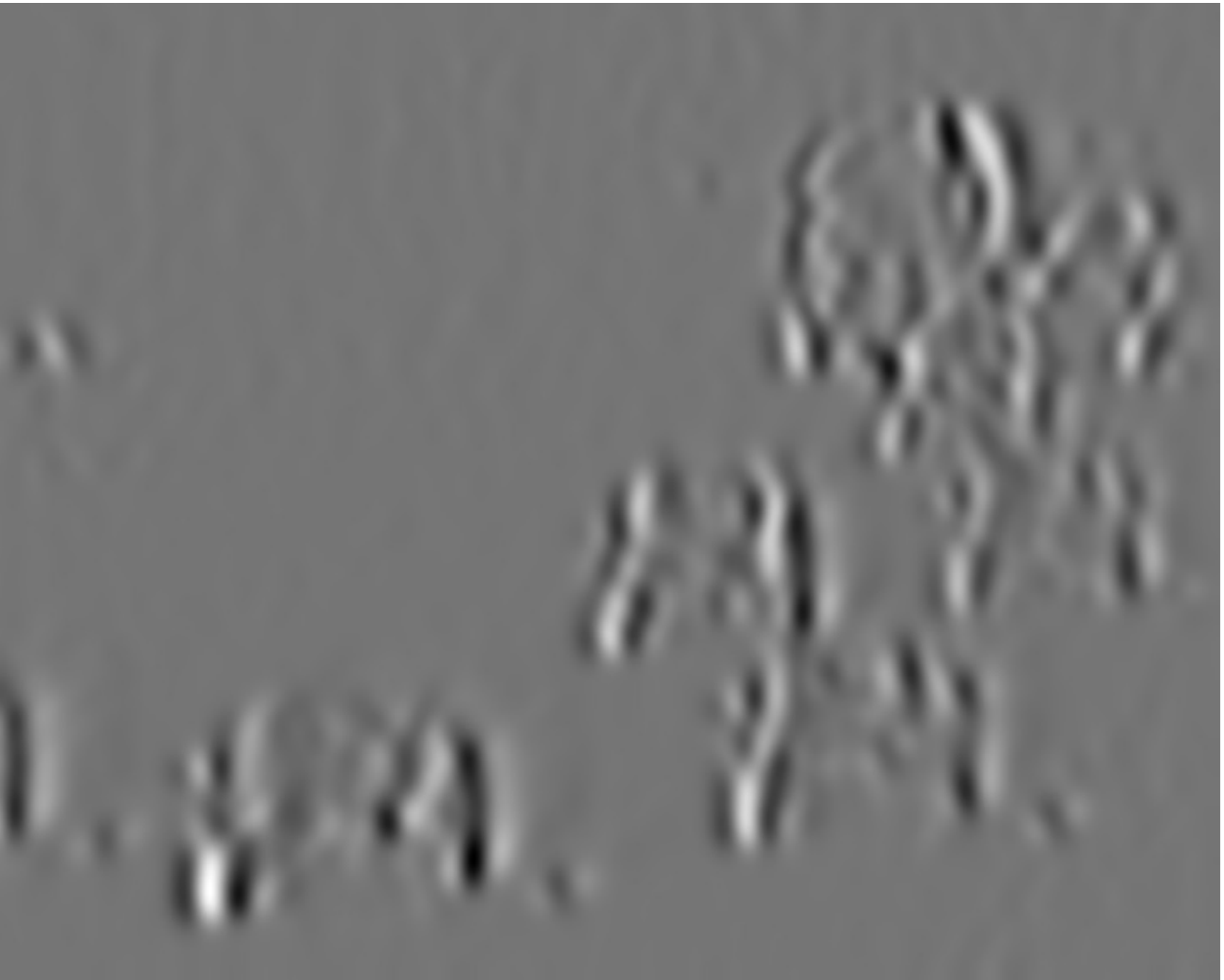}     &
       \includegraphics[width=0.15\textwidth]{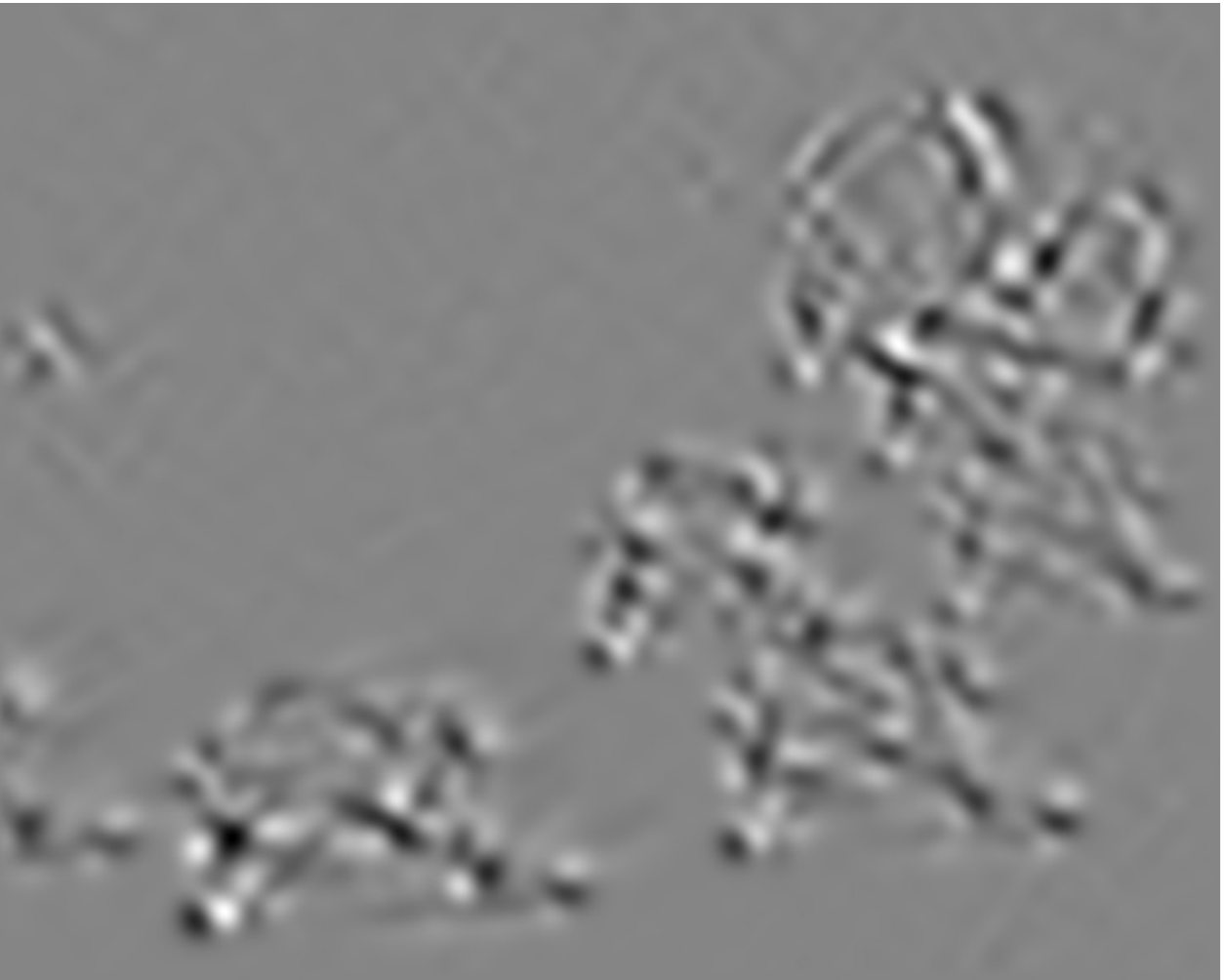}   &  
       \includegraphics[width=0.15\textwidth]{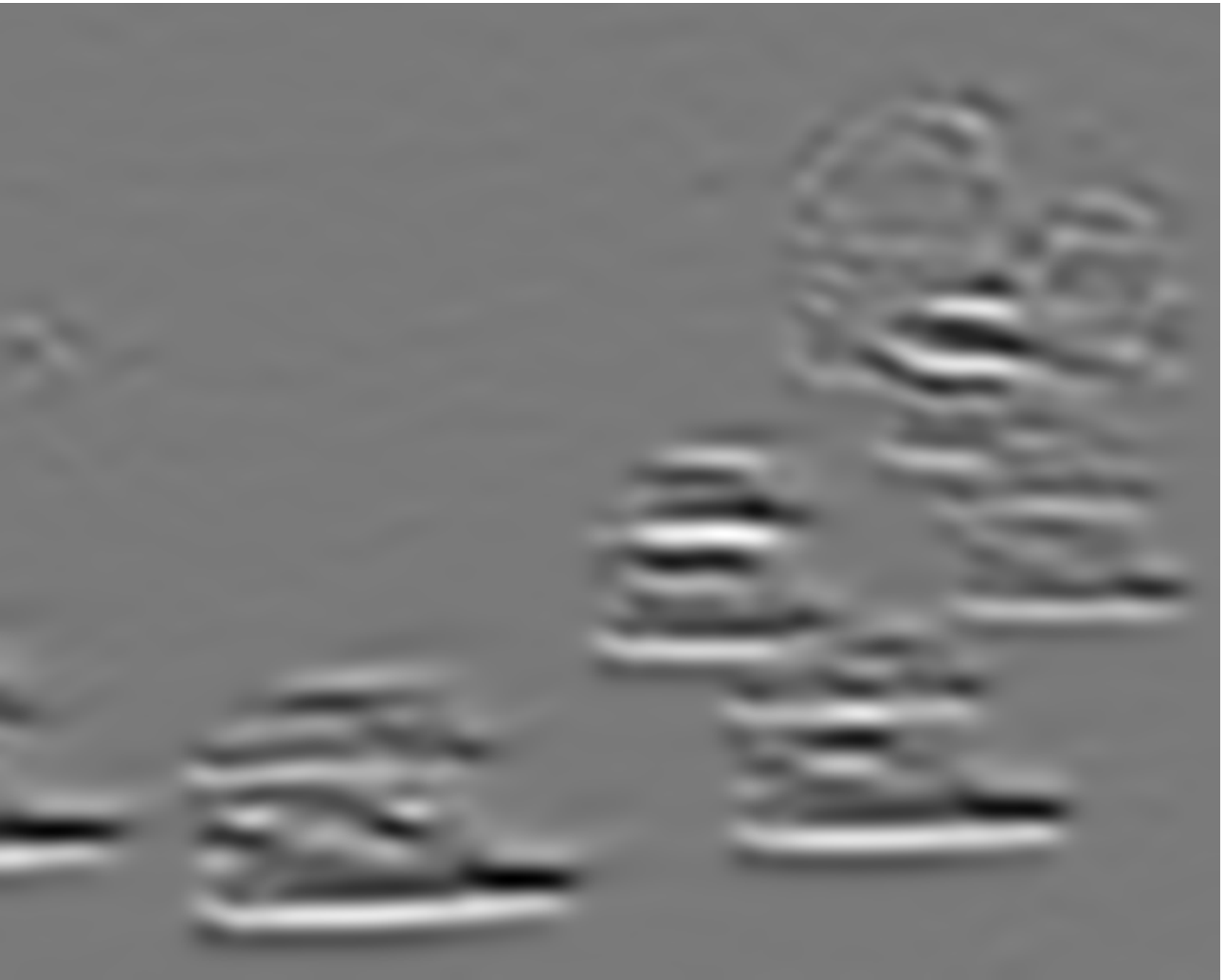}     \\
   
	\end{tabular}

   \end{center}
   
   \endgroup

   \caption{\emph{Spatio-temporal receptive field responses} for the set of spatio-temporal partial derivatives that the STRF $N$-jet descriptor is based on. Snapshots of receptive field responses are shown for two dynamic textures from the DynTex classes "waves" (top) and "traffic" (bottom) ($\sigma_s = 8$, $\sigma_\tau = 100$)}
\label{fig:strf-responses}
\end{figure*}

\begin{figure}[hpbt]   
\begingroup
    \begin{center}
    \scriptsize
    \setlength\tabcolsep{0.5mm}

    \begin{tabular}{c c c}

       $ | \nabla L |  $ & $| \nabla L_t | $ & $| \nabla L_{tt} |$   \\
       \includegraphics[width=0.15\textwidth]{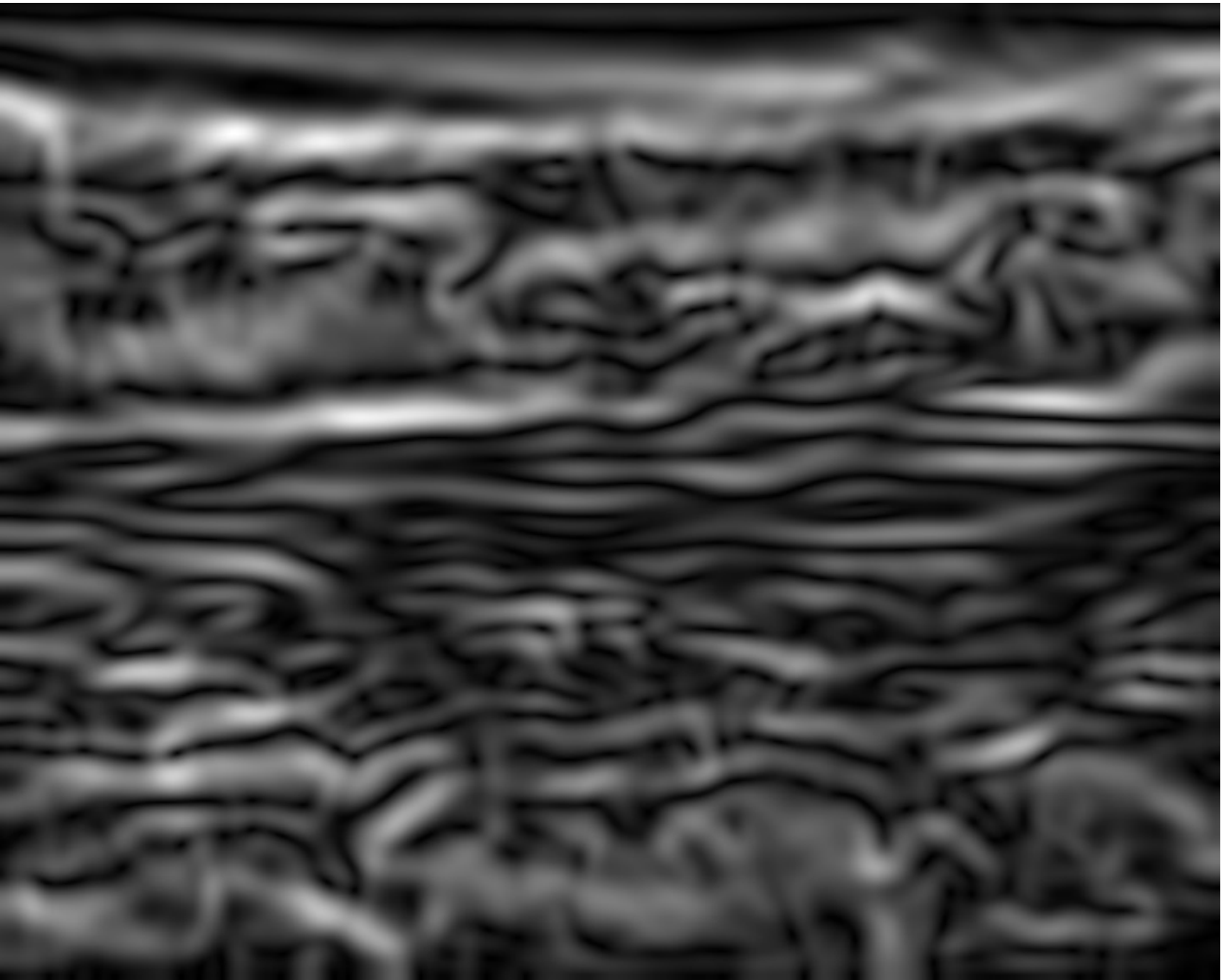} &
       \includegraphics[width=0.15\textwidth]{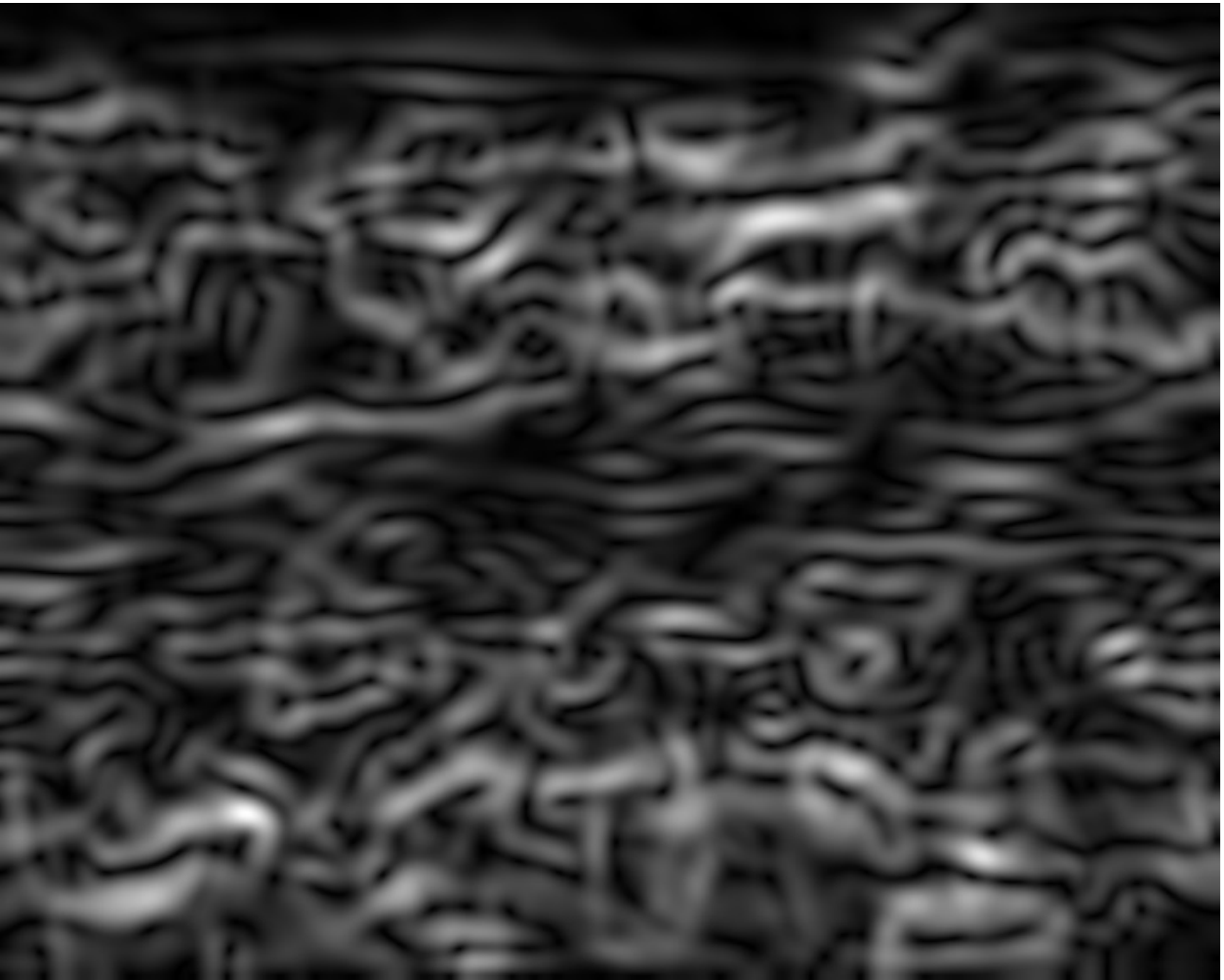} &
       \includegraphics[width=0.15\textwidth]{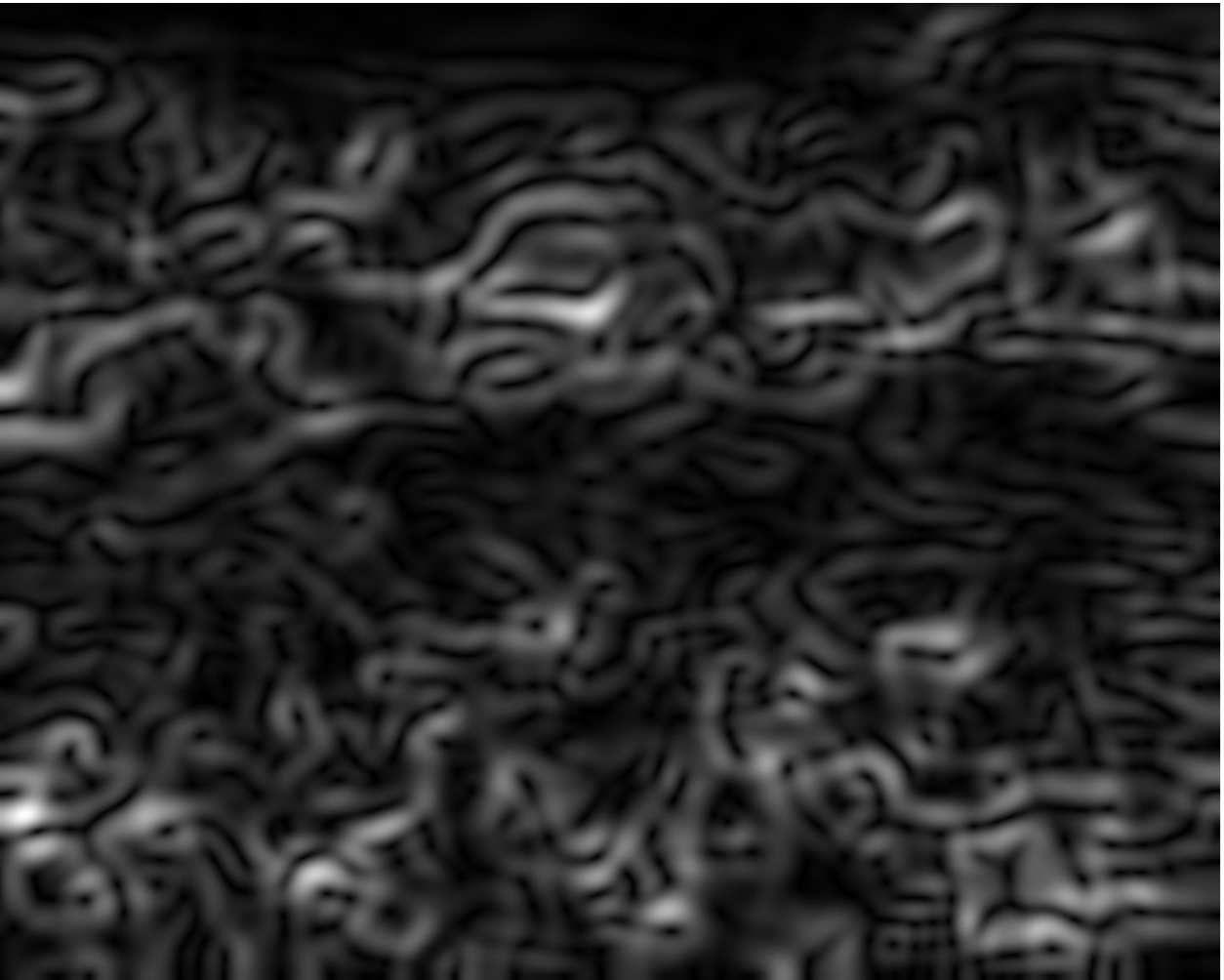}   \\       
       $ \nabla^2 L   $ & $ \nabla^2 L_t $ & $ \nabla^2 L_{tt} $   \\
       \includegraphics[width=0.15\textwidth]{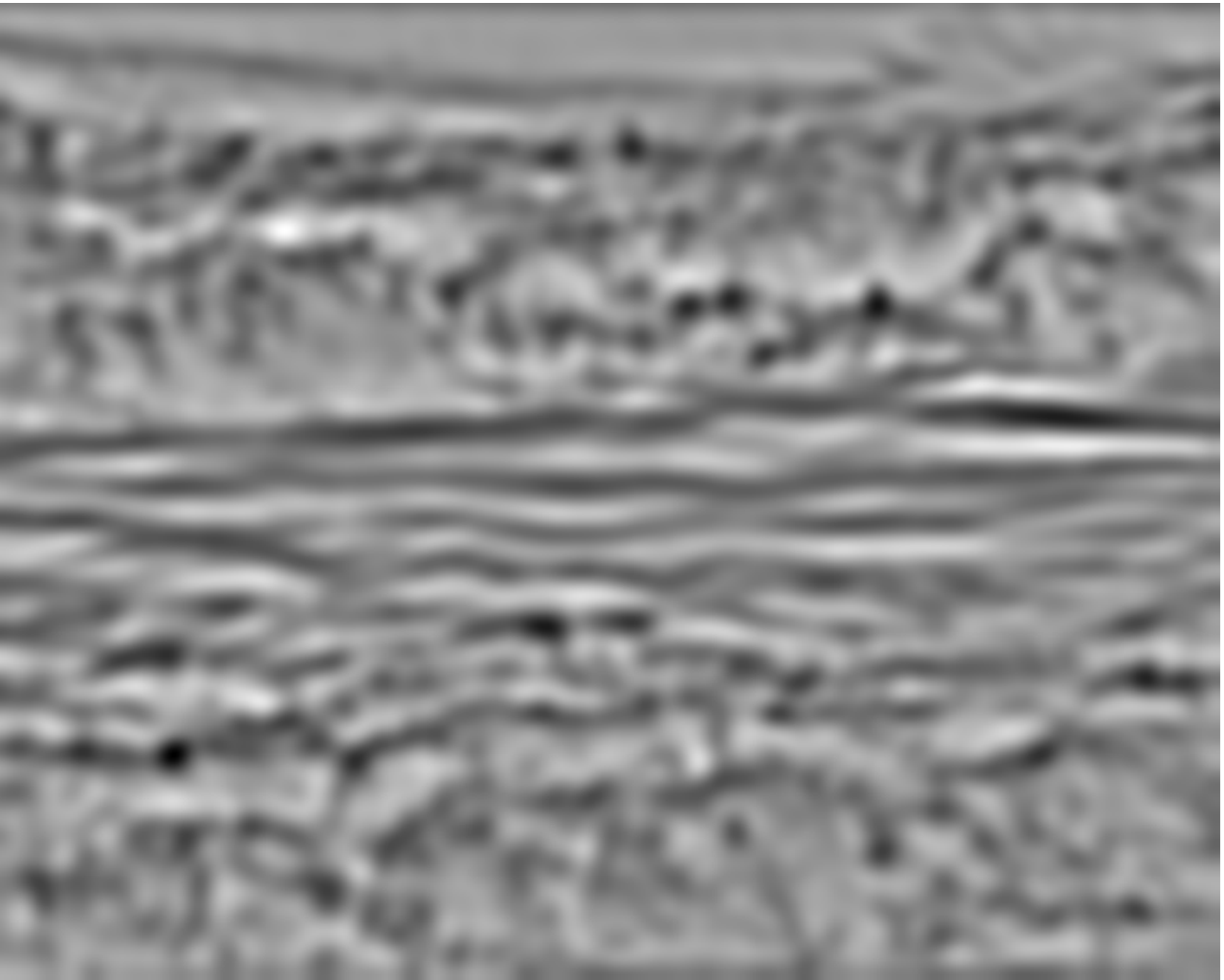} &
       \includegraphics[width=0.15\textwidth]{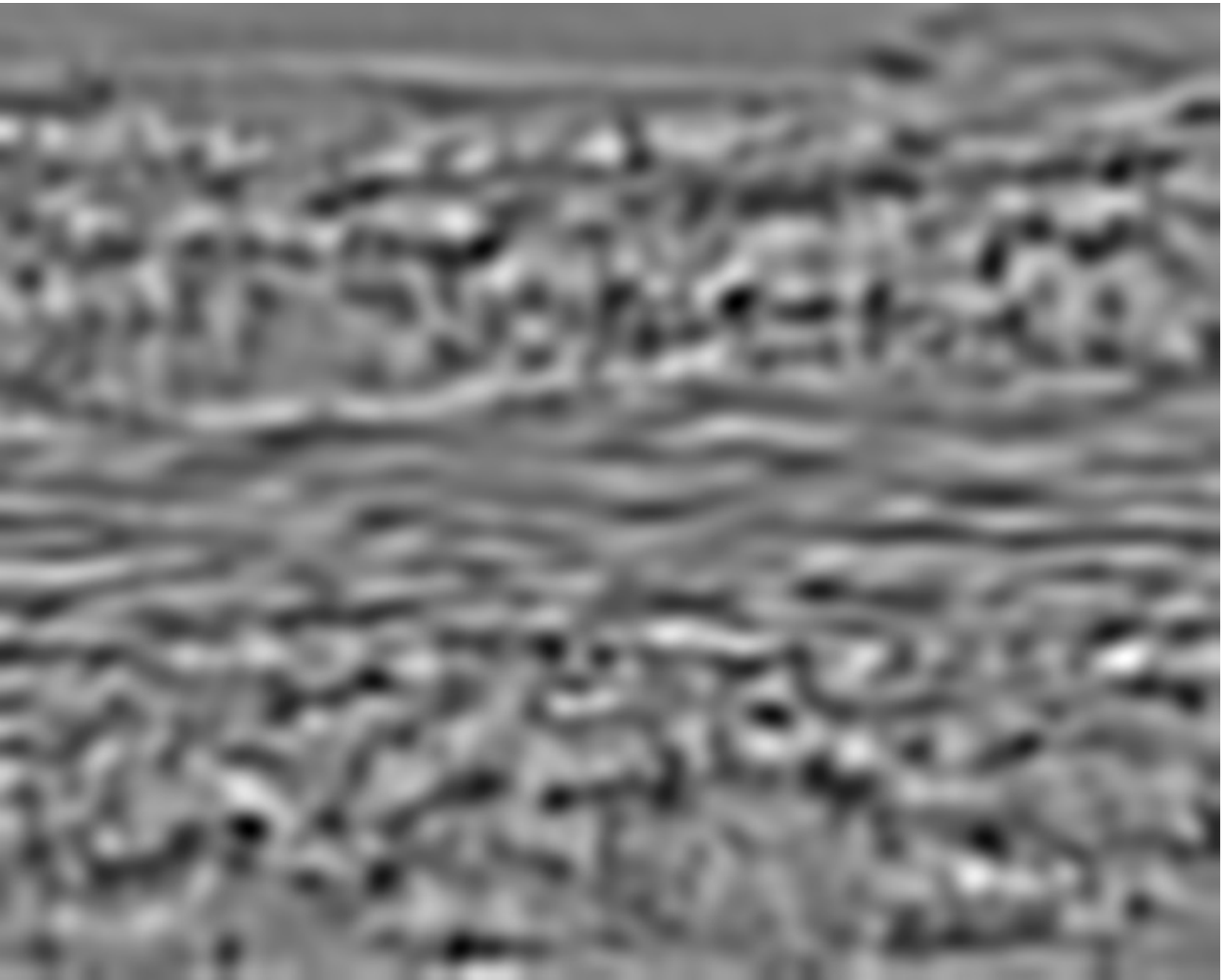} &
       \includegraphics[width=0.15\textwidth]{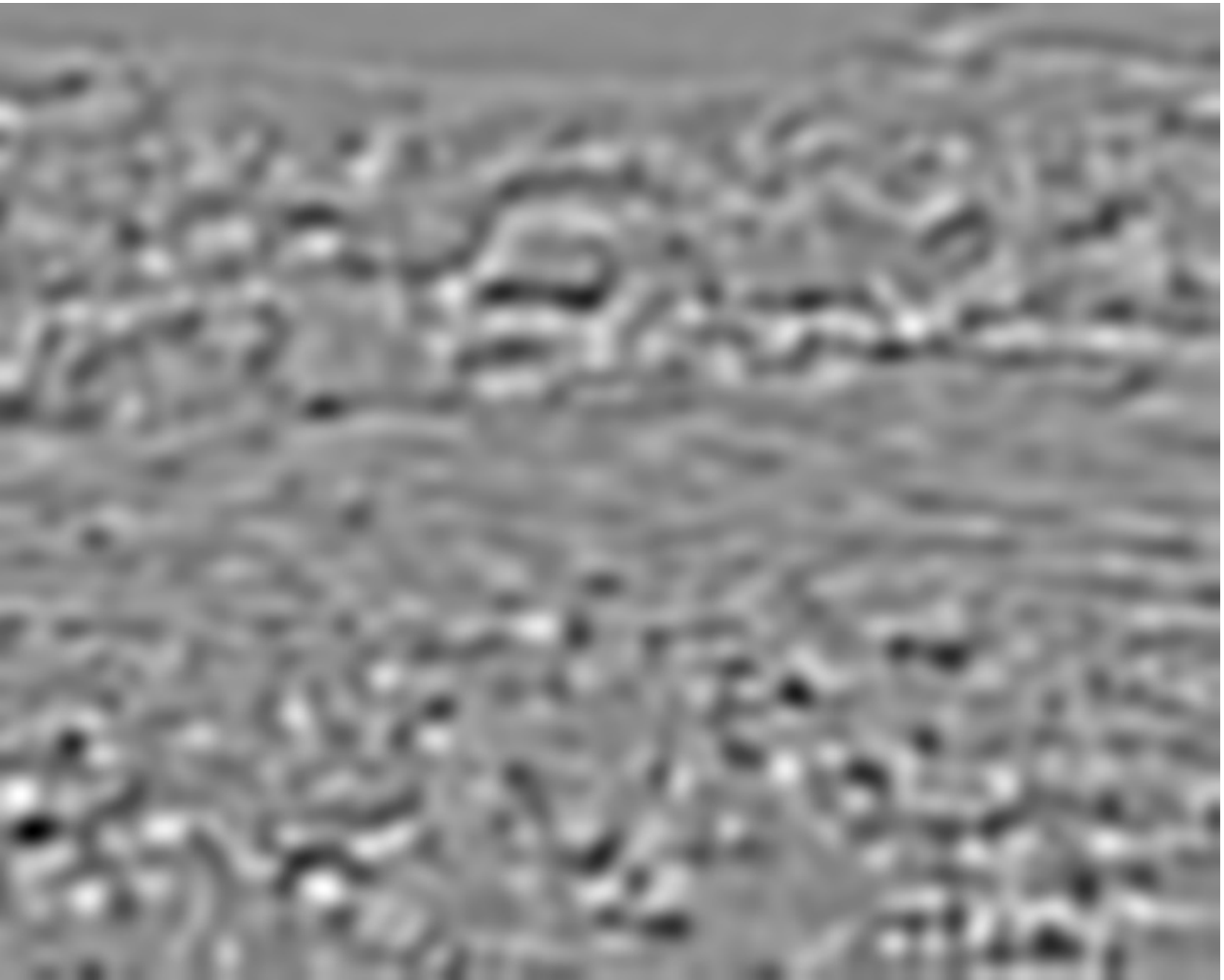}     \\

       $ \det \mathcal{H} L $ & $ \det \mathcal{H} L_t $ & $\det \mathcal{H} L_{tt} $  \\
       \includegraphics[width=0.15\textwidth]{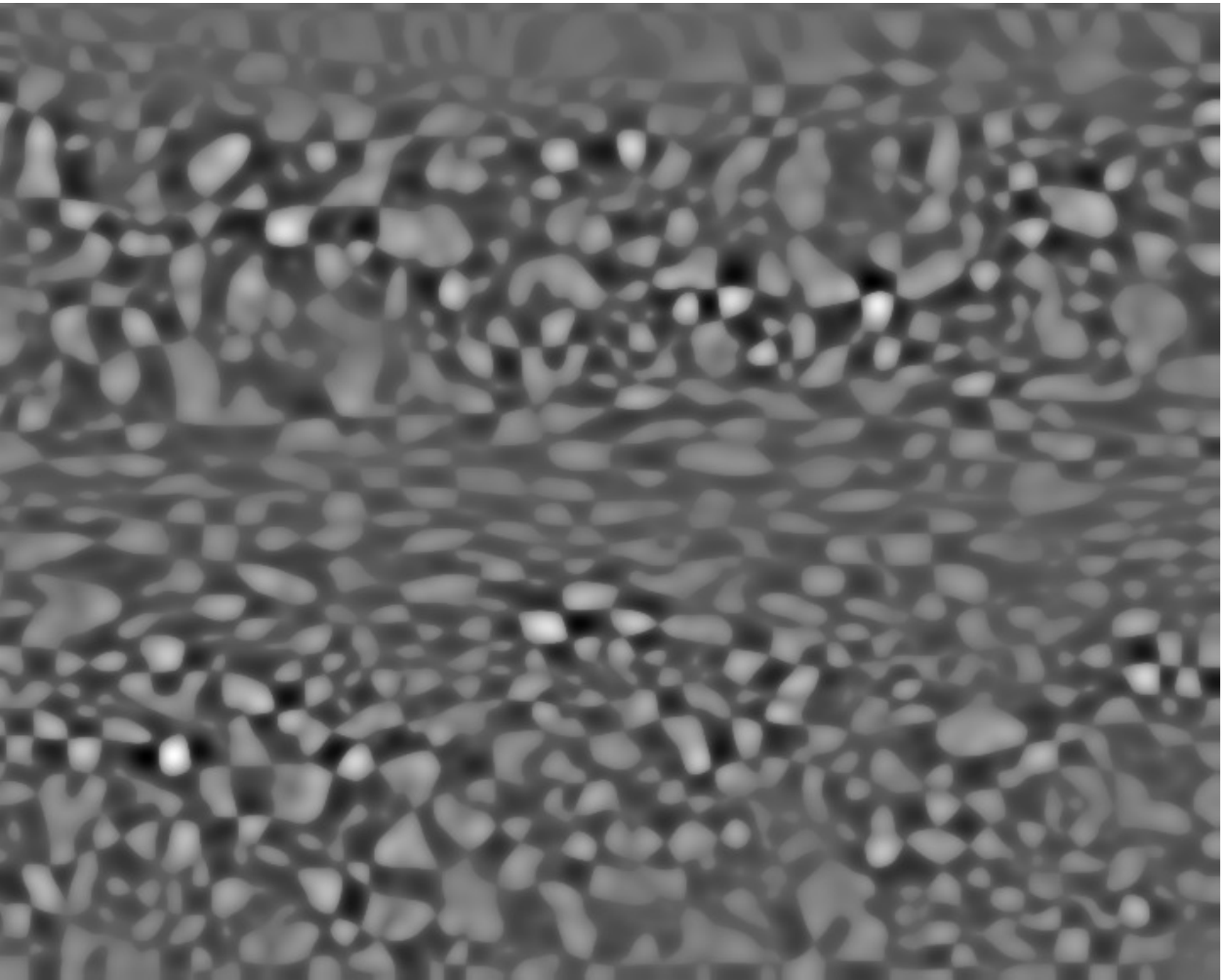} &
       \includegraphics[width=0.15\textwidth]{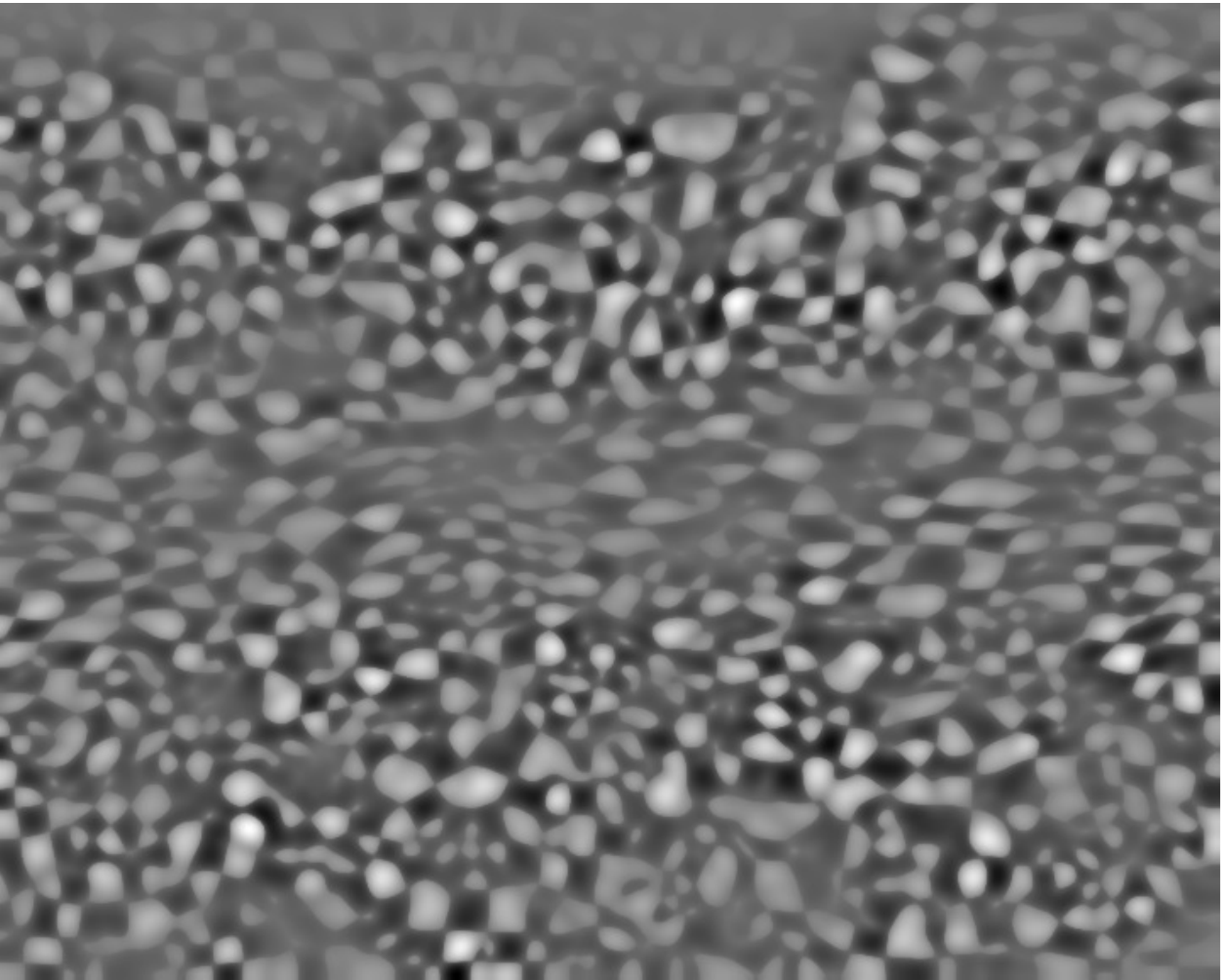} &
       \includegraphics[width=0.15\textwidth]{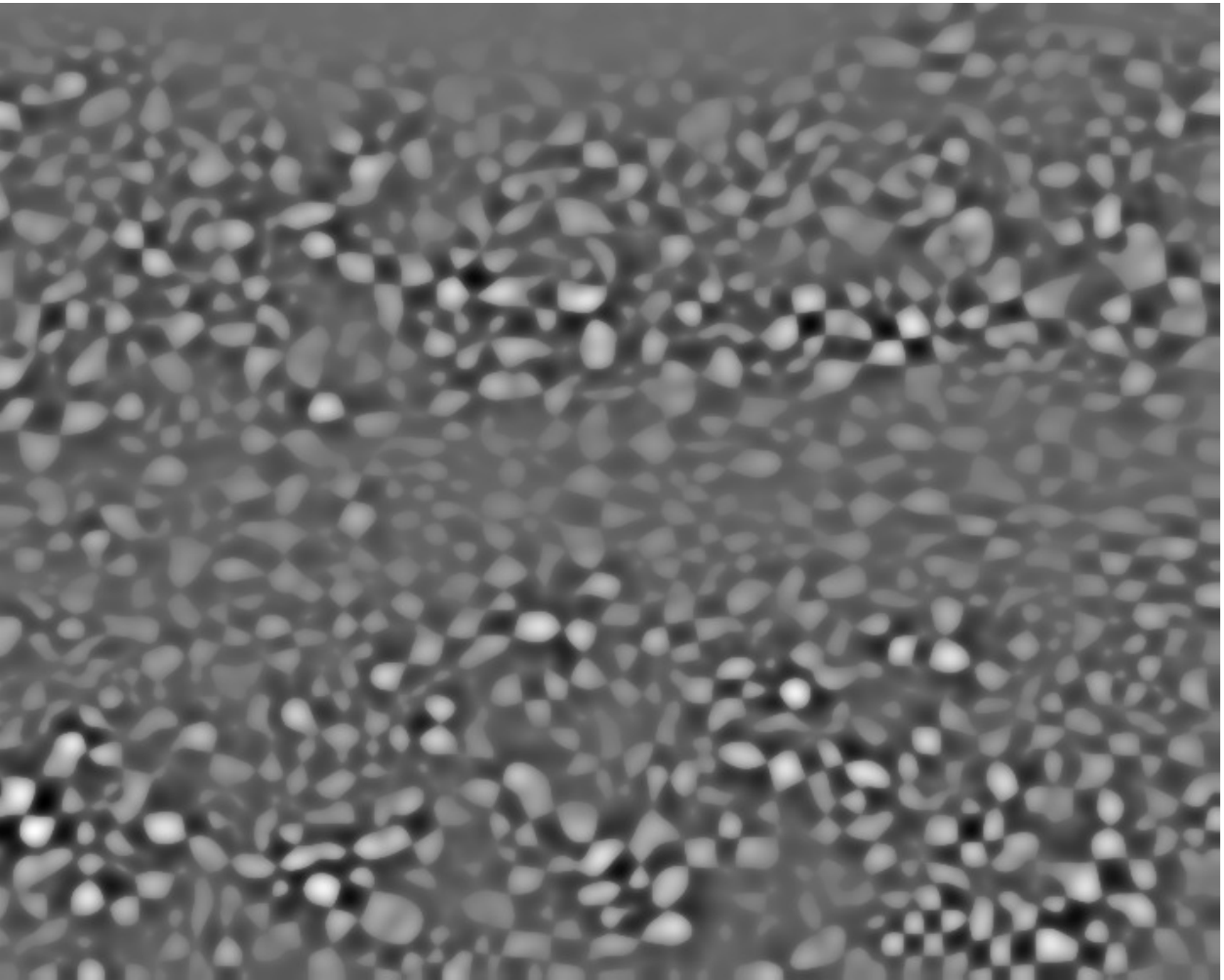}   \\
       
	\end{tabular}
	
   \end{center}
   
    \begin{center}
    \scriptsize
    \setlength\tabcolsep{0.5mm}
    \begin{tabular}{c c c}

       $ | \nabla L |  $ & $| \nabla L_t | $ & $| \nabla L_{tt} |$   \\
       \includegraphics[width=0.15\textwidth]{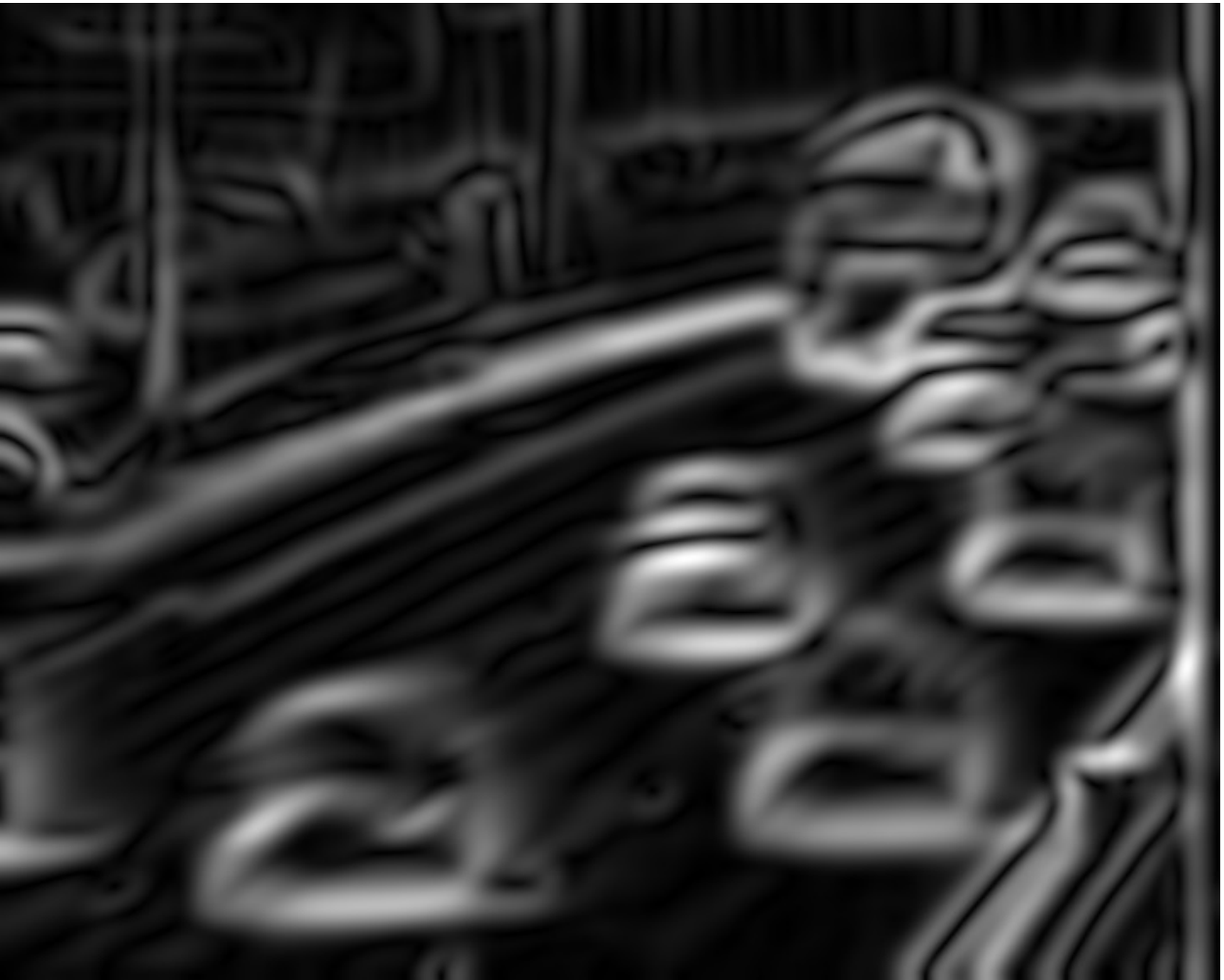} &
       \includegraphics[width=0.15\textwidth]{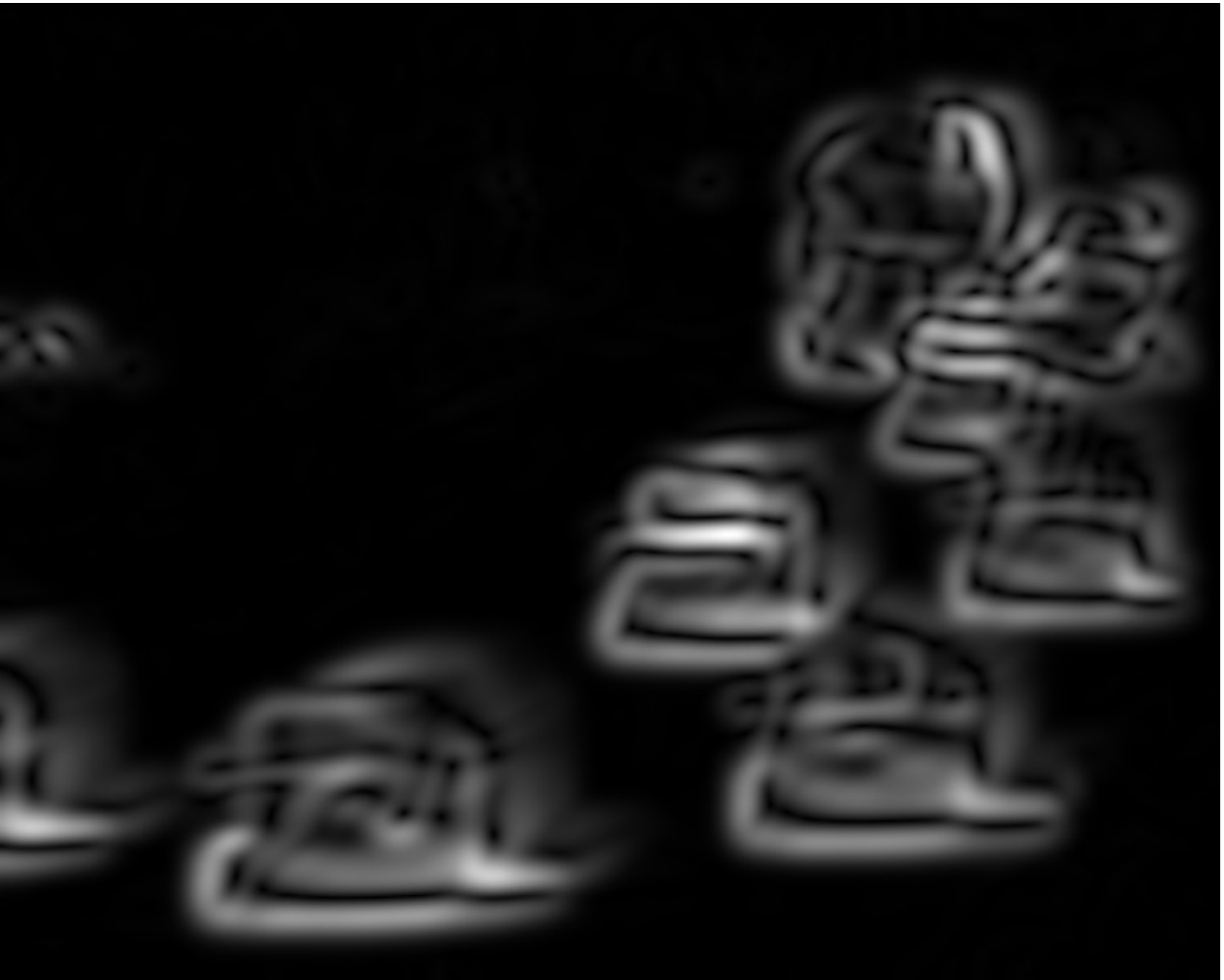} &
       \includegraphics[width=0.15\textwidth]{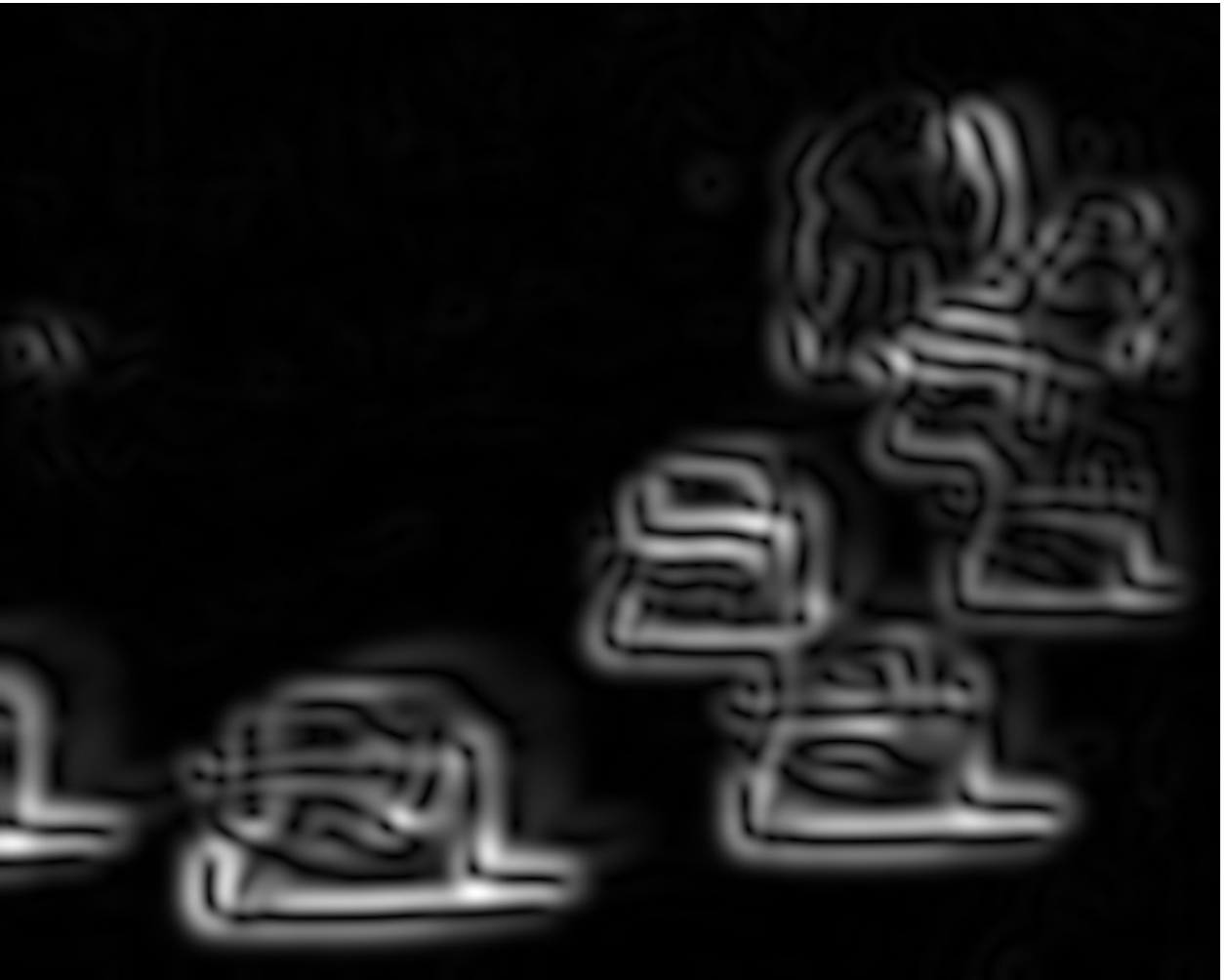}    \\       
       $ \nabla^2 L   $ & $ \nabla^2 L_t $ & $ \nabla^2 L_{tt} $   \\
       \includegraphics[width=0.15\textwidth]{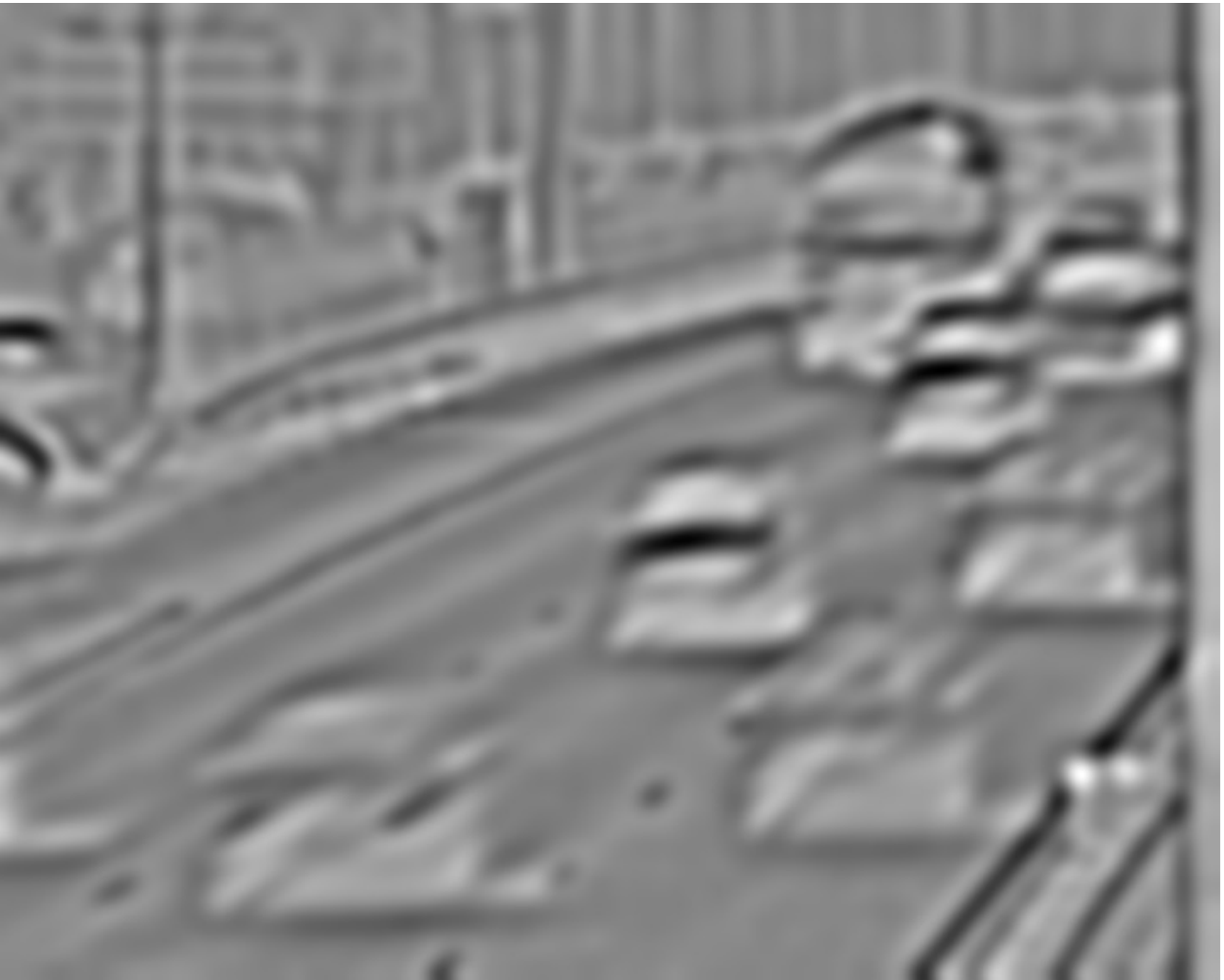} &
       \includegraphics[width=0.15\textwidth]{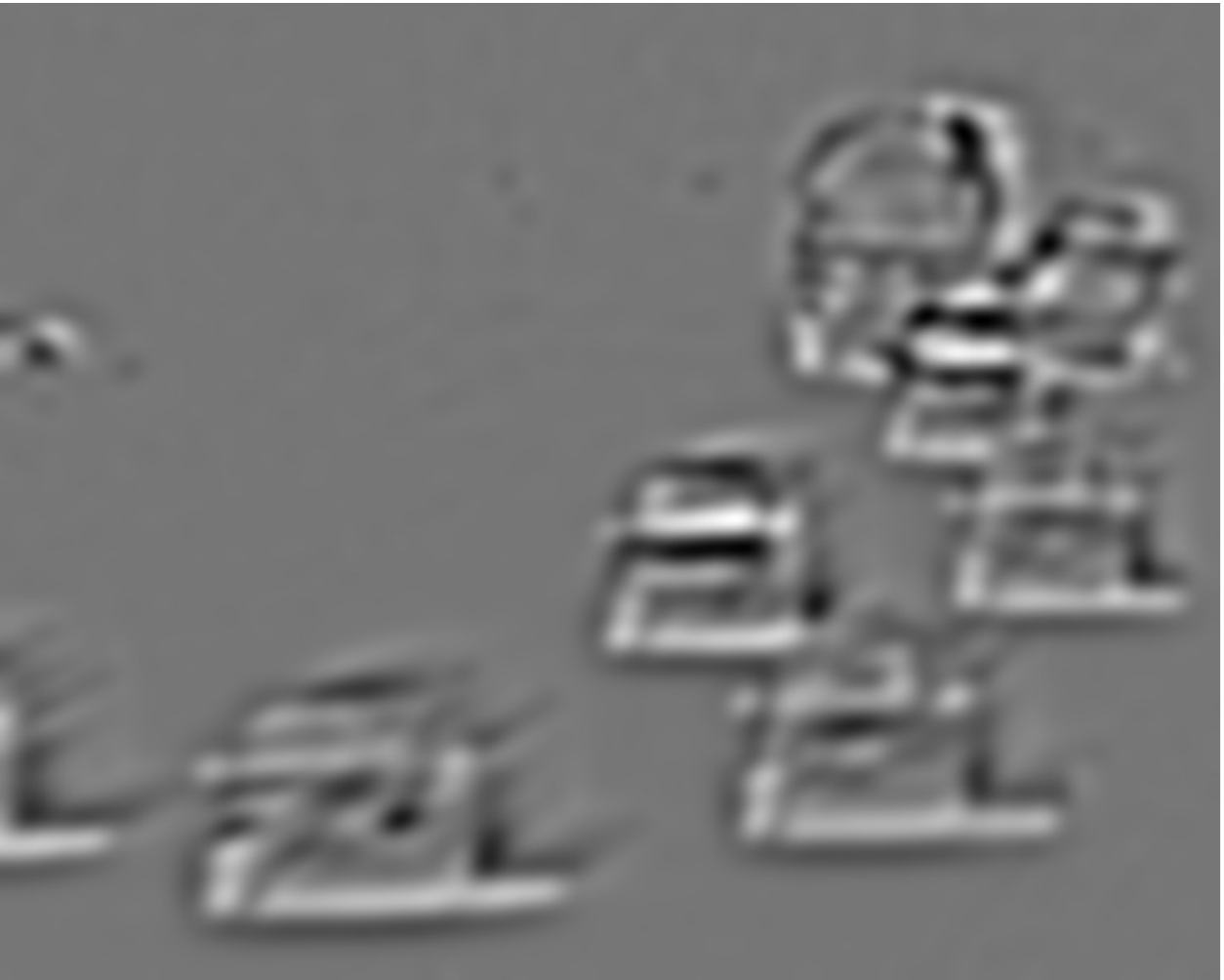} &
       \includegraphics[width=0.15\textwidth]{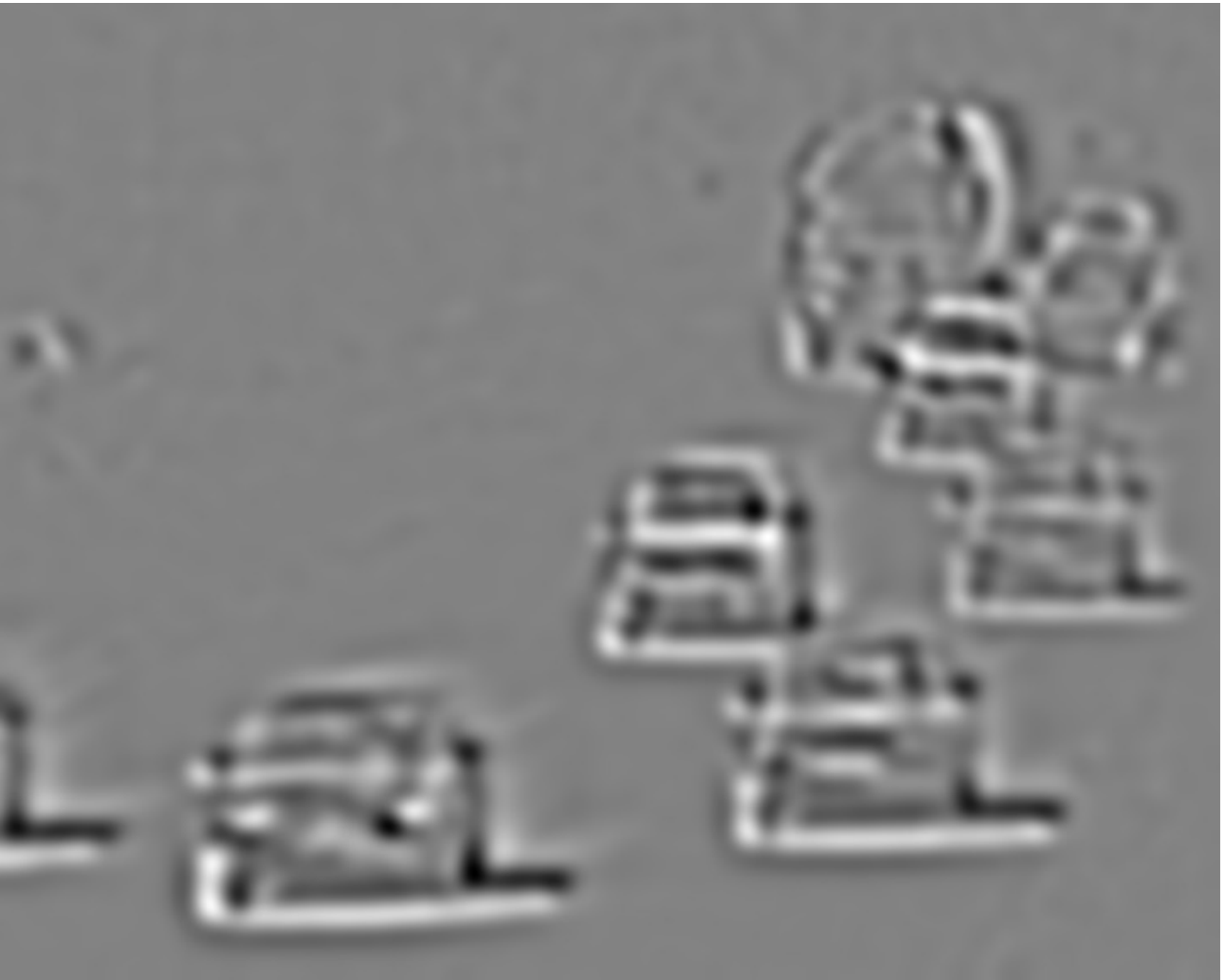}     \\

       $ \det \mathcal{H} L $ & $ \det \mathcal{H} L_t $ & $\det \mathcal{H} L_{tt} $  \\
       \includegraphics[width=0.15\textwidth]{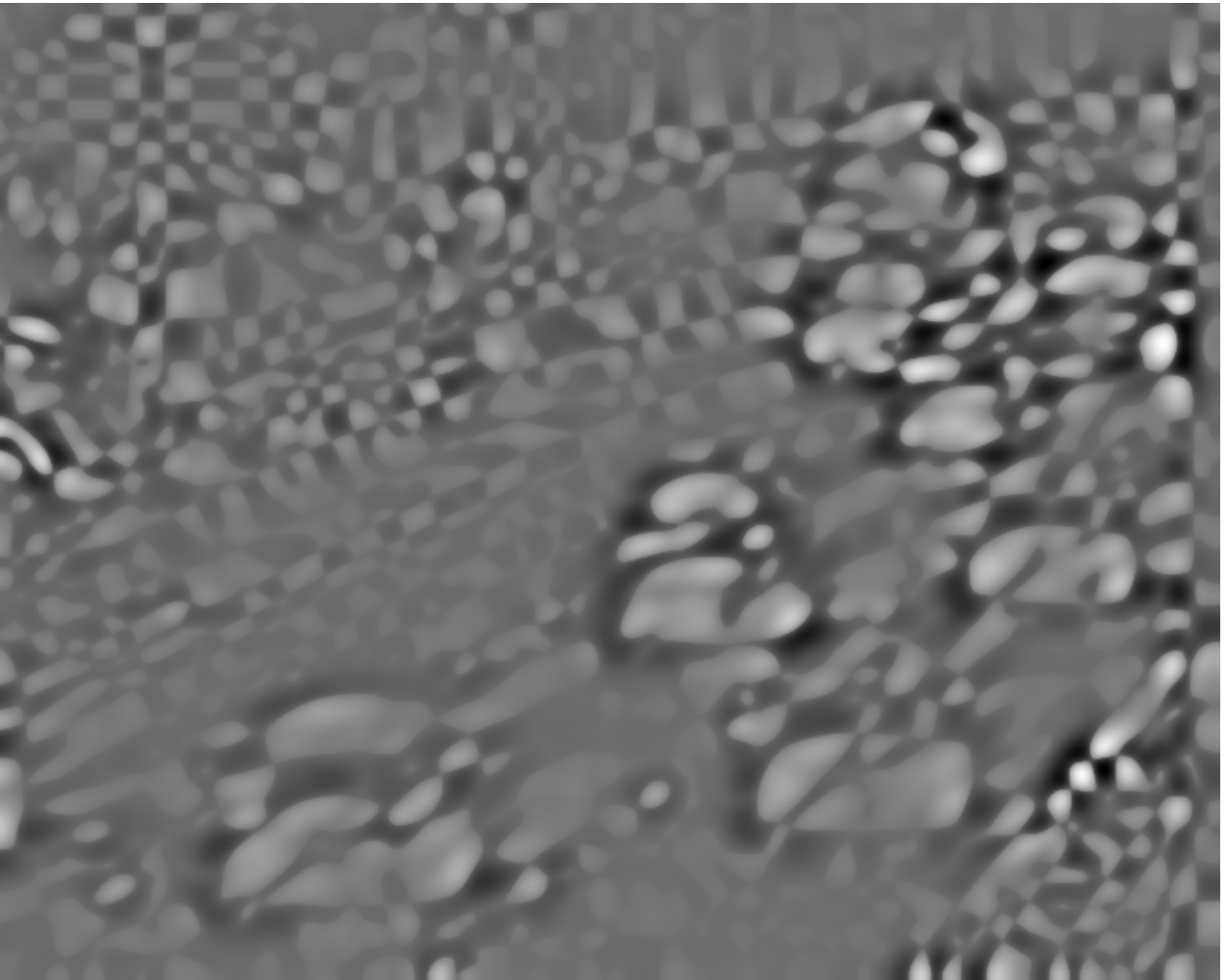} &
       \includegraphics[width=0.15\textwidth]{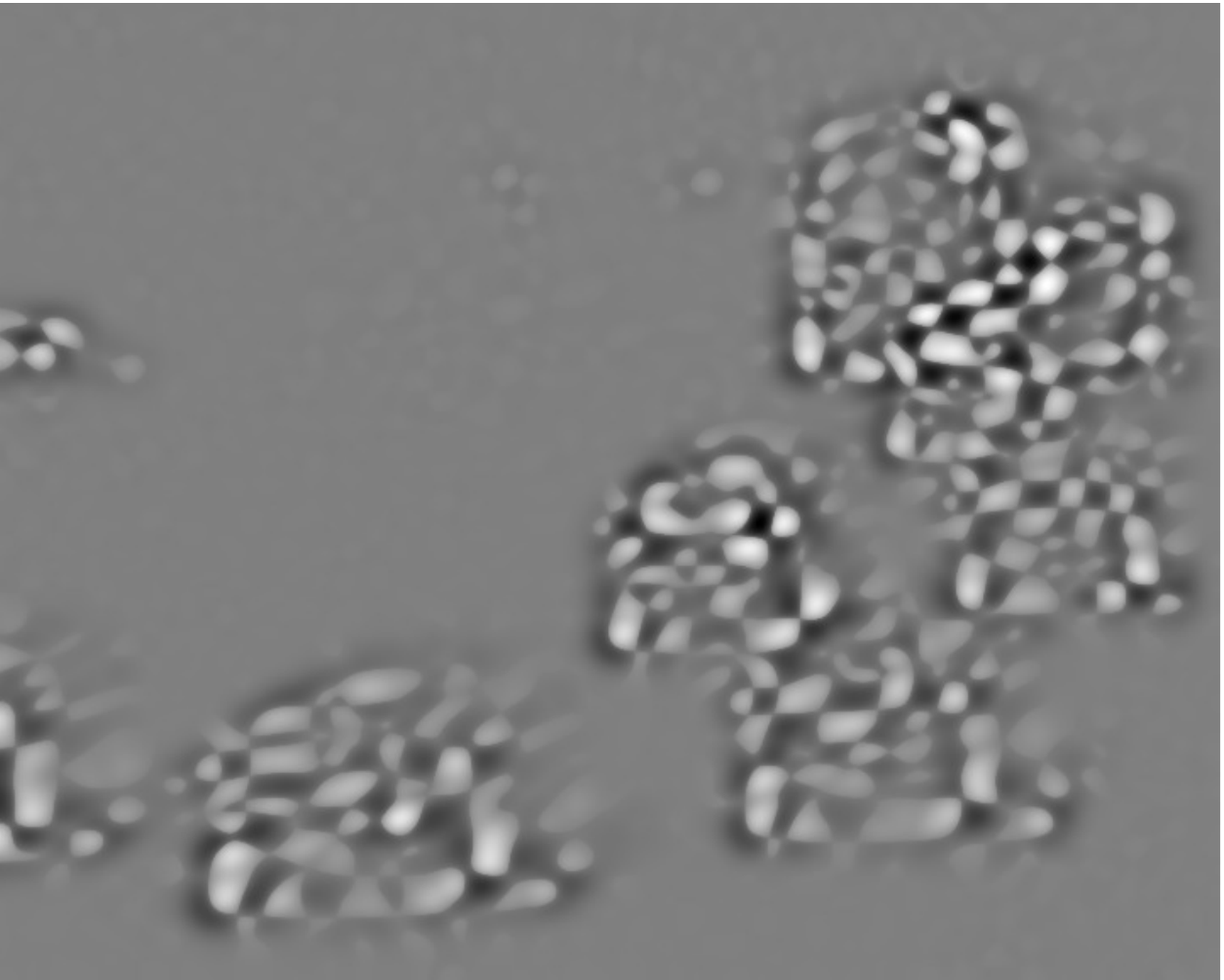} &
       \includegraphics[width=0.15\textwidth]{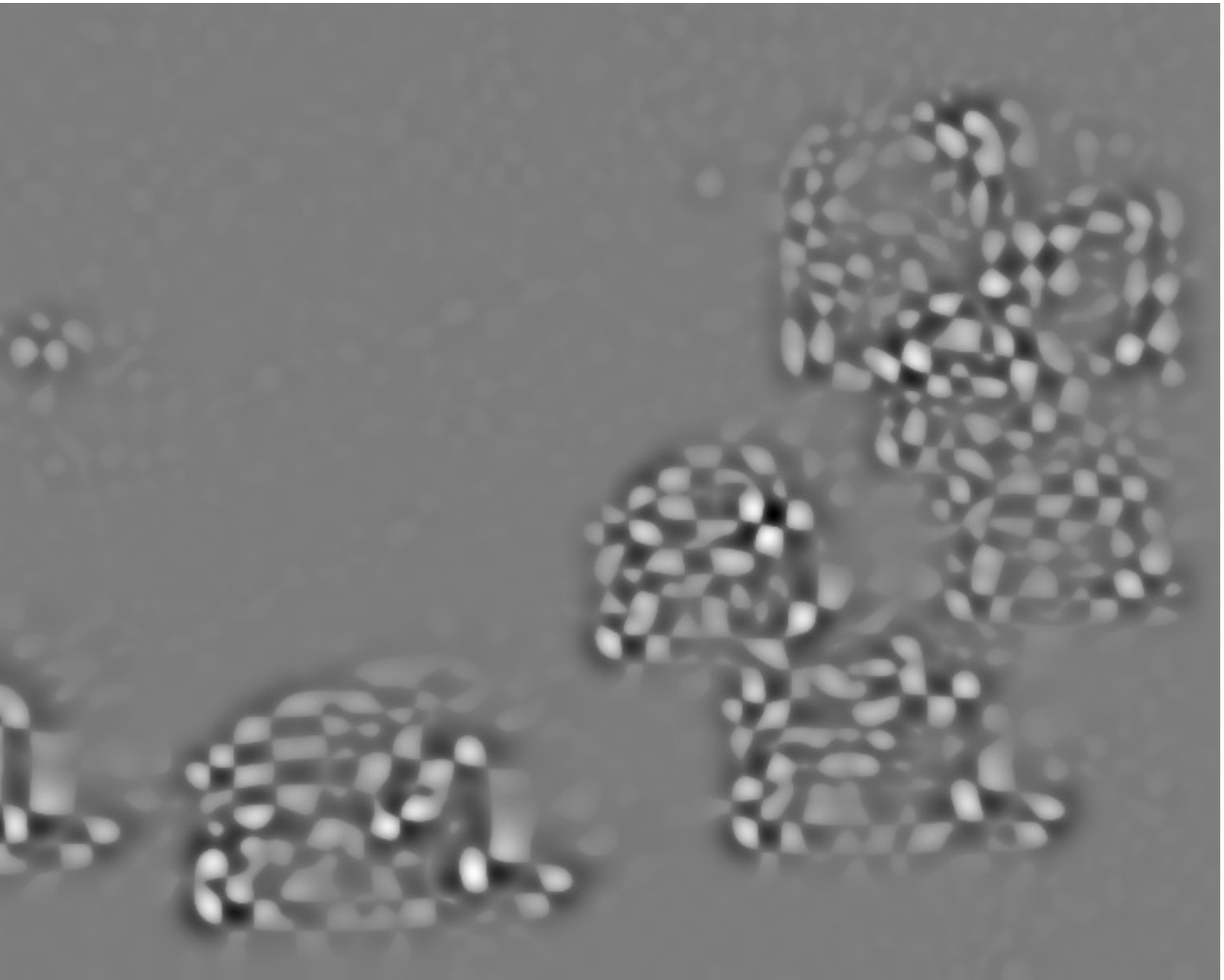}   \\
       
	\end{tabular}
	
   \end{center}
   \endgroup
   \caption{\emph{Rotationally invariant spatio-temporal receptive field responses} for the rotationally invariant differential invariants underlying the STRF RotInv video descriptor. Snapshots of receptive field responses are shown for two dynamic textures from the DynTex classes "waves" (top) and "traffic" (bottom) ($\sigma_s = 8$, $\sigma_\tau = 100$).}
\label{fig:strf-rotinv-responses}
\end{figure}

\subsubsection{Receptive field sets}
\label{sec:rfsets}
The set of receptive fields used as primitives for constructing the histogram will determine the type of information that is represented in the video descriptor. A straightforward example of this is that using rotationally invariant differential invariants will imply a rotationally invariant video descriptor. A second example is that including or excluding purely temporal derivatives will enable or disable capturing temporal intensity changes not mediated by spatial motion. We have chosen to compare video descriptors based on four different receptive field groups as summarised in Table~\ref{tab:rfsets}. 

\begin{table}[h!]
\caption{The video descriptors investigated in this paper and the receptive field sets they are based on.}
\begin{tabular}{lll}
\hline\noalign{\smallskip}
Name &  Receptive field set \\
\noalign{\smallskip}\hline\noalign{\smallskip}
\rule{0pt}{3ex}  RF Spatial & $\{ L_x, L_y, L_{xx}, L_{xy}, L_{yy} \}$ \\

\rule{0pt}{3ex}  STRF $N$-jet &  $ \{ L_x, L_y, L_{xx}, L_{xy}, L_{yy} \} , \{L_t, L_{tt} \} $, \\
& $\{ L_{xt}, L_{yt}, L_{xxt}, L_{xyt}, L_{yyt}\}, $ \\
& $\{ L_{xtt}, L_{ytt}, L_{xxtt}, L_{xytt}, L_{yytt}  \} $ \\

\rule{0pt}{3ex}  STRF RotInv  & $\{ |\nabla L|, |\nabla L_t|, |\nabla L_{tt}| \}$, \\
& $\{ \nabla^2 L, \nabla^2 L_t, \nabla^2 L_{tt} \}$, \\
& $\{ \det \mathcal{H} L, \det  \mathcal{H} L_t, \det  \mathcal{H} L_{tt}  \}$ \\

 \rule{0pt}{3ex}  STRF $N$-jet (previous \cite{JanLin-SSVM2017}) & $ \{ L_x, L_y, L_{xx}, L_{xy}, L_{yy} \} , \{L_t, L_{tt}\}, $ \\
 & $ \{ L_{xt}, L_{yt}, L_{xxt}, L_{xyt}, L_{yyt}\}$  \\

\noalign{\smallskip}\hline
\end{tabular}
\label{tab:rfsets}       
\end{table}
First, note that all video descriptors, except STRF N-jet (previous), include first- and second-order spatial and
temporal derivatives \emph{in pairs} $\{L_x, L_{xx}\}$, $\{L_t, L_{tt}\}$, $\{L_{xt}, L_{xtt}\}$, $\{ L_{xxt}, L_{xxtt}\}$ etc. The motivation for this is that first- and second-order derivatives provide complementary information and by including both, equal weight is put on first- and second-order
information. It has specifically been observed that biological receptive fields occur in pairs of odd-shaped and even-shaped receptive field profiles that can be well approximated by Gaussian derivatives (Koenderink and van Doorn \cite{KoeDoo87-BC}, De~Valois et al.\ \cite{ValCotMahElfWil00-VR}, Lindeberg \cite{Lin13-BICY}). In the following, we describe the four video descriptors in more detail and do further motivate the choice of their respective receptive field sets.

\paragraph{\textbf{RF Spatial}} is a purely spatial descriptor based on the full spatial $N$-jet up to order two. This descriptor will capture the spatial patterns in ``snapshots" of the scene (single frames) independent of the presence of movement. Using spatial derivatives up to order two means that each histogram cell template will represent a discretized second-order approximation of the local spatial image structure. An additional motivation for using this receptive field set is that this descriptor is one of the best performing spatial descriptors for the receptive field based object recognition method in \cite{LinLin12-CVIU}. This descriptor is primarily included as a baseline to compare the spatio-temporal descriptors against.

\paragraph{\textbf{STRF $N$-jet}} is a directionally selective spatio-temporal descriptor, where the first- and second-order spatial derivatives
are complemented with the first- and second-order temporal derivatives of these as well as the first- and second-order temporal
derivatives of the smoothed video $L$. Including purely temporal derivatives means that the descriptor can capture intensity
changes not mediated by spatial motion (flicker). 
The set of mixed spatio-temporal derivatives will on the other hand capture the interplay between changes over the spatial
and temporal domains, such as movements of salient spatial patterns. An additional motivation for including mixed spatio-temporal derivatives is that they represent features that are well localised with respect to joint spatio-temporal scales. This implies that when using multiple scales, a descriptor including mixed spatio-temporal derivatives will have better ability to separate spatio-temporal patterns at different spatio-temporal scales. 

\paragraph{\textbf{STRF RotInv}} is a rotationally invariant video descriptor based on a set of rotationally invariant features over the spatial domain: the spatial gradient magnitude $| \nabla L | = \sqrt{L_x^2 + L_y^2}$, the spatial Laplacian $\nabla^2 L = L_{xx} + L_{yy}$ and the determinant of the spatial Hessian $\det \mathcal{H} L = L_{xx}L_{yy} - L_{xy}^2$ 
\footnote{To transform the determinant of the spatial Hessian having the same dimensionality in terms of $[\mbox{intensity}]$ 
as the other spatial differential invariants, we transform the magnitude by a square root function while preserving its sign: 
$  (\det\mathcal{H} L)_{transf}  = \sign (\det \mathcal{H} L) \, | \det \mathcal{H} L |^{1/2} $.}. 
These are evaluated on the smoothed video $L$ directly and on the first- and second-order
temporal derivatives of the scale-space representation $L_t$  and $L_{tt}$. One motivation for choosing these spatial differential invariants is that
they are functionally independent and span the space of rotationally invariant first- and second-order differential invariants
over the spatial domain. This set of rotationally invariant features was also demonstrated to be the basis of one of
the best performing spatial  descriptors in \cite{LinLin12-CVIU}. 
By applying these differential operators to the first- and second-order temporal derivatives of the video, the interplay between temporal and spatial intensity variations is captured. 

\paragraph{\textbf{STRF $\mathbf{N}$-jet  (previous)}} is our previously published \cite{JanLin-SSVM2017} video descriptor. This descriptor is included for reference and differs from STRF $N$-jet by lacking the second-order temporal derivatives of the first- and second-order spatial derivatives. This descriptor was evaluated in \cite{JanLin-SSVM2017} without parameter tuning and we also here here keep the original parameters. 

\paragraph{}
~~~~It can be noted that none of these video descriptors makes use of the full spatio-temporal $4$-jet. This reflects the philosophy of treating space and time as distinct dimensions, where the most relevant information lies in the interplay between
spatial changes (here, of  first- and second-order) with temporal changes (here, of first- and second-order). Third- and fourth-order information with respect to either the spatial or the temporal domain is thus discarded. Receptive field responses for two videos of dynamic textures are shown for spatio-temporal partial derivatives in Figure~\ref{fig:strf-responses} and for rotational differential invariants in Figure~\ref{fig:strf-rotinv-responses}. 

It should be noted, that the recognition framework presented here also allows for using non-separable receptive fields with non-zero image velocities. Exploring this is, however, left for future work and in this study we instead focus on evaluating different sets of space-time separable receptive fields (see also the discussion in Section \ref{sec:spat-temp-RF}).

\subsubsection{Number of bins and principal components}
Different choices of the number of bins per feature dimension $n_{bins}$ and the number of principal components $n_{comp}$ will give rise to qualitatively different histogram descriptors. Using few principal components in combination with many bins will enable fine-grained recognition of a smaller number of similar pattern templates (separating patterns based on smaller magnitude differences in receptive field responses). On the other hand, using a larger number of principal components in combination with fewer bins will imply a histogram capturing a larger number of more varied but less "precise" patterns.
The different options that we have considered in this work are:
\begin{flalign*}
&n_{comp} \in [2,17] \\ 
&n_{bins} \in \{ 2,3,4,5,6,7,8,9,10,12,15,20,25\}  
\end{flalign*}
After a set of initial experiments, where we varied the number of bins and the number of principal components (presented in Section~\ref{sec:expm-binscomp}), we noted that binary histograms with 10-17 principal components achieve highly competitive results for all benchmarks. Binary histogram descriptors also have an appeal in simplicity and one less parameter to tune. Therefore, the subsequent experiments (Section~\ref{sec:expm-scales} and forward), were performed using binary histograms only. 

\subsubsection{Binary histograms}
When choosing $n_{bins}=2$ equivalent to a joint binary histogram, the local
image structure is described by only the sign of the different image
measurements. This will make the descriptor
invariant to uniform rescalings of the intensity values, such as multiplicative illumination transformations or indeed any 
change that does not affect the sign of the receptive field response. Binary
histograms in addition enable combining a larger number of image measurements without 
a prohibitive large descriptor dimensionality and have proven an effective approach 
by a large number of LBP-inspired methods.

\subsubsection{Spatial and temporal scales}
\label{sec:scales}
Using receptive fields at multiple spatial and temporal scales in the descriptor makes it possible to capture image structures of different spatial extent and temporal duration. Such a multi-scale descriptor will also comprise more complex image primitives, since the corresponding local space-time templates will represent combinations of patterns at different scales. 
The spatial and temporal scales, i.e. the standard deviation for the respective scale-space kernels, considered in this work are:
\begin{flalign*}
&\sigma_s \in \{1,2,4,8,16\}~\text{pixels}  \\
 &\sigma_\tau \in \{50,100,200,400\}~\text{ms}
\end{flalign*}
The lowest temporal scale 50 ms has been chosen to be of the same order as the time difference between adjacent frames for regular video frame rates around 25 fps. This temporal scale is also of the same order as the temporal scales of spatio-temporal receptive fields observed in the lateral geniculate nucleus (LGN) and the primary visual cortex (V1), where examples of receptive fields have been well modelled using time constants over the range 40-80 ms \cite[Figs. 3-4]{Lin16-JMIV}. 

In this study, we use receptive field responses at either a single spatial and temporal scale, or for a combination of pairs of adjacent spatial and temporal scales (2~x~2 scales): 
\begin{flalign*}
 &(\sigma_{s_1}, \sigma_{s_2})  \in \{(1,2),(2,4),(4,8),(8,16)\}~\text{pixels}  \\
  &(\sigma_{\tau_1}, \sigma_{\tau_2}) \in \{(50,100),(100,200),(200,400)\}~\text{ms}
  \end{flalign*}
The reason why we do not include combinations of more than two temporal scales and two spatial scales, is that an initial prestudy did not show any significant improvements from this on the benchmark datasets. The choice to use combinations of \emph{adjacent scales} is done mainly to limit the complexity of the study.
Thus, here, 20 combinations of a single spatial scale with a single temporal scale and 12 combinations of two spatial scales and two temporal scales are considered. 

A more general approach than using fixed pairs of adjacent scales is to operate on a wider range of spatio-temporal scales in parallel. For example, for breaking water waves that roll onto a beach, the coarser scale receptive fields will respond to the gross motion pattern of the water waves, whereas the finer scale receptive fields will respond to the detailed fine scale motion pattern of the water surface. A general purpose vision system should have the ability to dynamically operate over such different subsets of spatial and temporal scales, to extract maximum amount of relevant information about a dynamic scene. Specifically, there is interesting potential in determining local spatial and temporal scale levels adaptively from the video data, using recently developed methods for spatio-temporal scale selection \cite{Lin18-JMIV,Lin18-SIIMS}. We leave such extensions for future work.

\section{Datasets }
\label{sec:datasets}

We evaluate our proposed approach on six standard dynamic texture
recognition/classification benchmarks from two widely used dynamic texture
datasets: UCLA (Soatto et al.\ \cite{SoaDorWu-ICCV2001}) and DynTex (P{\'e}teri et al.\ \cite{PetRenFaz-PRL2010}). We here give
a brief description of the datasets and the benchmarks. Sample frames from the datasets are
shown in Figure \ref{fig:dataset-ucla} (UCLA) and Figure~\ref{fig:dataset-dyntex} (DynTex). 

\subsection{UCLA} 
The UCLA dataset was introduced by Soatto
et~al. \cite{SoaDorWu-ICCV2001} and is composed of 200 videos (160\,$\times$ 
110\,pixels, 15\,fps) featuring 50 different dynamic textures with 4
samples from each texture. The \textbf{UCLA50} benchmark \cite{SoaDorWu-ICCV2001} 
divides the 200 videos into 50 classes with one class per individual texture/scene. It should be noted that this
partitioning is not conceptual in the sense of the classes
constituting different types of textures such as ``fountains'', ``sea'' or ``flowers'' but instead targets
\emph{instance specific} and \emph{viewpoint specific} recognition. This means that not only different individual fountains 
but also the same fountain seen from two different viewpoints should be separated from each other.

Since for many applications it is more relevant to recognise different dynamic texture categories, a partitioning of the
UCLA dataset into \emph{conceptual classes}, \textbf{UCLA9}, 
was introduced by Ravichandran et~al. \cite{RavChaVid-CVPR2009}
with the following classes: boiling water\,(8),
fire\,(8), flowers\,(12), fountains\,(20), \\ plants\,(108), sea\,(12), smoke\,(4), water\,(12) and waterfall\,(16), where the numbers 
correspond to the number of samples from each class. Because of the large overrepresentation of plant videos for this benchmark, in the \textbf{UCLA8} benchmark, those are excluded to get a more balanced
dataset, leaving 92 videos from eight conceptual classes.

\subsection{DynTex}
A larger and more diverse dynamic texture dataset, \textbf{DynTex}, was introduced by
P{\'e}teri et~al. \cite{PetRenFaz-PRL2010}, featuring a larger variation of dynamic texture types recorded under more diverse conditions (720 $\times$  576 pixels, 25 fps). From this dataset,
three gradually larger and more challenging benchmarks have been compiled by Dubois et~al. \cite{DubSloetal-SIVP2015}. The \textbf{Alpha} benchmark includes 60 dynamic texture videos from three different
classes: sea, grass and trees. There are 20 examples of each
class and some variations in scale and viewpoint. The \textbf{Beta}
benchmark includes 162 dynamic texture videos from ten
classes: sea, vegetation, trees, flags, calm water, fountain, smoke,
escalator, traffic and rotation. There are 7 to 20 examples of each
class. The \textbf{Gamma} benchmark includes 264 dynamic texture videos from ten
classes: flowers, sea, trees without foliage,
dense foliage, escalator, calm water, flags, grass, traffic and fountains. There are 7 to 38 examples of
each class and this benchmark features the largest intraclass
variability in terms of scale, orientation, etc.

\section{Experimental setup }
\label{sec:expm-setup}
This section describes the cross-validation sche\-mes used for the different benchmarks,
the classifiers and the use of parameter tuning over the descriptor parameters. 

\subsection{Benchmark cross-validation schemes}
\label{sec:expm-setup-benchmark}
The standard test setup for the UCLA50 benchmark, which we adopt also here, is 4-fold cross-validation \cite{SoaDorWu-ICCV2001}. For each partitioning, three out of four samples from each dynamic texture instance are used for training, while the remaining one is held out for testing.  

The standard test setup for the UCLA8 and UCLA9 benchmarks is to report the average accuracy over 20 random partitions, with 50~\% data used for training and 50~\% for testing (randomly bisecting each class) \cite{GhaAhu-ECCV2010}. We use the same setup here, except that we report results as an average over 1000 trials to get more reliable statistics. This since we noted that, because of the small size of the dataset, the specific random partitioning will otherwise affect the result. For all the UCLA benchmarks, in contrast to the most common setup of using manually extracted patches, we use the \emph{non-cropped videos}, thus our setup could be considered a slightly harder problem. 

For the DynTex benchmarks, the experimental setup used is leave-one-out cross-validation as in \cite{HonRyuetal-MSSP2016,AraKit-TOM2014,YanXiaetal-NC2016,QiLietal-NC2016}. We perform no subsampling of videos but use the full 720 $\times$  576 pixels frames.

\subsection{Classifiers}
\label{sec:expm-setup-classifiers}
We present results of both using a support vector machine (SVM) classifier and a nearest neighbour (NN) classifier, the latter to evaluate the performance also of a simpler classifier without hidden tunable parameters. For NN we use the $\chi^2$-distance $d(x,y)= \sum_{i} (x_i-y_i)^2/ (x_i+y_i)$ to compute the distance between two histogram video descriptors and for SVM we use the $\chi^2$-kernel $e^{-\gamma d(x,y)}$. Results are quite robust to the choice of the SVM hyperparameters $\gamma$ and $C$. We here use $\gamma = 0.1$ and $C = 10,000$ for all experiments.

\subsection{Descriptor parameter tuning}
\label{sec:expm-ptuning}
Comparisons with state-of-the-art and between video descriptors based on different sets of receptive fields are made using binary descriptors. Parameter tuning is performed as a grid search over the number of principal components $n_{comp} \in [2,17]$,  spatial scales $\sigma_s \in \{1, 2, 4, 8, 16\}$ and temporal scales $ \sigma_\tau \in \{50,100, 200, 400\}$. For spatial and temporal scales, we consider both single scales and combinations of two adjacent spatial and temporal scales. The standard evaluation protocols for the respective benchmarks (i.e. slightly different cross validation schemes) are used for parameter selection and results are reported for each video descriptor using the optimal set of parameters.

This is the standard method for parameter selection used on these benchmarks. The reason for following it also here, is to be able to make a direct comparison with the already published results in the literature using the same experimental protocol. It should be noted, however, that this evaluation protocol, which does not fully separate the test and the train data, does introduce the risk for model selection bias. A better evaluation scheme would be to use \emph{a full nested cross-validation scheme}, although this would be quite computationally expensive. To give transparence to the effect of parameter tuning on the performance of our descriptors, we therefore also present results from parameter tuning of the different descriptor parameters in (Sections~\ref{sec:expm-binscomp}- \ref{sec:expm-scales}) showing how much the performance is affected by choosing non-optimal parameter values.

\section{Experiments}
\label{sec:expm}
Our first experiments consider a qualitative and quantitative evaluation of different versions of our video descriptors, where we present results on: (i) varying the number of bins and principal components, (ii) using different spatial and temporal scales for the receptive fields and (iii) comparing descriptors based on different sets of receptive fields. 
This is followed by (iv) a comparison with state-of-the-art dynamic texture recognition methods 
and finally (v) a qualitative analysis on reasons for errors.

\subsection{Number of bins and principal components}
\label{sec:expm-binscomp}
\begin{figure*}[hbpt]
  \begin{center}
      \begin{tabular}{cc}

       \includegraphics[width=0.5\textwidth]{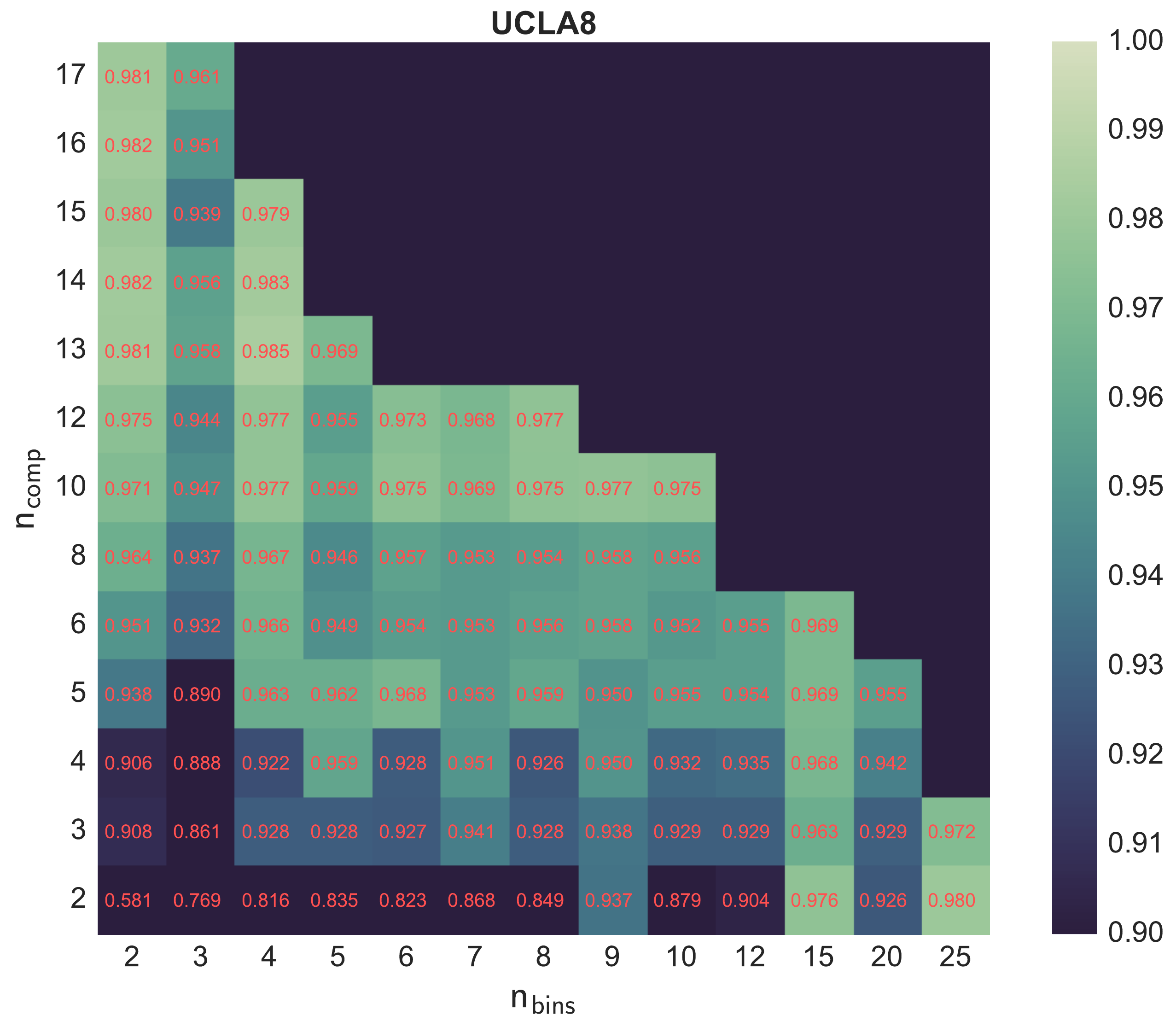} &
       \includegraphics[width=0.5\textwidth]{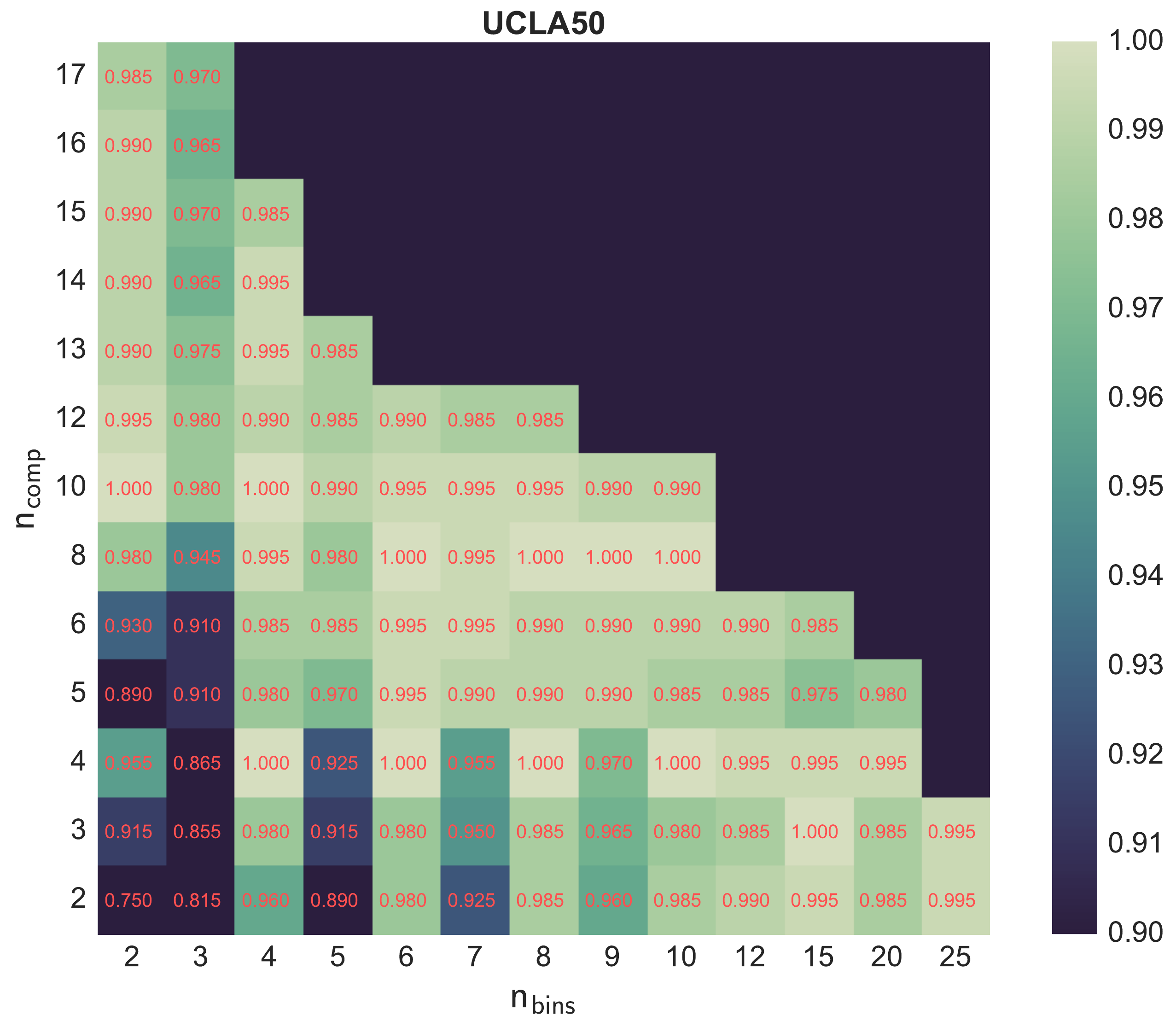}
     
           \end{tabular}
  \end{center}
  \caption{The classification performance of the STRF $N$-jet descriptor as function of the number of bins and the number of principal components for the UCLA8 and UCLA50 benchmarks. For the UCLA8 benchmark, the best results are obtained when using more than eight principal components in combination with binary or ternary histograms. For the UCLA50 benchmark, a large range of different configurations result in 0 \% error indicating that the task of recognising dynamic texture instances is less challenging compared to separating dynamic texture categories. Spatial scales: $(\sigma_{s_1}, \sigma_{s_2}) = (1, 2) \text{~pixels}$. Temporal scales: $(\sigma_{\tau_1}, \sigma_{\tau_2}) = (50, 100)$ ms. }
\label{fig:ucla-binscomp}

   \begin{center}
      \begin{tabular}{cc}

       \includegraphics[width=0.5\textwidth]{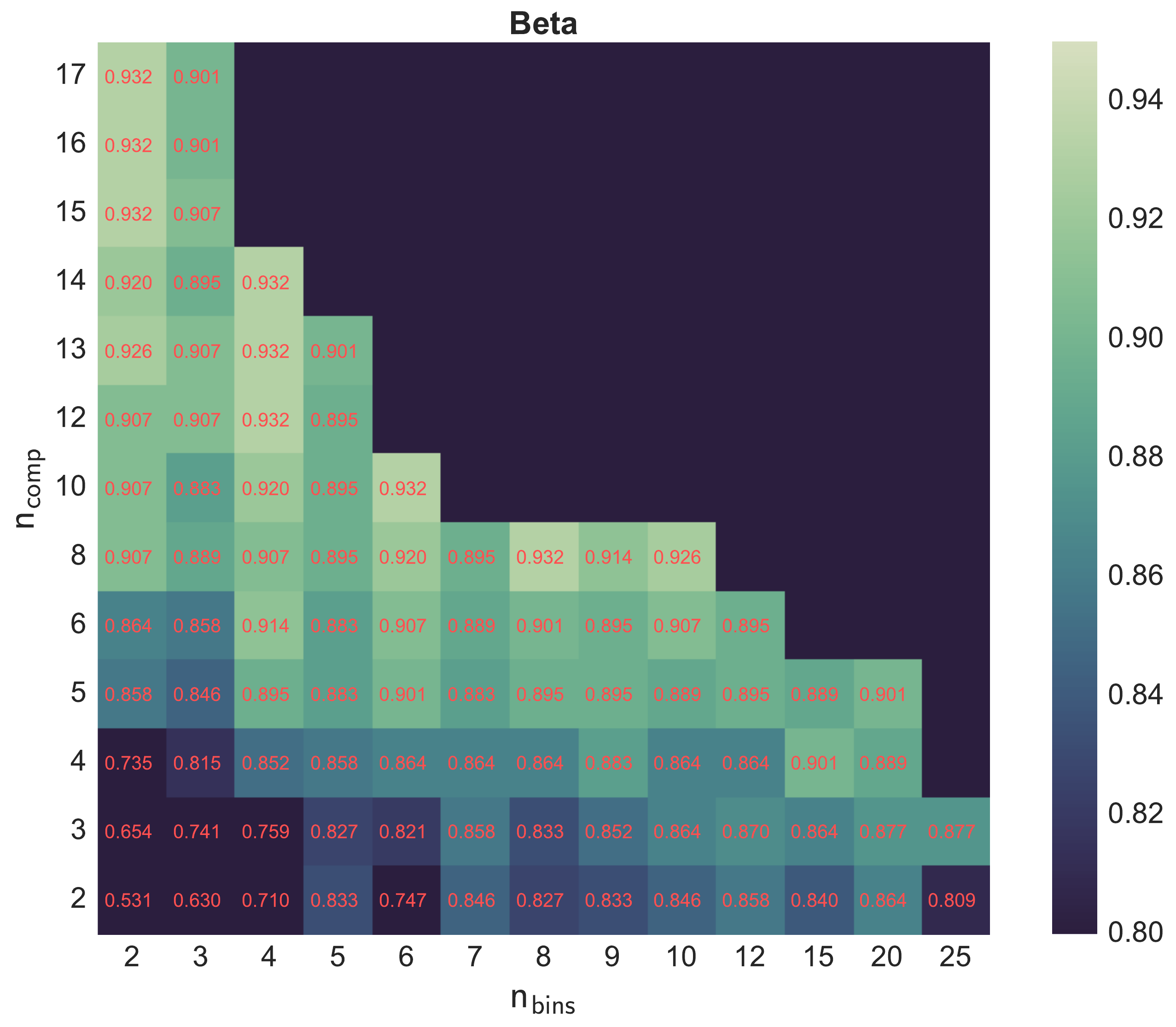} &
       \includegraphics[width=0.5\textwidth]{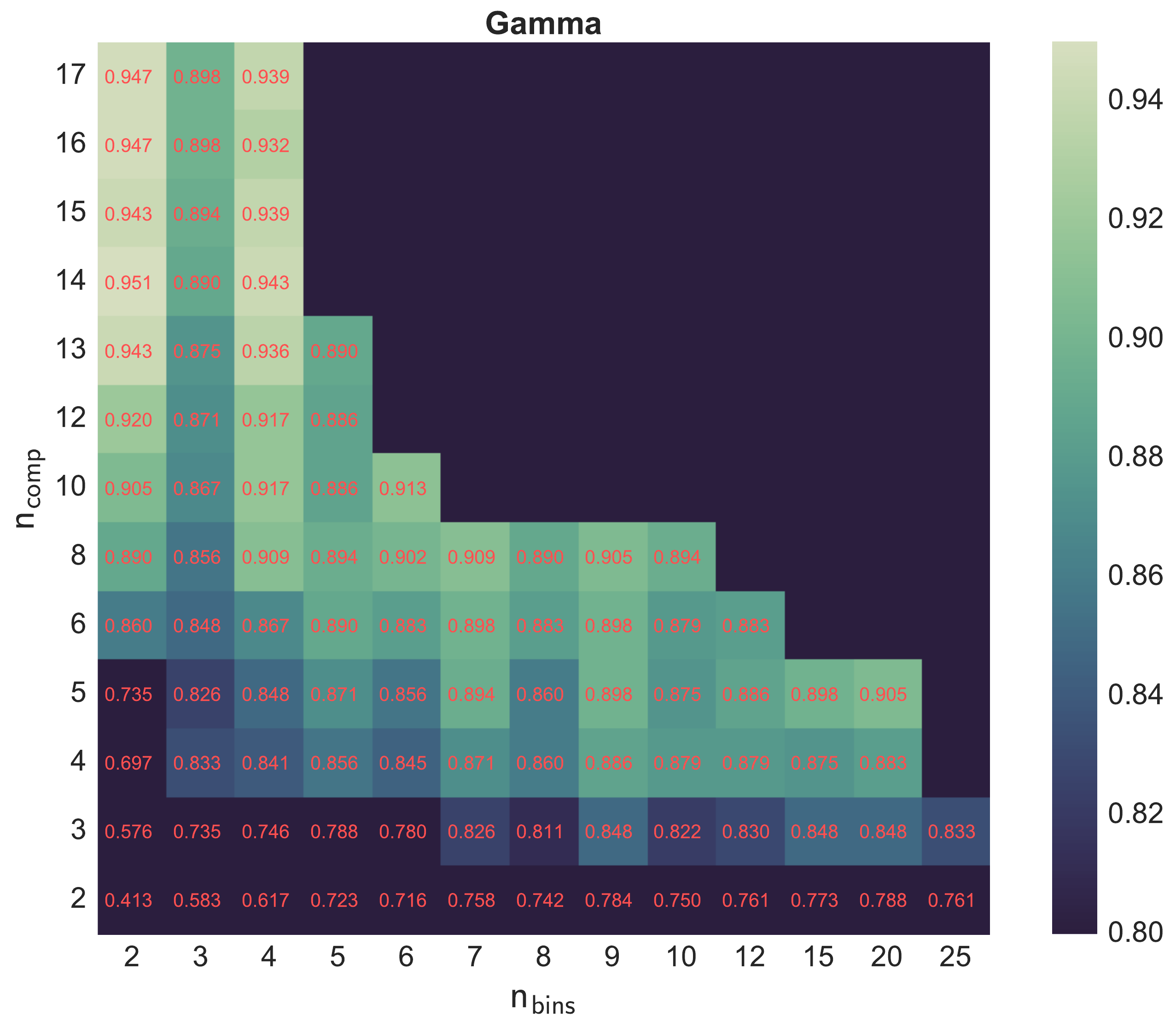}
     
           \end{tabular}
  \end{center}
  \caption{The classification performance of the STRF $N$-jet descriptor as function of the number of bins and the number of principal components for the Beta and Gamma benchmarks. The best results are obtained when using more than eight principal components in combination with binary or ternary histograms. Thus, for these more complex benchmarks, discarding too much information in the dimensionality reduction step impairs the performance. Spatial scales: $(\sigma_{s_1}, \sigma_{s_2}) = (4, 8) \text{~pixels}$. Temporal scales: $(\sigma_{\tau_1}, \sigma_{\tau_2}) = (100, 200)$ ms. }
\label{fig:dyntex-binscomp}
\end{figure*}

The classification performance of the STRF $N$-jet descriptor as function of the number of bins and the number of principal components used in the histogram descriptor are presented in Figure~\ref{fig:ucla-binscomp} for the UCLA8 and  UCLA50 benchmarks and in Figure~\ref{fig:dyntex-binscomp} for the Beta and Gamma benchmarks. 
A first observation is that, not surprisingly, when using a smaller number of principal components, each dimension needs to be divided into a larger number of bins to achieve good performance, e.g. for $n_{comp} \in{\{2, 3\}}$ the best performance is achieved for $n_{bins} \ge 15$ for all benchmarks. To discriminate between a large number of spatio-temporal patterns using only a few image measurements, these need to be more precisely recorded. A qualitative difference between using an odd or an even number of bins for $n_{bins} \le 8$ can also be noted. This can be explained by a qualitative difference in the handling of feature values close to zero. 

At the other end of the spectrum, it can be seen that when using a large number of principal components, fewer bins suffice. Using a large number of spatio-temporal primitives in combination with a small number of bins means that the different qualitative ``types" of patters are more diverse, while at the same time being less ``precise" in the sense of being unaffected by small changes in the magnitude of the filter responses. Binary or ternary descriptors are thus less sensitive to variations of the same rough type of space-time structure. Indeed, for binary descriptors only the sign of the receptive field response is recorded and a binary descriptor thus gives full invariance to e.g. multiplicative illumination transformations. 

For the larger Beta and Gamma benchmarks, it is clear that the descriptors that in this way combine a large number of image measurements with binary or ternary histograms achieve superior performance. This indicates that for these larger more complex datasets, capturing the essence of a local space-time pattern rather than its more precise appearance is the right trade-off. In fact, here, the best results can for all benchmarks be achieved using $n_{bins} = 2 $ and $n_{comp} \in [10,17]$, with the single exception of 0.3 percentage points lower error on the UCLA8 dataset if instead using a ternary descriptor. A similar observation that binary histograms perform very well was made in previous work on spatial recognition (\cite{LinLin12-CVIU}). Indeed, binary histograms based on the sign of the receptive field responses are independent on magnitude thresholds and invariant to multiplicative intensity transformations, which provides better robustness to illumination transformations. 

We thus conclude that binary histogram descriptors are a very useful option, combining top performance with simplicity. Therefore, we in the following investigate the effect of varying the remaining descriptor parameters using binary descriptors only. 


\subsection{Spatial and temporal scales}
\label{sec:expm-scales}
\begin{figure*}[hpbt]
  \begin{center}
      \begin{tabular}{ccc}

       \includegraphics[width=0.3\textwidth]{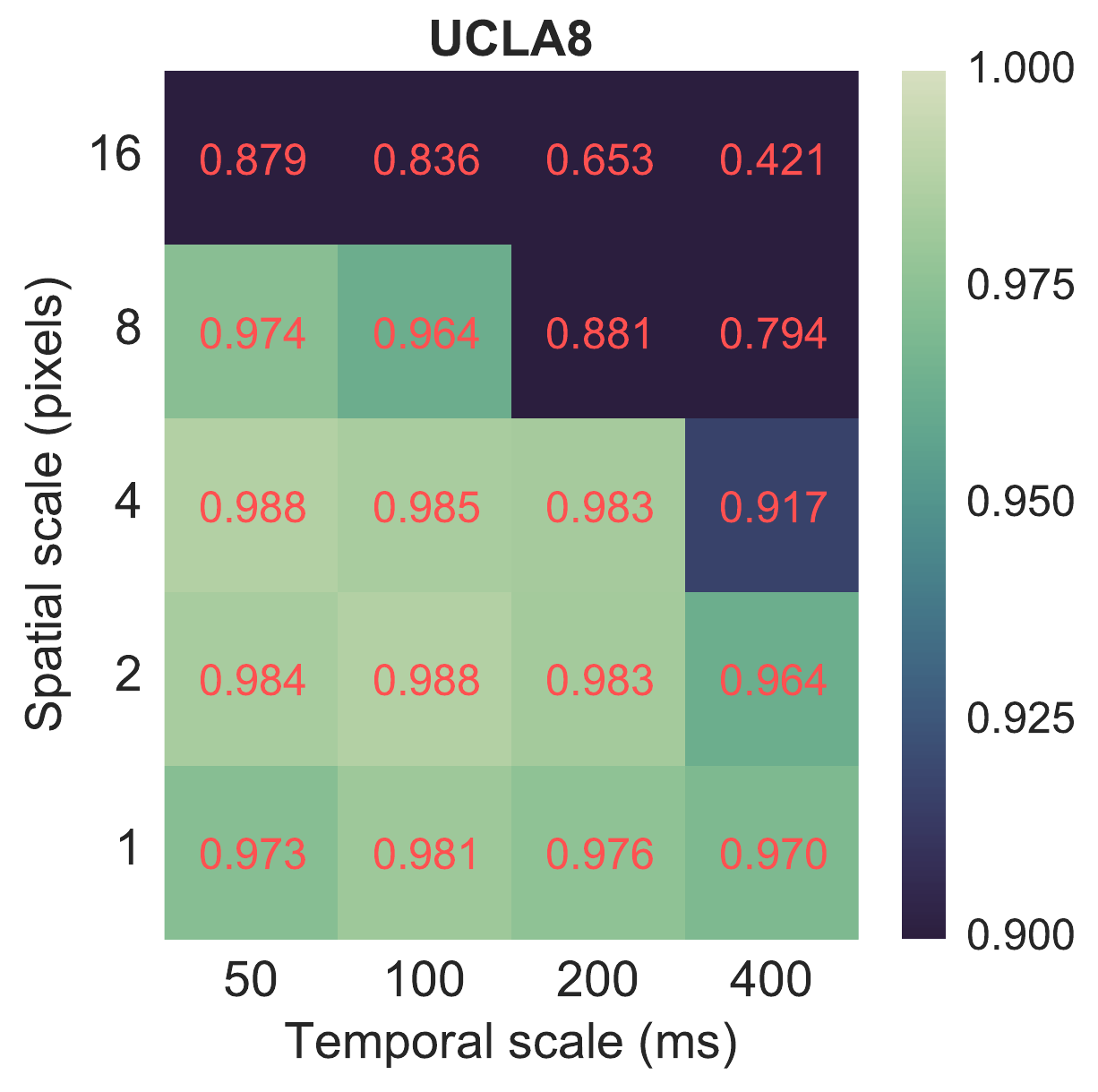} 
       &        \includegraphics[width=0.3\textwidth]{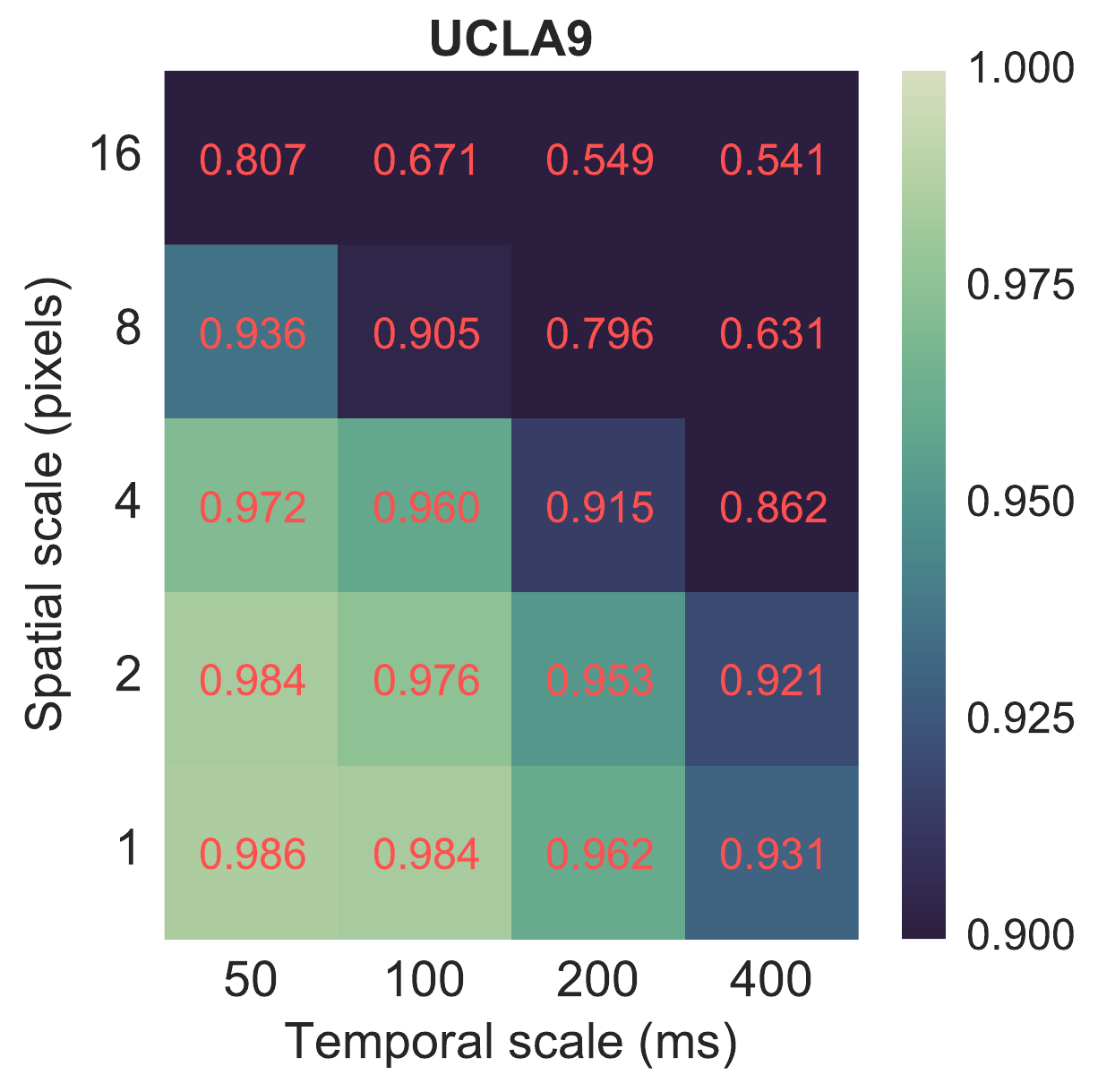} 
       &        \includegraphics[width=0.3\textwidth]{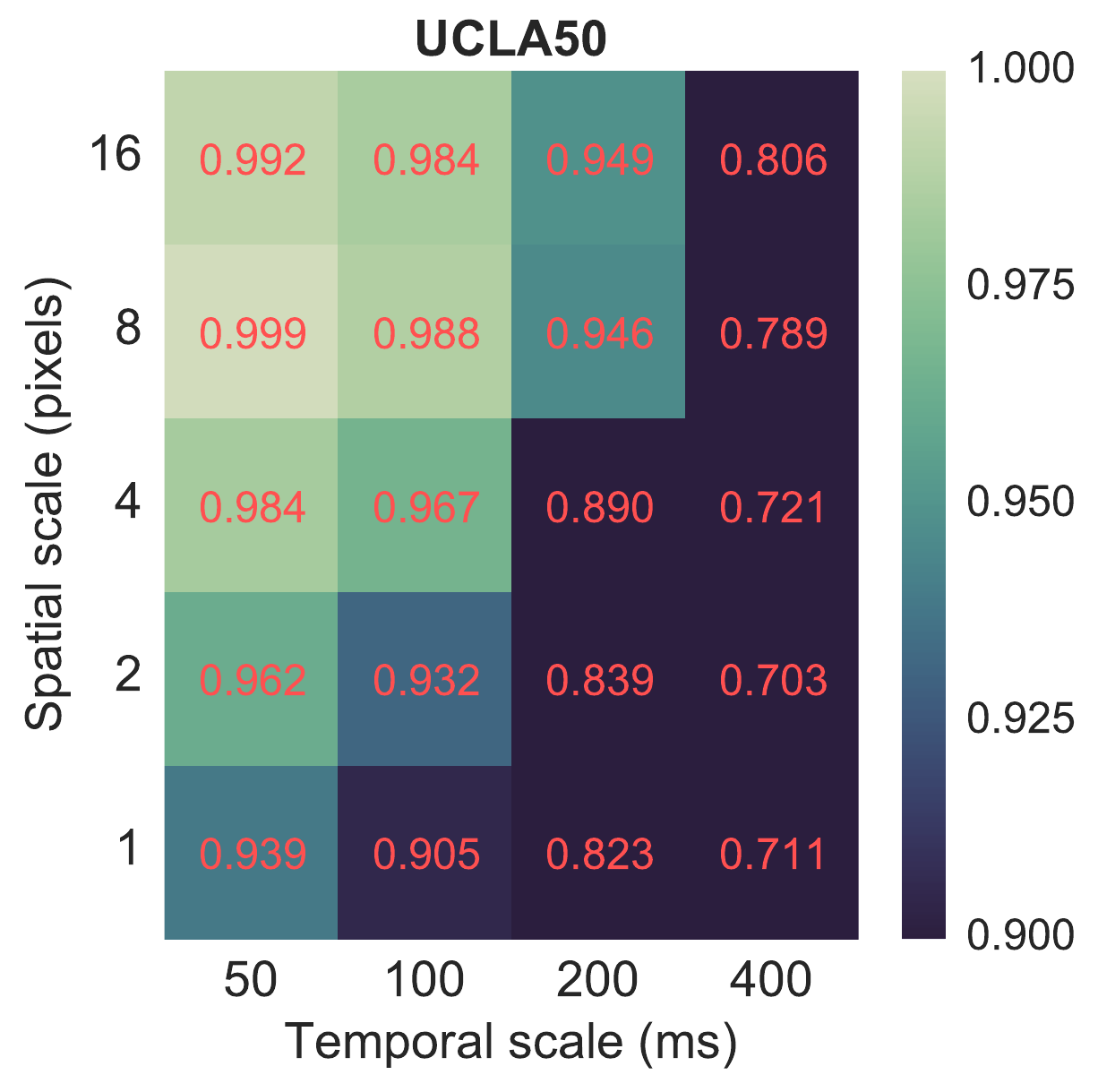} 
            
           \end{tabular}
             \caption{{\em The classification performance as function of the spatial and the temporal scales for the UCLA benchmarks}. The maps show the classification accuracy for different combinations of a single spatial scale $\sigma_s \in \{1,2,4,8,16 \}$ and a single temporal scale $\sigma_\tau \in \{50, 100, 200, 400\}$ for the descriptor STRF $N$-jet with $n_{bins} = 2$ and $n_{comp}=17$. For the UCLA8 and UCLA9 benchmarks, the best performance is obtained for a relatively small spatial scale in combination with a short temporal scale. For the UCLA50 benchmark, instead a combination of a relatively small spatial scale and a longer temporal scale gives the best results.}
\label{fig:ucla-scales1}

      \begin{tabular}{ccc}

       \includegraphics[width=0.3\textwidth]{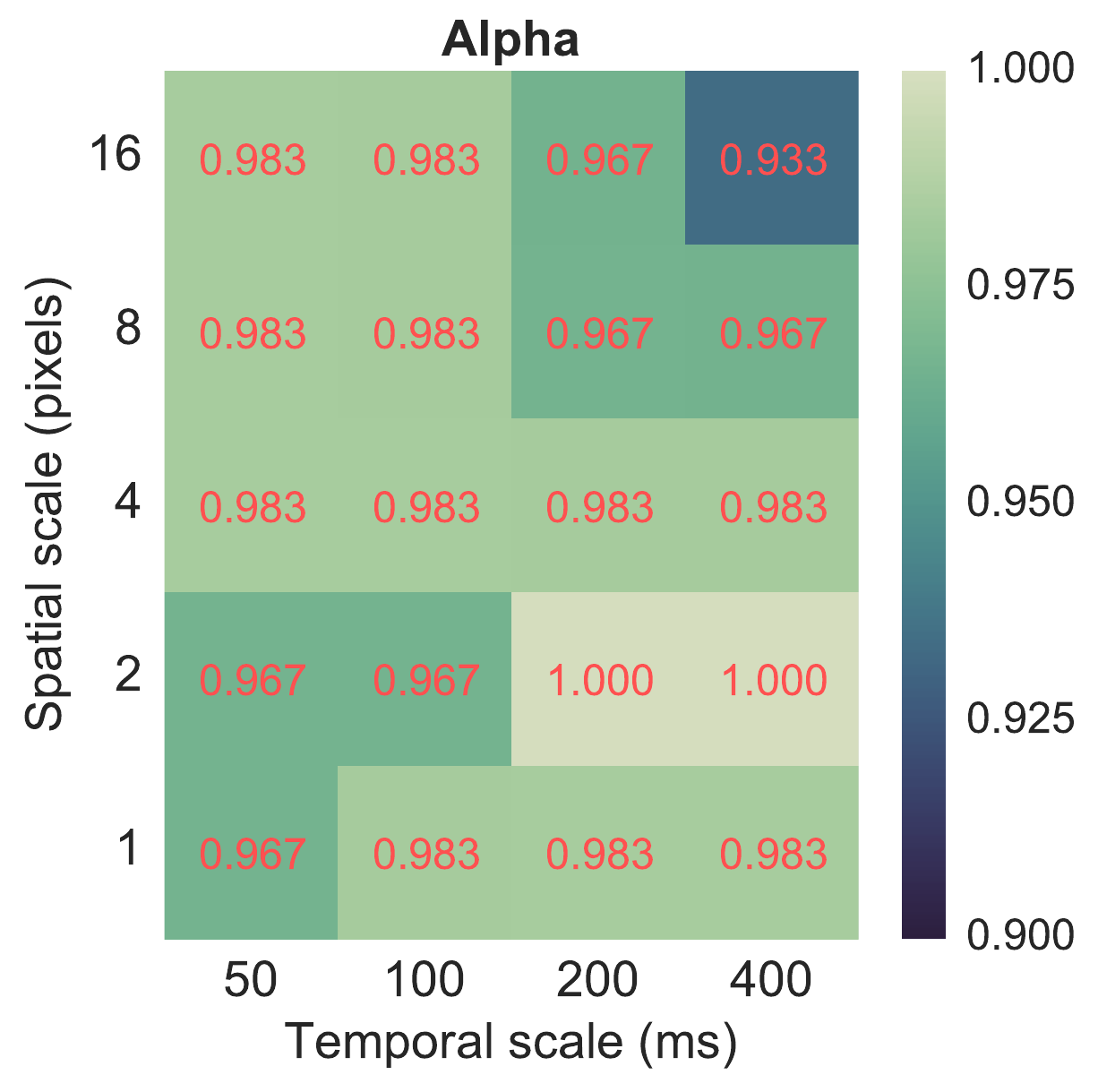} 
       &        \includegraphics[width=0.3\textwidth]{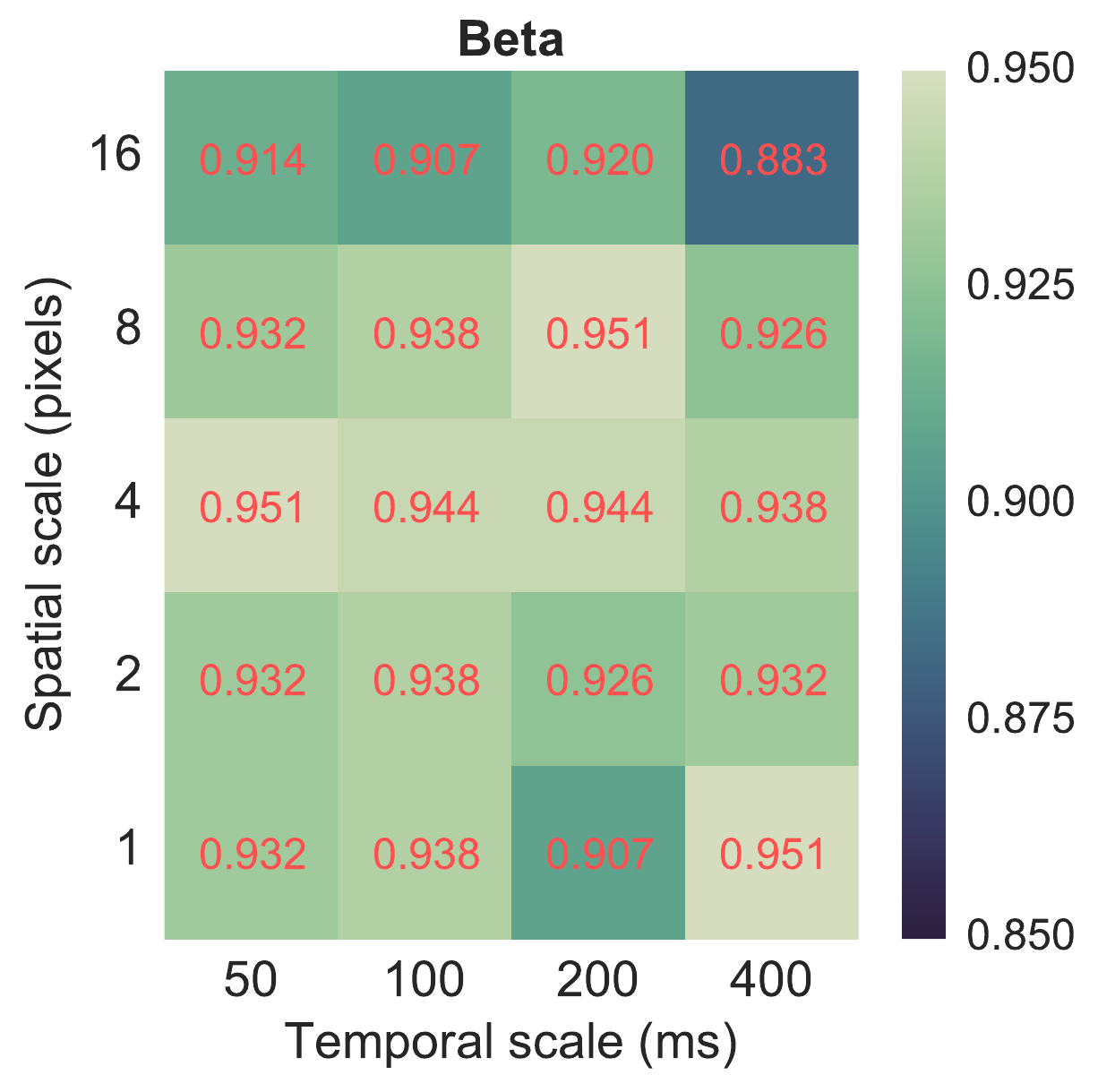} 
       &        \includegraphics[width=0.3\textwidth]{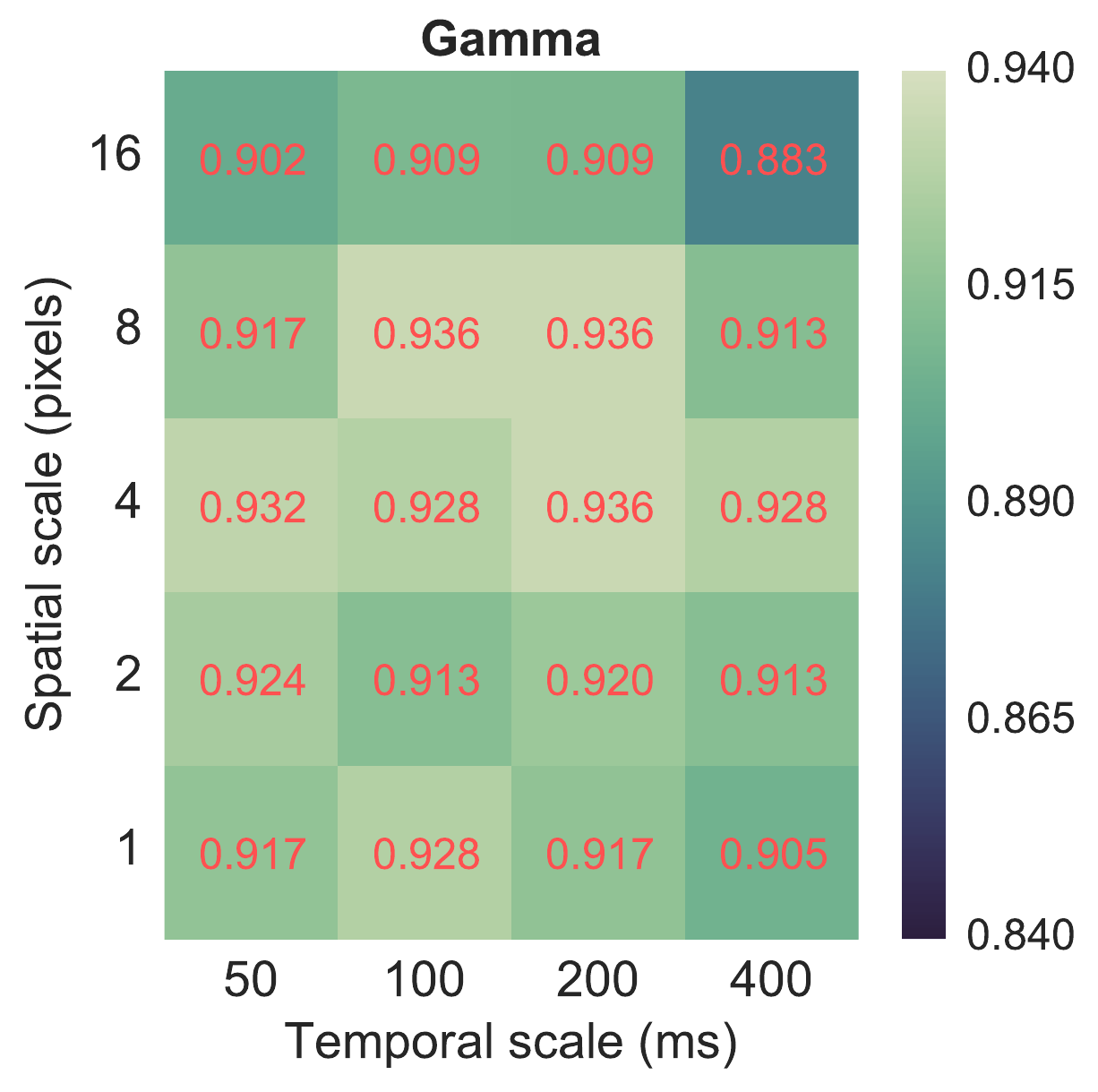} 
            
           \end{tabular}
  \end{center}
  \caption{{\em The classification performance as function of the spatial and the temporal scales for the DynTex benchmarks}. The maps show the classification accuracy for different combinations of a single spatial scale $\sigma_s \in \{1,2,4,8,16 \}$ and a single temporal scale $\sigma_\tau \in \{50, 100, 200, 400\}$ for the descriptor STRF $N$-jet with $n_{bins} = 2$ and $n_{comp}=17$. Note the different ranges used for colour coding the maps. Several local optima are obtained for all the three benchmarks. This suggests that different dynamic texture classes are best separated at different (non-adjacent) scales, which could be due to the larger diversity of the dynamic texture types present in these benchmarks.}

\label{fig:dyntex-scales1}
\end{figure*}

Each dataset will have a set of scales that are better for describing the spatial patterns and the motion patterns present in the videos. The classification performance of the STRF $N$-jet descriptor as function of the spatial and the temporal scales of the receptive fields for different combinations of a single spatial scale $\sigma_s \in \{1,2,4,8,16 \}$ and a single temporal scale $\sigma_\tau \in \{1,2,4,8,16 \}$ are shown in Figure \ref{fig:ucla-scales1} for the UCLA benchmarks and in Figure \ref{fig:dyntex-scales1} for the DynTex benchmarks. All results have been obtained with $n_{comp} = 15 $ and $n_{bins} = 2$. 

For all the UCLA benchmarks, an approximately unimodal maximum over scales is obtained. For the UCLA8 and UCLA9 benchmarks, the best performance is obtained when combining a smaller spatial scale with a shorter temporal scale. For the UCLA50 benchmark, the best results are instead achieved for shorter temporal scales in combination with larger spatial scales. The observation that
a short temporal scale works well for all benchmarks could indicate that fast motions are discriminative and that the best spatial scales are \emph{different} for UCLA50 is not strange, since this benchmark features instance recognition (e.g. separating 108 different plants) rather than generalising between classes. Although it might feel intuitive that small details should be useful for instance recognition, this will depend on the dataset. For example, plants with similar leaves but different global growth patterns could be easier to separate at a larger spatial scale.

For the DynTex benchmarks, the scale combinations that give the best results are scattered rather than showing an unimodal maximum. This could indicate that the different subsets of dynamic textures are best separated at different (and non-adjacent) scales. Since the DynTex dataset is quite diverse, this would not be strange. It should also be noted that the differences between the best and the second best results are here typically only one or two correctly classified videos. It is, however, clear that using the largest spatial scale in combination with the longest temporal scale gives markedly worse results.

When using 2 x 2 scales, we noted a similar performance pattern during scale tuning with unimodal maxima for the UCLA benchmarks and scattered maxima for the DynTex benchmarks (not shown). Comparing the absolute performance when using single vs multiple scales, it depends on the receptive field set if using multiple scales gives a consistent advantage. 
If inspecting the sets of optimal parameters found for the different benchmarks (presented in Appendix \ref{app:params}), it can be noted that, for the STRF $N$-jet descriptor, the best results are sometimes achieved using a single scale and sometimes when using 2 x 2 scales. However, STRF $N$-jet includes a quite large number (17) of receptive fields and when using a smaller set of receptive fields, such as in RF~Spatial (5 receptive fields) or STRF RotInv (9 receptive fields),  video descriptors using multiple spatio-temporal scales consistently have the best performance. This shows that receptive fields at different scales can contain complementary information. 

We conclude that, although results competitive with many state-of-the-art methods can be obtained for a heuristic choice of spatial and temporal scales, using parameter tuning to find an appropriate scale/set of scales may lead to improved performance of a few percentage points.


\begin{figure*}[hpbt]
\begin{center}
      \begin{tabular}{cc}

       \includegraphics[width=0.5\textwidth]{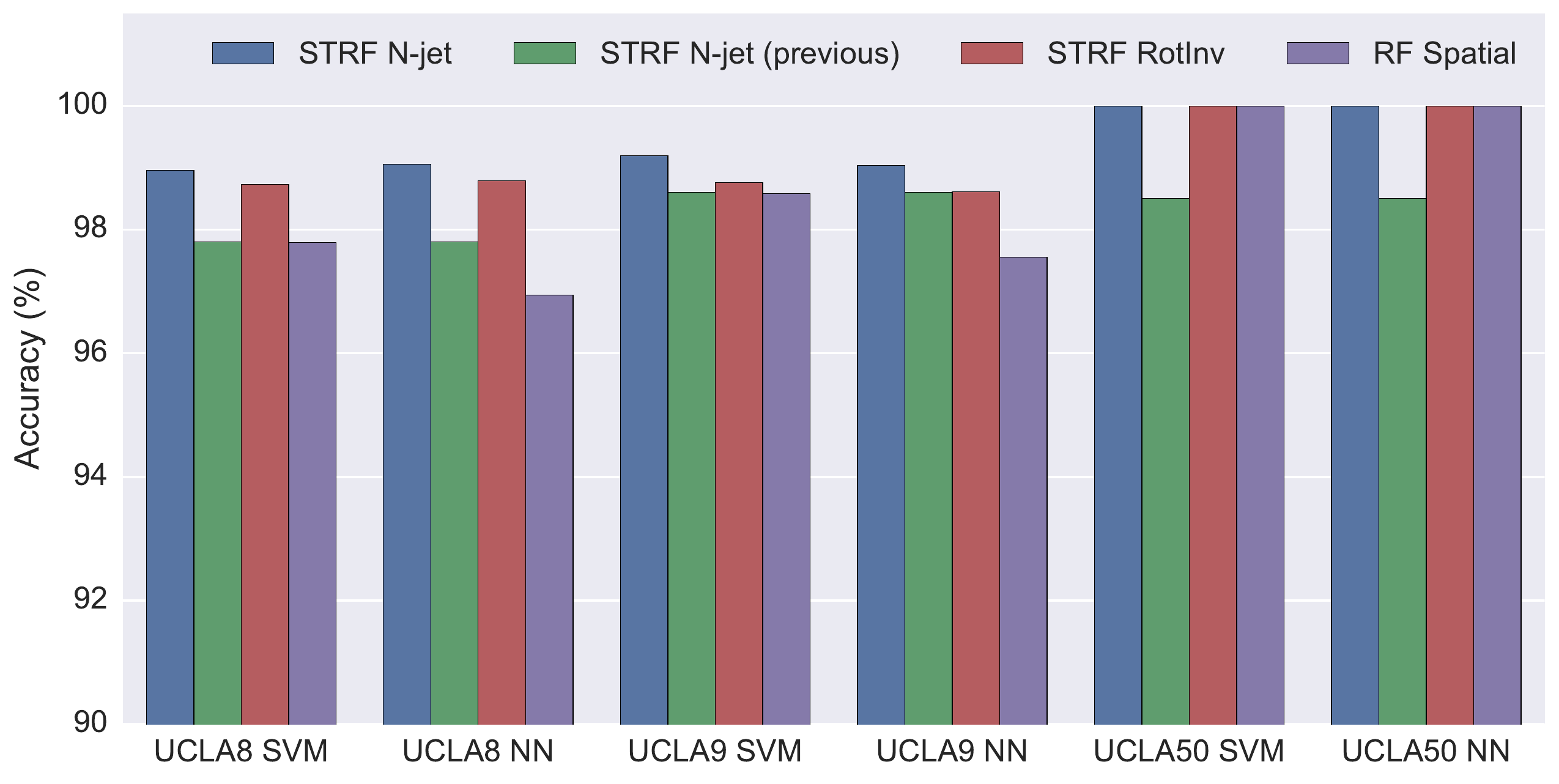} 
       \includegraphics[width=0.5\textwidth]{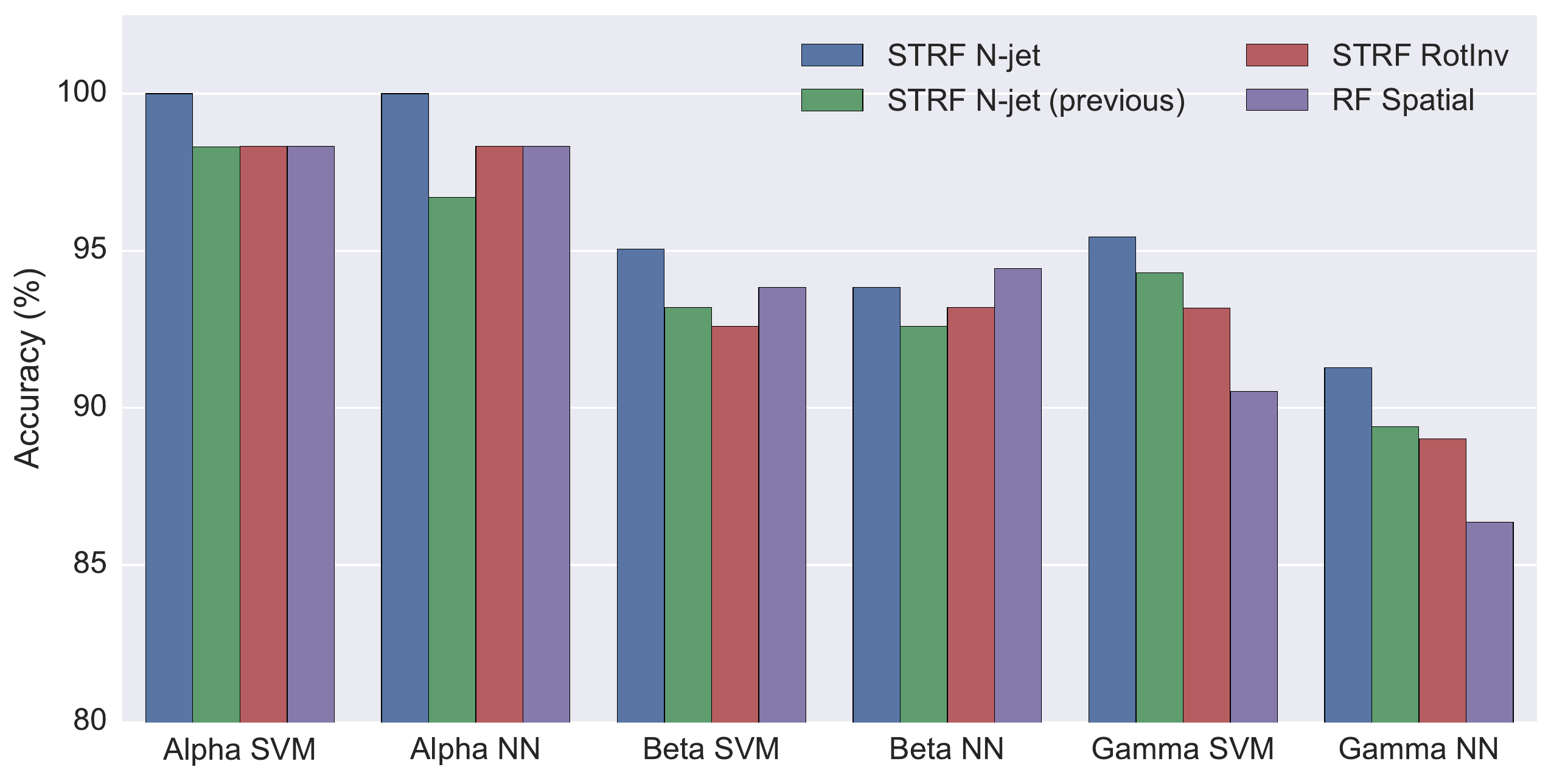} 
     
           \end{tabular}
  \caption{{\em The classification performance for video descriptors constructed from different sets of receptive fields}. Left: For the UCLA benchmarks. Right: For the DynTex benchmarks. 
The STRF $N$-jet descriptor achieves improved performance compared to RF Spatial (with the single exception of the Beta NN benchmark). This shows that the spatio-temporal receptive fields provide complementary information. The rotationally invariant STRF RotInv descriptor is also a competitive option. The new STRF $N$-jet achieves better results than STRF $N$-jet (previous) \cite{JanLin-SSVM2017}. All descriptors are binary. Additional parameters for each of these benchmark results are given in Appendix \ref{app:params}.}
\label{fig:rfsets}
\end{center}
\end{figure*}

\subsection{Receptive field sets}
\label{sec:expm-rfgroups}
In this section, we present results on relative performance between our four proposed video descriptors constructed from different sets of receptive fields (see Section \ref{sec:rfsets}): 

\begin{enumerate}[(i)]
\item The new spatio-temporal descriptor \emph{STRF $N$-jet}.
\item The rotationally invariant descriptor \emph{STRF RotInv}.
\item The purely spatial \emph{RF Spatial}.
\item The previous spatio-temporal descriptor \emph{STRF $N$-jet} (previous) \cite{JanLin-SSVM2017}.
\end{enumerate}
A comparison of the classification performance of these four video descriptors across all benchmarks is shown in Figure~\ref{fig:rfsets}. The performance of all four video descriptors is also compared to state-of-the-art in Table~\ref{tab:ucla-res} and Table~\ref{tab:dyntex-res}.

\subsubsection{STRF $N$-jet (previous) vs STRF $N$-jet} We note that parameter tuning and adding the second-order temporal derivatives of the spatial derivatives, result in improved performance for our new STRF $N$-jet descriptor compared to the STRF $N$-jet (previous) descriptor \cite{JanLin-SSVM2017}. The new descriptor shows improved accuracy for all the benchmarks. We have also observed an improvement from both these changes individually (not explicitly shown here). 

\subsubsection{Spatio-temporal vs spatial descriptors} 
A comparison between the STRF $N$-jet descriptor and RF Spatial reveals improved accuracy when including spatio-temporal receptive fields for the UCLA8, UCLA9, Alpha and Gamma benchmarks. Note that a comparison to the STRF $N$-jet (previous) descriptor is less relevant, since that descriptor is in contrast to the others not subject to parameter tuning. 

The largest improvement is obtained for the Gamma benchmark, where adding spatio-temporal receptive fields reduces the error from 9.5 \% to 4.5 \% when using an SVM classifier. Smaller improvements are obtained for the UCLA8 and UCLA9 benchmarks, with a reduction in error from 2.2 \% to 1 \% and from 1.4 \% to 0.8 \%, respectively. For the UCLA50 benchmarks, the performance saturates at 100 \% for both descriptor types (rather indicating the relative simplicity of this benchmark). The only exception where RF Spatial shows better performance is for the Beta benchmark using a NN classifier. Here, the purely spatial descriptor achieves 5.6 \% error vs 6.2 \% error for STRF $N$-jet. 

Competitive performance for purely spatial descriptors on the Beta benchmark has been reported previously \cite{QiLietal-NC2016} and we here make a similar observation. Thus, not surprisingly, for some settings settings genuine spatio-temporal information is of greater importance than for others. Here, the largest gain is indeed obtained for the most complex task. 

\subsubsection{Rotationally invariant descriptors} The rotationally invariant STRF $N$-jet RotInv descriptor does not achieve fully as good performance as the directionally dependent STRF $N$-jet descriptor for the tested benchmarks. The difference in classification accuracy in favour of the directionally selective descriptor is most pronounced for the more complex DynTex benchmarks: STRF RotInv achieves 7.4~\% and 6.8~\% error on the Beta and Gamma benchmarks using an SVM classifier, compared to STRF $N$-jet with 4.9~\% and 4.5~\% error, respectively. However, a comparison with state-of-the-art in Table~\ref{tab:ucla-res} and Table~\ref{tab:dyntex-res}, reveals that the STRF $N$-jet RotInv descriptor still achieves competitive performance compared to other dynamic texture recognition approaches. 

It is of conceptual interest that these good results can be obtained also when disregarding orientation information completely. Indeed, if considering marginal histograms of receptive field responses, the most striking differences between texture classes  such as waves, grass and foliage is the typical directions of change (waves show a stronger gradient in the vertical directions grass in the horizontal and foliage in both). A qualitative conclusion is that directional information is not the main mode of recognition here, instead the local space-time structure \emph{independent of orientation} is highly discriminative. We conclude that our proposed STRF RotInv descriptor could be a viable option for tasks where rotation invariance is of greater importance than for these benchmarks. However, the possible gain from enabling recognition of textures at orientations not present in the training data will have to be balanced against the possible gain from discriminative directional information.


\begin{table*}[hbpt]
  \setlength\tabcolsep{10pt}

    \caption{{\em Comparison to state-of-the-art for the UCLA benchmarks.} Our proposed \emph{STRF $N$-jet} descriptor shows consistently very competitive results for all these benchmarks, achieving the highest mean accuracy averaged over all benchmarks as well as the single best result on four out of the six benchmarks. All the STRF and RF descriptors are binary ($n_{bins} = 2$) and parameter tuning has been performed except for \emph{STRF $N$-jet (previous)}. The parameter values are given in Appendix \ref{app:params}. \textbf{Bold font} = highest accuracy for each single benchmark. Highlighted rows = our proposed descriptors.}   
 
\bigskip
\begin{center}
  \begin{tabular}{l l l l l l l l}
  \hline\noalign{\smallskip}
       & \multicolumn{2}{c}{\textbf{UCLA8}} & \multicolumn{2}{c}{\textbf{UCLA9}}
    & \multicolumn{2}{c}{\textbf{UCLA50}} \\
    & SVM & NN & SVM & NN  & SVM & NN \\
  \noalign{\smallskip}\hline\noalign{\smallskip}

\rowcolor{LightCyan}
\emph{STRF $N$-jet}  

     &  99.0 & \textbf{99.1}
     & 99.2 & \textbf{99.0} 
     & \textbf{100} & \textbf{100} & \emph{greyscale}\\
   
DNGP \cite{RivCha-TPAMI2015}
     & 99.4 & 97.0 
     &  \textbf{99.6} & 98.1
     & - & - & \emph{greyscale}\\
     
OTD \cite{QuaHuaJi-ICCV2015} 
      & \textbf{99.5} & 97.0
      & 98.2 & 97.5
      & 99.8 & 98.5 & \emph{greyscale}\\

DT-CNN \cite{AndWhe-arXiV2017} 
      & 99.0 & -
      & 98.4 & -
      & 99.5 & - & \emph{colour, deep learning} \\

\rowcolor{LightCyan}
\emph{STRF RotInv} 
     & 98.7 & 98.8
     & 98.8 & 98.6
     &  \textbf{100} & \textbf{100} & \emph{greyscale}\\

3D-OTF \cite{XuHuaJuFer-CVIU2012}
      &  \textbf{99.5} & 95.8
      &  97.2 &  96.3
      &  87.1 & 99.25 & \emph{greyscale}\\
     
Ensemble SVMs \cite{YanXiaetal-NC2016}
      &  - & -
      &  - & - 
      &  \textbf{100}& - & \emph{greyscale}\\
 
Enhanced LBP \cite{RenJiaYua-ICASSP2013} 
      & - & -
      & - & 98.2
      & - & \textbf{100} & \emph{greyscale}\\

MBSIF-TOP \cite{AraKit-TOM2014}  
      & - & 97.8
      & - & 98.8
      & - & 99.5 & \emph{greyscale}\\
 
\rowcolor{LightCyan}
\emph{STRF $N$-jet (previous)} \cite{JanLin-SSVM2017} 	
      & 97.8 & 97.5
      & 98.6 & 98.3
      & 98.5 & 97.0 & \emph{greyscale}\\
      
MEWLSP \cite{TiwTya-CEE2016} 
      & - & 98.0
      & - & 98.6
      & - & 96.5 & \emph{greyscale}\\

\rowcolor{LightCyan}
\emph{RF Spatial} 	
      & 97.8 & 96.9
      & 98.6 & 97.6
      & \textbf{100} & \textbf{100} & \emph{greyscale}\\

SKDL \cite{QuaCheHui-CVPR2016} 
      & 98.6 & -
      & - &   -
      & - & - & \emph{greyscale}\\     

HOG-NSP \cite{NorHaretal-ECCV2012} 
      & 98.7 & -
      & 98.1 & -
      & 97.2 & - & \emph{greyscale}\\

WMFS \cite{JiYanetal-TIP2013}      
      & 97.0 & 97.2
      & 97.1 & 97.0
      & 99.8 & 99.1 & \emph{greyscale}\\

PCA-net TOP \cite{AraAmiNor-JVCIR2017}
      &  - & -
      &  - & - 
      &  99.5 & - & \emph{greyscale}\\

CVLBP \cite{ZhaPie-WDV2006}, from \cite{TiwTya-CEE2016}  
      & - & 95.7
      & - & 96.9
      & - & 93.0 & \emph{greyscale}\\

DL-PEGASOS \cite{GhaAhu-ECCV2010} 
      & - & -
      & - & 95.6
      & - & 99.0 & \emph{greyscale}\\

Temporal dropout CNN \cite{CulSeb-ICM2014} 
      & - & -
      & - & -
      & - & 98.0 & \emph{greyscale, deep learning}\\

VLBP \cite{ZhaPie-WDV2006}, from \cite{TiwTya-CEE2016}  
       & - & 92.0
      & - & 96.3
      & - & 89.5 & \emph{greyscale}\\
      
Oriented energy rep.  \cite{DerWil-TPAMI2012} 
      & - & -
      & - & -
      & - &  81.0 & \emph{greyscale}\\
 
   \noalign{\smallskip}\hline\noalign{\smallskip}

  \end{tabular}
\end{center}

  \label{tab:ucla-res}

 \caption{
  {\em Comparison to state-of-the-art for the DynTex benchmarks}. Our proposed \emph{STRF $N$-jet} descriptor ranks at the very top among the grey-scale methods showing better performance than a large range of similar methods using different spatio-temporal primitives. All the STRF and RF descriptors are binary ($n_{bins} = 2$) and parameter tuning has been performed except for \emph{STRF $N$-jet (previous)}. The parameter values are given  in Appendix \ref{app:params}. \textbf{Bold font} = best greyscale descriptor for each single benchmark, \emph{italics font} = best colour descriptor, * indicates a different train/test partitioning for SVM and $\dagger$ the use of a nearest centroid classifier. Highlighted rows = our proposed descriptors.}
\medskip
\begin{center}
 \footnotesize
  \begin{tabular}{l l l l l l l l l}
	\hline\noalign{\medskip}

  & \multicolumn{2}{l}{\textbf{Alpha}} &
                                                \multicolumn{2}{l}{\textbf{Beta}}
    & \multicolumn{2}{l}{\textbf{Gamma}}  &  \\
    & SVM & NN & SVM & NN  & SVM & NN\\
  \noalign{\smallskip}\hline\noalign{\smallskip}
    
     DT-CNN \cite{AndWhe-arXiV2017} 
      & \textit{100} & -
      & \textit{100} & -
      & \textit{99.6} & - & \emph{colour, deep-learning} \\
    
    st-TCoF \cite{QiLietal-NC2016} 
      & \textit{100} & \textit{98.3}
      & \textit{100} & \textit{98.1}  
      & 98.1 & \textit{98.1} & \emph{colour, deep-learning} \\

     Deep Dual (D3)  \cite{HonRyuImYan-arXiv2017} 
      & \textit{100} & -
      & \textit{100} & - 
      & 98.1 & - & \emph{colour, deep-learning} \\ 
    
      Ensemble SVMs \cite{YanXiaetal-NC2016} 
      & - &  - 
      & - &  - 
      & \textit{99.5} & -  & \emph{colour} \\
      
      MR-SFA \cite{MiaXuXinTao-arXiv2017} 
      & - &  - 
      & \textbf{99.0} &  \textbf{98.1} 
      & - & -  & \emph{greyscale} \\

\rowcolor{LightCyan}
     \emph{STRF $N$-jet} 	
          & \textbf{100} & \textbf{100}
     & 95.1 & 93.8 
     & \textbf{95.5} & \textbf{91.2} & \emph{greyscale}\\     
 
 \rowcolor{LightCyan}     
    \emph{STRF $N$-jet (previous)} \cite{JanLin-SSVM2017}
     & 98.3 & 96.7
     & 93.2 & 92.6 
     & 94.3 & 89.4 & \emph{greyscale}\\

\rowcolor{LightCyan}     
     \emph{STRF RotInv} 
     & 98.3 & 98.3
     & 92.6 & 93.2
     & 93.2 & 89.0 & \emph{greyscale}\\

\rowcolor{LightCyan}
     \emph{RF Spatial} 	
     & 98.3 & 98.3
     & 93.8 & 94.4 
     & 90.5 & 86.4 & \emph{greyscale}\\
       
    SoB + Align \cite{SagKle-arXiv2017} 
      & - &  98.3 
      & -  & 90.1
      & - & 79.9  & \emph{greyscale} \\
     
    PCANet-TOP \cite{AraAmiNor-JVCIR2017} 
      & - & 96.7 
      & - & 90.7  
      & - & 89.4 & \emph{greyscale}\\   

     MBSIF-TOP \cite{AraKit-TOM2014} 
      & - & 90.0 
      & - & 90.7  
      & - & 91.3 & \emph{greyscale}\\
 
     AFS-TOP \cite{HonRyuetal-MSSP2016} 
      & 98.3 & 91.7 
      & 90.1 & 86.4 
      & 94.3 & 89.4 & \emph{greyscale}\\
    
    LBP-TOP \cite{ZhaGuoPie-TPAMI-2007}, from \cite{QiLietal-NC2016} 
      & 98.3 & 96.7
      & 88.9 & 85.8 
      & 94.2 & 84.9 & \emph{greyscale}\\

    ELM \cite{WanLiuSun-NC2016} 
      & - & - 
      & 93.8\textsuperscript{*} & - 
      & 88.3\textsuperscript{*}  & - & \emph{greyscale}\\

     SKDL  \cite{QuaCheHui-CVPR2016} 
      & 88.8\textsuperscript{*}  & -
      & 77.4\textsuperscript{*}  & -  
      & 75.6\textsuperscript{*}  & - & \emph{greyscale}\\

   2D+T curvelet \cite{DubSloetal-SIVP2015} 
      & - & 88.0\textsuperscript{$\dagger$} 
      & - & 70.0\textsuperscript{$\dagger$} 
      & - & 68.0\textsuperscript{$\dagger$}  & \emph{greyscale}\\
      
    OTD \cite{QuaHuaJi-ICCV2015}  
      & 87.8\textsuperscript{*}  & 86.6\textsuperscript{$\dagger$}
      & 76.7\textsuperscript{*}  & 69.0\textsuperscript{$\dagger$}
      & 74.8\textsuperscript{*}  & 64.2\textsuperscript{$\dagger$} & \emph{greyscale} \\
      
    DFS \cite{XuQuaetal-PR2015}  
      & 85.2\textsuperscript{*}  & - 
      & 76.9\textsuperscript{*}  & - 
      & 74.8\textsuperscript{*}  & - & \emph{greyscale}\\
            
  \noalign{\smallskip}\hline\noalign{\smallskip}
  \end{tabular}
\end{center}

  \label{tab:dyntex-res}
\end{table*}

\subsection{Comparison to state-of-the-art}
\label{sec:expm-comp-stateart}

This section presents a comparison between our proposed approach and state-of-the-art dynamic texture recognition methods. We include video descriptors constructed from four different sets of receptive fields (see Table \ref{tab:rfsets}) and compare against the best performing methods found in the literature for each benchmark. We also aim to include a range of \emph{different types} of approaches with an extra focus on methods similar to ours i.e. different LBP versions and relatively shallow (max 2 layers) spatio-temporal filtering based approaches using either handcrafted filters or filters learned from data. Results for all the other methods are taken from the literature, where the relevant references are indicated in the table. 

\subsubsection{UCLA datasets}
The UCLA benchmark results are presented in Table \ref{tab:ucla-res}. Our proposed STRF $N$-jet descriptor shows highly competitive performance compared to all the other methods, achieving the highest mean accuracy averaged over all the benchmarks and either the single best or the shared best result on four out of the six benchmarks. 

For the UCLA50 benchmark, our three new video descriptors achieve 0 \% error using both an SVM and a NN classifier. The main difference between these descriptors and the untuned STRF $N$-jet (previous) is the use of a larger spatial scale, which was seen in Section \ref{sec:expm-scales} to be more adequate for this benchmark. Enhanced LBP \cite{RenJiaYua-ICASSP2013} and Ensemble SVMs \cite{YanXiaetal-NC2016} also achieve 0 \% error rate and there are several methods with error rates below 0.5 \%. The main conclusions we draw from the UCLA50 results are that recognising the same dynamic texture instance from the same viewpoint is (not surprisingly) an in comparison easier task than separating conceptual classes and that our approach performs on par with the best state-of-the-art methods on this task. 

For the conceptual UCLA8 and UCLA9 benchmarks using an NN classifier, our STRF $N$-jet descriptor achieves 0.9 \% and 1.0 \% error, respectively, which are the single best results among all methods. This demonstrates that our approach is stable and works well with a simple classifier also for a quite high-dimensional descriptor. For the UCLA8 benchmark together with a NN classifier, the second best performing approach is our rotational-invariant descriptor STRF RotInv with 1.2 \% error and after that MEWLSP \cite{TiwTya-CEE2016} with 2 \% error. For UCLA9, the second best performing approach is MBSIF-TOP \cite{AraKit-TOM2014} with 1.2 \% error followed by STRF RotInv and MEWLSP, which both achieve 1.4 \% error. 

For the UCLA8 benchmark combined with an SVM classifier, the best performing approaches are OTD \cite{QuaHuaJi-ICCV2015} and 3D-OTF \cite{XuHuaJuFer-CVIU2012} both with 0.5 \% error. For UCLA9, the best method using an SVM classifier is DNGP \cite{RivCha-TPAMI2015}, which achieves 0.4 \% error. Our STRF $N$-jet descriptor achieves 1.0 \% error on the UCLA8 benchmark, and 0.8 \% error on the UCLA9 benchmark. It should be noted that OTD, 3D-OTF and DNGP simultaneously show considerably worse results on the NN benchmarks and that the standard UCLA protocol (average over 20 trials) can give quite variable results because of the limited number of samples in the benchmarks. Averaging over 1000 trials means that our results are more stable and less likely to include ``outliers" for some of the benchmarks. 

Our approach shows improved results on all the UCLA benchmarks compared to a large range of similar methods also based on gathering statistics of local space-time structure but using \emph{different spatio-temporal primitives}. This includes methods that are more complex in the sense of combining several different descriptors or a larger number of feature extracting steps (MEWLSP \cite{TiwTya-CEE2016}, HOG-NSP \cite{NorHaretal-ECCV2012}), methods learning higher-level hierarchical features (PCANet-TOP \cite{AraAmiNor-JVCIR2017}, SKDL \cite{QuaCheHui-CVPR2016}, temporal dropout DL, DT-CNN \cite{AndWhe-arXiV2017}) and improved and extended LBP-based methods (Enhanced LBP \cite{RenJiaYua-ICASSP2013}, MBSIF-TOP \cite{AraKit-TOM2014}, MEWLSP \cite{TiwTya-CEE2016}, CVLBP \cite{TiwTya-MSSP2016}) as well as the standard LBP-TOP \cite{ZhaGuoPie-TPAMI-2007} and VLBP \cite{ZhaPie-WDV2006} descriptors. An interesting observation is also that compared to VLBP and CVLBP, which similar to our approach use binary histograms and full 2D+T primitives, the performance of our approach is 2.1 to 10.5 percentage points better for all the benchmarks. The most important difference between these methods and our approach is indeed the spatio-temporal primitives used for computing the histogram. 

\subsubsection{DynTex datasets}
The DynTex benchmark results are presented in Table \ref{tab:dyntex-res}. For this larger and more complex dataset, it can be seen that utilising colour information and supervised hierarchical feature learning seems to give a clear advantage with three deep learning approaches on top. DT-CNN, trained from scratch to extract features on three orthogonal planes, demonstrates the best performance with 0 \% error on the Alpha and Beta benchmarks and 0.4 \% error on the Gamma benchmark. Two deep learning methods based on feature extraction using pretrained networks (Deep Dual descriptor and st-TCoF) also obtain very good results. 
 However, although included for reference, we do not directly aim here to compete with these more conceptually complex methods. The main focus of our work is instead to evaluate the usefulness of the time-causal spatio-temporal primitives without entanglement with a more complicated hierarchical framework.

Our proposed STRF $N$-jet descriptor achieves 0 \% error on the Alpha benchmark using both an SVM and a NN classifier, 4.9 \% (SVM) and 6.2 \% (NN) error on the Beta benchmark and 4.5 \% (SVM) and 8.8 \% error (NN) on the Gamma benchmark. This means that we achieve better results than all other non-deep learning methods utilising only grey scale information except one: MR-SFA \cite{MiaXuXinTao-arXiv2017} which achieves 1 \% error on the Beta SVM benchmark and 1.9 \% error on the Beta NN benchmark (this method has not been tested on the Alpha and Gamma benchmarks). It should, however, be noted that MR-SFA uses regional descriptors capturing the relative location of image structures and a bag-of-words framework on top of the histogram descriptor. This approach is thus significantly more complex compared to our method. Both these extensions would be relatively straightforward to implement also using our proposed video descriptors.

Our results can also be compared to the LBP-TOP extension AFS-TOP, which shows the most competitive results using an SVM classifier with 1.7 \%, 9.9 \% and 5.7 \% error, respectively, on the Alpha, Beta and Gamma benchmarks. Our approach thus achieves better performance on all the tested benchmarks, although AFS-TOP includes several added features, such as removing outlier frames. Improvements compared to the basic LBP-TOP descriptor are larger and this can be considered a more fair benchmark, since we are testing an early version of our approach. Compared to MBSIF-TOP and PCANet-TOP, which both learn 2D filters from data and apply those on three orthogonal planes, our approach also achieves better results on all the DynTex benchmarks.   

We also show notably better results (in the order of 10-20 percentage points) than those reported from using DFS \cite{XuQuaetal-PR2015}, OTD \cite{QuaHuaJi-ICCV2015}, SKDL \cite{QuaCheHui-CVPR2016} and the 2D+T curvelet transform \cite{DubSloetal-SIVP2015}. However, those use a nearest centroid classifier and a different SVM train-test partition, which means that a direct comparison is not possible. 
We also note that although STRF $N$-jet achieves the best results, the rotationally invariant descriptor version STRF RotInv, the RF Spatial descriptor and the untuned STRF $N$-jet (previous) descriptor also achieve competitive performance. This demonstrates the robustness and flexibility of our approach. 

In conclusion, our approach shows highly competitive performance for this larger and more complex benchmark, even though our proposed approach is a conceptually simple method utilising only local information. The STRF $N$-jet descriptor achieves better performance than all other grey-scale methods of similar conceptual complexity and better performance compared to both several more complex methods and two methods learning filters from data. We believe this should be considered as strong validation that these time-causal spatio-temporal receptive fields indeed capture useful spatio-temporal information.

\subsection{Descriptor size vs performance}
\label{sec:expm-size-performance}

The proposed family of video descriptors may vary substantially in descriptor size, depending on the parameter choices made. We here illustrate the trade-off between descriptor size and classification performance for binary descriptors on the most challenging Gamma benchmark. Size is here defined as the average number of non-empty cells for a certain descriptor type and benchmark. This also gives an estimate on the sparsity of the histograms. 

\begin{figure}[t]
  \begin{center}
       \includegraphics[width=0.40\textwidth]{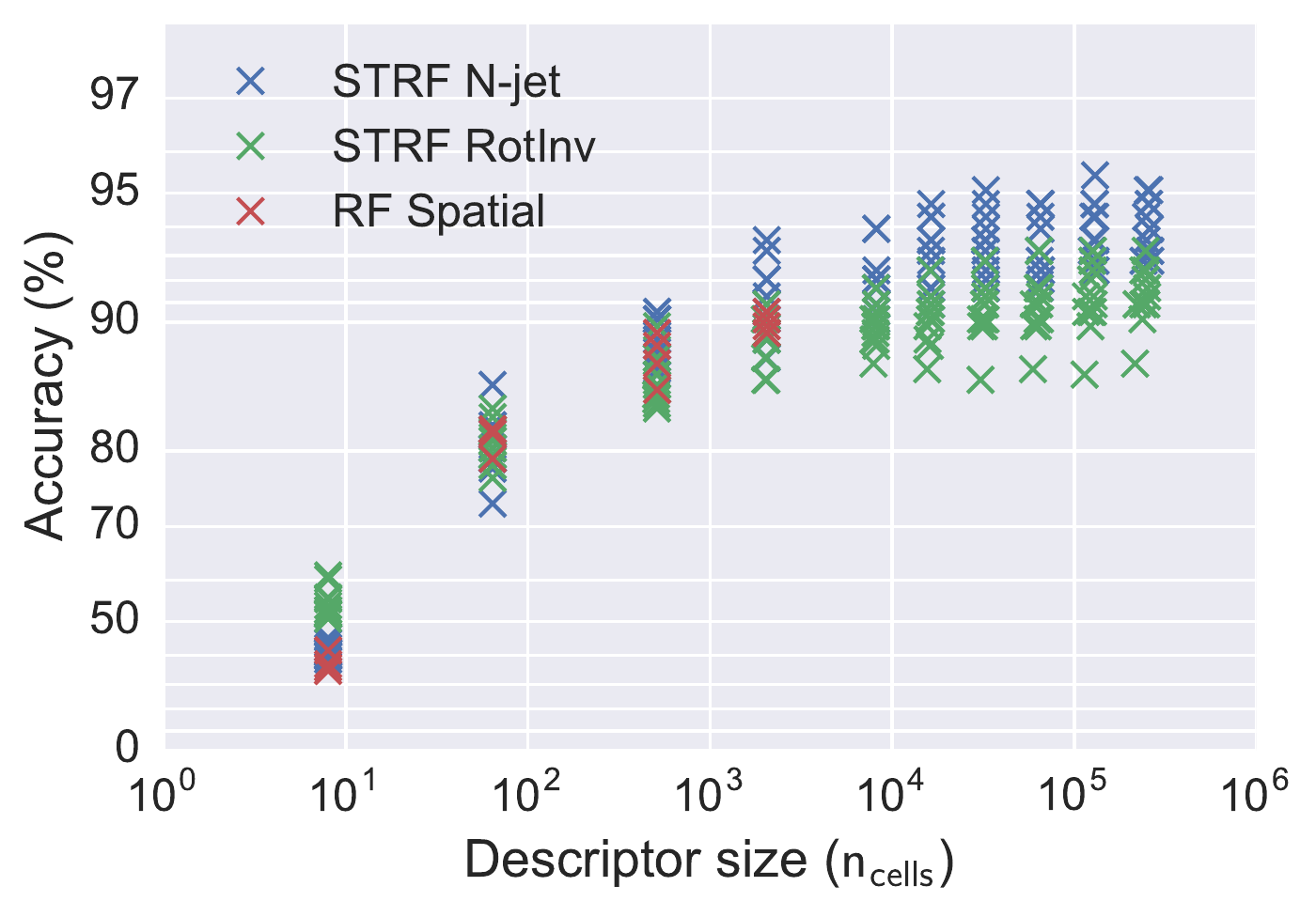}
  \end{center}
  \caption{\emph{Descriptor size vs classification accuracy} for the binary STRF $N$-jet, STRF RotInv and RF Spatial video  descriptors evaluated on the Gamma benchmark. 
The performance for each descriptor version is plotted against the average descriptor size for all different combinations of two spatial and two temporal scales (see Section \ref{sec:expm-ptuning}) and for different numbers of principal components (see Table \ref{tab:size-performance}). The accuracy is presented as $\log( \text{error}) = \log(100 - \text{accuracy})$ with the vertical axis reversed.}
\label{fig:size-performance}
\end{figure}

The classification accuracy together with the descriptor size for increasing numbers of principal components $n_{comp}$ for the binary STRF $N$-jet, STRF RotInv and RF Spatial descriptors is shown in Figure \ref{fig:size-performance}. We include all combinations of 2 x 2 scales (see Section \ref{sec:scales}) for each value of $n_{comp}$ to illustrate the overall trend rather than focusing on one specific choice of spatial and temporal scales. Table \ref{tab:size-performance} shows the best classification performance (after scale tuning) for the STRF $N$-jet descriptor on the Gamma benchmark for each value of $n_{comp}$ together with the average number of non-empty histogram cells. 

For all video descriptors, it can be seen that the performance first increases fast with the descriptor size (here, equivalently the number of principal components used) up to around $n_{comp} = 10$. For the STRF N-jet descriptor, some additional gains are then obtained, primarily between 10 and 14 principal components, after which the performance saturates. For the STRF RotInv descriptor, the performance saturates at a lower level than for the STRF $N$-jet descriptor indicating that by using only rotationally invariant receptive fields some discriminative information is lost. For the RF Spatial descriptor, the best performance is obtained using all principal components (5 receptive fields $\times$ 2 spatial scales). 

From Table \ref{tab:size-performance} it can further be seen that the best results are achieved for $8200 \le n_{cells}  \le 130 000$, with accuracy in the range 94.7 \% - 95.5 \%. However, also an even smaller descriptor using $n_{cells}=1024$ gives reasonably good performance. We conclude that good performance is not contingent on a very large descriptor size and that our approach can be viable also in settings where a descriptor with a small memory footprint is required.

Comparing the number of filled histogram cells with the corresponding powers of two, it can be seen that these binary histograms are almost filled. We found this to be true for the other binary histogram types tested here as well. In contrast with this result, we have noted as low as 0.01 \% filled histogram cells for some of the larger non-binary histograms tested in Section~\ref{sec:expm-binscomp}. 

Indeed, for non-binary histograms the more precise conditions on the magnitude of the receptive field responses will imply a larger set of very "rare" patterns. Thus, for the larger non-binary histograms there can be large benefits from using a sparse representation, while for the binary histograms tested here, a non-sparse representation is most advantageous.

\begin{table}[t]
\caption{\emph{Descriptor size vs classification accuracy} for the binary STRF $N$-jet descriptor evaluated on the Gamma benchmark. The table shows the average descriptor size (the number of non-zero histogram cells) together with the top accuracy obtained for video descriptors using different numbers of principal components.}
  \begin{center}
\begin{tabular}{lll}
\hline\noalign{\smallskip}
 $n_{comp}$ & $n_{cells}$ (filled) \text{~~~~~~~~~~} & accuracy (\%)  \\
\noalign{\smallskip}\hline\noalign{\smallskip}
2  &   $4.0 \times 10^0$  &    45.5 \\
5  & $ 3.2 \times  10^1$  &    86.0 \\
8  &$ 2.6 \times  10^2$ &     90.5 \\
10 &$ 1. 0 \times  10^3$  &    93.6 \\
12  &$ 4.1 \times  10^3 $ &    93.9 \\
13  &$ 8.2 \times  10^3 $&      94.7 \\
14  & $1.6 \times  10^4  $&    95.1 \\
15  &$ 3.3 \times  10^4  $&   94.7 \\
16 &$ 6.5 \times  10^4  $&    95.5 \\
17 &$ 1.3 \times  10^5 $&    95.1 \\
\noalign{\smallskip}\hline
\end{tabular}
  \end{center}

\label{tab:size-performance}      
\end{table}

\begin{figure}[hbpt]
  \begin{center}
       \includegraphics[width=0.45\textwidth]{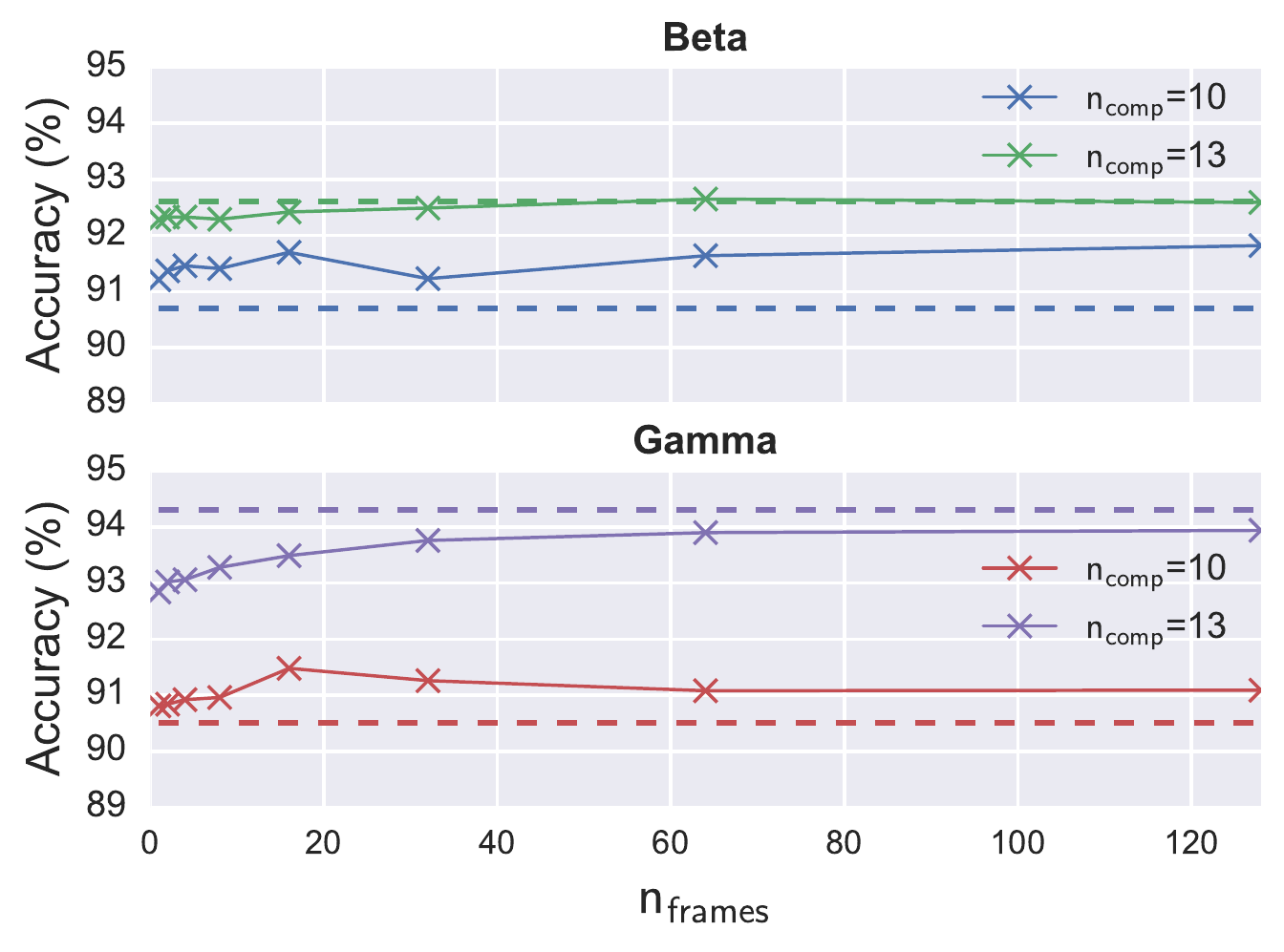}
  \end{center}
  \caption{\emph{Recognition from fewer frames} for the Beta and Gamma benchmarks using the binary STRF $N$-jet descriptor. The plots show the classification accuracy as a function of the number of frames used to gather statistics for the video descriptor. The dashed line represents the performance obtained when using the full video. For $n_{comp} = 13$, results close to baseline performance can be obtained also when using very few frames. For $n_{comp} = 10$, the performance is improved when training and testing using shorter sequences for all values of $n_{frames}$. Spatial scales: $(\sigma_{s_1}, \sigma_{s_2}) = (4, 8) \text{~pixels}$. Temporal scales: $(\sigma_{\tau_1}, \sigma_{\tau_2}) = (100, 200)$ ms.}
\label{fig:fewer-frames}
\end{figure}

\subsection{Recognition using fewer frames}
\label{sec:expm-frames}

In most real-world scenarios, an artificial or biological agent will get advantages from being able to interpret, and thus react, to its surroundings as fast as possible. 
We here investigate how the recognition performance of our proposed approach is affected by gathering statistics over a smaller interval of frames as opposed to over the full video. The experimental setup is chosen to reflect the most likely real world scenario -- full videos (or scenes) are available during training but classification of a new dynamic texture is made from a single shorter sequence. 

It should be noted that the stationarity properties of dynamic textures will often manifest over time-scales longer than a few frames. Thus, it is important that local (in time) parts of motion patterns can be matched to other local (in time) patterns, rather than forcing also a short sequence to match the statistics of a full video. We, here, extract a collection of shorter sequences of length $n_{frames} \in \{1,2,4,8,16, 32,\\ 64, 128\}$ from each video. 
Then, during training, the full set of shorter sequences for each value of $n_{frames}$ is used as training examples, while for testing the recognition is based on a single shorter sequence. 

We study the performance of the binary STRF $N$-jet descriptor with $n_{comp} \in \{10,13\}$ on the Beta and Gamma benchmarks. The results are presented in Figure \ref{fig:fewer-frames}. It can be seen that for $n_{comp} = 13$ performance close to the baseline can be achieved also using very few frames. For the Beta benchmark, there is 0.3 \% difference between using a single frame and the entire video (92.3 \% vs 92.6 \%), while for the Gamma benchmark this difference is 1.1 \% (92.8 \% vs 93.9~\%). When using a smaller descriptor with $n_{comp} = 10$, we note that the performance is instead \emph{improved} when training and testing using shorter sequences, where the best values of $n_{frames}$ gives accuracy 1 \% above the baseline for both benchmarks. Note that, in a real-time scenario, also the temporal delay of the time-causal filters will influence the reaction time and that at least three frames are needed to compute a discrete second-order temporal derivative approximation. 

The reason why using less data/fewer frames might give a drop in performance is clear -- there is simply less information that can be used for making a decision. 
The reason why using shorter sequences can improve the performance is most likely related to that the matching becomes more flexible -- each shorter sequence of a video only needs to have a similar representation to \emph{at least one} shorter sequence from another video in the same class, rather than requiring the full videos to match. 

In conclusion, these results demonstrate that our approach is robust to using descriptors computed from shorter video sequences and that it should thus be a viable option also for real-time scenarios where a quick decision is needed.

\subsection{Qualitative results}
\begin{figure*}[!t]
  \begin{center}
    \begin{tabular}{ccc}
       \includegraphics[width=0.305\textwidth]{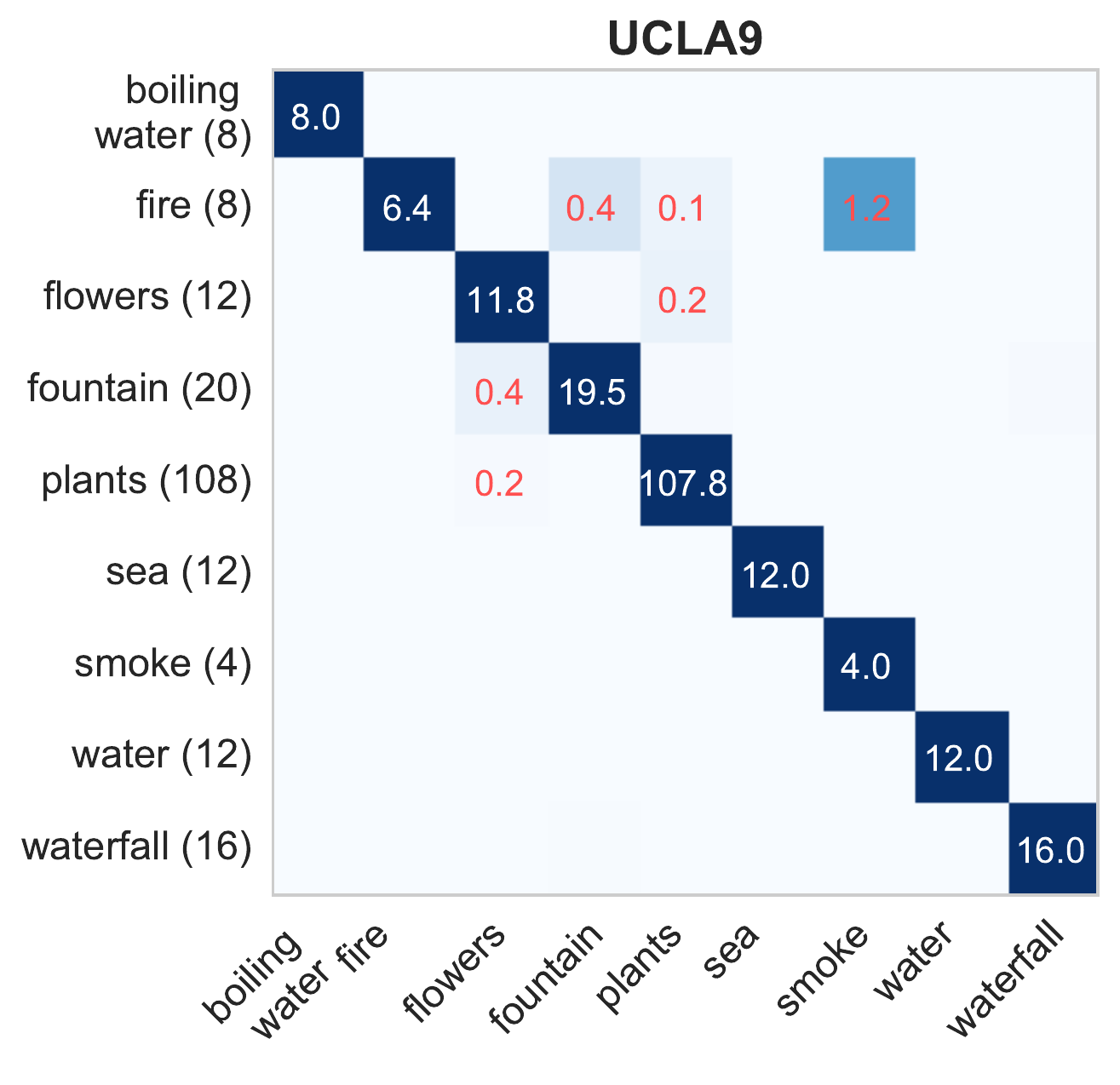}
        & \includegraphics[width=0.31\textwidth]{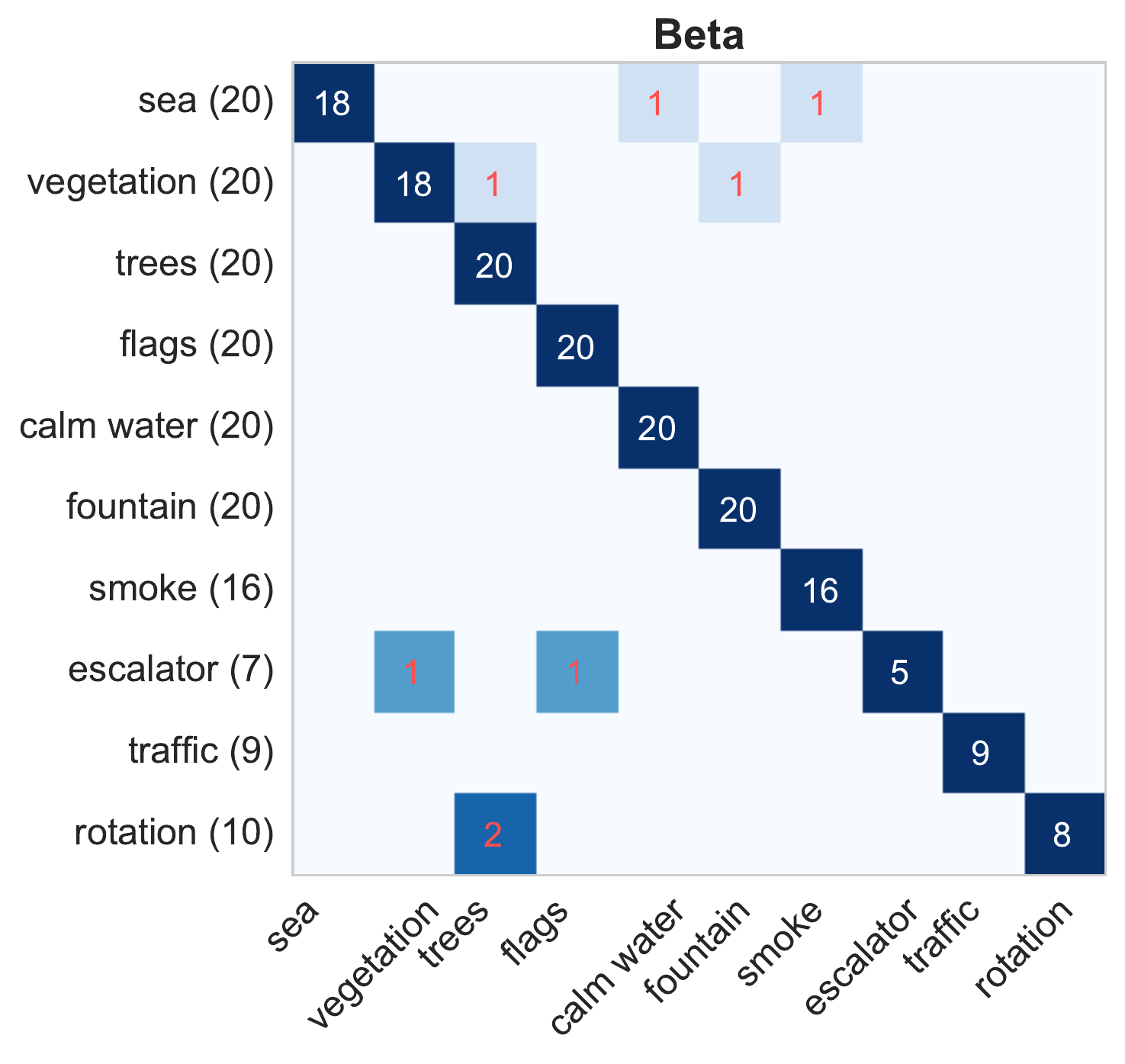}
        & \includegraphics[width=0.31\textwidth]{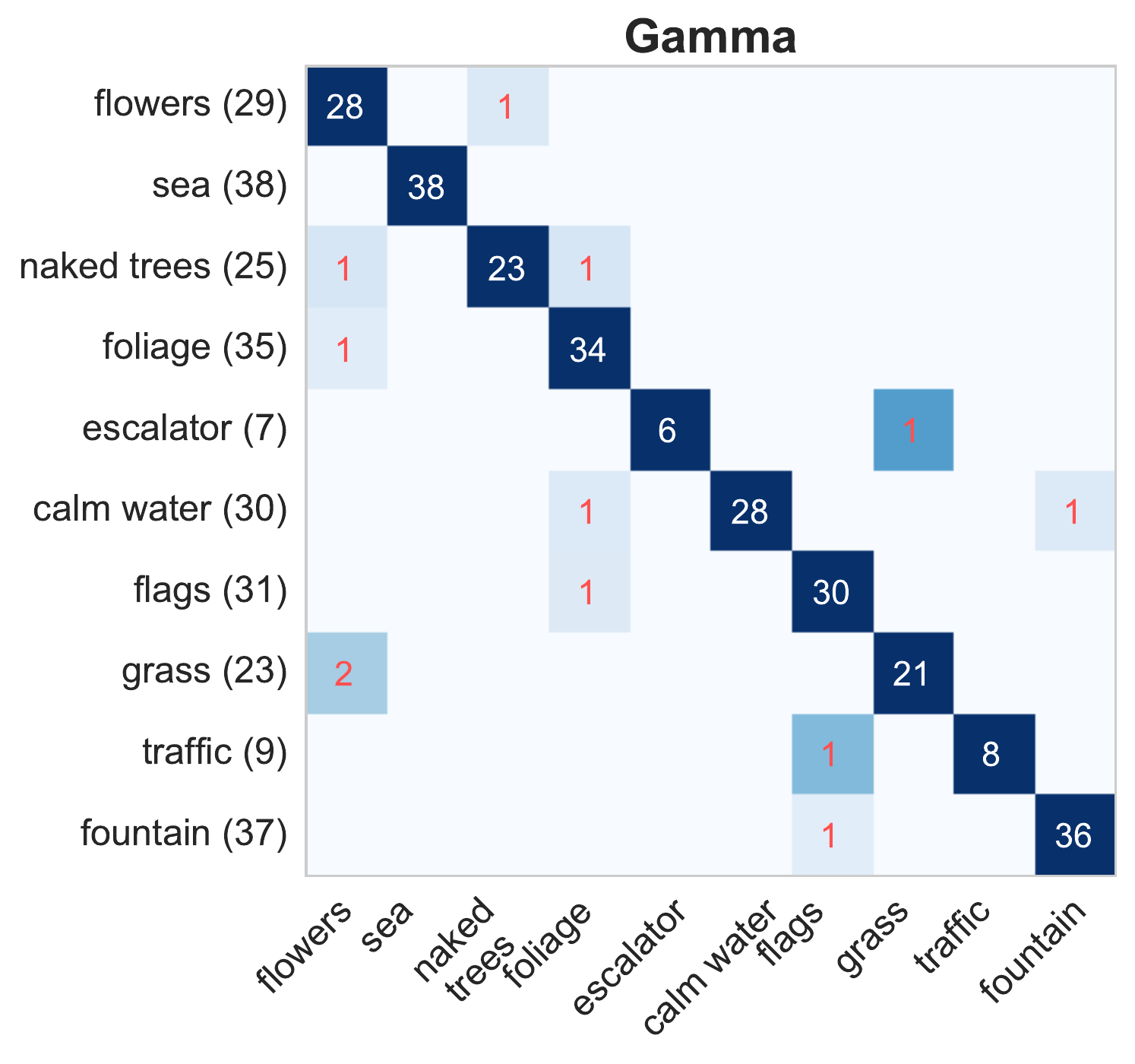}
     
  \end{tabular}
  \end{center}
  \caption{Confusion matrices for the STRF $N$-jet descriptor combined with an SVM classifier.
  Left: For the UCLA9 benchmark (averaged over trials). Parameters: $\sigma_s = (4, 8)$, $\sigma_{\tau} = (50, 100)$, $n_{comp} = 14$. Here, the largest confusion is between fire and smoke. 
  Middle: For the Beta benchmark. Parameters: $\sigma_s  = 8$, $\sigma_{\tau} = 200$, $n_{comp} = 17$. 
Right: For the Gamma benchmark. Parameters: $\sigma_{\tau} = (50, 100)$, $n_{comp} = 16$. Here, no single class is dominating among the confusions. See also Figure~\ref{fig:misclassified}.}
\label{fig:conf}
\end{figure*}

To gain more insight into the qualitative behaviour of our proposed family of video descriptors, we inspected the confusions and the closest neighbours to correctly classified and misclassified samples. Confusion matrices for the UCLA9, the Beta and the Gamma benchmarks for the STRF $N$-jet descriptor are presented in Figure~\ref{fig:conf}. We note that the main cause of error for UCLA9 is confusing fire with smoke. UCLA8 (not shown here) shares a similar pattern. There is indeed a similarity in dynamics between these textures in the presence of temporal intensity changes not mediated by spatial movements. Confusions between flowers and plants and between fountain and waterfall are most likely caused by similarities in the spatial appearance and the motion patterns of these dynamic texture classes. 

When inspecting the confusions between the different classes for the Beta and Gamma benchmarks, there is no clear pattern visible. This is probably partly due to the fact that these benchmarks contains larger intraclass variabilities.  Misclassifications seem to be caused by a single video in one class having some similarity to a single video of different class rather than certain classes being consistently mixed up. We note the largest ratio of misclassified samples for the escalator and traffic classes, which are also the classes with the fewest samples. 

\begin{figure*}[hbpt]
\begin{center}
         \includegraphics[width=0.95\textwidth]{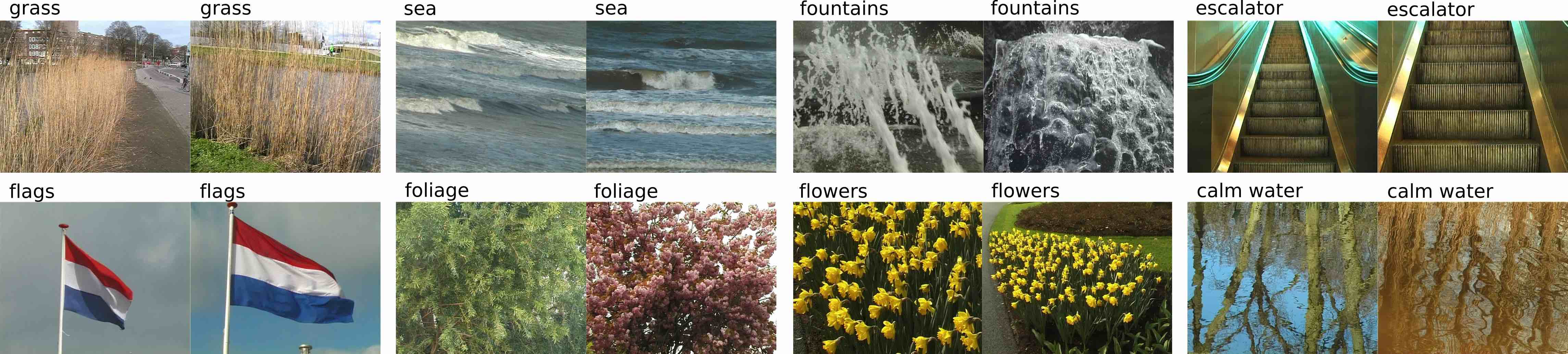}
         \end{center}
  \caption{A subset of correctly classified samples (left in each pair) for the Gamma benchmark together with their closest neighbours (right in each pair) when using the STRF $N$-jet descriptor together with an SVM classifier. Parameters: $\sigma_s  = (4, 8)$, $\sigma_{\tau} = (50, 100)$, $n_{comp} = 17$.}
\label{fig:correct}
\begin{center}
         \includegraphics[width=0.95\textwidth]{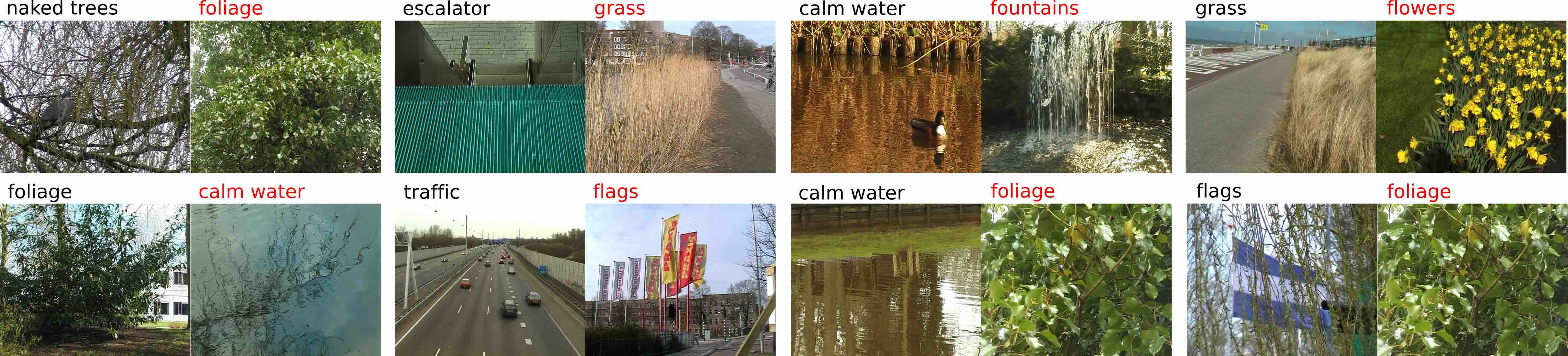}
\end{center}  
  \caption{A subset of misclassified samples (left in each pair with classname in black) for the Gamma benchmark together with their closest neighbours (right in each pair with classname in red) when using the STRF $N$-jet descriptor together with an SVM classifier. We note that sources of error are the presence of several dynamic textures in the same video (top row image pairs one, three and four and bottom row image pair four) and the presence of large static image regions. The remaining cases appear to be based on spatial or temporal similarity between individual texture instances, such as a bush reflected in the water being classified as foliage (bottom leftmost pair) or an escalator misclassified as grass (top row second image pair). Parameters: $\sigma_s  = (4, 8)$, $\sigma_{\tau} = (50, 100)$, $n_{comp} = 17$. }
\label{fig:misclassified}

\end{figure*}

A bit more light is shed on the reasons for misclassifications for the DynTex benchmarks when inspecting the closest neighbours in feature space for misclassified videos (Figure~\ref{fig:correct}) and for correctly classified videos (Figure~\ref{fig:misclassified}). We note that a frequent feature of the misclassified videos is the presence of multiple textures, such as a flag with light foliage in front of it (misclassified as foliage), or a fountain flowing into a pool of calm water (misclassified as calm water).
There are also examples of confusion caused by similarity in either the spatial appearance or the temporal dynamics between specific instances of different classes, such as calm water reflecting a small tree being misclassified as foliage or a field of grass waving in the wind misclassified as sea. 

A subset of the misclassifications can likely be resolved if utilising colour information, since colour can be highly discriminative for dynamic textures. When considering the state-of-the-art results summarised in Table \ref{tab:dyntex-res}, these also indicate that a ``deeper" framework, extracting more high-level abstract features would most likely improve the performance. Options that may be considered are adding additional layers of abstraction before or after the histograms, for example, by learning hierarchical features from data or using bag-of-words models.

\section{Summary and discussion}
We have presented a new family of video descriptors based on joint histograms of spatio-temporal receptive field responses and evaluated several members in this family on the problem of dynamic texture recognition. This is, to our knowledge, the first video descriptor that uses \emph{joint statistics} of a set of \emph{``ideal''} (in the sense of derived on the basis of pure mathematical reasoning) spatio-temporal scale-space filters for video analysis and the first quantitative performance evaluation of using the family of \emph{time-causal scale-space filters} derived by Lindeberg \cite{Lin16-JMIV} as primitives for video analysis. Our proposed approach generalises a previous method by Linde and Lindeberg \cite{LinLin04-ICPR,LinLin12-CVIU}, based on joint histograms of receptive field responses, from the spatial to the spatio-temporal domain and from object recognition to dynamic texture recognition. 

Our experimental evaluation on several benchmarks from two widely used dynamic texture datasets demonstrates competitive performance compared to state-of-the-art dynamic texture recognition methods. The best results are achiev\-ed for our binary STRF $N$-jet descriptor, which combines directionally selective spatial and spatio-temporal receptive field responses. For the UCLA benchmarks, the STRF $N$-jet descriptor achieves the highest mean accuracy averaged over all benchmarks as well as the shared best or single best result for several benchmarks. In addition, our approach achieves improved results on all the UCLA benchmarks compared to a large range of similar methods also based on gathering statistics of local space-time structure but using \emph{different spatio-temporal primitives}. 

For the larger more complex DynTex benchmarks, deep learning approaches come out on top. However, also here our proposed video descriptor achieves improved performance compared to a range of similar methods, such as local binary pattern based methods including recent extensions and improvements (Zhao and Pietikainen \cite{ZhaPie-WDV2006,ZhaGuoPie-TPAMI-2007}; Ren et al.\ \cite{RenJiaYua-ICASSP2013}, Hong et al. \cite{HonRyuetal-MSSP2016}) and two methods learning filters by means of unsupervised learning (Arashloo and Kittler \cite{AraKit-TOM2014}; Arashloo et al.\ \cite{AraAmiNor-JVCIR2017}). The improved performance compared to approaches that learn filters from data (where for the UCLA benchmarks this also includes two deep learning based approaches), shows that designing filters based on structural priors of the world can for these tasks be as effective as learning.

It should be noted that, we believe an extension of our framework, for example by complementing the current single layer of receptive fields with hierarchical features, would, indeed, be beneficial to address more complicated visual tasks. Therefore, the presented approach is not intended as a final point, but rather as a first conceptually simple approach for performing video-analysis based on these new spatio-temporal primitives. Using a set of localised histograms, which capture the relative locations of different image structures, or learning or designing higher level hierarchical features on top of such local spatio-temporal derivative responses, will most likely improve the performance. A single global histogram would of course not be appropriate for videos that are not pre-segmented or do not have the spatial and temporal stationarity properties of dynamic textures. To handle such scenes, regional histograms should instead be computed. Also, in the presence of global camera motion, velocity adaptation would be beneficial. However, the objective here has primarily been to use these new time-causal spatio-temporal primitives within a simple framework, to focus more on a first evaluation of the spatio-temporal primitives and less on what is to be built on top. Thus, the presented framework is not aimed at directly competing with such conceptually more complex methods. Our approach does not include combinations of different feature types, ensemble classifiers, regional pooling to capture the relative location of features or learned or handcrafted mid-level features. However, constructing a more complex framework on top of these spatio-temporal primitives is certainly possible and would with high probability result in additional performance gains. 

In summary, our conceptually simple video descriptor achieves highly competitive performance across all benchmarks compared to other grey-scale ``shallow" methods and improved performance compared to all other methods of similar conceptual complexity using different spatio-temporal primitives, either handcrafted or learned from data. We consider this as strong validation that these time-causal spatio-temporal receptive fields are highly descriptive for modelling the local space-time structure and as evidence in favour of their general applicability as primitives for other video analysis tasks.

Our approach could also be implemented using non-causal Gaussian spatio-temporal scale-space kernels. This might give somewhat improved results, since at each point in time, additional information from the future could then also be used. However, a time-delayed Gaussian kernel would imply longer temporal delays -- thereby making it less suited for time critical applications -- in addition to requiring more computations and larger temporal buffers. The computational advantages of these time-causal filters imply that they can, in fact, be preferable also in settings where the temporal causality might not be of primary interest.

The scale-space filters used in this work have a strong connection to biology in the sense that a subset of these receptive fields very well model both spatial and spatio-temporal receptive fields in the LGN and V1 (DeAngelis et al. \cite{DeAngOhzFre95-TINS}, Lindeberg \cite{Lin13-BICY,Lin16-JMIV}). The receptive fields in V1, V2, V4 and MT serve as input to a large number of higher visual areas. It does indeed appear attractive to be able to use similar filters for early visual processing over a wide range of visual tasks. This study can be seen as a first step in a more general investigation into what can be done with spatio-temporal features similar to those in the primate brain. 

Directions for future work include: (i) A multi-scale recognition framework, where each video is represented by a \emph{set of descriptors} computed at multiple spatial and temporal scales, both during training and testing. This  would enable scale-invariant recognition according to the theory presented in Appendix \ref{app:cov-properties}. (ii) An extension to colour by including spatio-chromo-temporal receptive fields would with high probability improve the performance for tasks where colour information is discriminative. (iii)~Including non-separable spatio-temporal receptive fields with non-zero image velocities in the video descriptor according to the more general receptive field model in \cite{Lin16-JMIV}. Such velocity-adapted receptive fields, in fact, constitute a dominant portion of the receptive fields in areas V1 and MT (DeAngelis et al. \cite{DeAngOhzFre95-TINS}). (iv) Using position-dependent histograms to take into account the relative locations of features. 
(v) Learning or designing higher level hierarchical features based on these spatio-temporal primitives. 

We propose that the spatio-temporal receptive field framework should be of more general applicability to other video analysis tasks. Time-causal spatio-temporal receptive fields are indeed the visual primitives used for solving a large range of visual tasks for biological agents. 
The theoretical properties of these spatio-temporal receptive fields imply that they can be used to design methods that are provably invariant or robust to different types of natural image transformations, where such invariances will reduce the sample complexity for learning. We thus see the possibility to both integrate time-causal spatio-temporal receptive fields into current video analysis methods and to design new types of methods based on these primitives.

\begin{acknowledgements}
We would like to thank Oskar Linde for providing access to his code for computing 
joint receptive field histograms for spatial object recognition, which  has influenced our 
implementation for spatio-temporal recognition. 
\end{acknowledgements}

\appendix

\section{Appendix}
\subsection{Covariance properties of partial derivatives, PCA
  components and differential invariants over spatio-temporal scales}
\label{app:cov-properties}
  
The video descriptors that we construct are based on combinations of
partial derivatives
\begin{equation}
  L_{x^{m_1} y^{m_2} t^n}
\end{equation}
of different orders $(m_1, m_2, n)$ 
computed at multiple spatio-temporal scales $(s, \tau)$, where
$s$ denotes the spatial scale parameter and $\tau$ the temporal scale parameter.
Specifically, all video descriptors are expressed in terms of
scale-normalized derivatives (Lindeberg \cite{Lin97-IJCV,Lin16-JMIV})
\begin{equation}
  \partial_{\xi} = s^{\gamma_s/2} \partial_x, \quad
  \partial_{\eta} = s^{\gamma_s/2} \partial_y, \quad
   \partial_{\zeta} = \tau^{\gamma_{\tau}/2} \partial_t
\end{equation}
written
\begin{equation}
  L_{\xi^{m_1} \eta^{m_2} \zeta^n} 
  = s^{(m_1 + m_2) \gamma_s/2} \, \tau^{n \gamma_{\tau}/2} \, L_{x^{m_1} y^{m_2} t^n}
\end{equation}
where $\gamma_s > 0$ and $\gamma_{\tau} > 0$ are scale normalization
powers of the spatial and the temporal domains, respectively.
 
Consider an independent scaling transformation of the spatial and the
temporal domains
\begin{equation}
  f'(x_1', x_2', t') = f(x_1, x_2, t)
\end{equation}
for
\begin{equation}
  \label{eq-indep-sc-trans-spat-temp}
  (x_1', x_2', t') = (S_s \, x_1, S_s \, x_2, S_{\tau} \, t)
\end{equation}
where $S_s$ and $S_{\tau}$ denote the spatial and temporal scaling
factors, respectively. Define the space-time separable
spatio-temporal scale-space representations $L$ and $L'$ 
of $f$ and $f'$, respectively, according to
\begin{align}
  \begin{split}
      & L(x_1, x_2, t;\; s, \tau) =
  \end{split}\nonumber\\
  \begin{split}
      & \quad\quad
         \left( T(\cdot, \cdot, \cdot;\; s, \tau) * f(\cdot, \cdot, \cdot) \right)(x_1, x_2, t;\; s, \tau)
   \end{split}\\
  \begin{split}
      & L'(x'_1, x'_2, t';\; s', \tau')  = 
   \end{split}\nonumber\\
  \begin{split}
     & \quad\quad
     \left( T(\cdot, \cdot, \cdot;\; s', \tau') * f'(\cdot, \cdot, \cdot) \right)(x'_1, x'_2, t';\; s', \tau')
  \end{split}
\end{align}
Then, for the spatio-temporal scale-space kernel defined from the
Gaussian kernel $g(x, y;\; s)$ over the spatial domain and the time-causal limit
kernel $\Psi(t;\; \tau, c)$ over the temporal domain
\begin{equation}
  T(x, y, t;\; s, \tau) = g(x, y;\; s) \, \Psi(t;\; \tau, c)
\end{equation}
these spatio-temporal scale-space representations will be closed under
independent scaling transformations of the spatial and the temporal
domains
\begin{equation}
   L'(x'_1, x'_2, t';\; s', \tau') = L(x_1, x_2, t;\; s, \tau)
\end{equation}
provided that the spatio-temporal scale levels are appropriately
matched \cite[Equation (26)]{Lin18-JMIV}
\begin{equation}
  \label{eq-match-spat-temp-scales}
  s' = S_s^2 \, s, \quad\quad \tau' = S_{\tau}^2 \, \tau
\end{equation}
This property holds for all non-zero spatial scaling factors $S_s$.
Because of the discrete nature of the temporal scale levels 
$\tau_k = \tau_0 \, c^{2k}$ in the time-causal temporal scale-space
representation obtained by convolution with the time-causal limit
kernel, this closedness property does, however, only hold for temporal scaling
factors $S_{\tau}$ that are integer powers of the distribution
parameter $c$ of the time-causal limit kernel
\begin{equation}
  S_{\tau} = c^j \quad \mbox{for} \quad j \in \bbbz
\end{equation}
Specifically, under an independent scaling transformation of the
spatial and the temporal domains (\ref{eq-indep-sc-trans-spat-temp}),
the partial derivatives transform according to \cite[Equation~(9)]{Lin17-SSVM}
\begin{multline}
  L'_{\xi'^{m_1} \eta'^{m_2} t'^n}(x', y', t';\; s', \tau') = \\ 
  = S_s^{(m_1 + m_2)(\gamma_s - 1)} \, 
     S_{\tau}^{n (\gamma_{\tau} - 1)} \, 
     L_{\xi^{m_1} \eta^{m_2} \zeta^n}(x, y, t;\; s, \tau)
\end{multline}
In particular, for the specific choice of $\gamma_s = 1$ and $\gamma_{\tau} = 1$
these partial derivatives will be equal
\begin{equation}
  L'_{\xi'^{m_1} \eta'^{m_2} t'^n}(x', y', t';\; s', \tau')
  = L_{\xi^{m_1} \eta^{m_2} \zeta^n}(x, y, t;\; s, \tau)
\end{equation}
when computed at matching spatio-temporal scales
(\ref{eq-match-spat-temp-scales}).

Concerning the dimensionality reduction step, where a large set of
scale-normalised partial derivatives $L_{\xi^{m_1} \eta^{m_2} \zeta^n}$
is replaced by a lower-dimensional set of PCA-components
\begin{equation}
  \tilde{F}_i = \sum_j w_{ij} \, L_{\xi^{m_{1,i,j}} \eta^{m_{2,i,j}} \zeta^{n_{i,j}}}
\end{equation}
this implies that also the PCA components will be equal at matching spatio-temporal scales
(\ref{eq-match-spat-temp-scales}), although constituting linear
combinations of partial derivatives of a different spatial orders 
$(m_{1,i,j}, m_{2,i,j})$ and temporal orders $n_{i,j}$

This property does also extend to scale-normalised spatio-temporal differential
invariants computed for $\gamma_s = 1$ and $\gamma_{\tau} = 1$ 
as well as to PCA components defined from linear 
combinations of such scale-normalised spatio-temporal differential invariants.

In these respects, our video descriptors are truly scale covariant
under independent scaling transformations of the spatial and the
temporal domains.

Corresponding spatio-temporal scale-covariance properties do also hold
for video descriptors defined from the non-causal Gaussian spatio-temporal scale-space
representation computed based on spatial smoothing with the regular
two-dimensional Gaussian kernel and temporal smoothing with the
one-dimensional non-causal temporal Gaussian kernel.
Then, the scale covariance properties hold for all non-zero spatial
scaling constants $S_s$ and all non-zero temporal scaling constants $S_{\tau}$.

\subsection{Parameters for benchmark results}
\label{app:params}
This section presents the parameter settings used for comparing descriptors constructed from different sets of receptive fields (Section \ref{sec:expm-rfgroups}) and for comparing our approach with state-of-the-art (Section \ref{sec:expm-comp-stateart}). The parameters are
shown for the UCLA benchmarks in Table \ref{tab:benchmark-params-ucla} and for the DynTex benchmarks in Table \ref{tab:benchmark-params-dyntex}. For the STRF $N$-jet, STRF RotInv and RF Spatial descriptors, the parameters have been determined by cross-validation (see Section~\ref{sec:expm-ptuning}) while the STRF $N$-jet (previous) descriptor is tested with the heuristically chosen parameters used in \cite{JanLin-SSVM2017} (and included here for completeness). 
In the quite frequent case that several sets of parameters give the same classification performance, only one of these parameter settings is reported here.

\begin{table*}[]
 \caption{The parameter values selected for the UCLA benchmark results. Note that for the STRF $N$-jet, STRF RotInv and RF Spatial descriptors, parameter tuning has been performed according to Section \ref{sec:expm-ptuning},
while the STRF $N$-jet descriptor has been computed with the heuristically chosen parameters used in 
\cite{JanLin-SSVM2017}.}
\begin{center}
 \footnotesize
  \begin{tabular}{l l l l l l l l}
	\hline\noalign{\smallskip}
    & Descriptor & Benchmark & Classifier & $n_{bins}$ & $n_{comp}$  & Spatial scales (pixels) & Temporal scales (ms) \\
\noalign{\smallskip}\hline\noalign{\smallskip}

     & STRF $N$-jet & UCLA8 & SVM & 2 & 16 & (4, 8) & (100, 200) \\ 	
     & STRF $N$-jet & UCLA8 & NN & 2 & 16 & (4) &  (50) \\ 			
     
     & STRF $N$-jet & UCLA9 & SVM & 2 & 14 & (1, 2) & (50, 100) \\ 		
     & STRF $N$-jet & UCLA9 & NN & 2 & 14 & (1) & (50) \\ 		
     
     & STRF $N$-jet & UCLA50 & SVM & 2 & 13 & (4, 8) & (50, 100) \\ 	
     & STRF $N$-jet & UCLA50 & NN & 2 & 12 & (8, 16) & (50, 100) \\ 	
 

     & STRF RotInv & UCLA8 & SVM & 2 & 14 & (4, 8) & (50, 100) \\ 	
     & STRF RotInv & UCLA8 & NN & 2 & 13 & (4, 8) &  (50, 100) \\ 	
     	
     & STRF RotInv & UCLA9 & SVM & 2 & 12 & (4, 8) & (50, 100) \\ 	
     & STRF RotInv & UCLA9 & NN & 2 & 13 & (1, 2)  & (100, 200) \\ 	

     & STRF RotInv & UCLA50 & SVM & 2 & 6 & (4, 8) & (50, 100) \\ 	
     & STRF RotInv & UCLA50 & NN & 2 & 5 & (4, 8) & (50, 100) \\

     & RF Spatial & UCLA8 & SVM & 2 & 10 & (8, 16) & - \\ 	
     & RF Spatial & UCLA8 & NN & 2 & 9 & (8, 16) & - \\ 	
 	
     & RF Spatial & UCLA9 & SVM & 2 & 9 & (4, 8) & - \\ 	
     & RF Spatial & UCLA9 & NN & 2 & 6 & (4, 8) & - \\ 		

     & RF Spatial & UCLA50 & SVM & 2 & 5 & (4, 8) & - \\ 	
     & RF Spatial & UCLA50 & NN & 2 & 5 & (4, 8) & - \\ 	
       
     & STRF $N$-jet (previous) & UCLA8 & SVM & 2 & 15 & $(1, 2)$ & (50, 100)\\ 	
     & STRF $N$-jet (previous) & UCLA8 & NN & 2 & 15 & (1, 2) & (50, 100)\\ 		
 
     & STRF $N$-jet (previous) & UCLA9 & SVM & 2 & 15 & (1, 2) & (50, 100)\\ 		
     & STRF $N$-jet (previous) & UCLA9 & NN & 2 & 15 & (1, 2) & (50, 100)\\ 		
 
     & STRF $N$-jet (previous) & UCLA50 & SVM & 2 & 15 & (1, 2) & (50, 100)\\ 		
     & STRF $N$-jet (previous) & UCLA50 & NN & 2 & 15 & (1, 2) & (50, 100)\\ 		

  \noalign{\smallskip}\hline\noalign{\smallskip}
  \end{tabular}
\end{center}

  \label{tab:benchmark-params-ucla}

 \caption{The parameter values selected for the DynTex benchmark results. Note that for the STRF $N$-jet, STRF RotInv and RF Spatial descriptors, 
 parameter tuning has been performed according to Section \ref{sec:expm-ptuning}, while the STRF $N$-jet descriptor has been computed with the heuristically chosen parameters used in \cite{JanLin-SSVM2017}.}

\begin{center}
 \footnotesize
  \begin{tabular}{l l l l l l l l l}
	\hline\noalign{\smallskip}
    & Descriptor & Benchmark & Classifier & $n_{bins}$ & $n_{comp}$  & Spatial scales (pixels) & Temporal scales (ms) \\
\noalign{\smallskip}\hline\noalign{\smallskip}

     & STRF $N$-jet & Alpha & SVM & 2 & 17 & (2) & (200) \\ 		
     & STRF $N$-jet & Alpha & NN & 2 & 11 & (1) &  (400) \\ 		

     & STRF $N$-jet & Beta & SVM & 2 & 17 & (8) & (200) \\ 		
     & STRF $N$-jet & Beta & NN & 2 & 13 & (2, 4) & (200, 400) \\ 	

     & STRF $N$-jet & Gamma & SVM & 2 & 16 & (4, 8) & (50, 100) \\ 	
     & STRF $N$-jet & Gamma & NN & 2 & 13 & (2, 4) & (200, 400) \\

     & STRF RotInv & Alpha & SVM & 2 & 5 & (8,  16) & (100, 200) \\	
     & STRF RotInv & Alpha & NN & 2 & 5 & (8, 16) &  (100, 200) \\ 	

     & STRF RotInv & Beta & SVM & 2 & 15 & (4, 8) & (200, 400)  \\ 	
     & STRF RotInv & Beta & NN & 2 & 17 & (8, 16)  & (100, 200) \\ 	
     
     & STRF RotInv & Gamma & SVM & 2 & 15 & (2, 4) & (100, 200) \\  
     & STRF RotInv & Gamma & NN & 2 & 16 & (2, 4) & (50, 100) \\

     & RF Spatial  & Alpha & SVM & 2 & 8 & (8, 16) & - \\	
     & RF Spatial  & Alpha & NN & 2 & 5 & (4, 8) & - \\ 		
    
     & RF Spatial  & Beta & SVM & 2 & 10 & (2, 4) & - \\ 		
     & RF Spatial  & Beta & NN & 2 & 10 & (4, 8) & - \\ 		
     
     & RF Spatial  & Gamma & SVM & 2 & 10 & (2, 4) & - \\ 	
     & RF Spatial  & Gamma & NN & 2 & 10 & (2, 4) & - \\

     & STRF $N$-jet (previous) & Alpha & SVM & 2 & 15 & (2, 4) & (200, 400) \\ 		
     & STRF $N$-jet (previous) & Alpha & NN & 2 & 15 & (2, 4) & (200, 400) \\ 		
  
     & STRF $N$-jet (previous) & Beta & SVM & 2 & 15 & (2, 4) & (200, 400) \\ 		
     & STRF $N$-jet (previous) & Beta & NN & 2 & 15 & (2, 4) & (200, 400) \\ 		
  
     & STRF $N$-jet (previous) & Gamma & SVM & 2 & 15 & (2, 4) & (200, 400) \\ 	
     & STRF $N$-jet (previous) & Gamma & NN & 2 & 15 & (2, 4) & (200, 400) \\ 		

  \noalign{\smallskip}\hline\noalign{\smallskip}
  \end{tabular}
\end{center}

  \label{tab:benchmark-params-dyntex}
\end{table*}

\bibliographystyle{spmpsci}      
\bibliography{defs,tlmac,yjrefs2}   

\end{document}